%% file: main.tex
\documentclass[11pt]{article}

\usepackage[final]{acl}

\usepackage{times}
\usepackage{latexsym}

\usepackage[T1]{fontenc}

\usepackage[utf8]{inputenc}

\usepackage{microtype}

\usepackage{inconsolata}

\usepackage{graphicx}

\usepackage[utf8]{inputenc} 
\usepackage[T1]{fontenc}    
\usepackage[table]{xcolor}
\definecolor{mygreen}{RGB}{34,139,34} 
\usepackage{url}            
\usepackage{booktabs}       
\usepackage{amsfonts}       
\usepackage{nicefrac}       
\usepackage{microtype}      
\usepackage{tcolorbox}
\usepackage{amsmath}
\usepackage{array}
\usepackage{makecell}
\usepackage{longtable}
\usepackage{caption}
\usepackage{pifont}
\usepackage{inconsolata}
\usepackage{enumitem}
\usepackage{titletoc}
\usepackage{amssymb}
\usepackage{multirow}
\usepackage[normalem]{ulem}
\usepackage{tabularx}
\usepackage{xurl}

\definecolor{red}{RGB}{255,222,222} 
\definecolor{bred}{RGB}{230,57,70} 
\definecolor{purp}{RGB}{169, 57, 230} 

\title{Grounded or Guessing? LVLM Confidence Estimation via\\Blind-Image Contrastive Ranking}

\author{
  \textbf{Reza Khanmohammadi\textsuperscript{\textnormal{1}}\thanks{Corresponding author: \href{mailto:khanreza@msu.edu}{\texttt{khanreza@msu.edu}}}}\textnormal{,}
  \textbf{Erfan Miahi\textsuperscript{\textnormal{2}}}\textnormal{,}
  \textbf{Simerjot Kaur\textsuperscript{\textnormal{3}}}\textnormal{,}
\\
  \textbf{Charese H. Smiley\textsuperscript{\textnormal{3}}}\textnormal{,}
  \textbf{Ivan Brugere\textsuperscript{\textnormal{3}}}\textnormal{,}
  \textbf{Kundan Thind\textsuperscript{\textnormal{4,\textdagger}}}\textnormal{,}
  \textbf{Mohammad M. Ghassemi\textsuperscript{\textnormal{1},}\thanks{Shared senior authorship}}
\\
\\
  \textsuperscript{1}Michigan State University \quad
  \textsuperscript{2}Independent AI Researcher \\
  \textsuperscript{3}JPMorgan AI Research \quad
  \textsuperscript{4}Henry Ford Health
\\
      \normalsize\textsuperscript{1} \texttt{\{khanreza, ghassem3\}@msu.edu} \quad \textsuperscript{2} \texttt{mhi.erfan1@gmail.com} \\
      \normalsize\textsuperscript{3} \texttt{\{simerjot.kaur, charese.h.smiley, ivan.brugere\}@jpmchase.com} \quad \textsuperscript{4} \texttt{kthind1@hfhs.org}
}

\begin{document}
\maketitle





\input{sections/abstract}
\input{sections/introduction}
\input{sections/related_work}
\input{sections/dataset}
\input{sections/method}
\input{sections/results}
\input{sections/discussion}
\input{sections/conclusion}
\input{sections/limitations}
\input{sections/ethical_considerations}
\input{sections/acknowledgments}
\input{sections/disclaimer}
\bibliography{references}
\clearpage
\appendix
\onecolumn

\begin{center}
\huge\textbf{Appendix}
\end{center}

\begin{center}
\Large\textbf{Table of Contents}
\end{center}

\rule{\linewidth}{0.6pt}

{\hypersetup{linkcolor=black}
\startcontents[sections]
\printcontents[sections]{}{1}{}}

\rule{\linewidth}{0.6pt}

\twocolumn

\input{sections/appendix_dataset}
\input{sections/appendix_baselines}
\input{sections/appendix_novelty}
\input{sections/appendix_metrics}
\input{sections/appendix_validation}
\input{sections/appendix_optuna}
\input{sections/appendix_params}
\input{sections/appendix_design}
\input{sections/appendix_extended_results}
\input{sections/appendix_grounding_eval}
\end{document}

%% file: sections/abstract.tex
\begin{abstract}
Large vision-language models (LVLMs) suffer from visual ungroundedness: they can produce a fluent, confident, and even correct response driven entirely by language priors, with the image contributing nothing to the prediction. Existing confidence estimation methods cannot detect this, as they observe model behavior under normal inference with no mechanism to determine whether a prediction was shaped by the image or by text alone. We introduce \textbf{BICR} (\textbf{B}lind-\textbf{I}mage \textbf{C}ontrastive \textbf{R}anking), a model-agnostic confidence estimation framework. The LVLM stays entirely frozen; instead, we extract its hidden states twice per sample, once with the real image-question pair and once with the image blacked out while the question is held fixed. A small, separate probe is then trained on the real-image hidden state, regularized by a ranking loss that penalizes higher confidence on the blacked-out view, teaching it to treat visual grounding as a signal of reliability. At inference only the single real-image pass is used, so BICR adds zero cost beyond a standard single-view probe. Evaluated across five modern LVLMs and seven baselines on a benchmark covering visual question answering, object hallucination detection, medical imaging, and financial document understanding, BICR achieves the best cross-LVLM average on both calibration and discrimination simultaneously, with statistically significant discrimination gains robust to cluster-aware analysis, at 4--18$\times$ fewer parameters than the strongest probing baseline.
\end{abstract}

\noindent\begin{minipage}{\linewidth}
{\normalsize \textbf{Code} --- {\small \href{https://github.com/Ledengary/BICR}{https://github.com/Ledengary/BICR}}} \\
{\normalsize \textbf{Data} --- {\small \href{https://huggingface.co/datasets/ledengary/VLCB}{https://huggingface.co/datasets/ledengary/VLCB}}}
\end{minipage}
\vspace{-10pt}

%% file: sections/introduction.tex
\section{Introduction}
\label{sec:introduction}

\begin{figure*}[t]
\centering
\includegraphics[width=\textwidth]{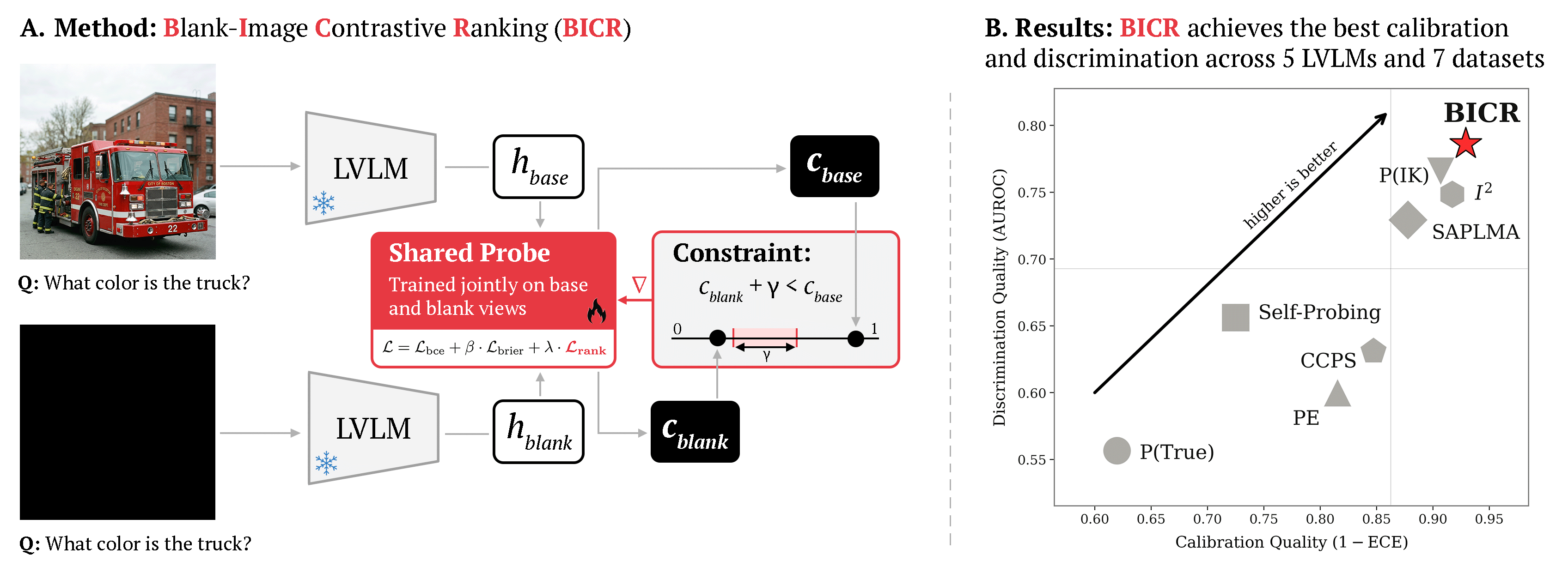}
\vspace{-10pt}
\caption{\textbf{Overview of our method (\textbf\texttt{BICR}) and headline results.} \textbf{(A)} \texttt{BICR} pairs each question with two views, the real image and a blank counterfactual, and trains a shared probe on top of a frozen large vision-language model (LVLM). The ranking loss $\mathcal{L}_{rank}$ enforces that the real-view confidence $c_{base}$ exceeds the blank-view confidence $c_{blank}$ by a margin $\gamma$, teaching the probe that confidence must be grounded in the visual input. Only the base view is used at inference. \textbf{(B)} Each marker is one confidence estimator's cross-LVLM average across our 5 LVLMs and 7 datasets, plotting calibration quality, $1 - \mathrm{ECE}$ (Expected Calibration Error), against discrimination quality, AUROC (Area Under the ROC Curve). \texttt{BICR} (red star) is the only method in the upper-right region, achieving both high calibration and high discrimination simultaneously.}
\label{fig:main}
\vspace{-10pt}
\end{figure*}

\noindent\textbf{In Large Vision-Language Models (LVLMs), an answer can be confident and correct while being entirely driven by language priors.} LVLMs function by prepending a sequence of visual tokens, produced by a vision encoder, to the language model's input context before any text generation begins. The language backbone receives both image and text as input, but \textit{this does not mean the answer is driven by both}. Specific attention heads in LVLMs attend up to five times more strongly to text than to visual tokens~\citep{yang2025mitigating}, and answer distributions are nearly indistinguishable whether the image is provided or replaced with a blank input~\citep{Woo2025AVISC,Hidden}. A model can therefore produce a fluent, confident, and even correct response based almost entirely on learned linguistic priors, with the image contributing nothing to the prediction. This failure mode, which we term \textit{visual ungroundedness}, is distinct from ordinary model error: the model is not reasoning incorrectly about the image, it is bypassing the image altogether.

\noindent\textbf{Existing confidence estimation methods cannot tell visually ungrounded predictions apart from grounded ones.} Hidden-state probing methods do carry useful discriminative signal in the LVLM setting~\citep{li2024referencefree}, but they share a common blind spot: a probe trained on representations extracted under normal inference sees only a single snapshot of the model's internal state, from which it cannot determine whether that state was shaped by the image or by text alone. Without exposure to that contrast during training, such a probe will assign indistinguishable confidence scores to grounded and ungrounded predictions, systematically overstating confidence in the latter.

\noindent\textbf{Detecting visual ungroundedness still requires getting the two basic properties of a confidence score right, and doing so simultaneously remains an open challenge.} A useful confidence estimator must satisfy two properties at once: \textit{calibration} (expressed confidence matches empirical accuracy, so that a model assigning 80\% confidence to a set of answers is correct on roughly 8 of 10 of them~\citep{tempscaling}), and \textit{discrimination} (confidence scores meaningfully separate correct predictions from incorrect ones). LVLMs do not naturally exhibit either: when asked a question they cannot reliably answer, LVLMs respond with the same fluent confidence as when they are correct, expressing no meaningful uncertainty~\citep{tempscaling, dang2026instinct, chen2025unveilinguncertainty}. Prior methods have made progress on each property individually (see \S\ref{sec:related_work}), but jointly satisfying both, especially under a visual modality, remains an open question.

\noindent\textbf{\textbf{\color{bred}B}lind-\textbf{\color{bred}I}mage \textbf{\color{bred}C}ontrastive \textbf{\color{bred}R}anking ({\color{bred}\texttt{BICR}}) addresses visual ungroundedness directly by teaching a confidence probe to distinguish image-driven predictions from language-prior-driven ones.} For each sample, \texttt{BICR} extracts hidden states from a frozen LVLM twice, once with the original image-question pair and once with the image blacked out. The gap between these two hidden states is an operational proxy for visual grounding, which we use to predict correctness rather than to measure grounding directly. A lightweight probe is trained on real-image hidden states with a ranking loss penalizing higher confidence on the blacked-out view, teaching the probe to treat this gap as a reliability signal at no inference cost. Across five LVLMs and seven baselines on a benchmark spanning visual question answering (VQA), object hallucination detection, medical imaging, and financial document understanding (Figure~\ref{fig:main}), \texttt{BICR} achieves the best cross-LVLM averages on calibration and discrimination simultaneously, with statistically significant discrimination gains robust to cluster-aware analysis and 4--18$\times$ fewer parameters than the strongest probing baseline, InternalInspector~\citep{internalinspector}. The contributions of this work are as follows:
\begin{itemize}
    \item \texttt{\textbf{BICR}}: a model-agnostic confidence estimation framework for LVLMs that trains a lightweight probe via a blind-image contrastive ranking objective to use visual grounding as a reliability signal, achieving the best cross-LVLM average on calibration (ECE, Brier) and discrimination (AUCPR, AUROC).
    \item \texttt{\textbf{VLCB}}: a benchmark aggregating seven public VQA sources with model responses and correctness labels across five modern LVLMs, spanning general, medical, and financial reasoning domains, released publicly to support reproducibility and future work on confidence estimation in multimodal models.
    \item \textbf{Comprehensive evaluation} of seven confidence estimation baselines (P(True), Self-Probing, Prompt Ensembles, P(I~Know), SAPLMA, InternalInspector, CCPS) across five LVLM architectures and diverse high-stakes domains, providing the first systematic benchmark of confidence estimation in LVLMs under a unified protocol that jointly reports calibration and discrimination.
    \item \textbf{Empirical evidence} that the gap between real and blank-image hidden states, our proxy for visual grounding, gives a reliable and discriminative signal of correctness across models, tasks, and domains.
\end{itemize}

%% file: sections/related_work.tex
\section{Related Work}
\label{sec:related_work}

\noindent\textbf{Confidence Estimation in Language Models.} Confidence estimation methods for language models fall into four families. \emph{Prompt-based} methods elicit a confidence signal through additional queries: verbalized confidence asks the model to state its certainty numerically~\citep{self-probing,tian-etal-2023-just,zeng-etal-2025-thinking}, and self-consistency samples multiple responses and measures agreement~\citep{self-consistency}. \emph{Logit-based} methods read the output distribution directly, using token-level log-probabilities~\citep{zhou2025hademif} often rescaled via temperature scaling~\citep{jiang-etal-2021-know,tempscaling}. \emph{Internal-state probing} methods train classifiers on hidden activations: P(IK)~\citep{kadavath} probes the state before generation, SAPLMA~\citep{saplma} reads hidden-layer activations on the generated response, and InternalInspector~\citep{internalinspector} pools attention, feed-forward, and residual states across all layers through a learned encoder. \emph{Internal-stability} methods probe representational response to perturbation: CCPS~\citep{ccps} applies targeted adversarial perturbations to the final hidden state and trains a classifier on the resulting trajectory. We benchmark \texttt{BICR} against these seven methods in \S\ref{sec:results}.

\noindent\textbf{Calibration and Grounding in Vision-Language Models.} LVLMs prepend visual tokens to the language model's input, but this architectural integration does not guarantee that visual content drives the answer~\citep{yang2025mitigating,Woo2025AVISC,Hidden}. Even when output is correct, meaning-preserving image perturbations induce substantial shifts in internal representations~\citep{Wani2024HiddenInstability}, and LVLMs are persistently miscalibrated, with verbalized confidence poorly tracking correctness in ways prompting and temperature scaling do not correct~\citep{dang2026instinct,chen2025unveilinguncertainty,xuan2025seeingbelieving}. Recent LVLM-specific methods address adjacent symptoms without producing a per-sample calibrated score: CSP~\citep{Zhao2025SemanticCalibration} and medical VQA frameworks~\citep{Du2026MedicalVQA} target output-level calibration, while VL-Calibration~\citep{vl-calibration} decomposes confidence into visual and reasoning components via reinforcement learning fine-tuning. A parallel line targets hallucination rather than confidence: VCD~\citep{10657718} contrasts output distributions from original and noise-distorted images at every decoding step, MemVR~\citep{MemVR} re-injects visual tokens as key-value memory at a middle layer when the model is uncertain, and VL-Uncertainty~\citep{VL-Uncertainty} clusters responses to perturbed inputs and reports hallucination-detection accuracy. These interventions all rewrite the generated answer rather than emit a per-sample confidence score, and they move in different directions on the visual signal: VCD degrades the view, MemVR supplements it, and \texttt{BICR} removes it. None targets the joint calibration-and-discrimination protocol we adopt, and they collectively retrain the model, work at the output level, or add inference cost without addressing whether internal representations are anchored in visual evidence.

\noindent\textbf{Hidden-State and Attention Probing in LVLMs.} Hidden-state probing carries useful signal in LVLMs: classifiers on final-layer representations achieve strong hallucination detection across architectures and tasks~\citep{li2024referencefree}. Two adjacent lines read related internal signals for different purposes than per-sample confidence: SVAR~\citep{Devils} uses the middle-layer visual attention ratio to classify object-level hallucination during generation, and TVI~\citep{Chain-of-Embedding} contrasts hidden states with and without the image to localize a visual integration point and quantify language prior, reported as a population-level Spearman correlation rather than a per-sample probability. However, when the visual input is degraded, probe performance degrades sharply with it~\citep{li2024referencefree}, revealing that these probes read the quality of the visual signal in the hidden state rather than whether any visual signal was used. A probe reading an ungrounded prediction and a grounded one may therefore see nearly identical representations, since neither was ever exposed to that contrast.

\noindent\textbf{The missing contrastive signal.} Across all four paradigms, no method trains a per-sample confidence estimator on the contrast between grounded and ungrounded representations. Probing methods (P(IK), SAPLMA, InternalInspector, SVAR) read snapshots under normal inference; output-level methods (CSP, VL-Calibration, VCD, VL-Uncertainty) do not expose internal grounding; TVI performs the contrast but as a population diagnostic, not a per-sample correctness-trained score. \texttt{BICR} fills this gap: it is the only method that is at once a frozen-model probe, read from a single last-prompt-token state before generation, supervised on correctness, built on a real-versus-blank contrast, and evaluated jointly on calibration and discrimination. Appendix~\ref{app:novelty} positions \texttt{BICR} against each method family in detail (Table~\ref{tab:novelty}).

%% file: sections/dataset.tex
\section{\texorpdfstring{\textbf{\color{purp}V}ision-\textbf{\color{purp}L}anguage Model \textbf{\color{purp}C}onfidence Estimation \textbf{\color{purp}B}enchmark (\textbf{\color{purp}\texttt{VLCB}})}{Vision-Language Model Confidence Estimation Benchmark (VLCB)}}
\label{sec:benchmark}

Evaluating confidence estimation methods requires a benchmark that provides correctness labels for model responses across diverse tasks and model families, supports a clear separation between training and test distributions, and spans the domains where reliable confidence scores matter most. To our knowledge, no existing resource satisfies all three, which is what \texttt{VLCB} is designed to address. We evaluate \texttt{BICR} alongside seven baselines spanning three paradigm families: prompt-based methods that treat the LVLM as a black box (P(True)~\citep{kadavath}, Self-Probing~\citep{self-probing}, Prompt Ensembles~\citep{promptensemble}), internal-state probes that train lightweight classifiers on hidden-state snapshots (P(I~Know)~\citep{kadavath}, SAPLMA~\citep{saplma}, InternalInspector~\citep{internalinspector}), and one internal-stability method that reads representational robustness under perturbation (CCPS~\citep{ccps}). Full baseline descriptions are provided in Appendix~\ref{app:baselines}.

\noindent\textbf{Design principle.} The central design choice in \texttt{VLCB} is deliberate distribution shift between training and evaluation. Confidence probe training uses GQA~\citep{GQA} exclusively: 20,000 training and 5,000 validation samples stratified by question type, with short, unambiguous answers. The test set is detailed in Appendix~\ref{app:appendix_dataset} and combines a held-out GQA test split for in-distribution reference with six additional datasets unseen during training: POPE~\citep{POPE} for object hallucination detection, GMAI-MMBench~\citep{GMAI-MMBench} for medical multimodal reasoning, MME-Finance~\citep{MME-Finance} for financial chart understanding, MMMU-Pro~\citep{MMMU} in 4-option and 10-option configurations for college-level reasoning, and LLaVA-in-the-Wild~\citep{LlavaWild} for Open-ended visual dialogue. Table~\ref{tab:vlcb_splits} summarizes the splits. This out-of-domain setup makes performance meaningful: a confidence estimator that succeeds only on its training distribution offers no deployment guarantee.

\begin{table}[t]
\centering
\caption{\texttt{VLCB} train, validation, and test splits.}
\label{tab:vlcb_splits}
\setlength{\tabcolsep}{5pt}
\resizebox{\columnwidth}{!}{%
\begin{tabular}{llrr}
\toprule
\textbf{Split} & \textbf{Source} & \textbf{N} & \textbf{\%} \\
\midrule
Train & GQA (VQA)                                  & 20,000 & 100.0 \\
Val   & GQA (VQA)                                  &  5,000 & 100.0 \\
\midrule
Test  & GQA (VQA)                                  & 12,568 & 41.2 \\
      & POPE (Object Hallucination)                &  9,000 & 29.5 \\
      & GMAI-MMBench (Medical Reasoning)           &  4,549 & 14.9 \\
      & MMMU-Pro 4-option (College Reasoning)      &  1,720 &  5.6 \\
      & MMMU-Pro 10-option (College Reasoning)     &  1,725 &  5.7 \\
      & MME-Finance (Chart Understanding)          &    892 &  2.9 \\
      & LLaVA-Wild (Visual Dialogue)               &     60 &  0.2 \\
\midrule
      & \textbf{Total}                             & \textbf{30,514} & \textbf{100.0} \\
\bottomrule
\end{tabular}%
}
\vspace{-10pt}
\end{table}

\noindent\textbf{Model coverage and response generation.} We evaluate five open-weight instruction-tuned LVLMs: Qwen3-VL-8B, LLaVA-NeXT-13B, InternVL3.5-14B, Gemma-3-27B, and DeepSeek-VL2. Together they cover 4.5B to 27B active parameters, three distinct vision encoder lineages, and both dense and mixture-of-experts language model architectures. Full architectural specifications are provided in Appendix~\ref{app:lvlms}. All five models are run on the complete train, validation, and test splits under identical generation conditions: greedy decoding with a maximum of 64 new tokens and images downscaled to a maximum of 2,048 pixels on the longer edge. Each model receives a semantically uniform prompt instructing it to answer briefly and completely; prompt delivery details vary by model chat template and are described in Appendix~\ref{app:gen_extract}.

\noindent\textbf{Correctness labeling.} Each generated response is assigned a binary correctness label by a \texttt{gpt-5-mini} judge applied uniformly across all seven source datasets and all five LVLMs. Using a single judge across task types, including multiple-choice questions where string matching would in principle suffice, ensures that formatting variation across LVLM chat templates does not introduce grading artifacts and that all correctness labels are produced by the same protocol. This practice is well established in the confidence estimation literature, and our approach follows the same protocols adopted by prior work~\citep{calibration-tuning, ccps, rmcb}. We further validate the judge against a stratified human-annotated audit (350 items, 98.3\% agreement, Cohen's $\kappa = 0.966$), detailed in Appendix~\ref{app:human_audit}. Per-LVLM correctness statistics broken down by dataset and split are reported in Appendix~\ref{app:appendix_dataset}.

%% file: sections/method.tex
\section{\texorpdfstring{\textbf{\color{bred}B}lind-\textbf{\color{bred}I}mage \textbf{\color{bred}C}ontrastive \textbf{\color{bred}R}anking (\textbf{\color{bred}\texttt{BICR}})}{Blind-Image Contrastive Ranking (BICR)}}
\label{sec:method}

\noindent\textbf{Problem setup.} Let $\mathcal{M}$ denote a frozen LVLM with hidden dimension $d_h$. Given a visual question $(q, v)$ with question text $q$ and image $v$, we perform a prompt-only forward pass through $\mathcal{M}$ and extract the hidden state at the last prompt token position, the point at which the model's representation of the full input is formed and generation would begin. We denote this hidden state $\mathbf{h}_{\mathrm{base}} \in \mathbb{R}^{d_h}$. The goal is to learn a function $f: \mathbb{R}^{d_h} \to \mathbb{R}$ whose sigmoid $\sigma(f(\mathbf{h}_{\mathrm{base}}))$ estimates the probability that the model's generated answer $a$ is correct, using only $\mathbf{h}_{\mathrm{base}}$ at inference time.

\noindent\textbf{Two-view hidden state extraction.} For each training sample, given question $q$, image $v$, and binary correctness label $y \in \{0,1\}$, where $y=1$ when the model's generated answer is judged correct against the ground truth and $y=0$ otherwise, we extract hidden states from $\mathcal{M}$ under two input conditions. The base view uses the original image: $\mathbf{h}_{\mathrm{base}}$ is the hidden state at the last prompt token of the forward pass over $(q, v)$. The blank view substitutes a solid black image $v_{\varnothing}$ (RGB $(0,0,0)$, matching the spatial dimensions of $v$) while holding $q$ fixed: $\mathbf{h}_{\mathrm{blank}}$ is the hidden state at the same position from the forward pass over $(q, v_{\varnothing})$. Both states come from the final decoder layer, so any difference between them is attributable solely to the visual input. The blank-image pass is computed once as a preprocessing step and reused across all training runs and seeds.

\noindent\textbf{Probe architecture.} The confidence probe $f$ is a multi-layer perceptron with ReLU activations and dropout between layers, mapping $\mathbf{h} \in \mathbb{R}^{d_h}$ to a scalar logit. The same MLP is shared across both views during training: $\mathbf{h}_{\mathrm{base}}$ and $\mathbf{h}_{\mathrm{blank}}$ pass through $f$ with identical parameters. This weight sharing is what allows the contrastive objective introduced below to shape the learned representation rather than merely fitting a separate decision threshold per view: a single probe must simultaneously produce high confidence for grounded real-image predictions and suppressed confidence for blank-image ones. The depth and width of the MLP are selected by Optuna hyperparameter search; full details are in Appendix~\ref{app:optuna}.

\noindent\textbf{Training objective.} The training loss combines three terms. The supervised loss is binary cross-entropy with positive-class weighting to handle label imbalance, where $n_+$ and $n_-$ are the counts of correct and incorrect samples in training:
\begin{equation}
\mathcal{L}_{\mathrm{bce}}
=
\mathrm{BCE}(f(\mathbf{h}_{\mathrm{base}}), y; w_+),
\quad
w_+ = n_-/n_+
\end{equation}

The calibration loss is a Brier score penalty on base-view predictions:
\begin{equation}
\resizebox{0.95\columnwidth}{!}{$
\mathcal{L}_{\mathrm{brier}}
=
\frac{1}{n}\sum_{i=1}^{n}
(\hat{p}_i^{\mathrm{base}} - y_i)^2,
\quad
\hat{p}_i^{\mathrm{base}}
=
\sigma(f(\mathbf{h}_{\mathrm{base},i}))
$}
\end{equation}

The visual grounding ranking loss is the core contribution of \texttt{BICR}. For each correctly answered sample ($y=1$), the probe is required to assign higher confidence when the real image is present than when it is absent, with margin $\gamma > 0$ enforcing a minimum gap between the two scores rather than merely requiring one to exceed the other:
\begin{equation}
\mathcal{L}_{\mathrm{rank}}
=
\frac{
\sum_i
\mathrm{ReLU}\!\big(
\gamma -
(\hat{p}_i^{\mathrm{base}} -
\hat{p}_i^{\mathrm{blank}})
\big)
y_i
}{
\sum_i y_i + \epsilon
}
\label{eq:rank}
\end{equation}
where $\hat{p}_i^{\mathrm{blank}} = \sigma(f(\mathbf{h}_{\mathrm{blank},i}))$ and $\epsilon = 10^{-8}$. The constraint is restricted to correct samples because the ranking direction has a clear semantic interpretation there: a correct answer that relied on the image should produce higher confidence with the image present than without it. For incorrect answers no directional constraint is warranted, as the answer is wrong regardless of visual grounding. The three terms are combined as:
\begin{equation}
\mathcal{L}
=
\mathcal{L}_{\mathrm{bce}}
+
\beta \mathcal{L}_{\mathrm{brier}}
+
\lambda \mathcal{L}_{\mathrm{rank}}
\label{eq:total_loss}
\end{equation}
where $\beta$, $\lambda$, and $\gamma$ are selected by Optuna. The contribution of each term is validated empirically in Appendix~\ref{app:design_loss}.

\noindent\textbf{Inference.} At test time, \texttt{BICR} requires a single prompt-only forward pass through the frozen LVLM to obtain $\mathbf{h}_{\mathrm{base}}$, followed by a pass through the probe:
\begin{equation}
\hat{c}
=
\sigma\!\big(
f(\mathbf{h}_{\mathrm{base}})
\big)
\end{equation}
The blank view's only consumer is $\mathcal{L}_{\mathrm{rank}}$ during training; because the probe is shared across both views, the contrast is absorbed into its weights, and the trained probe already suppresses confidence on blank-image-like states without a second pass. Inference therefore needs only the single real-image pass, so \texttt{BICR} adds zero inference overhead relative to any single-view probe. Trainable parameter counts are compared in Appendix~\ref{app:params}.

%% file: sections/results.tex
\section{Results}
\label{sec:results}

All methods are evaluated on the \texttt{VLCB} test split across five LVLMs under a shared protocol. We report four primary metrics: Expected Calibration Error (ECE) and Brier Score (BS) for calibration, and Area Under the Precision--Recall Curve (AUCPR) and Area Under the ROC Curve (AUROC) for discrimination. Full definitions and additional metrics are in Appendix~\ref{app:evaluation_metrics}. Trained methods report mean performance across five random seeds, each with 50 Optuna hyperparameter trials, selected via a composite validation score that jointly optimizes discrimination and calibration; the full training and selection protocol is described in Appendix~\ref{app:validation}.

\input{tables/main_results}

\begin{table}[t]
\centering
\footnotesize
\caption{Loss ablation for \texttt{BICR} (cross-LVLM average, 5 seeds $\times$ 5 LVLMs).}
\label{tab:ablation_main}
\setlength{\tabcolsep}{4pt}

\resizebox{\columnwidth}{!}{%
\begin{tabular}{lcccc}
\toprule
\textbf{Variant} &
\textbf{ECE}$\downarrow$ &
\textbf{BS}$\downarrow$ &
\textbf{AUCPR}$\uparrow$ &
\textbf{AUROC}$\uparrow$ \\
\midrule

\rowcolor{red}
\textbf{Full (\texttt{BICR})}
& \textbf{7.09}
& \textbf{18.41}
& \textbf{87.47}
& \textbf{78.63} \\

$-\mathcal{L}_{\mathrm{brier}}$
& 8.48 {\scriptsize(+1.39)}
& 19.04 {\scriptsize(+0.63)}
& 87.10 {\scriptsize(-0.37)}
& 78.02 {\scriptsize(-0.61)} \\

$-\mathcal{L}_{\mathrm{rank}}$
& 8.13 {\scriptsize(+1.04)}
& 19.63 {\scriptsize(+1.22)}
& 85.52 {\scriptsize(-1.95)}
& 75.31 {\scriptsize(-3.32)} \\

$\mathcal{L}_{\mathrm{bce}}$ only
& 9.15 {\scriptsize(+2.06)}
& 19.91 {\scriptsize(+1.50)}
& 85.48 {\scriptsize(-1.99)}
& 75.25 {\scriptsize(-3.38)} \\

\bottomrule
\end{tabular}%
}

\end{table}

\noindent\textbf{Main results.} Table~\ref{tab:main_results} presents the pooled aggregate performance of all methods across the five LVLMs. \texttt{BICR} achieves the best cross-LVLM average on all four metrics: ECE of 7.09\%, BS of 18.41\%, AUCPR of 87.47\%, and AUROC of 78.63\%. Among per-LVLM results, \texttt{BICR} leads on AUCPR and AUROC for every LVLM (five of five), and on ECE and BS for three of five LVLMs.

The two strongest baselines are P(I~Know) and InternalInspector. P(I~Know) is architecturally identical to \texttt{BICR}, an Optuna-tuned MLP over a single hidden-state vector, but is trained with BCE loss alone. Because P(I~Know) has no canonical probe architecture at this scale, we give it the identical Optuna search space and budget as \texttt{BICR}, so the two compete on the same architectural and optimization footing and their comparison isolates the training objective; the remaining trained baselines retain their published, validated configurations (Appendices~\ref{app:baselines},~\ref{app:optuna},~\ref{app:params}). \texttt{BICR}'s improvement over P(I~Know) ($+2.1$ AUROC, $+1.2$ AUCPR, $-2.2$ ECE on the cross-LVLM average) is therefore attributable to the training-time auxiliary losses (primarily the ranking signal from the blank-image comparison), not to architectural differences. InternalInspector achieves competitive ECE on two LVLMs (Qwen and Gemma) using a ResNet18-based CNN encoder with $11.3$M parameters, $7\times$ more than \texttt{BICR}'s average of $1.6$M, but falls behind on discrimination across all LVLMs. A detailed comparison of trainable parameter counts is provided in Appendix~\ref{app:params}. The three prompt-based methods (P(True), Self-Probing, Prompt Ensemble) trail the internal-state probes by sizable margins on the cross-LVLM average, with the smallest gap (Self-Probing to SAPLMA) at roughly 7 AUROC points and the largest exceeding 20, confirming that verbalized or logit-based confidence signals are insufficient for reliable confidence estimation without access to internal representations. All discrimination improvements of \texttt{BICR} over trained baselines are statistically significant under both pooled and cluster-aware analyses, and calibration improvements are significant against three of the four trained baselines (P(I~Know), SAPLMA, CCPS) under both analyses; full significance results are reported in Appendix~\ref{app:results_significance}.

\begin{figure*}[t]
\centering
\includegraphics[width=0.95\textwidth]{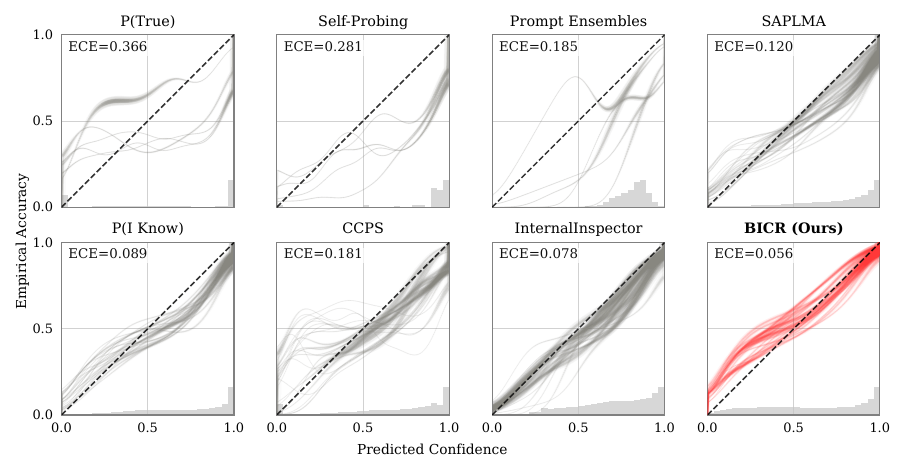}
\vspace{-10pt}
\caption{Cross-LVLM reliability diagrams for all eight methods. ECE (top-left of each panel) is computed on samples pooled across all 5 LVLMs and 5 seeds, matching the reliability curves; these values differ slightly from the per-LVLM-then-averaged ECE in Table~\ref{tab:main_results} because ECE is non-linear in the per-bin frequencies. Bar height is the empirical accuracy for that confidence bin, bar opacity is proportional to the fraction of samples in the bin, and the dashed diagonal represents perfect calibration.}
\vspace{-10pt}
\label{fig:calibration}
\end{figure*}

\noindent\textbf{Ablation study.} Table~\ref{tab:ablation_main} reports the contribution of each loss component to \texttt{BICR}'s training objective. $\mathcal{L}_{\mathrm{rank}}$ is the critical component: removing it degrades AUROC by $3.32$ points and AUCPR by $1.95$ ($p < 0.001$, paired Wilcoxon and cluster-aware bootstrap), confirming that the blank-image contrastive signal is the primary driver of \texttt{BICR}'s discriminative gain. $\mathcal{L}_{\mathrm{brier}}$ provides a smaller but consistent calibration benefit ($\Delta$ECE $= -1.39$, $\Delta$BS $= -0.63$). Removing both auxiliary losses produces the worst configuration on every metric ($p < 0.005$ on all discrimination metrics). Beyond the loss components, three additional design studies are reported in the appendix: a behavioral analysis showing that $\mathcal{L}_{\mathrm{rank}}$ broadens the probe's confidence distribution and tightens calibration in the high-confidence range where overconfidence is most consequential (Appendix~\ref{app:design_behavior}); a comparison of five null-image strategies (black, white, Gaussian noise, blurred original, pixel-shuffled), in which the solid-black null is the strongest training signal on every metric (Appendix~\ref{app:design_blank_color}); and an analysis of the Optuna-selected loss coefficients, which finds the rank weight $\lambda$ consistently positive and non-trivial across LVLMs and seeds (Appendix~\ref{app:design_hparams}). Per-LVLM ablation breakdowns and full significance results are in Appendix~\ref{app:design_loss}.

\noindent\textbf{Calibration analysis.} Figure~\ref{fig:calibration} presents the cross-LVLM reliability diagram for all eight methods. Three patterns are evident. Prompt-based methods exhibit severe miscalibration: P(True) concentrates most of its mass in the extreme bins ($[0, 0.2)$ and $[0.8, 1.0]$) with empirical accuracies far from the predicted values (ECE $= 0.366$); Self-Probing pushes nearly all predictions above $0.8$ regardless of correctness (ECE $= 0.281$). Trained baselines improve substantially, with InternalInspector reaching ECE $= 0.078$, but concentrate predictions in the mid-to-high range, underutilizing the low-confidence region where uncertain predictions should fall. \texttt{BICR} achieves the best calibration (ECE $= 0.056$) with confidence mass distributed across the full range: the $[0.8, 1.0]$ bin achieves $91.3\%$ empirical accuracy, and substantial mass at $[0.2, 0.4)$ reflects $42.1\%$ accuracy, indicating that \texttt{BICR} has learned to express genuine uncertainty when visual evidence is weak. A bin-level analysis showing that $\mathcal{L}_{\mathrm{rank}}$ is responsible for this behavior is provided in Appendix~\ref{app:design_behavior}.

\noindent\textbf{Per-dataset analysis.} \texttt{BICR} achieves the best pooled performance across all datasets combined (Table~\ref{tab:main_results}), and this advantage holds under per-dataset equal-weight aggregation as well: averaging metrics with equal weight per dataset (rather than per sample) widens \texttt{BICR}'s ECE and BS gaps over every trained baseline (Appendix~\ref{app:results_unweighted}), confirming that the headline result is not an artifact of the larger datasets dominating the pooled view. The per-dataset breakdown (Appendix~\ref{app:extended_results}) further shows that \texttt{BICR}'s gains decompose into two regimes. On the larger datasets (GQA, POPE), \texttt{BICR}'s discrimination advantage is consistent across LVLMs, sharpening the correct--incorrect ranking on tasks where representations are already informative. On the harder grounding-bound datasets (GMAI-MMBench, MMMU-Pro, MME-Finance), discrimination is more contested, but \texttt{BICR}'s calibration advantage is largest there: it is the best-calibrated method on four of seven datasets (\S\ref{app:results_calibration_perdataset}), reflecting that the blank-image contrastive signal prevents the probe from collapsing into overconfident outputs in regimes where every method struggles to reason about the image. This split between where discrimination concentrates (large, easier datasets) and where calibration concentrates (harder grounding-bound datasets) is what produces \texttt{BICR}'s joint advantage on both axes when aggregated across the benchmark. A direct mechanism check on the sub-population the LVLM behaviorally treats as image-invariant (Appendix~\ref{app:grounding_detection}) confirms that \texttt{BICR}'s confidence is the most accurate of any method evaluated on samples the LVLM behaviorally treats as image-invariant, with paired-bootstrap BS gains significant on $31$ of $35$ (LVLM, baseline) pairs.

%% file: tables/main_results.tex
\begin{table*}[t]
\centering
\caption{Pooled aggregate performance across all LVLMs. Metrics reported as percentages (\%). Arrows indicate direction of improvement. Best values per LVLM are \textbf{bolded}. Trained methods report mean across 5 seeds (50 Optuna trials each). Bottom-right: cross-VLM average. Extended per-dataset results in Appendix~\ref{app:extended_results}.}
\label{tab:main_results}
\resizebox{\textwidth}{!}{
\begin{tabular}{lcccc|lcccc}
\toprule
\multicolumn{5}{c}{\textit{DeepSeek-VL2}} & \multicolumn{5}{c}{\textit{Qwen3-VL-8B-Instruct}} \\
\textbf{\textit{Method}} & \textbf{\textit{ECE $\downarrow$}} & \textbf{\textit{BS $\downarrow$}} & \textbf{\textit{AUCPR $\uparrow$}} & \textbf{\textit{AUROC $\uparrow$}} & \textbf{\textit{Method}} & \textbf{\textit{ECE $\downarrow$}} & \textbf{\textit{BS $\downarrow$}} & \textbf{\textit{AUCPR $\uparrow$}} & \textbf{\textit{AUROC $\uparrow$}} \\
\midrule
P(True) & 34.07 & 37.43 & 68.09 & 52.66 & P(True) & 43.88 & 44.16 & 76.18 & 54.62 \\
Self-Probing & 35.07 & 37.35 & 74.19 & 62.22 & Self-Probing & 24.39 & 27.47 & 77.00 & 59.49 \\
Prompt Ensembles & 16.30 & 24.72 & 72.49 & 73.46 & Prompt Ensembles & 19.80 & 25.96 & 66.20 & 52.90 \\
SAPLMA & 12.83 & 21.51 & 79.69 & 77.14 & SAPLMA & 10.57 & 19.43 & 86.41 & 74.22 \\
P(I~Know) & 8.50 & 19.26 & 84.58 & 78.49 & P(I~Know) & 7.54 & 17.99 & 88.43 & 77.19 \\
CCPS & 7.70 & 22.14 & 76.97 & 70.94 & CCPS & 28.67 & 45.64 & 66.18 & 44.93 \\
InternalInspector & 7.39 & 18.95 & 84.54 & 79.31 & InternalInspector & \textbf{5.43} & \textbf{16.95} & 89.75 & 79.60 \\
\rowcolor{red}
\textbf{\texttt{BICR (Ours)}} & \textbf{6.02} & \textbf{17.90} & \textbf{86.19} & \textbf{81.11} & \textbf{\texttt{BICR (Ours)}} & 8.91 & 17.40 & \textbf{90.26} & \textbf{80.14} \\
\midrule
\midrule
\multicolumn{5}{c}{\textit{LLaVA-NeXT-Vicuna-13B}} & \multicolumn{5}{c}{\textit{InternVL3.5-14B}} \\
\textbf{\textit{Method}} & \textbf{\textit{ECE $\downarrow$}} & \textbf{\textit{BS $\downarrow$}} & \textbf{\textit{AUCPR $\uparrow$}} & \textbf{\textit{AUROC $\uparrow$}} & \textbf{\textit{Method}} & \textbf{\textit{ECE $\downarrow$}} & \textbf{\textit{BS $\downarrow$}} & \textbf{\textit{AUCPR $\uparrow$}} & \textbf{\textit{AUROC $\uparrow$}} \\
\midrule
P(True) & 26.23 & 30.93 & 67.91 & 54.73 & P(True) & 41.19 & 41.45 & 78.02 & 59.49 \\
Self-Probing & 28.92 & 29.97 & 81.87 & 67.30 & Self-Probing & 21.51 & 24.71 & 76.77 & 70.75 \\
Prompt Ensembles & 12.77 & 23.49 & 77.31 & 68.47 & Prompt Ensembles & 16.70 & 25.90 & 60.26 & 43.37 \\
SAPLMA & 16.52 & 22.99 & 82.01 & 72.47 & SAPLMA & 16.58 & 23.27 & 76.21 & 65.42 \\
P(I~Know) & 10.81 & 19.87 & 87.06 & 77.32 & P(I~Know) & 10.83 & 20.16 & 86.35 & 73.78 \\
CCPS & 16.27 & 21.98 & 76.00 & 72.89 & CCPS & 14.72 & 23.83 & 71.64 & 58.22 \\
InternalInspector & 13.79 & 22.31 & 81.42 & 71.87 & InternalInspector & 10.59 & 21.48 & 82.64 & 69.31 \\
\rowcolor{red}
\textbf{\texttt{BICR (Ours)}} & \textbf{5.65} & \textbf{18.16} & \textbf{87.74} & \textbf{78.94} & \textbf{\texttt{BICR (Ours)}} & \textbf{7.90} & \textbf{19.04} & \textbf{88.04} & \textbf{76.39} \\
\midrule
\midrule
\multicolumn{5}{c}{\textit{Gemma-3-27B-IT}} & \multicolumn{5}{c}{\textit{\textbf{\textit{Cross-LVLM Average}}}} \\
\textbf{\textit{Method}} & \textbf{\textit{ECE $\downarrow$}} & \textbf{\textit{BS $\downarrow$}} & \textbf{\textit{AUCPR $\uparrow$}} & \textbf{\textit{AUROC $\uparrow$}} & \textbf{\textit{Method}} & \textbf{\textit{ECE $\downarrow$}} & \textbf{\textit{BS $\downarrow$}} & \textbf{\textit{AUCPR $\uparrow$}} & \textbf{\textit{AUROC $\uparrow$}} \\
\midrule
P(True) & 44.80 & 45.05 & 74.01 & 56.81 & P(True) & 38.04 & 39.80 & 72.84 & 55.66 \\
Self-Probing & 27.67 & 29.46 & 79.20 & 68.39 & Self-Probing & 27.51 & 29.79 & 77.81 & 65.63 \\
Prompt Ensembles & 26.72 & 30.29 & 70.13 & 61.82 & Prompt Ensembles & 18.46 & 26.07 & 69.28 & 60.00 \\
SAPLMA & 4.61 & \textbf{19.42} & 83.83 & 75.45 & SAPLMA & 12.22 & 21.32 & 81.63 & 72.94 \\
P(I~Know) & 8.88 & 19.86 & 85.07 & 76.08 & P(I~Know) & 9.31 & 19.43 & 86.30 & 76.57 \\
CCPS & 8.98 & 22.14 & 73.37 & 68.49 & CCPS & 15.27 & 27.15 & 72.83 & 63.10 \\
InternalInspector & \textbf{4.29} & 19.68 & 83.17 & 74.15 & InternalInspector & 8.30 & 19.88 & 84.31 & 74.85 \\
\rowcolor{red}
\textbf{\texttt{BICR (Ours)}} & 6.98 & 19.56 & \textbf{85.10} & \textbf{76.56} & \textbf{\texttt{BICR (Ours)}} & \textbf{7.09} & \textbf{18.41} & \textbf{87.47} & \textbf{78.63} \\
\bottomrule
\end{tabular}
}
\end{table*}

%% file: sections/discussion.tex
\section{Discussion}
\label{sec:discussion}

\noindent\textbf{Our operational signal of visual grounding is readable from hidden states, and image-invariant behavior is detectable.} The gap between hidden states under a real versus blank image is a reliable signal of correctness, even though a probe trained only on real-image states has no basis to distinguish a grounded prediction from one shaped entirely by language priors. The $3.32$ AUROC drop when $\mathcal{L}_{\mathrm{rank}}$ is removed ($p < 0.001$ across 25 runs and all five LVLMs) confirms the signal is not incidental, and a behavioral test on the sub-population the LVLM treats as image-invariant (Appendix~\ref{app:grounding_detection}) shows \texttt{BICR}'s calibration advantage concentrates on exactly the population the rank loss targets.

\noindent\textbf{Confidence estimation can precede response generation entirely.} \texttt{BICR} operates at the last prompt token, after the model has formed its full internal representation of the task but before generation begins. Confidence is therefore estimated from how the model represented the input, not from what it said. In deployment, a pipeline can flag low-confidence inputs before paying generation cost, enabling triage, escalation, or human review without waiting for a response. P(True) and Self-Probing require the model to generate an answer and then a second response expressing certainty; SAPLMA and CCPS are bound to a completed generation pass. \texttt{BICR} has no such dependency.

\noindent\textbf{Training-time contrast, no inference-time cost.} The contrastive signal that drives \texttt{BICR}'s gains could in principle be read at inference by running both passes per query. \texttt{BICR} instead absorbs the contrast into the probe's weights during training: the blank-image pass is preprocessing, and inference is a single forward pass. Methods like Prompt Ensembles, Self-Probing, and CCPS pay their cost at every query; \texttt{BICR} pays it once.

\noindent\textbf{The calibration-discrimination trade-off yields to the right training signal.} Prior work identifies a tension where optimizing one metric degrades the other~\citep{rmcb}. \texttt{BICR} is the only method in our benchmark with the best cross-LVLM average on all four metrics simultaneously, and the ablation reveals why: $\mathcal{L}_{\mathrm{rank}}$ improves discrimination by suppressing overconfident scores on ungrounded predictions, while $\mathcal{L}_{\mathrm{brier}}$ penalizes the gap between predicted scores and true correctness frequencies. The two losses are complementary because they target distinct failure modes: ranking correct above incorrect, versus producing scores reliable as probability estimates.

\noindent\textbf{\texttt{BICR}'s advantage concentrates where it matters most.} Gains are largest on GMAI-MMBench and MMMU-Pro, the two hardest and most visually demanding benchmarks in \texttt{VLCB}, and smallest on POPE, where binary hallucination detection is learnable largely from language priors. This validates the design: the contrastive signal is most informative where the image is genuinely necessary. An ungrounded correct guess on a yes/no probe is an edge case; on a medical diagnosis or financial chart, the same failure carries real cost.

%% file: sections/conclusion.tex
\section{Conclusion}
\label{sec:conclusion}
We introduced \texttt{BICR}, an LVLM confidence estimation framework built on a targeted training-time intervention: replace the image with black, extract the hidden state at the same prompt position, and regularize a lightweight probe to assign lower confidence when the visual input disappears. This single contrastive signal, applied before generation begins at zero additional inference overhead relative to a single-view probe, is sufficient to achieve the best performance in our benchmark on calibration and discrimination simultaneously across five LVLMs and seven baselines, with statistically significant gains at 4--18$\times$ fewer parameters than the strongest probing baseline. The findings suggest that a useful proxy for visual grounding is legible in hidden states, that the calibration-discrimination trade-off yields to the right training signal, and that higher confidence quality does not require more inference compute. We release \texttt{VLCB}, our LVLM confidence estimation benchmark spanning general, medical, and financial visual reasoning, together with all evaluation code to support future research on trustworthy LVLM deployment.

%% file: sections/limitations.tex
\section*{Limitations}
\label{sec:limitations}
Despite \texttt{BICR}'s strong empirical performance across five LVLMs and seven datasets, several limitations bound the scope of our findings. First, \texttt{BICR} requires access to the LVLM's internal hidden states, so it applies to open-weight models but not to closed-weight LVLMs (e.g., GPT-4V, Claude, Gemini) served only through APIs; extending the blank-image contrastive principle to API-only settings is an open problem. Second, our use of ``visual grounding'' and ``visual ungroundedness'' refers to an operational proxy rather than a direct measurement: we treat a prediction as grounded to the extent that the LVLM's hidden state differs between the real-image and blank-image views, a necessary but not sufficient condition for grounding in the stronger semantic sense. This contrastive-proxy framing is standard in the LVLM grounding literature, including VCD~\citep{10657718}, VL-Uncertainty~\citep{VL-Uncertainty}, and SVAR~\citep{Devils}, none of which measure grounding directly either; more direct measurement (e.g., causal interventions on individual visual tokens) is an open problem. Third, our evaluation is VQA-centric: although \texttt{VLCB} spans diverse domains (general, medical, financial, and college-level reasoning), the tasks are predominantly question-answering in structure, and open-ended items are a small fraction of the benchmark (only the 60 LLaVA-in-the-Wild samples are open-ended). This protocol may not fully reflect \texttt{BICR}'s effectiveness on other critical LVLM tasks, such as open-ended generation, long-form image captioning, or complex multi-turn reasoning, where the dynamics of visual ungroundedness may manifest differently, and it does not cover video, 3D, or embodied settings; whether the real-versus-blank contrast transfers to these regimes is an open question we leave to future work. Fourth, the rank loss $\mathcal{L}_{\mathrm{rank}}$ constrains only correctly-answered samples ($y=1$), since the contrastive direction encodes \emph{grounded correctness}, which has no analogue for incorrect responses where the answer is wrong regardless of grounding; a symmetric or class-conditional formulation that also constrains the $y=0$ direction is a natural extension we did not investigate. We believe these limitations do not detract from our core findings but instead point to natural directions for future work on the reliability of LVLMs.

%% file: sections/ethical_considerations.tex
\section*{Ethical Considerations}
\label{sec:ethics}

While \texttt{BICR} is developed with the goal of improving the reliability of large vision-language models, several ethical considerations are relevant. The primary concern is over-reliance on automated confidence scores. Our results show that even the best methods carry trade-offs and no method is perfectly calibrated across every dataset and LVLM combination. In high-stakes domains such as medicine, finance, or law, accepting an LVLM's output simply because its associated confidence score is high, without independent human judgment and oversight, could lead to adverse outcomes when an LVLM error is not flagged by the confidence estimator. This concern is especially acute in the medical imaging and document understanding settings represented in our benchmark, where confidence scores might inform downstream decisions with material consequences. A second concern is fairness across diverse populations and data distributions. The LVLMs we evaluate carry whatever biases were present in their training data, and any confidence estimator built on top of those LVLMs, including \texttt{BICR}, may inherit or even amplify those biases. As a result, confidence scores could be systematically less reliable for certain demographic groups, image styles, or question types, potentially leading to inequitable downstream outcomes. Therefore, any deployment of \texttt{BICR} or related methods, particularly in sensitive applications, should be preceded by thorough fairness testing across relevant subgroups, accompanied by ongoing monitoring, and framed explicitly as a tool that assists human experts rather than replacing their critical judgment.

%% file: sections/acknowledgments.tex
\section*{Acknowledgments}

This work was supported by the JPMorgan Chase AI Research Faculty Research Award. The authors are solely responsible for the contents of this paper; the opinions expressed do not necessarily reflect those of the funding organizations. The authors also acknowledge the use of Large Language Models to assist in polishing the language and grammar of this manuscript.

%% file: sections/disclaimer.tex
\section*{Disclaimer}

This paper was prepared for informational purposes by the Artificial Intelligence Research group of JPMorgan Chase \& Co and its affiliates (``JP Morgan''), and is not a product of the Research Department of JP Morgan. JP Morgan makes no representation and warranty whatsoever and disclaims all liability, for the completeness, accuracy or reliability of the information contained herein. This document is not intended as investment research or investment advice, or a recommendation, offer or solicitation for the purchase or sale of any security, financial instrument, financial product or service, or to be used in any way for evaluating the merits of participating in any transaction, and shall not constitute a solicitation under any jurisdiction or to any person, if such solicitation under such jurisdiction or to such person would be unlawful.

%% file: sections/appendix_dataset.tex
\section{Large Vision-Language Model Backbones}
\label{app:lvlms}

{\looseness=-1
The five LVLMs benchmarked in this work were selected to make our cross-model claims meaningful rather than to maximize coverage of any single architectural axis. Together they span four open-weight model families with dense language backbones (Qwen3-VL, LLaVA-NeXT, InternVL3.5, Gemma-3) and one mixture-of-experts design (DeepSeek-VL2); a parameter range from 4.5B activated to 27B; three distinct vision-encoder lineages (CLIP-derived ViT-L/14, the SigLIP family, and the Qwen3-VL ViT with DeepStack); and language context lengths from 4K to 256K tokens. The intent of this selection is that any confidence estimation method evaluated across all five must demonstrate it generalizes across visual encoders, language backbones, and parameter regimes rather than exploiting properties of a particular architecture.
\par}

Tables~\ref{tab:vlcb_llm} and~\ref{tab:vlcb_vit} report the LLM-side and vision-encoder configurations respectively, taken directly from the official model configs on the Hugging Face Hub at the time of writing. Two structural points in these tables are worth noting because they shape how \texttt{BICR}'s hidden-state extraction (\S\ref{sec:method}) interacts with each model. First, hidden size $H$ varies from 2{,}560 (DeepSeek-VL2) to 5{,}376 (Gemma-3), which directly drives the trainable parameter count of every probe-based confidence estimator we evaluate; the exact per-model parameter counts are reported in Appendix~\ref{app:params}. Second, the vision encoders differ not only in lineage but in input resolution policy: Qwen3-VL accepts dynamic resolutions, SigLIP variants are fixed at 384 or 896 pixels, and CLIP ViT-L/14 is fixed at 336 pixels, which is what motivates the uniform 2{,}048-pixel cap applied at the input side of our generation pipeline (\S\ref{app:gen_extract}) before each model's own preprocessor takes over.

\begin{table*}[ht]
\centering
\footnotesize
\setlength{\tabcolsep}{4pt}
\caption{
LLM-side architecture of the five LVLMs benchmarked in \texttt{VLCB}. Columns report the Hugging Face model identifier (Hub ID), language backbone (Backbone), parameter count in billions ($P$, with activated parameters reported for DeepSeek-VL2), hidden size ($H$), number of transformer layers ($L$), number of attention heads as query/key-value heads ($H\,(Q/KV)$), and maximum context length ($C$). Numbers are taken from official model configs on the Hugging Face Hub at the time of writing. $^\dagger$DeepSeek-VL2 uses an MoE LM; activated parameters are reported.
}
\label{tab:vlcb_llm}

\begin{tabular}{
>{\raggedright\arraybackslash}p{0.38\textwidth}
>{\raggedright\arraybackslash}p{0.2\textwidth}
>{\raggedleft\arraybackslash}p{0.04\textwidth}
>{\raggedleft\arraybackslash}p{0.04\textwidth}
>{\raggedleft\arraybackslash}p{0.04\textwidth}
>{\raggedleft\arraybackslash}p{0.04\textwidth}
>{\raggedleft\arraybackslash}p{0.1\textwidth}
}
\toprule
\textbf{Hub ID} & \textbf{Backbone} & \textbf{$P$} & \textbf{$H$} & \textbf{$L$} & \textbf{$H\,(Q/KV)$} & \textbf{$C$} \\
\midrule
\href{https://huggingface.co/Qwen/Qwen3-VL-8B-Instruct}{Qwen/Qwen3-VL-8B-Instruct} \cite{qwen3}
& Qwen3-VL-Text
& 8.0
& 4096
& 36
& 32\,/\,8
& 256K \\

\href{https://huggingface.co/llava-hf/llava-v1.6-vicuna-13b-hf}{llava-hf/llava-v1.6-vicuna-13b-hf} \cite{llava}
& LLaMA (Vicuna-13B)
& 13.0
& 5120
& 40
& 40\,/\,40
& 4K \\

\href{https://huggingface.co/OpenGVLab/InternVL3_5-14B-HF}{OpenGVLab/InternVL3\_5-14B-HF} \cite{internvl3_5}
& Qwen3-14B
& 15.1
& 5120
& 40
& 40\,/\,8
& 40K \\

\href{https://huggingface.co/google/gemma-3-27b-it}{google/gemma-3-27b-it} \cite{gemma3}
& Gemma-3
& 27.0
& 5376
& 62
& 32\,/\,16
& 128K \\

\href{https://huggingface.co/deepseek-ai/deepseek-vl2}{deepseek-ai/deepseek-vl2} \cite{deepseek-vl2}
& DeepSeek-VL2 (MoE)$^\dagger$
& 4.5
& 2560
& 30
& 32\,/\,32
& 4K \\
\bottomrule
\end{tabular}
\end{table*}

\begin{table*}[ht]
\centering
\footnotesize
\caption{Vision-encoder architecture of the five LVLMs benchmarked in \texttt{VLCB}. Numbers are taken from official model configs on the Hugging Face Hub at the time of writing.}
\label{tab:vlcb_vit}
\begin{tabular}{l l r r r r r}
\toprule
\textbf{Hub ID} & \textbf{Vision Encoder} &
\makecell{\textbf{ViT}\\\textbf{Hidden}} &
\makecell{\textbf{ViT}\\\textbf{Layers}} &
\makecell{\textbf{ViT}\\\textbf{Heads}} &
\makecell{\textbf{Patch}} &
\makecell{\textbf{Input}\\\textbf{Res.}} \\
\midrule
\href{https://huggingface.co/Qwen/Qwen3-VL-8B-Instruct}{Qwen/Qwen3-VL-8B-Instruct}
  & Qwen3-VL ViT & 1152 & 27 & 16 & 16 & dynamic \\
\href{https://huggingface.co/llava-hf/llava-v1.6-vicuna-13b-hf}{llava-hf/llava-v1.6-vicuna-13b-hf}
  & CLIP ViT-L/14            & 1024 & 24 & 16 & 14 & 336 \\
\href{https://huggingface.co/OpenGVLab/InternVL3_5-14B-HF}{OpenGVLab/InternVL3\_5-14B-HF}
  & InternViT-300M           & 1024 & 24 & 16 & 14 & 448 \\
\href{https://huggingface.co/google/gemma-3-27b-it}{google/gemma-3-27b-it}
  & SigLIP                   & 1152 & 27 & 16 & 14 & 896 \\
\href{https://huggingface.co/deepseek-ai/deepseek-vl2}{deepseek-ai/deepseek-vl2}
  & SigLIP-SO400M            & 1152 & 27 & 16 & 14 & 384 \\
\bottomrule
\end{tabular}
\end{table*}

\paragraph{Compute environment.} The pipeline uses a two-tier setup that separates expensive LVLM inference from lightweight confidence-estimator training. LVLM inference, including hidden-state extraction, response generation for \texttt{VLCB} construction, and the inference-only baselines (Self-Probing, Prompt Ensembles), runs on NVIDIA H200 GPU, with response generation served through vLLM~\citep{vllm} for high throughput. Four of the five LVLMs (Qwen3-VL-8B, LLaVA-NeXT-13B, InternVL3.5-14B, Gemma-3-27B) run in full precision; DeepSeek-VL2 runs in half precision with reduced batch size due to known numerical instabilities (precision settings are detailed in Appendix~\ref{app:appendix_dataset}). Confidence-estimator training and evaluation (\texttt{BICR}, P(I~Know), SAPLMA, InternalInspector, CCPS) operate on cached hidden states rather than the live LVLM and run on a cluster of 8$\times$NVIDIA A100 40GB GPUs. Each (method, LVLM, seed) configuration fits on a single A100, so we shard the (LVLM, seed) grid across the 8 GPUs; the 50 Optuna trials per (LVLM, seed) tuple for \texttt{BICR} and P(I~Know) are the dominant training cost.

\section{VLCB Benchmark Construction}
\label{app:appendix_dataset}

The main body (\S\ref{sec:benchmark}) frames \texttt{VLCB} by its design principle: training and validation are drawn from a single source on purpose, and evaluation spans heterogeneous out-of-domain task formats on purpose, so that performance numbers reflect generalization rather than within-distribution fitting. This appendix documents the engineering that operationalizes that principle. We describe the seven public source datasets aggregated into \texttt{VLCB}, the unified record schema that absorbs their heterogeneity into a single training and evaluation interface, the train, validation, and test assembly together with the quality-control checks that guarantee its splits are disjoint and reproducible, the response generation pipeline used to elicit answers from each evaluated LVLM under semantically uniform conditions, and the LLM-judge protocol used to label every generated response as correct or incorrect. All datasets are in English. For comprehensive information regarding the original construction and domain coverage of each source benchmark, we refer the reader to their respective publications.

\subsection{Data Curation and Standard Schema}

The seven source datasets aggregated into \texttt{VLCB} span grounded visual reasoning, hallucination probing, multimodal multiple-choice exams, medical VQA, financial chart understanding, and open-ended instruction following. Each was originally released with its own record format, image storage convention, and label schema, none of which agree across sources. To support a unified training and evaluation harness across this heterogeneity, every raw sample is processed into the following standardized HuggingFace Dataset record:

\begin{itemize}
    \item \textbf{\texttt{hash\_id}} (\texttt{str}): a deterministic, dataset-specific MD5 hash over
    \path{{dataset}[SEP]{category}[SEP]{question}[SEP]{answer}[SEP]{image_key}},
    where \texttt{image\_key} is whichever per-source identifier (image filename, base64 payload, image-slot list) uniquely disambiguates samples that would otherwise share question and answer text.
    \item \textbf{\texttt{image}} (\texttt{PIL.Image.Image}, RGB): the visual input. Variable resolution is preserved at curation time; resizing is applied only at inference (\S\ref{app:gen_extract}).
    \item \textbf{\texttt{question}} (\texttt{str}): the input question presented to the LVLM, with task-specific multiple-choice options and any required context already inlined.
    \item \textbf{\texttt{answer}} (\texttt{str}): the ground-truth answer in the canonical form expected for that source dataset (e.g., a single option letter for multiple-choice tasks; a short answer span for GQA; the textual answer for MME-Finance and LLaVA-in-the-Wild).
    \item \textbf{\texttt{category}} (\texttt{str}): a dataset-specific sub-category label (e.g., GQA's \texttt{detailed} question type, POPE's negative-sampling regime, GMAI-MMBench's clinical VQA task, MMMU-Pro's subject and topic difficulty). Defaults to \texttt{"N/A"} when no taxonomy is available.
    \item \textbf{\texttt{dataset}} (\texttt{str}): the source dataset identifier.
\end{itemize}

\begin{figure*}[t]
\centering
\footnotesize

\tcbset{
  boxsep=2pt,
  left=3pt,
  right=3pt,
  top=2pt,
  bottom=2pt,
  arc=1mm
}

\begin{tcolorbox}[
title=GQA --- Compositional Visual Reasoning,
fonttitle=\bfseries\small,
fontupper=\footnotesize
]
\begin{minipage}[c]{0.20\linewidth}
  \includegraphics[width=0.92\linewidth]{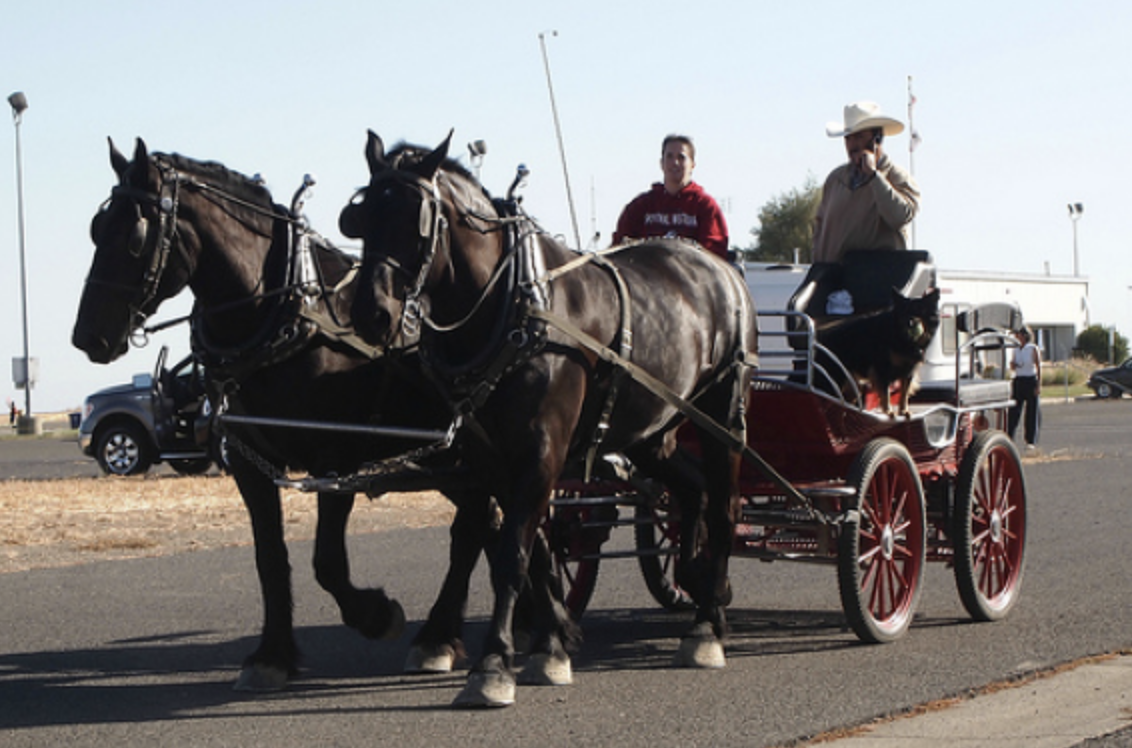}
\end{minipage}%
\hspace{4pt}%
\begin{minipage}[c]{0.75\linewidth}
  \setlength{\parskip}{2pt}
  \textbf{Category:} relChooser\\[1pt]
  \textbf{Question:} Is the brown horse to the right or to the left of the person that is standing on the road?\\[1pt]
  \textbf{Answer:} left
\end{minipage}
\end{tcolorbox}

\vspace{-4pt}

\begin{tcolorbox}[
title=POPE --- Object Hallucination Probing,
fonttitle=\bfseries\small,
fontupper=\footnotesize
]
\begin{minipage}[c]{0.20\linewidth}
  \includegraphics[width=0.92\linewidth]{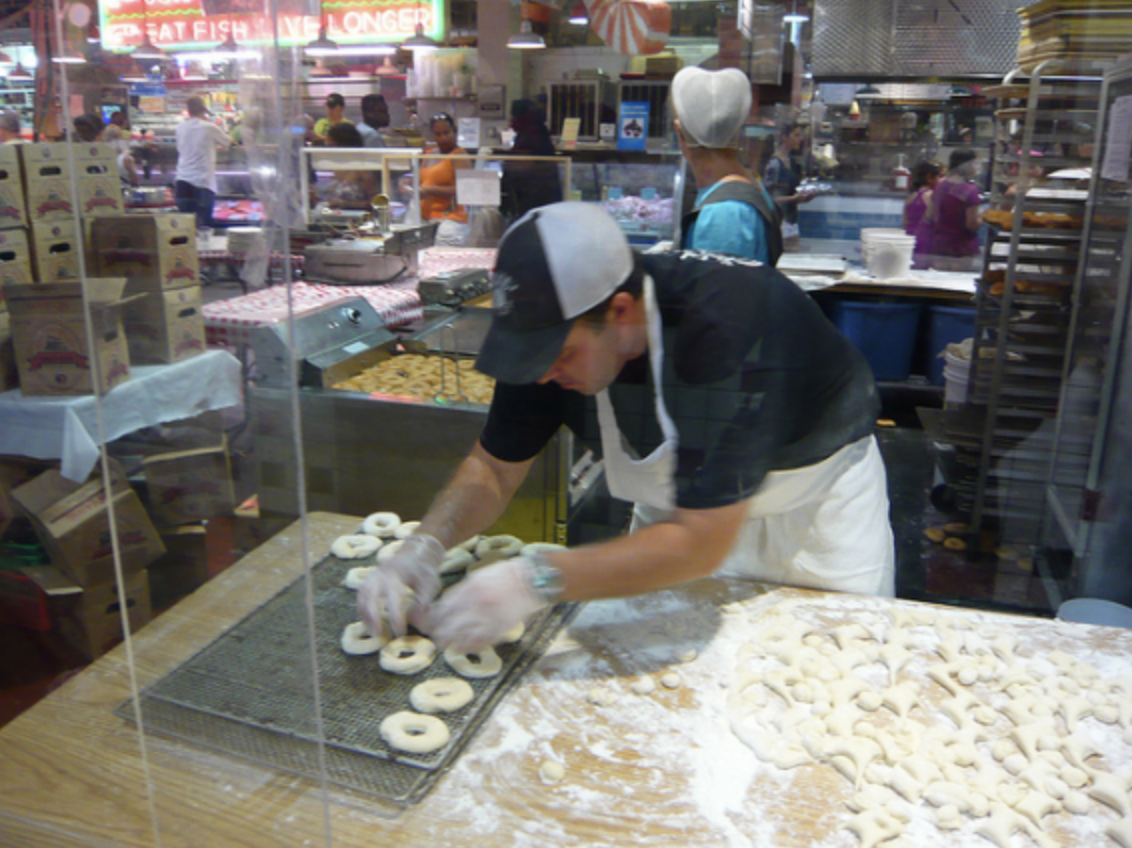}
\end{minipage}%
\hspace{4pt}%
\begin{minipage}[c]{0.75\linewidth}
  \setlength{\parskip}{2pt}
  \textbf{Category:} adversarial\\[1pt]
  \textbf{Question:} Is there a spoon in the image?\\[1pt]
  \textbf{Answer:} no
\end{minipage}
\end{tcolorbox}

\vspace{-4pt}

\begin{tcolorbox}[
title=GMAI-MMBench --- Medical Multimodal Reasoning,
fonttitle=\bfseries\small,
fontupper=\footnotesize
]
\begin{minipage}[c]{0.20\linewidth}
  \includegraphics[width=0.92\linewidth]{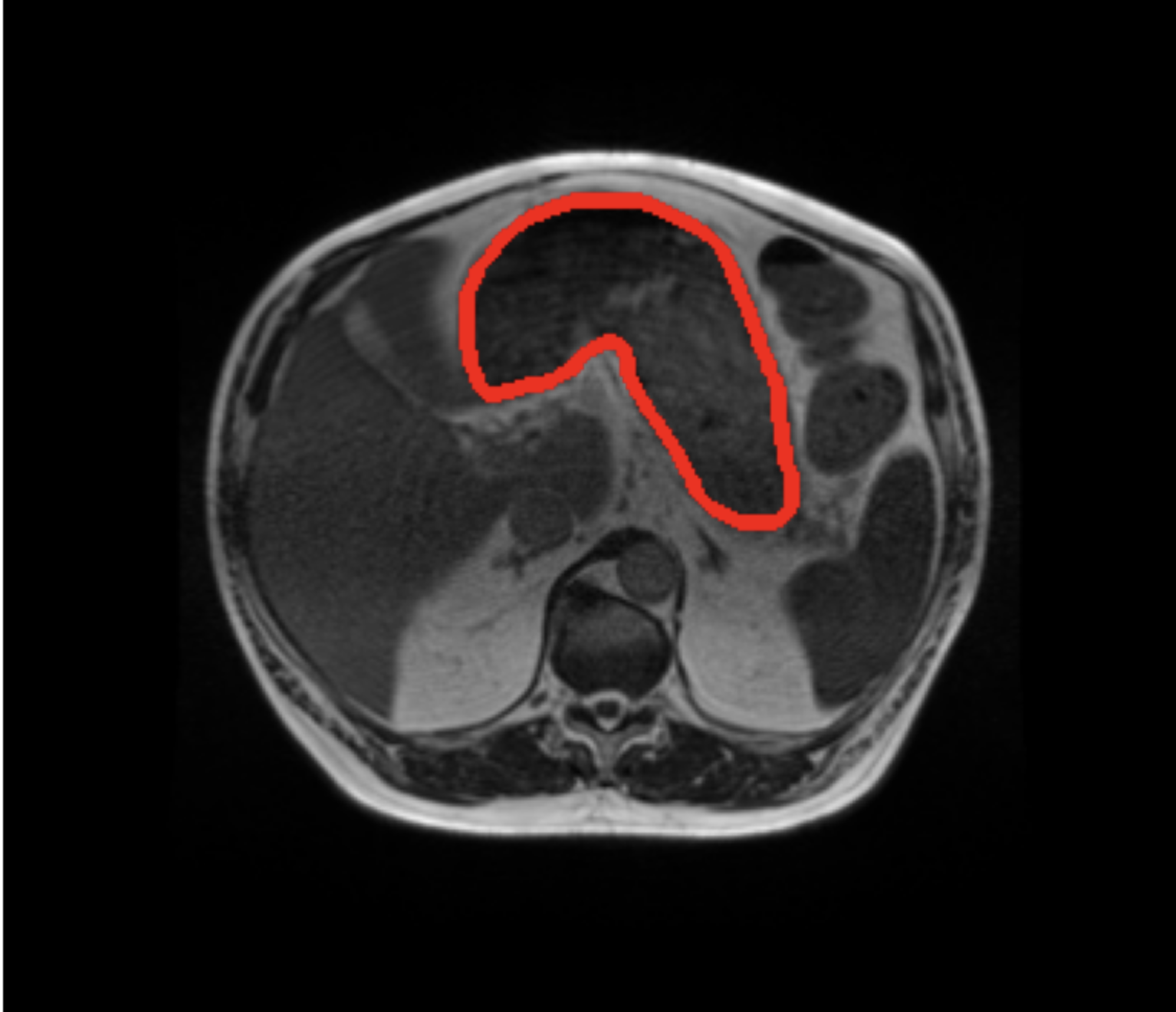}
\end{minipage}%
\hspace{4pt}%
\begin{minipage}[c]{0.75\linewidth}
  \setlength{\parskip}{2pt}
  \textbf{Category:} Organ Recognition -- Abdomen\\[1pt]
  \textbf{Question:} This is an MRI image. Which option best describes the marked area?\\[1pt]
  A) heart, B) gallbladder, C) stomach, D) liver, E) necrotic tissue\\[1pt]
  \textbf{Answer:} C
\end{minipage}
\end{tcolorbox}

\vspace{-4pt}

\begin{tcolorbox}[
title=MME-Finance --- Financial Chart VQA,
fonttitle=\bfseries\small,
fontupper=\footnotesize
]
\begin{minipage}[c]{0.20\linewidth}
  \includegraphics[width=0.92\linewidth]{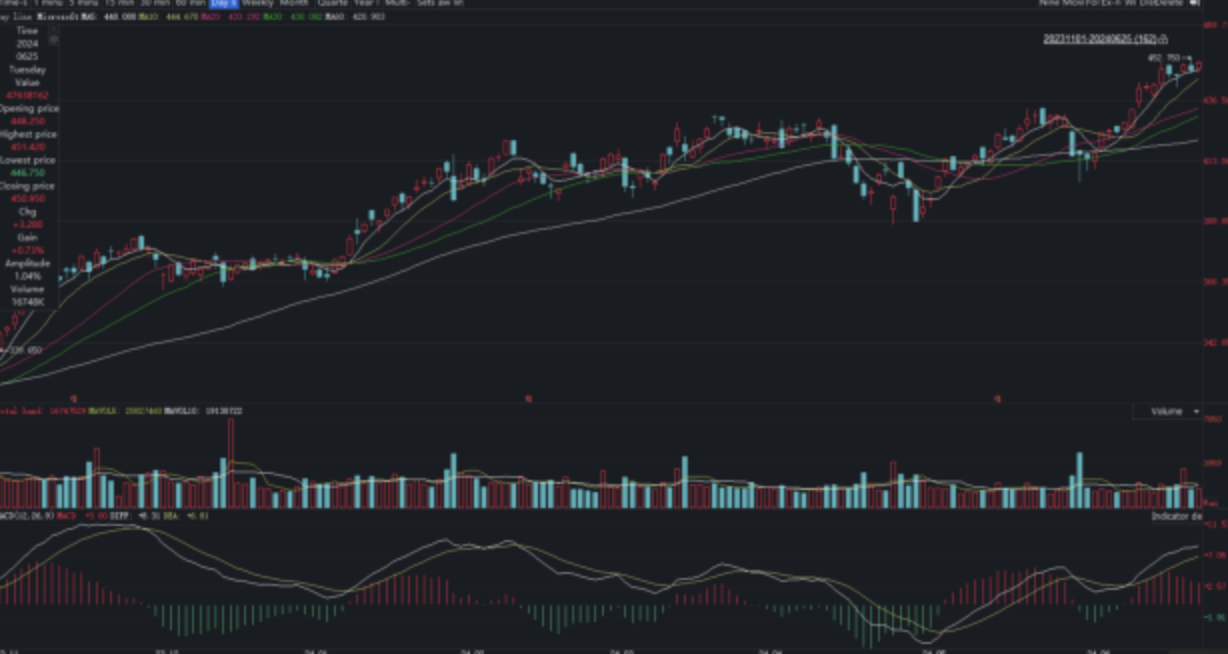}
\end{minipage}%
\hspace{4pt}%
\begin{minipage}[c]{0.75\linewidth}
  \setlength{\parskip}{2pt}
  \textbf{Category:} Financial Knowledge\\[1pt]
  \textbf{Question:} What company is represented in the chart and what market is it listed on?\\[1pt]
  \textbf{Answer:} Microsoft Corporation, NASDAQ.
\end{minipage}
\end{tcolorbox}

\vspace{-4pt}

\begin{tcolorbox}[
title=MMMU-Pro --- Multidisciplinary College-Level VQA,
fonttitle=\bfseries\small,
fontupper=\footnotesize
]
\begin{minipage}[c]{0.20\linewidth}
  \includegraphics[width=0.92\linewidth]{figures/MMMU-Pro-example.png}
\end{minipage}%
\hspace{4pt}%
\begin{minipage}[c]{0.75\linewidth}
  \setlength{\parskip}{2pt}
  \textbf{Category:} Agriculture[SEP]Hard[SEP]Photographs\\[1pt]
  \textbf{Question:} What could be the reason behind the browning on this potato leaf?\\[1pt]
  A) Overwatering, B) Mite feeding, C) Bacterial infection, D) Fungal infection,\\
  E) Lack of sunlight, F) Don't know, G) Sunburn, H) Nutrient deficiency,\\
  I) Ozone damage, J) Viral infection\\[1pt]
  \textbf{Answer:} B
\end{minipage}
\end{tcolorbox}

\vspace{-4pt}

\begin{tcolorbox}[
title=LLaVA-in-the-Wild --- Open-Ended Instruction Following,
fonttitle=\bfseries\small,
fontupper=\footnotesize
]
\begin{minipage}[c]{0.20\linewidth}
  \includegraphics[width=0.92\linewidth]{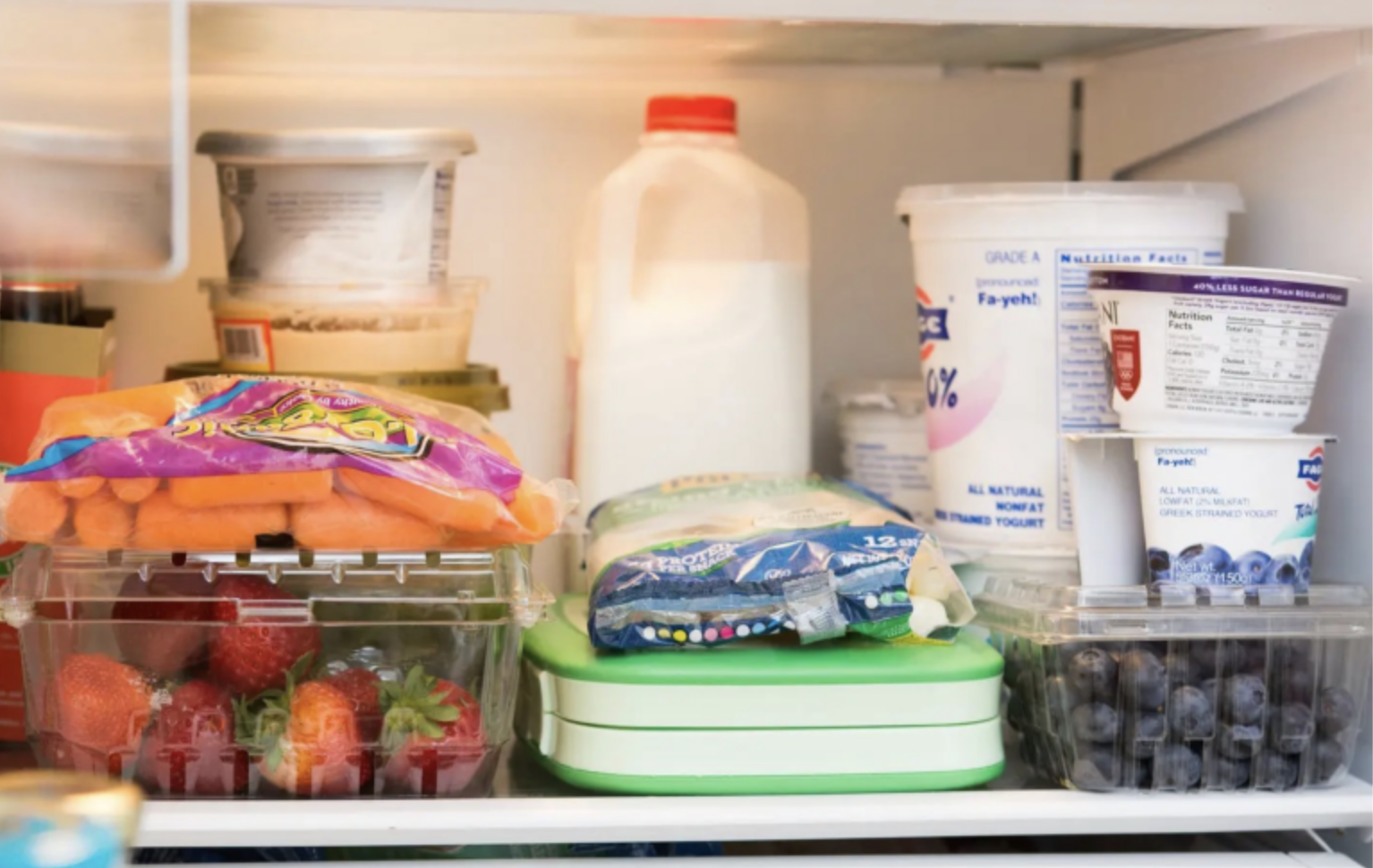}
\end{minipage}%
\hspace{4pt}%
\begin{minipage}[c]{0.75\linewidth}
  \setlength{\parskip}{2pt}
  \textbf{Category:} conv\\[1pt]
  \textbf{Question:} Is there any strawberry-flavored yogurt in the fridge?\\[1pt]
  \textbf{Answer:} No. The fridge contains Fage non-fat yogurt, blueberry yogurt, and another yogurt cup with an unknown flavor.
\end{minipage}
\end{tcolorbox}

\vspace{-2pt}

\caption{One representative sample from each source dataset in \texttt{VLCB}. Each entry shows the image, task category, and question--answer pair.}
\label{fig:vlcb_examples}

\end{figure*}

The \texttt{hash\_id} field is the mechanism that makes \texttt{VLCB} reproducible by independent users. Because the licensing terms of several source datasets prevent us from redistributing the assembled benchmark as a single archive (\S\ref{app:availability}), reproducibility relies on the property that any user who has independently obtained the source datasets can recompute the same hashes and recover the same splits. To support this, all curation procedures fix the random seed at \texttt{SEED=23} so that subset selection, stratified splitting, and ordering are fully deterministic. Figure~\ref{fig:vlcb_examples} shows one representative sample from each source dataset to illustrate the visual and textual variety the schema absorbs.

\subsection{Source Datasets and Versions}

Each source dataset is loaded from its official distribution and processed independently into a per-source artifact before any cross-source aggregation. Table~\ref{tab:vlcb_dataset_sources} lists every source, its access path, the split used, and its original license. 

\begin{table*}[ht]
\centering
\small
\setlength{\tabcolsep}{4pt}
\caption{Source datasets used to construct \texttt{VLCB}, with split selection and license.}
\label{tab:vlcb_dataset_sources}

\begin{tabular}{
>{\raggedright\arraybackslash}p{0.16\textwidth}
>{\raggedright\arraybackslash}p{0.30\textwidth}
>{\raggedright\arraybackslash}p{0.46\textwidth}
}
\toprule
\textbf{Dataset} & \textbf{Source} & \textbf{Split / License} \\
\midrule
GQA \cite{GQA} 
& HuggingFace \texttt{lmms-lab/GQA} 
& balanced train, val, testdev / MIT \\

POPE \cite{POPE} 
& HuggingFace \texttt{lmms-lab/POPE} 
& test (adversarial / popular / random) / CC BY 4.0 \\

GMAI-MMBench \cite{GMAI-MMBench} 
& HuggingFace \texttt{OpenGVLab/GMAI-MMBench} 
& VAL TSV (only split with answers) / Apache 2.0 \\

MME-Finance \cite{MME-Finance} 
& Official MME-Finance release (TSV + image archive) 
& CC BY-NC-SA 4.0 \\

MMMU-Pro \cite{MMMU} 
& HuggingFace \texttt{MMMU/MMMU\_Pro} 
& \texttt{standard (4 options)}, \texttt{standard (10 options)} / Apache 2.0 \\

LLaVA-in-the-Wild \cite{LlavaWild} 
& Official LLaVA-Bench-in-the-Wild release 
& test / - \\
\bottomrule
\end{tabular}
\end{table*}

\subsubsection{GQA (Compositional Visual Reasoning)}
GQA \cite{GQA} is a large-scale benchmark for compositional visual question answering over real-world scene graphs, and is the source of every training and validation sample in \texttt{VLCB}; the held-out \texttt{testdev} split appears in the test set as the in-distribution reference point. We use the \emph{balanced} \texttt{train}, \texttt{val}, and \texttt{testdev} splits released through \texttt{lmms-lab/GQA}, which contain 943{,}000 / 132{,}062 / 12{,}578 instructions over 72{,}140 / 10{,}234 / 398 images respectively. From each split we discard instructions whose \texttt{detailed} question type is missing or appears in fewer than two samples (a stratification requirement), then draw a stratified subsample on \texttt{detailed}. The starting subsample sizes are 20{,}000 (train), 5{,}001 (val), and \texttt{min(20000, available)} = 12{,}574 (test); these targets were chosen to give the confidence probe a substantial training pool (20K samples is large enough for stable Optuna search across our five seeds while remaining tractable to extract hidden states for) and a validation pool roughly one quarter that size, while the test target is bounded by what GQA testdev makes available. After hash-based deduplication this yields \textbf{20{,}000 / 5{,}000 / 12{,}568} clean records spanning 104 / 101 / 91 \texttt{detailed} categories.

The standardized record stores the original GQA question verbatim, the short ground-truth answer, the scene image attached by image\_id, and the \texttt{detailed} field as \texttt{category}. The five most frequent training categories are \texttt{relS} (10.1\%), \texttt{categoryRelS} (9.3\%), \texttt{relO} (8.9\%), \texttt{positionQuery} (4.9\%), and \texttt{relChooser} (4.0\%). Question length is moderate (mean $8.8$, median $8.0$ tokens) and answers are essentially single-token (mean and median $1.0$), reflecting the short-answer style that makes GQA suitable as a clean training distribution for confidence probes.

\subsubsection{POPE (Object Hallucination Probing)}
POPE \cite{POPE} probes object hallucination through yes/no questions of the form ``Is there a \{object\} in the image?'' grounded on COCO \texttt{val2014} images \cite{MSCOCO}, and serves as the near-ceiling binary detection task in \texttt{VLCB}. The \texttt{lmms-lab/POPE} test split contains 9{,}000 questions split equally among three negative-sampling regimes exposed as \texttt{category}: \texttt{adversarial} (objects co-occurring with scene content actually in the image, the hardest of the three), \texttt{popular} (frequent COCO objects), and \texttt{random}. The split is exactly answer-balanced (4{,}500 \texttt{yes}, 4{,}500 \texttt{no}). We retain all 9{,}000 questions; each becomes a one-token-answer record. The category distribution is reported in Table~\ref{tab:vlcb_pope}.

\begin{table}[ht]
\small
\centering
\caption{POPE category distribution (test).}
\label{tab:vlcb_pope}
\begin{tabular}{lrr}
\toprule
\textbf{Category} & \textbf{Samples} & \textbf{\%} \\
\midrule
adversarial & 3{,}000 & 33.3 \\
popular     & 3{,}000 & 33.3 \\
random      & 3{,}000 & 33.3 \\
\midrule
\textbf{Total} & \textbf{9{,}000} & \textbf{100.0} \\
\bottomrule
\end{tabular}
\end{table}

\subsubsection{GMAI-MMBench (Medical Multimodal Reasoning)}
GMAI-MMBench \cite{GMAI-MMBench} is a large-scale multimodal medical benchmark covering radiology, pathology, dermatology, ophthalmology, and surgery, and is the principal high-stakes setting in \texttt{VLCB}. The HuggingFace \texttt{train} placeholder aggregates 25{,}831 questions but does not preserve the official split structure. The official splits are released as TSVs: \texttt{GMAI\_mm\_bench\_VAL.tsv} (4{,}550 questions \emph{with} answers) and \texttt{GMAI\_mm\_bench\_TEST\_part\_*.tsv} ($\sim$21{,}281 questions, answers withheld for the leaderboard). Because the test partitions ship without ground truth, we use the official VAL split. One of the 4{,}550 raw images exceeds PIL’s maximum allowed image size during decompression and is discarded, leaving \textbf{4{,}549} samples.

For each retained sample we (i) decode the base64 image, (ii) format the question by appending options ``A) $o_A$ B) $o_B$ \dots'' (option E is omitted when null), and (iii) recover the option-letter ground truth by matching the textual \texttt{category} field against the option strings (case-insensitive, with a partial-match fallback). The standardized \texttt{category} field stores the source \texttt{clinical VQA task} (over 40 distinct values, with \emph{Disease Diagnosis} dominating at 46.4\% of the upstream pool). The answer-letter distribution after mapping is A:1{,}150 / B:1{,}175 / C:1{,}055 / D:977 / E:192, close to uniform across the four primary options with a long tail in option E.

\subsubsection{MME-Finance (Financial Chart VQA)}
MME-Finance \cite{MME-Finance} evaluates LVLMs on financial charts (candlestick, line, bar, table) rendered across four display styles (PC, photography, mobile vertical, mobile horizontal), and serves as the financial-document setting in \texttt{VLCB}. The release ships 1{,}171 question--image pairs across nine task categories. We retain only the six numerical, perceptual, OCR, and domain-knowledge categories that admit a single textual answer compatible with our pipeline, yielding \textbf{892} samples; the three categories dropped are open-ended explanation tasks for which a single textual ground truth is not well defined. The \texttt{question} field is taken verbatim, the \texttt{answer} preserves the dataset's free-form (often multi-line) reference text, and \texttt{category} stores the \texttt{task\_category} field. The retained category distribution is reported in Table~\ref{tab:vlcb_mmefinance}.

\begin{table}[ht]
\small
\centering
\caption{MME-Finance task category distribution (test, after filtering).}
\label{tab:vlcb_mmefinance}
\begin{tabular}{lrr}
\toprule
\textbf{Category} & \textbf{Samples} & \textbf{\%} \\
\midrule
Spatial Awareness                & 229 & 25.7 \\
OCR                              & 178 & 20.0 \\
Entity Recognition               & 163 & 18.3 \\
Financial Knowledge              & 147 & 16.5 \\
Accurate Numerical Calculation   & 133 & 14.9 \\
Numerical Calculation            &  42 &  4.7 \\
\midrule
\textbf{Total}                   & \textbf{892} & \textbf{100.0} \\
\bottomrule
\end{tabular}
\end{table}

\subsubsection{MMMU-Pro (Multidisciplinary College-Level VQA)}
MMMU-Pro \cite{MMMU} is a hardened version of MMMU spanning 30 college-level subjects across art, business, science, health, humanities, and engineering, with up to seven images per problem, and is the multi-choice reasoning setting in \texttt{VLCB}. We process the \texttt{standard} configurations released at 4 and 10 answer options (1{,}730 questions each); the screenshot-based \texttt{vision} configuration is also curated for completeness but is not used in the main evaluation suite. After hash-deduplication we retain \textbf{1{,}720} (4-option) and \textbf{1{,}725} (10-option) samples; the small drop relative to the 1{,}730 raw count reflects a handful of question--image pairs that hash identically across the two configurations. The \texttt{category} field is the joined string
\path{{subject}[SEP]{topic\_difficulty}[SEP]}\mbox{\texttt{\{img\_type joined by [LSEP]\}}},
preserving subject (e.g.\ Math, Clinical Medicine), topic difficulty, and the list of contained image modalities. Subjects are approximately uniform (50--60 questions each); the full distribution is reported in Table~\ref{tab:vlcb_mmmupro}. We include both 4-option and 10-option configurations because they share a question pool but differ in distractor count, which lets us examine whether confidence estimators degrade gracefully as the answer space widens.

\begin{table*}[ht]
\small
\centering
\caption{MMMU-Pro subject distribution (4-option and 10-option configurations are nearly identical pre-deduplication).}
\label{tab:vlcb_mmmupro}
\begin{tabular}{lr lr lr}
\toprule
\textbf{Subject} & \textbf{n} & \textbf{Subject} & \textbf{n} & \textbf{Subject} & \textbf{n} \\
\midrule
Agriculture & 60 & Diagnostics \& Lab.\ Medicine & 60 & Public Health & 58 \\
Design & 60 & Psychology & 60 & Accounting & 58 \\
Finance & 60 & Clinical Medicine & 59 & Energy \& Power & 58 \\
Physics & 60 & Biology & 59 & Pharmacy & 57 \\
Architecture \& Engineering & 60 & Marketing & 59 & History & 56 \\
Electronics & 60 & Economics & 59 & Art Theory & 55 \\
Computer Science & 60 & Mechanical Engineering & 59 & Sociology & 54 \\
Math & 60 & & & Art & 53 \\
Music & 60 & & & Literature & 52 \\
Materials & 60 & & & Basic Medical Science & 52 \\
Chemistry & 60 & & & Geography & 52 \\
& & & & Manage & 50 \\
\bottomrule
\end{tabular}
\end{table*}

\subsubsection{LLaVA-in-the-Wild (Open-Ended Multimodal Instruction Following)}
LLaVA-Bench-in-the-Wild \cite{LlavaWild} is a 60-question benchmark of Open-ended visual dialogue spanning 24 in-the-wild images and three answer styles: \texttt{conv} (short conversational), \texttt{detail} (detailed description), and \texttt{complex} (multi-step reasoning). It is the smallest source in \texttt{VLCB} and serves as a stress test for confidence estimation under open-ended generation; per-bin reliability estimates on this dataset should be read with the small-sample caveat developed in \S\ref{app:results_calibration_perdataset}. We pair each question with the GPT-4 reference response distributed alongside the benchmark and use it as the textual ground truth for our LLM-judge correctness grading (\S\ref{app:correctness}). 

\subsection{Final Train, Validation, and Test Assembly}
\label{app:assembly}

The per-source artifacts are merged into three splits used downstream: a training set, a validation set, and a test set. The test split is the union of the seven test artifacts above, with duplicates by \texttt{hash\_id} removed (none were found, by design of the per-dataset hashing scheme). Training and validation are reserved exclusively from GQA. This is the operational realization of \texttt{VLCB}'s design principle: a confidence estimator trained on a single, well-conditioned VQA distribution must generalize to test settings spanning medical imaging, financial chart understanding, multi-choice reasoning, and open-ended dialogue, none of which it has seen during training. Final sizes are reported in Table~\ref{tab:vlcb_split_summary} and the test split's per-source composition in Table~\ref{tab:vlcb_test_distribution}.

\begin{table*}[ht]
\small
\centering
\caption{\texttt{VLCB} split sizes and composition.}
\label{tab:vlcb_split_summary}
\begin{tabular}{lrl}
\toprule
\textbf{Split} & \textbf{Samples} & \textbf{Composition} \\
\midrule
\texttt{\texttt{VLCB}\_train\_raw} & 20{,}000 & GQA train (stratified on \texttt{detailed}) \\
\texttt{\texttt{VLCB}\_val\_raw}   &  5{,}000 & GQA val   (stratified on \texttt{detailed}) \\
\texttt{\texttt{VLCB}\_test\_raw}  & 30{,}514 & 7-source union, deduplicated by \texttt{hash\_id} \\
\bottomrule
\end{tabular}
\end{table*}

\begin{table}[ht]
\small
\centering
\caption{Per-source composition of \texttt{\texttt{VLCB}\_test\_raw}.}
\label{tab:vlcb_test_distribution}
\begin{tabular}{l r r}
\toprule
\textbf{Source} & \textbf{Samples} & \textbf{\%} \\
\midrule
GQA               & 12{,}568 & 41.19 \\
POPE              &  9{,}000 & 29.50 \\
GMAI-MMBench      &  4{,}549 & 14.91 \\
MMMU-Pro (10-opt) &  1{,}725 &  5.65 \\
MMMU-Pro (4-opt)  &  1{,}720 &  5.64 \\
MME-Finance       &     892  &  2.92 \\
LLaVA-in-the-Wild &      60  &  0.20 \\
\midrule
\textbf{Total}    & \textbf{30{,}514} & \textbf{100.00} \\
\bottomrule
\end{tabular}
\end{table}

GQA and POPE together account for roughly 71\% of the test split. This concentration reflects their established role as the two most widely used benchmarks in the LVLM confidence estimation and hallucination literature, where they have served as the de facto evaluation substrates for prior work; their prominence in \texttt{VLCB} therefore preserves comparability with that literature rather than reflecting a curation choice. The skew it introduces nonetheless motivates the two complementary aggregation modes used throughout our results: pooled aggregation (which weights every test sample equally and is therefore dominated by these two sources) and unweighted dataset averaging (which assigns equal weight to each of the seven sources irrespective of their sample counts). Both views are reported in Appendix~\ref{app:extended_results}, and the contrast between them is what allows us to disentangle aggregate performance from per-domain robustness.

\begin{table*}[ht]
\centering
\footnotesize
\caption{Image dimension statistics per source dataset, computed over every (question, image) pair in the union of train, val, and test splits. Images are counted once per pair, so an image associated with multiple questions contributes its dimensions multiple times; this matches the distribution of image sizes the inference pipeline actually encounters during evaluation. Width and height are reported separately as images are non-square.}
\label{tab:vlcb_image_dims}
\begin{tabular}{l r r r r r r r r r}
\toprule
\textbf{Dataset} & \textbf{$N$} & \textbf{W\,min} & \textbf{W\,max} & \textbf{W\,med} & \textbf{W\,mean} & \textbf{H\,min} & \textbf{H\,max} & \textbf{H\,med} & \textbf{H\,mean} \\
\midrule
GMAI-MMBench      & 4{,}549  &   55 & 6{,}824 & 512   & 857.4  &  54 & 8{,}686 & 434  & 681.0  \\
GQA               & 37{,}568 &   72 & 1{,}280 & 500   & 523.3  &  51 & 1{,}280 & 406  & 433.5  \\
LLaVA-Wild        &     60   &  505 & 4{,}800 & 1{,}214 & 1{,}416.9 & 374 & 3{,}203 & 1{,}152 & 1{,}173.4 \\
MME-Finance       &    892   &  674 & 3{,}648 & 1{,}604 & 1{,}612.4 & 195 & 3{,}648 & 1{,}368 & 1{,}193.3 \\
MMMU-Pro (10-opt) & 1{,}725  &   43 & 2{,}954 & 602   & 725.3  &  26 & 2{,}560 & 357  & 484.0  \\
MMMU-Pro (4-opt)  & 1{,}720  &   43 & 2{,}954 & 604   & 725.9  &  26 & 2{,}560 & 357  & 484.2  \\
POPE              & 9{,}000  &  333 &   640   & 640   & 584.7  & 234 &   640   & 480  & 478.8  \\
\bottomrule
\end{tabular}
\end{table*}

\paragraph{Quality control.} Every assembled split is verified against a battery of assertions before any downstream use: every required field must be present and non-empty; every \texttt{hash\_id} must be unique within its split; the pairwise intersection of \texttt{hash\_id}s across train, val, and test must be empty; and the train and val splits must contain exclusively GQA samples. To further rule out any cross-source leakage that could in principle slip past a hash defined per-source, we additionally compute a \emph{content-only} fingerprint over the canonicalized question text and a perceptual hash of the image, and intersect this fingerprint between (i) the training split and each of the seven test sources, and (ii) the validation split and each of the seven test sources. Across all 14 cross-source intersections, zero (image, question) pairs collide, confirming that no test sample has a content-equivalent twin in training or validation. As a looser diagnostic, we also intersect on image fingerprint alone: the only non-zero result is 5 GQA-train images that also appear in the POPE test set, attributable to the well-known Visual Genome--COCO image-pool overlap; the corresponding question texts differ (POPE's templated yes/no probes vs.\ GQA's scene-graph-derived compositional queries), so the (image, question) pairs do not collide. Image dimension statistics across source datasets are summarized in Table~\ref{tab:vlcb_image_dims}; the wide range of resolutions, from $43 \times 26$ pixels in MMMU-Pro to $6{,}824 \times 8{,}686$ pixels in GMAI-MMBench, reflects the heterogeneous visual content of the benchmark and motivates the 2{,}048-pixel downscaling policy applied at inference time (\S\ref{app:gen_extract}).

\subsection{Response Generation}
\label{app:gen_extract}

All five LVLMs are run on the full training, validation, and test splits under as uniform a generation protocol as their differing chat templates allow. Inference runs in \texttt{float32} with one exception: DeepSeek-VL2's vision encoder is numerically unstable outside \texttt{bfloat16} and is therefore loaded and evaluated in \texttt{bfloat16} throughout. This is a stability concession rather than a design choice and is the only point at which the generation conditions differ across models. Images whose larger dimension exceeds 2{,}048 pixels are downscaled while preserving aspect ratio; upsampling is never applied. Each sample is queried using the model's native chat and processor template, with generation fully deterministic (greedy decoding, \texttt{max\_new\_tokens=64}).

\paragraph{Prompt construction.}
Every model receives the same semantic instruction: answer the visual question briefly and completely, conditioned on the provided image. The instruction itself is held constant across models; only the delivery channel differs to accommodate each chat template. Qwen3-VL, Gemma-3, and InternVL3.5 use a dedicated system turn:

\begin{tcolorbox}[title=Qwen3-VL / Gemma-3 / InternVL3.5 --- Prompt,fonttitle=\bfseries,fontupper=\small]
\textbf{System:} You are a vision language assistant. Provide brief, complete answers.

\textbf{User:} \{image\} \{question\}
\end{tcolorbox}

LLaVA-NeXT does not reliably honour a system turn, so the instruction is appended directly to the user message:

\begin{tcolorbox}[title=LLaVA-NeXT --- Prompt,fonttitle=\bfseries,fontupper=\small]
\textbf{User:} \{image\} \{question\}

Provide a brief, complete answer.
\end{tcolorbox}

DeepSeek-VL2 uses a bespoke conversation schema where the system instruction is passed as an out-of-band processor argument rather than a chat turn:

\begin{tcolorbox}[title=DeepSeek-VL2 --- Prompt,fonttitle=\bfseries,fontupper=\small]
\textbf{System (out-of-band):} You are a vision language assistant. Provide brief, complete answers.

\textbf{User:} \{image\} \{question\}
\end{tcolorbox}

\subsection{Response Grading}
\label{app:correctness}
We label every sample in \texttt{VLCB}, on every dataset, with a binary correctness score $y_i \in \{0,1\}$ produced by a single LLM judge. The reason for using one protocol uniformly is that any alternative would mix grading rules across the benchmark: multiple-choice subsets could in principle be graded by string matching against the option key, and open-ended subsets necessarily require an LLM judge, but combining the two would mean systematic differences in measured confidence quality across datasets risk being attributable to grading artifacts rather than to genuine differences in model behavior. We therefore route the multiple-choice subsets (POPE, GMAI-MMBench, MMMU-Pro) through the judge as well, so the entire benchmark is graded under one rule. For the open-ended subsets (GQA short answers, MME-Finance, LLaVA-in-the-Wild) the LLM judge is in any case the only viable adjudicator, and this practice has been validated and widely adopted in recent calibration literature \cite{calibration-tuning, PHSV, emit, ccps, rmcb}.

\paragraph{Judge model and protocol.}
We use \texttt{gpt-5-mini} (with \texttt{reasoning\_effort=low}) accessed through the OpenAI Responses API. For each sample, the judge receives the question, the ground-truth answer from the dataset, the LVLM's generated response, and the corresponding image as a separate multimodal input, and is asked to determine whether the generated answer is semantically equivalent to the ground truth. The reliability of using a powerful LLM for this constrained equivalence task has been demonstrated by \citet{calibration-tuning}, who found that GPT-4 judgments exhibit a low average absolute difference of 4.5\% in accuracy estimation compared to human annotators. Building on their findings, and given the availability of more capable models since that study, we used \texttt{gpt-5-mini} to ensure high-quality labels at lower cost. We directly verify the reliability of this judge on \texttt{VLCB} through a stratified human-annotated audit, reported in Appendix~\ref{app:human_audit}. The system and user prompts are given below.

\begin{tcolorbox}[title=Response Grading --- System Prompt,fonttitle=\bfseries,fontupper=\small]
You are an expert answer evaluator. Your task is to determine if a student's answer to a question is correct by comparing it to the ground truth answer. \\ \\
1. Read the question carefully. \\
2. Compare the student's answer to the ground truth answer. \\
3. Consider semantic equivalence --- answers that mean the same thing should be considered correct even if worded differently. \\
4. Return ONLY ``yes'' if the answer is correct, or ``no'' if it is incorrect. \\
5. Be lenient with minor variations in wording, capitalization, or punctuation.
\end{tcolorbox}

\begin{tcolorbox}[title=Response Grading --- User Prompt,fonttitle=\bfseries,fontupper=\small]
Question: \{question\}

Ground Truth Answer: \{ground\_truth\_answer\}

Student Answer: \{generated\_response\}

Is the student's answer correct? (yes/no):
\end{tcolorbox}

\paragraph{Multimodal input format.} The image is passed to the judge as a separate multimodal input alongside the textual user prompt rather than being serialized into the prompt text, which is why the template above contains no image placeholder.

The resulting correctness and response-length statistics for each LVLM, broken down by dataset and split, are reported in Table~\ref{tab:vlcb_correctness}. Two patterns in the table are worth flagging because they shape how our results in \S\ref{sec:results} should be read. First, within-LVLM accuracy varies substantially across datasets: for Qwen3-VL-8B, for example, accuracy ranges from 37.2\% on MMMU-Pro 10-option to 88.7\% on POPE. This is the empirical reflection of the difficulty heterogeneity that motivates reporting an unweighted dataset average alongside pooled aggregation in Appendix~\ref{app:extended_results}. Second, the cross-LVLM accuracy spread on the harder datasets, for example GMAI-MMBench (35.3\% on LLaVA-NeXT to 60.0\% on InternVL3.5), is the source of the variance against which the per-method confidence estimation results in the main paper must be read.

\clearpage
\onecolumn
\begingroup
\small
\setlength{\LTleft}{\fill}
\setlength{\LTright}{\fill}
\setlength{\LTcapwidth}{\textwidth}
\begin{longtable}{llrrrll}
\caption{Per-VLM correctness breakdown by dataset and split. Train and validation splits contain only GQA, while the test split covers all seven source datasets. For each subset, we report total samples, correct and incorrect predictions (count and percentage), together with minimum, mean, and maximum word counts for the question and generated response. The aggregate row is reported only for the test split as \textbf{Total}.}
\label{tab:vlcb_correctness}\\
\toprule
\textbf{Split} & \textbf{Dataset} & \textbf{Total} & \textbf{Correct (\#, \%)} & \textbf{Incorrect (\#, \%)} & \textbf{Q words} & \textbf{R words} \\
\midrule
\endfirsthead

\multicolumn{7}{l}{\footnotesize\emph{(Continued from previous page)}}\\
\toprule
\textbf{Split} & \textbf{Dataset} & \textbf{Total} & \textbf{Correct (\#, \%)} & \textbf{Incorrect (\#, \%)} & \textbf{Q words} & \textbf{R words} \\
\midrule
\endhead

\midrule
\multicolumn{7}{r}{\footnotesize\emph{(Continued on next page)}}\\
\endfoot

\bottomrule
\endlastfoot

\multicolumn{7}{c}{\small\textbf{\texttt{Qwen/Qwen3-VL-8B-Instruct}}} \\
\addlinespace[2pt]
Train & GQA & 20{,}000 & 15{,}495 (77.5\%) & 4{,}505 (22.5\%) & 3/8.8/24 & 1/1.1/42 \\
\midrule
Val & GQA & 5{,}000 & 3{,}785 (75.7\%) & 1{,}215 (24.3\%) & 3/8.8/23 & 1/1.1/36 \\
\midrule
Test & GMAI-MMBench & 4{,}549 & 2{,}378 (52.3\%) & 2{,}171 (47.7\%) & 16/30.3/189 & 1/3.6/50 \\
 & GQA & 12{,}568 & 8{,}670 (69.0\%) & 3{,}898 (31.0\%) & 3/8.5/25 & 1/1.1/53 \\
 & LLaVA-Wild & 60 & 28 (46.7\%) & 32 (53.3\%) & 5/11.2/33 & 1/29.6/54 \\
 & MME-Finance & 892 & 461 (51.7\%) & 431 (48.3\%) & 3/11.3/21 & 1/9.9/54 \\
 & MMMU-Pro (10-opt) & 1{,}725 & 641 (37.2\%) & 1{,}084 (62.8\%) & 9/77.8/704 & 1/5.5/55 \\
 & MMMU-Pro (4-opt) & 1{,}720 & 810 (47.1\%) & 910 (52.9\%) & 8/54.1/582 & 1/5.3/55 \\
 & POPE & 9{,}000 & 7{,}979 (88.7\%) & 1{,}021 (11.3\%) & 7/7.2/8 & 1/1.0/1 \\
 & \textbf{Total} & \textbf{30{,}514} & \textbf{20{,}967 (68.7\%)} & \textbf{9{,}547 (31.3\%)} & \textbf{3/17.9/704} & \textbf{1/2.2/55} \\
\midrule
\midrule

\multicolumn{7}{c}{\small\textbf{\texttt{llava-hf/llava-v1.6-vicuna-13b-hf}}} \\
\addlinespace[2pt]
Train & GQA & 20{,}000 & 15{,}850 (79.2\%) & 4{,}150 (20.8\%) & 3/8.8/24 & 1/2.7/56 \\
\midrule
Val & GQA & 5{,}000 & 3{,}847 (76.9\%) & 1{,}153 (23.1\%) & 3/8.8/23 & 1/2.7/55 \\
\midrule
Test & GMAI-MMBench & 4{,}549 & 1{,}605 (35.3\%) & 2{,}944 (64.7\%) & 16/30.3/189 & 2/9.2/53 \\
 & GQA & 12{,}568 & 8{,}848 (70.4\%) & 3{,}720 (29.6\%) & 3/8.5/25 & 1/3.7/58 \\
 & LLaVA-Wild & 60 & 18 (30.0\%) & 42 (70.0\%) & 5/11.2/33 & 1/36.5/56 \\
 & MME-Finance & 892 & 194 (21.7\%) & 698 (78.3\%) & 3/11.3/21 & 1/21.3/58 \\
 & MMMU-Pro (10-opt) & 1{,}725 & 318 (18.4\%) & 1{,}407 (81.6\%) & 9/77.8/704 & 1/22.9/60 \\
 & MMMU-Pro (4-opt) & 1{,}720 & 470 (27.3\%) & 1{,}250 (72.7\%) & 8/54.1/582 & 1/22.5/60 \\
 & POPE & 9{,}000 & 7{,}952 (88.4\%) & 1{,}048 (11.6\%) & 7/7.2/8 & 1/6.6/25 \\
 & \textbf{Total} & \textbf{30{,}514} & \textbf{19{,}405 (63.6\%)} & \textbf{11{,}109 (36.4\%)} & \textbf{3/17.9/704} & \textbf{1/8.1/60} \\
\midrule
\midrule

\multicolumn{7}{c}{\small\textbf{\texttt{OpenGVLab/InternVL3\_5-14B-HF}}} \\
\addlinespace[2pt]
Train & GQA & 20{,}000 & 15{,}155 (75.8\%) & 4{,}845 (24.2\%) & 3/8.8/24 & 1/29.6/62 \\
\midrule
Val & GQA & 5{,}000 & 3{,}750 (75.0\%) & 1{,}250 (25.0\%) & 3/8.8/23 & 1/29.5/61 \\
\midrule
Test & GMAI-MMBench & 4{,}549 & 2{,}730 (60.0\%) & 1{,}819 (40.0\%) & 16/30.3/189 & 1/10.4/57 \\
 & GQA & 12{,}568 & 8{,}594 (68.4\%) & 3{,}974 (31.6\%) & 3/8.5/25 & 1/31.4/62 \\
 & LLaVA-Wild & 60 & 15 (25.0\%) & 45 (75.0\%) & 5/11.2/33 & 8/42.1/58 \\
 & MME-Finance & 892 & 460 (51.6\%) & 432 (48.4\%) & 3/11.3/21 & 1/25.2/59 \\
 & MMMU-Pro (10-opt) & 1{,}724 & 541 (31.4\%) & 1{,}183 (68.6\%) & 9/77.9/704 & 1/33.5/59 \\
 & MMMU-Pro (4-opt) & 1{,}720 & 654 (38.0\%) & 1{,}066 (62.0\%) & 8/54.1/582 & 1/35.4/60 \\
 & POPE & 9{,}000 & 7{,}683 (85.4\%) & 1{,}317 (14.6\%) & 7/7.2/8 & 1/23.5/59 \\
 & \textbf{Total} & \textbf{30{,}513} & \textbf{20{,}677 (67.8\%)} & \textbf{9{,}836 (32.2\%)} & \textbf{3/17.9/704} & \textbf{1/26.1/62} \\
\midrule
\midrule

\multicolumn{7}{c}{\small\textbf{\texttt{google/gemma-3-27b-it}}} \\
\addlinespace[2pt]
Train & GQA & 20{,}000 & 13{,}122 (65.6\%) & 6{,}878 (34.4\%) & 3/8.8/24 & 1/4.4/42 \\
\midrule
Val & GQA & 5{,}000 & 3{,}244 (64.9\%) & 1{,}756 (35.1\%) & 3/8.8/23 & 1/4.4/40 \\
\midrule
Test & GMAI-MMBench & 4{,}549 & 2{,}225 (48.9\%) & 2{,}324 (51.1\%) & 16/30.3/189 & 2/14.5/54 \\
 & GQA & 12{,}568 & 7{,}479 (59.5\%) & 5{,}089 (40.5\%) & 3/8.5/25 & 1/5.3/52 \\
 & LLaVA-Wild & 60 & 31 (51.7\%) & 29 (48.3\%) & 5/11.2/33 & 1/31.0/56 \\
 & MME-Finance & 892 & 381 (42.7\%) & 511 (57.3\%) & 3/11.3/21 & 1/11.8/51 \\
 & MMMU-Pro (10-opt) & 1{,}717 & 618 (36.0\%) & 1{,}099 (64.0\%) & 9/76.3/558 & 1/21.9/55 \\
 & MMMU-Pro (4-opt) & 1{,}718 & 822 (47.8\%) & 896 (52.2\%) & 8/53.6/508 & 1/20.4/55 \\
 & POPE & 9{,}000 & 7{,}565 (84.1\%) & 1{,}435 (15.9\%) & 7/7.2/8 & 1/1.3/23 \\
 & \textbf{Total} & \textbf{30{,}504} & \textbf{19{,}121 (62.7\%)} & \textbf{11{,}383 (37.3\%)} & \textbf{3/17.8/558} & \textbf{1/7.5/56} \\
\midrule
\midrule

\multicolumn{7}{c}{\small\textbf{\texttt{deepseek-ai/deepseek-vl2}}} \\
\addlinespace[2pt]
Train & GQA & 20{,}000 & 12{,}883 (64.4\%) & 7{,}117 (35.6\%) & 3/8.8/24 & 1/8.4/64 \\
\midrule
Val & GQA & 5{,}000 & 3{,}154 (63.1\%) & 1{,}846 (36.9\%) & 3/8.8/23 & 1/8.3/64 \\
\midrule
Test & GMAI-MMBench & 4{,}549 & 1{,}670 (36.7\%) & 2{,}879 (63.3\%) & 16/30.3/189 & 0/11.2/61 \\
 & GQA & 12{,}568 & 6{,}757 (53.8\%) & 5{,}811 (46.2\%) & 3/8.5/25 & 1/9.1/64 \\
 & LLaVA-Wild & 60 & 16 (26.7\%) & 44 (73.3\%) & 5/11.2/33 & 1/32.9/64 \\
 & MME-Finance & 892 & 265 (29.7\%) & 627 (70.3\%) & 3/11.3/21 & 0/7.3/64 \\
 & MMMU-Pro (10-opt) & 1{,}725 & 185 (10.7\%) & 1{,}540 (89.3\%) & 9/77.8/704 & 0/31.8/64 \\
 & MMMU-Pro (4-opt) & 1{,}720 & 316 (18.4\%) & 1{,}404 (81.6\%) & 8/54.1/582 & 0/30.5/64 \\
 & POPE & 9{,}000 & 7{,}573 (84.1\%) & 1{,}427 (15.9\%) & 7/7.2/8 & 1/6.5/35 \\
 & \textbf{Total} & \textbf{30{,}514} & \textbf{16{,}782 (55.0\%)} & \textbf{13{,}732 (45.0\%)} & \textbf{3/17.9/704} & \textbf{0/11.1/64} \\

\end{longtable}
\endgroup
\twocolumn

\subsection{Human Agreement Audit of the Correctness Judge}
\label{app:human_audit}

\begin{table*}[t]
\centering
\small
\setlength{\tabcolsep}{6pt}
\caption{\textbf{Human agreement audit of the \texttt{gpt-5-mini} correctness judge.} A seed-42 stratified sample of 350 \texttt{VLCB} responses (50 per dataset, balanced 25 judge-correct / 25 judge-incorrect) was re-annotated blind to the judge's label. Two items carry no binary human judgment and are excluded, so agreement and $\kappa$ are computed on $n = 348$ (the two affected datasets scored on $n = 49$). \emph{Agreements}: items where human and judge coincide. \emph{Judge stricter}: judge incorrect, human correct. \emph{Judge more lenient}: judge correct, human incorrect.}
\label{tab:human_audit}
\resizebox{\textwidth}{!}{%
\begin{tabular}{lrrrrrrr}
\toprule
Dataset & $n$ & Agreements & Agreement \% & Cohen's $\kappa$ & Disagreements & Judge stricter & Judge more lenient \\
\midrule
GQA & 50 & 50 & 100.0 & 1.000 & 0 & 0 & 0 \\
POPE & 50 & 50 & 100.0 & 1.000 & 0 & 0 & 0 \\
GMAI-MMBench & 50 & 50 & 100.0 & 1.000 & 0 & 0 & 0 \\
MMMU-Pro-4 & 50 & 50 & 100.0 & 1.000 & 0 & 0 & 0 \\
MMMU-Pro-10 & 50 & 48 & 96.0 & 0.920 & 2 & 0 & 2 \\
MME-Finance & 49 & 46 & 93.9 & 0.877 & 3 & 3 & 0 \\
LLaVA-in-the-Wild & 49 & 48 & 98.0 & 0.959 & 1 & 0 & 1 \\
\midrule
\textbf{All} & \textbf{348} & \textbf{342} & \textbf{98.3} & \textbf{0.966} & \textbf{6} & \textbf{3} & \textbf{3} \\
\bottomrule
\end{tabular}%
}
\end{table*}

All correctness labels in \texttt{VLCB} are produced by the \texttt{gpt-5-mini} judge described in Appendix~\ref{app:correctness}. To verify that this automated protocol tracks human judgment, one author re-annotated a stratified sample of the benchmark and we compared the two label sets.

\paragraph{Sampling design.} We drew 50 items from each of the seven \texttt{VLCB} datasets (350 total) with a fixed random seed (42). Within each dataset the draw is balanced across the judge's decision, 25 items the judge labelled correct and 25 it labelled incorrect, so the audit stress-tests both the judge's positive and negative calls equally rather than oversampling whichever class dominates a given dataset.

\paragraph{Annotation protocol.} One author annotated every sampled item blind to the judge's label: the annotation interface showed only the image, question, model answer, and ground truth, with the judge's call hidden, and applied the same correctness criterion used by the judge (semantic equivalence to the ground truth). The human and judge labels were compared only afterwards. Two items received no binary human judgment; both are long-reference open-ended items (one LLaVA-in-the-Wild, one MME-Finance), where the ground truth did not admit a clean correct/incorrect call. These two are excluded from the agreement computation, which is therefore over $n = 348$; the two affected datasets are scored on $n = 49$.

\paragraph{Results.} Table~\ref{tab:human_audit} reports per-dataset agreement. Human and judge agree on 342 of 348 items (98.3\%), with Cohen's $\kappa = 0.966$, in the ``almost perfect'' range. Agreement is 100\% on the four datasets with objective answer keys (GQA, POPE, GMAI-MMBench, MMMU-Pro-4), and all six disagreements fall on the three datasets where semantic judgment is genuinely hard (MMMU-Pro-10, MME-Finance, LLaVA-in-the-Wild), with per-dataset $\kappa$ between 0.877 and 0.959. The disagreements are balanced in direction: three where the judge was stricter than the human (judge incorrect, human correct) and three where it was more lenient (judge correct, human incorrect), indicating no systematic directional bias in the judge.

\paragraph{Interpretation.} This is a two-rater agreement study with a single annotator and one judgment per item. It yields no gold label and no accuracy figure for either rater, so the six disagreements are not judge errors; they mark items on which two competent raters applying the same criterion diverged. The high agreement is consistent with, and stronger than, the reliability basis cited in Appendix~\ref{app:correctness}: \citet{calibration-tuning} report a 4.5\% average absolute difference between GPT-4 judgments and human annotators on this kind of constrained equivalence task. Two design choices likely contribute, a newer-generation judge (\texttt{gpt-5-mini}) than the GPT-4 used in that study, and grading against the actual image rather than the text alone. The main limitation of this audit is its scale and its use of a single annotator; domain-specialist grading of the medical and financial subsets at larger scale remains a valuable further validation step.

\subsection{Benchmark Availability and Licensing}
\label{app:availability}

\texttt{VLCB} is a composite resource over seven public datasets, each governed by its own license; licenses range from permissive (Apache 2.0, MIT, CC BY) to restrictive (CC BY-NC-SA, research-use-only). We release the components needed to reconstruct it: the per-source curation code and the aggregator that assembles the three splits, the generation driver and the LLM-judge grading code, and the deterministic \texttt{hash\_id} construction. Any user who has independently obtained the source datasets from their official distributors can therefore reproduce the \texttt{VLCB} splits bit-for-bit by running our code; the assembled benchmark is not redistributed as a single archive because doing so would conflict with the more restrictive source licenses. All released code is licensed under MIT. The reconstructed \texttt{VLCB} benchmark is a derivative work and inherits the most restrictive terms of its constituent sources; it is therefore intended for \textbf{non-commercial research use only} and is subject to all applicable ShareAlike provisions inherited from GMAI-MMBench. Users are solely responsible for acquiring the source datasets from their official distributors and for adhering to their original license terms.

%% file: sections/appendix_baselines.tex
\section{Baseline Confidence Estimation Methods}
\label{app:baselines}

This appendix documents the seven established confidence estimation baselines benchmarked in this work. Each method operates on the language model backbone and reads one of three signals to estimate confidence in a generated answer: output logits, verbalized confidence scores, or hidden states. We organize the methods into three families along this axis: prompt-based methods (\S\ref{app:baselines_bb}), internal-state probing methods (\S\ref{app:baselines_probing}), and internal-stability methods (\S\ref{app:baselines_stability}). For trainable methods, we adhere to the architectural descriptions of the original publications wherever they are specified, with deviations flagged in the relevant subsections. All baselines operate on the generation pass described in \S\ref{app:gen_extract}; hidden states and logits are extracted from the same frozen LVLM checkpoints listed in Tables~\ref{tab:vlcb_llm}--\ref{tab:vlcb_vit}, with no additional fine-tuning of the underlying LVLM. Throughout this appendix, each method is referred to by the canonical name we use in the rest of the paper; Table~\ref{tab:baseline_overview} also lists the short-form abbreviation used in figure legends and per-method labels in the results.

\begin{table*}[ht]
\centering
\resizebox{\textwidth}{!}{
\begin{tabular}{llllc}
\toprule
\textbf{Method} & \textbf{Citation} & \textbf{Model Access} & \textbf{Input Signal} & \makecell[c]{\textbf{Training}\\\textbf{Required}} \\
\midrule
\multicolumn{5}{c}{\small\textbf{Prompt-Based}} \\
\addlinespace[2pt]
P(True)
  & \cite{kadavath}
  & Token-logit access
  & Output token scores
  & No \\
Self-Probing
  & \cite{self-probing}
  & Black-box
  & Verbalized confidence
  & No \\
Prompt Ensembles (PE)
  & \cite{promptensemble}
  & Black-box
  & Aggregated output probabilities
  & No \\
\midrule
\multicolumn{5}{c}{\small\textbf{Internal-State Probing}} \\
\addlinespace[2pt]
P(I Know)
  & \cite{kadavath}
  & White-box (hidden states)
  & First-token final-layer hidden state
  & Yes \\
SAPLMA
  & \cite{saplma}
  & White-box (hidden states)
  & Last-token final-layer hidden state
  & Yes \\
InternalInspector ($\text{I}^2$)
  & \cite{internalinspector}
  & White-box (all-layer hidden states)
  & Per-layer activation, attention, and FFN states
  & Yes \\
\midrule
\multicolumn{5}{c}{\small\textbf{Internal-Stability}} \\
\addlinespace[2pt]
CCPS
  & \cite{ccps}
  & White-box (hidden states + gradients)
  & Perturbation trajectory statistics
  & Yes \\
\bottomrule
\end{tabular}
}
\caption{Overview of the seven confidence estimation baselines benchmarked in
this work, organized by the family of signal they exploit, the degree of model
access they require, and whether they involve any training.}
\label{tab:baseline_overview}
\end{table*}

\subsection{Prompt-Based Methods}
\label{app:baselines_bb}

Prompt-based methods treat the LVLM as an externally queried system, deriving confidence through prompt design rather than access to internal representations. The three methods in this family differ in how they extract a confidence signal: P(True) reads a token logit from a self-evaluation query, Self-Probing parses a verbalized confidence score, and Prompt Ensembles aggregates output probabilities across paraphrased questions. All three are training-free, and all three require at least one additional inference pass beyond the original answer-generation pass (Self-Probing and P(True) require one additional generation each, Prompt Ensembles requires ten).

\subsubsection{P(True)}
P(True), introduced by \citet{kadavath}, assesses the LVLM's self-evaluation of its own generated answer through a token-logit readout on a binary self-query. P(True) does not access hidden representations or activations; it does require token-level output scores for the auxiliary response token, which is a standard decoding output exposed by most local inference frameworks and by several API providers via log-probability endpoints, so no internal model access is implied. After the LVLM produces a response to the original visual question, the (image, question, generated answer) triple is fed back with the following uncertainty query:

\begin{tcolorbox}[title=P(True) --- Uncertainty Query,fonttitle=\bfseries,fontupper=\small]
Is the proposed answer correct? \\
A) no \\
B) yes \\
Reply with A or B only. \\
Answer:
\end{tcolorbox}

\paragraph{Implementation.} The model generates a single token under greedy decoding and the output logits at that position are recorded. To be robust to tokenizer variation across LVLMs, we collect logits for all \texttt{A}/\texttt{B} token variants (upper- and lowercase, with and without a leading space, with and without surrounding parentheses) and retain the maximum within each equivalence class. Let $\ell_A$ and $\ell_B$ denote the resulting scalar logits for the ``no'' and ``yes'' options respectively. The final confidence score is
\begin{equation}
p_{\text{True}} = \mathrm{softmax}([\ell_A,\, \ell_B])[1].
\end{equation}

\subsubsection{Self-Probing (SP)}
Self-Probing, proposed by \citet{self-probing}, prompts the LVLM to verbalize a numerical confidence in its own answer through a free-text generation pass. After the LVLM produces a response, it is queried with:

\begin{tcolorbox}[title=Self-Probing --- Verbalization Prompt,fonttitle=\bfseries,fontupper=\small]
Question: \{question\} \\
Possible Answer: \{generated\_answer\} \\[4pt]
Q: How likely is the above answer to be correct? give your confidence in the following format: \\
Confidence: <number from 0 to 100>\% \\
Note: The confidence indicates how likely you think the answer is true.
\end{tcolorbox}

\paragraph{Implementation.} The verbalized response is parsed with a regular expression to extract the confidence value, which is then normalized to $[0, 1]$.

\subsubsection{Prompt Ensembles (PE)}

Prompt Ensembles, formalized by \citet{promptensemble}, estimates confidence by averaging predictions across $N$ meaning-preserving rewrites of the input question. The LVLM independently answers the original question and each of the $N$ paraphrases under greedy decoding, yielding $N+1$ generated responses per sample. A per-prompt confidence is computed as the length-normalized sequence likelihood of the response, and the ensemble confidence is the arithmetic mean of the $N+1$ per-prompt scores. The intuition is that an answer the model is sure of should remain stable across semantically equivalent rephrasings, while an answer driven by surface-level cues should fluctuate; averaging across paraphrases therefore acts as a soft consistency check at the output-distribution level.

\begin{tcolorbox}[title=PE --- System Prompt,fonttitle=\bfseries,fontupper=\small]
You are an expert linguist and domain specialist generating alternative phrasings
for a Visual Question Answering (VQA) task. Your goal is to generate variations
of a question while preserving its exact semantic meaning and expected answer.

\textbf{Instructions:}
\begin{itemize}[nosep,leftmargin=*]
  \item Generate the requested number of alternative ways to phrase the question.

  \item \textbf{Preserve the Answer:} Ensure that for any given image, the answer
  to your new questions would be \emph{identical} to the answer of the original
  question.

  \item \textbf{Maintain Domain Precision:}
  \begin{itemize}[nosep,leftmargin=1.5em]
    \item If the question is \textbf{Medical}, preserve the correct clinical terminology.
    \item If the question is \textbf{Financial}, keep the technical intent clear.
  \end{itemize}

  \item \textbf{Maintain Question Type:}
  \begin{itemize}[nosep,leftmargin=1.5em]
    \item If the original is a \textbf{Yes/No} question, your rephrasings must remain Yes/No questions.
    \item If the original asks for a \textbf{Count}, your rephrasings must ask for a number.
  \end{itemize}

  \item \textbf{Multiple Choice Handling (Crucial):}
  \begin{itemize}[nosep,leftmargin=1.5em]
    \item If the original question includes multiple options (e.g., A, B, C, D):
    \begin{enumerate}[nosep,leftmargin=1.5em]
      \item Rephrase \textbf{only} the question stem (the text asking the question).
      \item \textbf{Append the EXACT same options} in the \textbf{EXACT same order}
      to the end of your rephrased question.
      \item Do \textbf{not} shuffle, reword, or modify the options in any way.
    \end{enumerate}
  \end{itemize}

  \item \textbf{Allowed Changes:} You may vary word order, sentence structure,
  and use strict synonyms for the question text.

  \item \textbf{Prohibited Changes:} Do \textbf{not} add new constraints, remove
  location details, or introduce ambiguity.
\end{itemize}

\textbf{Output Format:}

Each rephrased question should be wrapped in numbered tags like this:

\texttt{[question\_1] Rephrased question stem? (A) Option 1 (B) Option 2... [/question\_1]}\\
\texttt{[question\_2] Rephrased question stem? (A) Option 1 (B) Option 2... [/question\_2]}\\
\texttt{...and so on for each question.}
\end{tcolorbox}

\begin{tcolorbox}[title=PE --- User Prompt,fonttitle=\bfseries,fontupper=\small]
\textbf{Original Question:} \texttt{'\{question\}'}

Please generate \{N\} alternative phrasings for this question.

\textbf{Output Format:}

\texttt{[question\_1] ... [/question\_1]}\\
\texttt{[question\_2] ... [/question\_2]}\\
\texttt{...}\\
\texttt{[question\_\{N\}] ... [/question\_\{N\}]}
\end{tcolorbox}

\paragraph{Implementation.} We use $N=10$, the smallest ensemble size at which \citet{promptensemble} report near-saturated calibration gains, balancing the per-paraphrase inference cost against the marginal benefit of adding another rewrite. Paraphrases are generated via the OpenAI \texttt{gpt-5-mini} API with reasoning effort set to \texttt{medium}, under the system and user prompts shown above. For each of the eleven (image, question) pairs (the original plus ten paraphrases) we record per-token log-probabilities $\{\log p_t\}_{t=1}^{T}$ of the generated response and compute a per-prompt confidence as the length-normalized sequence likelihood,
\begin{equation}
c^{(i)} = \exp\!\left(\frac{1}{T}\sum_{t=1}^{T} \log p_t\right),
\end{equation}
equivalent to the geometric mean of per-token probabilities. This follows standard practice for sequence-likelihood confidence in LLM uncertainty estimation \citep{geometric-mean1,geometric-mean2}. The ensemble confidence is the arithmetic mean of the eleven per-prompt scores, $c^{\text{PE}} = \tfrac{1}{N+1}\sum_{i=0}^{N} c^{(i)}$.

\subsection{Internal-State Probing Methods}
\label{app:baselines_probing}

Probing methods access the LVLM's internal activations and train a lightweight classifier head to predict answer correctness from static snapshots of those activations. The three methods in this family differ in which forward pass they read from (prompt-only or extended through the generated response) and in how the resulting hidden states are aggregated; the specifics are given per method below. Across all three methods the LVLM weights are held fixed; with methods using early stopping on the composite validation score defined in Appendix~\ref{app:validation}. All reported metrics are computed on the test set.

\subsubsection{P(I~Know) (P(IK))}
P(I~Know), also from \citet{kadavath}, estimates the probability that the LVLM will produce a correct answer \emph{before} any response is generated. We implement P(IK) as a lightweight classifier head on the frozen LVLM: during the extraction forward pass we capture the final-layer hidden state at the last token of the input context (the system prompt followed by the user message containing the image and the question), yielding a vector $h \in \mathbb{R}^{d}$ that summarizes the model's representation of the full input at the moment it would otherwise begin generating. The probe therefore reads only the prompt, with no generated response in the forward pass; this contrasts with SAPLMA and InternalInspector, which both extend the pass through the generated response and read from its final token. P(IK) is the natural architectural counterpart to \texttt{BICR}: both train an MLP over a single hidden-state vector from a prompt-only pass at the same token position, and \texttt{BICR}'s improvement over P(IK) in the main results (\S\ref{sec:results}) is therefore attributable to the blank-image ranking signal rather than to any architectural difference.

\paragraph{Implementation.} The classifier is a multi-layer perceptron with ReLU activations, trained with \texttt{BCEWithLogitsLoss} using a positive-class reweighting factor to correct for class imbalance and optimized with Adam. Because the original publication does not specify head hyperparameters at the LVLM scale we work with, we tune them per LVLM via an Optuna search on the validation set; the full search space and training protocol are documented in Appendix~\ref{app:optuna}. At inference, the confidence score is $\sigma(f_\theta(h))$.

\subsubsection{SAPLMA}
SAPLMA \cite{saplma} trains a lightweight feedforward classifier on the final-layer hidden state at the last token of the model's generated response to predict answer correctness. The original method reads this hidden state at the end of the generated text alone, which is appropriate in text-only settings where the response is a self-contained statement: ``the capital of the US is DC'' carries its own truth conditions and a representation of just that statement is enough to assess it. LVLM responses are typically not self-contained in this way. A one-word answer like ``red'' to ``what color is this'' is uninterpretable in isolation; whether it is correct depends entirely on the image and the question that prompted it. We therefore extract the hidden state at the last token of the full sequence consisting of the input context (system prompt, image, question) followed by the generated response, so that the probe reads a representation reflecting the joint (image, question, response) context after the model has committed to its answer. This is the only departure from the original SAPLMA convention.

\paragraph{Implementation.} The classifier is a four-layer MLP with hidden widths $d \to 256 \to 128 \to 64 \to 1$, ReLU activations, and a final linear projection to a scalar logit, following the architecture of \citet{saplma}. Training uses \texttt{BCEWithLogitsLoss} with a positive-class reweighting factor to correct for class imbalance, optimized with Adam at batch size $32$ for at most $200$ epochs with early stopping on the composite validation score (Appendix~\ref{app:validation}) with patience $20$. To reduce seed sensitivity we train five independent classifiers with seeds $\{23, 42, 137, 2024, 3407\}$ and report the mean of all metrics across runs. At inference, the confidence score is $\sigma(f_\theta(h))$.

\subsubsection{\texorpdfstring{InternalInspector ($I^2$)}{InternalInspector (I2)}}

\citet{internalinspector} argue that useful correctness signal is distributed across the full depth of the model rather than concentrated in a single layer. For each sample we extract three per-layer representations at the last token of the full sequence consisting of the input context (system prompt, image, question) followed by the generated response, via forward hooks on every transformer block: the post-residual activation state $h^{(l)}$, the pre-residual multi-head self-attention output $a^{(l)}$, and the pre-residual feed-forward output $m^{(l)}$. Stacking these across $L$ layers yields a per-sample tensor of shape $[L, d, 3]$, which is treated as a 3-channel image with $L$ rows (one per layer), $d$ columns (one per hidden dimension), and one channel for each of the three state types. We benchmark the strongest variant reported in the original work, a CNN-based encoder over all three state types, which outperforms alternatives that use a Transformer encoder or a subset of state types. The trainable parameter count of this variant is fixed at approximately $11.3$M regardless of the underlying LVLM, because the CNN encoder reduces the input to a fixed spatial footprint via adaptive pooling before the projection head; the full parameter accounting is in Appendix~\ref{app:params}.

\paragraph{Implementation.} The $[L, d, 3]$ tensor is passed through a ResNet18 CNN encoder (stem followed by four residual stages with channel progression $64 \to 128 \to 256 \to 512$) and adaptive average pooling, yielding a $512$-dimensional embedding. A linear projection $512 \to 128$ followed by a four-layer MLP classifier ($128 \to 256 \to 128 \to 64 \to 1$) with ReLU activations and dropout $0.1$ produces the correctness logit. The encoder and classifier are trained jointly from scratch using a supervised contrastive loss combined with \texttt{BCEWithLogitsLoss} (with positive-class reweighting for class imbalance) following \citet[Eq.~4]{internalinspector}, with Adam at learning rate $10^{-3}$, weight decay $10^{-4}$, and contrastive temperature $\tau=0.1$. Training runs for at most $200$ epochs with early stopping on the composite validation score (Appendix~\ref{app:validation}) with patience $20$.

\subsection{Internal-Stability Methods}
\label{app:baselines_stability}

Stability-based methods probe the LVLM's internal representations not by reading static activations but by measuring how those representations \emph{respond} to controlled perturbations. The confidence signal is representational robustness: hidden states that retain their predictive content under targeted intervention are treated as reliable, while those whose predictions shift easily under the same intervention are treated as fragile. We evaluate one such method.

\subsubsection{CCPS}

CCPS \cite{ccps} estimates confidence by probing the stability of the LVLM's final hidden states under targeted adversarial perturbations: hidden states behind correct predictions should resist small interventions, while those behind incorrect ones should shift easily. For each token in the generated response, the hidden state is perturbed along the unit-normalized gradient direction in $S{=}5$ equal-sized steps up to $\varepsilon_{\max}{=}20.0$ (matching the original publication's settings), and per-token statistics are recorded across three groups, namely original-state features, perturbation-trajectory features, and comparison features quantifying the distributional shift between original and perturbed states (Table~\ref{tab:all_features}), yielding a $75$-channel per-token feature sequence over the full response.

\paragraph{Implementation.} CCPS uses a two-stage head. \emph{Stage~1} pre-trains a convolutional encoder, \texttt{Conv1d}$(75 \to 64, k{=}3) \to$ ReLU $\to$ \texttt{Conv1d}$(64 \to 32, k{=}3) \to$ ReLU $\to$ \texttt{AdaptiveMaxPool1d} $\to$ \texttt{Linear}$(32 \to 16)$, with a margin-based contrastive loss (margin $1.0$, class-agnostic by construction) for $5{,}000$ steps. \emph{Stage~2} appends \texttt{Linear}$(16 \to 32) \to$ ReLU $\to$ \texttt{Linear}$(32 \to 2)$ and jointly fine-tunes under cross-entropy (with positive-class reweighting for class imbalance) for a further $5{,}000$ steps. Both stages use Adam at learning rate $10^{-4}$, weight decay $0.1$, and batch size $32$. At inference, the confidence score is the softmax probability that the generated response is correct (i.e., the probability mass on the positive class of the binary classifier).

\begin{table*}[t]
\centering
\footnotesize
\resizebox{\textwidth}{!}{
\begin{tabular}{p{0.45\textwidth}p{0.5\textwidth}}
\toprule
\multicolumn{2}{c}{\textbf{Original State Features}} \\
\midrule
\texttt{original\_log\_prob\_actual} & Log-probability of the actual token under
the model's unperturbed output distribution. \\
\texttt{original\_prob\_actual} & Probability of the actual token under the
unperturbed distribution. \\
\texttt{original\_logit\_actual} & Raw logit of the actual token prior to any
perturbation. \\
\texttt{original\_prob\_argmax} & Highest probability assigned to any token by
the unperturbed model. \\
\texttt{original\_logit\_argmax} & Highest logit value assigned to any token
prior to perturbation. \\
\texttt{original\_entropy} & Entropy of the unperturbed predictive distribution:
$-\sum_i P_{\text{orig}}(i)\log P_{\text{orig}}(i)$. \\
\texttt{original\_margin\_logit\_top1\_top2} & Logit gap between the top-1 and
top-2 tokens before perturbation. \\
\texttt{original\_margin\_prob\_top1\_top2} & Probability gap between the top-1
and top-2 tokens before perturbation. \\
\texttt{original\_norm\_logits\_L2} & L2 norm of the unperturbed logit vector. \\
\texttt{original\_std\_logits} & Standard deviation of the unperturbed logit
values. \\
\texttt{original\_norm\_hidden\_state\_L2} & L2 norm of the unperturbed last
hidden state vector. \\
\texttt{is\_actual\_token\_original\_argmax} & Binary indicator of whether the
actual token is the argmax under the unperturbed model. \\
\midrule
\multicolumn{2}{c}{\textbf{Perturbation Trajectory Features}} \\
\midrule
\texttt{jacobian\_norm\_token} & L2 norm of the Jacobian of the token's
log-probability with respect to the hidden state, measuring local sensitivity. \\
\texttt{epsilon\_to\_flip\_token} & Smallest perturbation magnitude along the
gradient direction sufficient to change the argmax prediction. \\
\texttt{pei\_value\_token} & Perturbation Energy Integral (PEI): cumulative
normalized drop in the actual token's log-probability across all perturbation
steps. \\
\midrule
\multicolumn{2}{c}{\textbf{Comparison Features (Original vs.\ Perturbed)}} \\
\midrule
\texttt{perturbed\_log\_prob\_actual} & Log-probability of the actual token after
hidden-state perturbation. \\
\texttt{perturbed\_prob\_actual} & Probability of the actual token after
perturbation. \\
\texttt{perturbed\_logit\_actual} & Logit of the actual token after perturbation. \\
\texttt{perturbed\_prob\_argmax} & Highest probability assigned to any token after
perturbation. \\
\texttt{perturbed\_logit\_argmax} & Highest logit value after perturbation. \\
\texttt{perturbed\_entropy} & Entropy of the perturbed predictive distribution. \\
\texttt{perturbed\_margin\_logit\_top1\_top2} & Logit gap between top-1 and top-2
tokens after perturbation. \\
\texttt{perturbed\_norm\_logits\_L2} & L2 norm of the perturbed logit vector. \\
\texttt{delta\_log\_prob\_actual\_from\_original} & Absolute drop in log-probability
of the actual token after perturbation. \\
\texttt{did\_argmax\_change\_from\_original} & Binary indicator of whether the
argmax token shifted after perturbation. \\
\texttt{kl\_div\_perturbed\_from\_original} & KL divergence from the original to
the perturbed output distribution. \\
\texttt{js\_div\_perturbed\_from\_original} & Jensen-Shannon divergence between
the original and perturbed distributions. \\
\texttt{cosine\_sim\_logits\_perturbed\_to\_original} & Cosine similarity between
logit vectors before and after perturbation. \\
\texttt{cosine\_sim\_hidden\_perturbed\_to\_original} & Cosine similarity between
hidden-state vectors before and after perturbation. \\
\texttt{l2\_dist\_hidden\_perturbed\_from\_original} & L2 distance between
hidden-state vectors before and after perturbation. \\
\bottomrule
\end{tabular}
}
\caption{Feature groups and definitions used by CCPS \cite{ccps} to characterize
hidden-state stability under targeted perturbation.}
\label{tab:all_features}
\end{table*}

%% file: sections/appendix_novelty.tex
\section{Positioning Relative to Related Methods}
\label{app:novelty}

The literature on visual reliability shares the premise that grounding matters, but it contains several distinct \emph{kinds} of method, and \texttt{BICR} belongs to none of them. \texttt{BICR} is the only approach that is at once a frozen-model probe, read from a single last-prompt-token hidden state before generation, supervised on answer correctness, built on a real-versus-blank visual contrast, and evaluated jointly on calibration and discrimination. Table~\ref{tab:novelty} organizes the field into method families along the axes that separate them from \texttt{BICR}.

\begin{table*}[t]
\centering
\small
\renewcommand{\arraystretch}{1.3}
\setlength{\tabcolsep}{5pt}
\caption{\textbf{Method families in the LVLM visual-reliability literature, positioned against \texttt{BICR}.} \emph{Real-vs-null visual contrast} indicates whether the method compares representations or outputs under the real image against a perturbed or absent visual input. \emph{Frozen} indicates whether the base LVLM is left unchanged.}
\label{tab:novelty}
\begin{tabular}{@{}p{2.55cm}p{3.15cm}p{3.05cm}p{3.65cm}p{1.85cm}@{}}
\toprule
\textbf{Method family} & \textbf{Example(s)} & \textbf{What it outputs} & \textbf{Real-vs-null visual contrast} & \textbf{Frozen?} \\
\midrule
Decoding / hallucination mitigation & VCD~\citep{10657718}, MemVR~\citep{MemVR} & a modified answer & VCD: logit-level; MemVR: none & frozen \\
Interpretability diagnostic & TVI~\citep{Chain-of-Embedding}, Chen et al.~\citep{chen2025unveilinguncertainty} & grounding magnitude / calibration study & TVI: hidden-state, multi-layer & frozen \\
Backbone retraining & VL-Calibration~\citep{vl-calibration}, CSP~\citep{Zhao2025SemanticCalibration} & verbalized confidence & no & \textbf{retrains the LVLM} \\
Sampling / orchestration detector & VL-Uncertainty~\citep{VL-Uncertainty}, MS-FBI~\citep{Du2026MedicalVQA} & uncertainty from sampled / multi-round answers & no & frozen \\
Output-fusion estimator & Dang et al.~\citep{dang2026instinct} & fused P(correct) & no visual intervention & frozen \\
Single-view internal detector & SUQ~\citep{li2024referencefree}, SVAR~\citep{Devils} & truthfulness / hallucination score & none & frozen (aux head) \\
\midrule
\textbf{Confidence estimator (ours)} & \textbf{\texttt{BICR}} & \textbf{calibrated per-sample P(correct); answer untouched} & \textbf{hidden-state ranking, real vs blank} & \textbf{frozen probe} \\
\bottomrule
\end{tabular}
\end{table*}

\noindent\textbf{Three separations do the most work.} First, only VCD~\citep{10657718}, TVI~\citep{Chain-of-Embedding}, and \texttt{BICR} use a real-versus-null visual contrast at all, and even these move in different directions: VCD degrades the view with noise, while TVI and \texttt{BICR} remove it. Most of the field performs no such contrast. Second, \texttt{BICR} is the only method that produces its signal before any answer token exists, enabling pre-generation triage; every other method operates during or after generation. Third, our own null-image ablation (Appendix~\ref{app:design_blank_color}) shows that the information-different perturbations several of these works use (noise, blur) are weaker nulls for a grounding-contrast probe than a solid black image, evidence that \texttt{BICR}'s null is a validated design choice rather than an arbitrary variant.

\noindent\textbf{The closest relatives.} We are precise about the methods nearest to \texttt{BICR} rather than overclaim. TVI~\citep{Chain-of-Embedding} is the nearest relative and reached hidden-state grounding contrast first, but as an unsupervised, multi-layer, population-level diagnostic, validated by Spearman correlation with no per-sample calibrated score; \texttt{BICR} operationalizes a related contrast as a trained per-sample estimator. Dang et al.~\citep{dang2026instinct} share our task framing (post-hoc calibrated P(correct)) but fuse two answer-position channels that never consult the image, so they cannot flag a confident language-prior guess, which is exactly \texttt{BICR}'s target. The supervised uncertainty quantification (SUQ) family surveyed by \citet{li2024referencefree} is the closest in architecture, a supervised probe over hidden-state activations, but it reads a single snapshot with no contrast and, by that study's own visual-degradation experiment, degrades when the image degrades; this is the blind spot \texttt{BICR}'s real-versus-blank contrast is designed to fill. For the two decoding methods VCD~\citep{10657718} and MemVR~\citep{MemVR}, we additionally evaluate \texttt{BICR} against both directly on their own models and benchmarks, on the task of spotting hallucinated answers; Appendix~\ref{app:vcd_memvr} reports that \texttt{BICR} outperforms faithful confidence signals derived from each, flags more of the model's errors than either intervention catches, and composes on top of their outputs.

%% file: sections/appendix_metrics.tex
\section{Evaluation Metrics}
\label{app:evaluation_metrics}

We evaluate confidence estimation quality along two complementary axes: \emph{calibration}, which measures how well predicted confidence scores reflect true correctness frequencies, and \emph{discrimination}, which measures how well they separate correct from incorrect predictions. The metrics reported throughout this work are organized below.

\subsection{Calibration Metrics}
\label{app:metrics_calibration}

\subsubsection{Expected Calibration Error (ECE)}
A well-calibrated confidence estimator should assign a score of $p$ to predictions that are correct $p$ of the time in expectation. We measure calibration via ECE, which partitions all $n$ samples into $b$ equal-width bins $\{B_j\}_{j=1}^{b}$ over $[0,1]$ and computes the weighted average absolute deviation between mean predicted confidence and empirical accuracy:
\[
\text{ECE} = \sum_{j=1}^{b} \frac{|B_j|}{n} \left| \text{conf}(B_j) - \text{acc}(B_j) \right|
\]
where $\text{conf}(B_j)$ and $\text{acc}(B_j)$ denote the average confidence and observed accuracy within bin $B_j$, respectively. We use $b=10$ equal-width bins throughout. Lower ECE indicates better alignment between predicted scores and actual correctness rates.

\subsubsection{Brier Score (BS)}
The Brier Score measures the mean squared error between each predicted confidence $p_k$ and the binary correctness label $o_k \in \{0,1\}$:
\[
\text{BS} = \frac{1}{N} \sum_{k=1}^{N} (p_k - o_k)^2
\]
It reflects both calibration and the ability to assign informative probabilities, penalizing estimators that are overconfident, underconfident, or stuck near $0.5$ regardless of correctness. Lower values reflect higher overall reliability.

\subsection{Discrimination Metrics}
\label{app:metrics_discrimination}

The first two discrimination metrics (Accuracy and F1) are threshold-based and require binarizing the confidence score. We use a fixed default threshold of $0.5$ on the probe's sigmoid output for every method, with no per-method or per-dataset threshold tuning. This avoids any test-set-derived threshold selection and keeps the comparison protocol uniform across methods. The remaining two metrics (AUCPR and AUROC) summarize performance across all thresholds and require no such selection. Model selection during training is governed by a separate composite validation score described in Appendix~\ref{app:validation} and is not used to tune any test-time threshold.

\subsubsection{Accuracy (ACC)}
Accuracy measures the proportion of samples for which the binarized confidence prediction agrees with the ground-truth correctness label:
\[
\text{ACC} = \frac{\text{TP} + \text{TN}}{\text{TP} + \text{FP} + \text{FN} + \text{TN}}
\]
Here, the positive class is a correct LVLM response and the negative class is an incorrect one. We report Accuracy to contextualize the difficulty of each evaluation setting and to enable direct comparison with methods that report threshold-based performance.

\subsubsection{F1 Score (F1)}
F1 is the harmonic mean of precision ($\text{TP}/(\text{TP}+\text{FP})$) and recall ($\text{TP}/(\text{TP}+\text{FN})$), again with the positive class defined as a correct LVLM response:
\[
\text{F1} = \frac{2 \cdot \text{Precision} \cdot \text{Recall}}{\text{Precision} + \text{Recall}}
\]
Relative to Accuracy, F1 is more informative when the correct/incorrect class distribution is skewed, since it weights the joint quality of identifying correct predictions and avoiding false alarms on incorrect ones, while ignoring true negatives entirely.

\subsubsection{Area Under the Precision--Recall Curve (AUCPR)}
AUCPR summarizes precision against recall across all confidence thresholds, with the positive class again defined as a correct LVLM response. Relative to AUROC, it places greater weight on performance over the positive class, making it more informative when the class distribution is skewed.

\subsubsection{Area Under the ROC Curve (AUROC)}
AUROC measures the probability that a randomly drawn correct prediction receives a higher confidence score than a randomly drawn incorrect one, $P(s^{+} > s^{-})$, where $s^{+}$ and $s^{-}$ denote scores assigned to correct and incorrect predictions respectively. It is threshold-independent and relatively insensitive to class prevalence compared with threshold-based metrics. A score of $1.0$ reflects perfect separation; $0.5$ corresponds to chance.

%% file: sections/appendix_validation.tex
\section{Validation Monitoring and Model Selection}
\label{app:validation}

Training the methods in our benchmark raises a non-trivial model-selection question, since confidence estimation is a two-objective problem: a confidence score must \emph{discriminate} correct from incorrect answers and must also be \emph{calibrated} in an absolute sense. A training run that minimizes a cross-entropy loss does not directly favour either property and can drift between them from epoch to epoch. To make checkpoint selection principled and consistent across methods, we monitor a single composite validation score throughout training and use it as the unifying signal for model selection across every trainable method in our benchmark. All five trainable methods (SAPLMA, P(I~Know), InternalInspector, CCPS, and our proposed \texttt{BICR}) use this score for early stopping with a patience of $20$ validation steps applied uniformly across all five. Two of these (P(I~Know) and \texttt{BICR}) additionally use the same score as the Optuna optimization objective; see Appendix~\ref{app:optuna}. Across all five methods, training uses positive-class reweighted binary cross-entropy in which the positive label corresponds to a correct LVLM response. Class imbalance is handled by setting $w_{+} = n_{-}/n_{+}$, where $n_{+}$ and $n_{-}$ are the counts of correct and incorrect samples in the training split respectively, so that the loss contribution from each class is balanced regardless of which class is the majority.

Following the protocol established by \citet{rmcb}, the composite validation score is a convex combination of AUROC and $(1-\text{ECE})$:
\begin{equation}
\text{CompositeScore}
=
\alpha \cdot \text{AUROC}
+
(1-\alpha)\cdot(1-\text{ECE})
\end{equation}

and we fix $\alpha = 0.6$ across every trainable method in the benchmark. We adopt this weighting for the same reason articulated by the original work: ranking correctness reliably is the primary practical requirement of a confidence estimator in deployment, so a slight preference for AUROC over calibration error is warranted, but a score that is discriminatively strong yet grossly miscalibrated cannot be interpreted as a probability and is therefore of limited utility. The $0.6/0.4$ split penalizes miscalibration sharply enough to discourage degenerate solutions that collapse onto a narrow confidence distribution while still allowing ranking quality to break ties between otherwise comparable checkpoints.

\paragraph{Use in training.} For all five trainable methods, we evaluate the composite score on the validation split at every validation step and retain the checkpoint with the highest composite score as the final model. The early-stopping patience of $20$ validation steps means training terminates when the composite has not improved for $20$ consecutive checks. For the two Optuna-tuned methods (P(I~Know) and \texttt{BICR}), the composite score at the best epoch of each trial is returned as the trial's objective value, so both per-trial checkpoint selection and cross-trial hyperparameter selection are driven by the same quantity.

%% file: sections/appendix_optuna.tex
\texttt{BICR}\section{Hyperparameter Search with Optuna}
\label{app:optuna}

Of the seven baselines documented in Appendix~\ref{app:baselines}, three are inference-only and therefore not subject to training (P(True), Self-Probing, Prompt Ensembles), and three further methods (SAPLMA, InternalInspector, CCPS) are trained with the exact architectures and hyperparameters prescribed by their original publications. Our proposed method \texttt{BICR} and one baseline, P(I~Know), are trained with an Optuna hyperparameter search. This appendix describes the search protocol shared by both methods and the search space each one explores.

\subsection{Optimization Protocol}
\label{app:optuna_protocol}

We use the \texttt{Optuna} framework \citep{optuna} with a Tree-structured Parzen Estimator (TPE) sampler, seeded by the same random seed that drives the training data pipeline so that search behaviour is reproducible. The protocol below applies identically to \texttt{BICR} and P(I~Know).

Each (method, LVLM, seed) tuple is optimized for $50$ trials, with five independent seeds $\{23,\,42,\,137,\,2024,\,3407\}$ run per (method, LVLM) pair, and all downstream evaluation metrics reported as the mean across seeds. Each trial trains for at most $200$ epochs at batch size $32$, with early stopping on the composite validation score (Appendix~\ref{app:validation}) at patience $20$. The trial's objective value is the composite score at its best epoch, and the trial with the highest objective is selected as the final configuration for that (LVLM, seed). To accelerate the search, Optuna's Median Pruner is applied with $5$ start-up trials, $10$ warm-up steps, and an interval of $5$ steps, terminating unpromising trials early on the same intermediate composite score. To prevent model capacity from being conflated with raw parameter count when comparing architectural efficiency, all configurations are constrained to a maximum of $5{,}000{,}000$ trainable parameters; any trial suggesting a model outside this budget is pruned before training.

\subsection{Hyperparameter Search Space}
\label{app:optuna_space}

The search spaces for \texttt{BICR} and P(I~Know) are summarized in Table~\ref{tab:hparams}. Both methods tune the MLP classifier architecture (depth, width, dropout) and optimizer settings (learning rate, weight decay). \texttt{BICR} additionally tunes three loss-coefficient hyperparameters that control the auxiliary training objectives coupling its two extraction views ($\mathbf{h}_{\mathrm{base}}$, $\mathbf{h}_{\mathrm{blank}}$): $\beta$ (weight on the Brier calibration term $\mathcal{L}_{\mathrm{brier}}$ on $\mathbf{h}_{\mathrm{base}}$), $\lambda$ (weight on the visual-grounding ranking loss $\mathcal{L}_{\mathrm{rank}}$ that contrasts $\mathbf{h}_{\mathrm{base}}$ against $\mathbf{h}_{\mathrm{blank}}$), and $\gamma$ (the ranking-loss margin in probability space; see Eq.~\ref{eq:rank} in \S\ref{sec:method}). The fixed-architecture methods (SAPLMA, InternalInspector, CCPS) are not listed in the table, since their architectures and hyperparameters are taken verbatim from the original publications and are documented in Appendix~\ref{app:baselines}.

\begin{table*}[t]
\centering
\caption{Hyperparameter search space for the Optuna-tuned methods. Both methods share the classifier-architecture and optimizer search space; \texttt{BICR} additionally tunes its three loss coefficients. Notation \texttt{a,b} inside a set denotes a single categorical choice corresponding to an MLP with hidden widths $(a, b)$.}
\label{tab:hparams}
\footnotesize
\begin{tabular}{@{}l p{0.72\textwidth}@{}}
\toprule
\textbf{Component} & \textbf{Search space} \\
\midrule
\multicolumn{2}{@{}l}{\textbf{Shared (BICR and P(I~Know))}} \\
\addlinespace[2pt]
\texttt{classifier\_layers} & $\{\texttt{0};\; \texttt{256};\; \texttt{512};\; \texttt{128,64};\; \texttt{256,128};\; \texttt{512,256};\; \texttt{1024,512};\; \texttt{1024,512,256}\}$ \\
\texttt{classifier\_dropout} & $\{0.0,\; 0.1,\; 0.3,\; 0.5\}$ \\
\texttt{learning\_rate} & $[10^{-5},\; 10^{-3}]$, log-uniform \\
\texttt{weight\_decay} & $[10^{-6},\; 10^{-3}]$, log-uniform \\
\midrule
\multicolumn{2}{@{}l}{\textbf{BICR-only (ours)}} \\
\addlinespace[2pt]
$\beta$ (Brier weight) & $[0.0,\; 0.5]$, uniform \\
$\lambda$ (rank weight) & $[0.01,\; 0.3]$, uniform \\
$\gamma$ (margin) & $[0.05,\; 0.25]$, uniform \\
\bottomrule
\end{tabular}
\end{table*}

\paragraph{Note on the loss-coefficient search.} The four shared rows in Table~\ref{tab:hparams} cover the classifier architecture (depth, width, dropout) and optimizer settings, and are identical between \texttt{BICR} and P(I~Know) so that the two methods compete on the same architectural and optimization footing. \texttt{BICR} alone tunes the three loss-coefficient hyperparameters in the second row group ($\beta$, $\lambda$, $\gamma$), since these control the auxiliary objectives that couple its two extraction views and have no analogue in P(I~Know)'s BCE-only training. The analysis of the Optuna-selected values for these three coefficients across LVLMs and seeds is provided in Appendix~\ref{app:design_hparams}.

%% file: sections/appendix_params.tex
\section{Analysis of Additional Trainable Parameters}
\label{app:params}

This appendix quantifies and compares the \emph{additional} learnable parameters introduced by each evaluated confidence estimation method, including our proposed \texttt{BICR}, when applied to a frozen base LVLM. We first report the architectural dimensions of the LVLMs that drive the parameter counts of the linear-probe methods (\S\ref{app:params_dims}), then provide the formulas for the additional trainable parameters of each method (\S\ref{app:params_formulas}), and finally report the exact parameter counts used in our experiments (\S\ref{app:params_exact}), followed by a short discussion (\S\ref{app:params_discussion}). All counts are of \emph{trainable} parameters (i.e., parameters returned by \texttt{nn.Module.parameters()} with \texttt{requires\_grad=True}); batch-normalization running statistics and other registered buffers are excluded. All counts include biases unless otherwise noted.

\subsection{Base LVLM Architectural Parameters}
\label{app:params_dims}

The key architectural dimensions of the base LVLMs used in this study that influence the number of trainable parameters of probe-style methods are summarized in Table~\ref{tab:app_model_dims}: the language-model hidden size $d_h$ and the number of decoder layers $L$. We also list the total number of parameters of each LVLM (read directly from its \texttt{model.safetensors.index.json}); the percentages reported in \S\ref{app:params_exact} use this column as denominator. The full model cards (vision-encoder details, number of attention heads, etc.) are given in Appendix~\ref{app:lvlms}.

\begin{table*}[t]
\centering
\small
\caption{Architectural dimensions and total parameter counts of the base LVLMs used in this study. The total-parameters column is obtained by summing the product of tensor shape dimensions over every tensor listed in \texttt{model.safetensors.index.json} and is used as the denominator for the \% column in Table~\ref{tab:app_exact_param_counts}.}
\label{tab:app_model_dims}
\begin{tabular}{lrrr}
\toprule
\textbf{Base LVLM} & \textbf{$d_h$} & \textbf{$L$} & \textbf{Total parameters} \\
\midrule
\texttt{Qwen/Qwen3-VL-8B-Instruct}           & 4{,}096 & 36 & $8{,}767{,}123{,}696$ \,(8.77\,B) \\
\texttt{llava-hf/llava-v1.6-vicuna-13b-hf}   & 5{,}120 & 40 & $13{,}351{,}499{,}776$ \,(13.35\,B) \\
\texttt{OpenGVLab/InternVL3\_5-14B-HF}       & 5{,}120 & 40 & $15{,}119{,}523{,}840$ \,(15.12\,B) \\
\texttt{google/gemma-3-27b-it}               & 5{,}376 & 62 & $27{,}432{,}406{,}640$ \,(27.43\,B) \\
\texttt{deepseek-ai/deepseek-vl2}            & 2{,}560 & 30 & $27{,}480{,}134{,}248$ \,(27.48\,B) \\
\bottomrule
\end{tabular}
\end{table*}

\subsection{Formulation of Additional Trainable Parameters}
\label{app:params_formulas}

Table~\ref{tab:app_param_formulas} lists the trainable components and the closed-form parameter expressions for each method. For InternalInspector and CCPS the count is independent of the base LVLM, so no formula in $d_h$ or $L$ is needed. For the Optuna-tuned methods (P(I~Know) and \texttt{BICR}), the selected depth and widths of the classifier head differ per (LVLM, seed); we write $(H_1, \ldots, H_k)$ for the tuple of hidden widths that Optuna selects for a given run.

\begin{table*}[t]
\centering
\footnotesize
\caption{Formulas for additional trainable parameters introduced by each method. $d_h$ is the LVLM hidden size from Table~\ref{tab:app_model_dims}; $(H_1, \ldots, H_k)$ is the Optuna-selected tuple of hidden widths of the classifier head for a given run.}
\label{tab:app_param_formulas}
\begin{tabular}{p{2.2cm} p{5.0cm} p{5.2cm}}
\toprule
\textbf{Method} & \textbf{Trainable component(s)} & \textbf{Formula (incl.\ biases)} \\
\midrule
P(True) & None (prompting only) & $0$ \\
Self-Probing & None (prompting only) & $0$ \\
Prompt Ensembles & None (inference only) & $0$ \\
\midrule
SAPLMA & MLP $d_h \to 256 \to 128 \to 64 \to 1$ & $256\,d_h + 41{,}473$ \\
\addlinespace[2pt]
InternalInspector & ResNet18-style CNN encoder (3-channel input) + projection $512 \to 128$ + MLP $128 \to 256 \to 128 \to 64 \to 1$ & $11{,}316{,}417$ \;(independent of $d_h$, $L$) \\
\addlinespace[2pt]
CCPS & \textbf{Stage 1:} Conv1d$(75{\to}64, k{=}3)$ + Conv1d$(64{\to}32, k{=}3)$ + Linear$(32{\to}16)$ \newline \textbf{Stage 2:} encoder fine-tuned jointly + Linear$(16{\to}32)$ + Linear$(32{\to}2)$ & $21{,}778$ \;(independent of $d_h$, $L$; encoder embedded in classifier) \\
\midrule
P(I~Know) & MLP $d_h \to H_1 \to \cdots \to H_k \to 1$ & $d_h H_1 + H_1 + \sum_{i=1}^{k-1}\big(H_i H_{i+1} + H_{i+1}\big) + (H_k + 1)$ \\
\addlinespace[2pt]
\texttt{BICR} (ours) & MLP $d_h \to H_1 \to \cdots \to H_k \to 1$ \newline (shared across base and blank views; blank view used only at training time via $\mathcal{L}_{\mathrm{rank}}$) & $d_h H_1 + H_1 + \sum_{i=1}^{k-1}\big(H_i H_{i+1} + H_{i+1}\big) + (H_k + 1)$ \\
\bottomrule
\end{tabular}
\end{table*}

\paragraph{Note on InternalInspector.} The ResNet18 encoder used by InternalInspector takes a 3-channel input corresponding to the stacked (activation, attention, feed-forward) states at each layer. The spatial dimensions of its input are $(L, d_h)$, but because every downstream layer is followed by adaptive average pooling to a fixed $(1,1)$ spatial footprint before the $512 \to 128$ projection, the total trainable-parameter count is independent of both $L$ and $d_h$.

\paragraph{Note on CCPS.} CCPS uses a two-stage training pipeline: a contrastive encoder is first pre-trained with $21{,}168$ parameters (Stage~1), and then a $610$-parameter classification head ($\text{Linear}(16{\to}32)+\text{Linear}(32{\to}2)$) is attached on top of the encoder, with the entire stack fine-tuned end-to-end (Stage~2). The final deployed model is the Stage~2 classifier checkpoint, which contains all encoder weights plus the classification head, totalling $21{,}778$ parameters. The head operates on a per-token sequence of $75$ trajectory features computed from the frozen LVLM's hidden states, and no part of the dimension scales with $d_h$; the parameter count is therefore identical across all LVLMs.

\paragraph{Note on \texttt{BICR}.} \texttt{BICR} uses two views: the base hidden state $\mathbf{h}_{\mathrm{base}}$ and the blank-image hidden state $\mathbf{h}_{\mathrm{blank}}$. Both views pass through the \emph{same} MLP at training time, but only $\mathbf{h}_{\mathrm{base}}$ is processed at inference. The blank view serves exclusively as a training-time regularizer through the ranking loss $\mathcal{L}_{\mathrm{rank}}$, adding zero parameters and zero inference cost beyond the standard MLP probe. The parameter formula is therefore identical to that of P(I~Know); the difference between the two methods lies entirely in the training objective, not the architecture.

\subsection{Exact Additional Trainable Parameter Counts}
\label{app:params_exact}

Table~\ref{tab:app_exact_param_counts} reports the exact number of additional trainable parameters introduced by each method when applied to each base LVLM, per benchmark seed. Each cell reports the trainable parameter count of the saved checkpoint and, in parentheses, the count expressed as a percentage of the total parameter count of the corresponding base LVLM (Table~\ref{tab:app_model_dims}).

The fixed-architecture methods (SAPLMA, InternalInspector, CCPS) have a single count per (method, LVLM); we still tabulate all five seeds so that readers can verify the per-seed determinism of the reconstruction, and so that the layout of the table is uniform across methods. The two Optuna-tuned methods (P(I~Know) and \texttt{BICR}) re-run the architecture search independently for each (method, LVLM, seed) tuple, as documented in Appendix~\ref{app:optuna}, and consequently the selected \texttt{classifier\_layers} (and therefore the trainable-parameter count) can differ across seeds; for these two methods the cell additionally reports the Optuna-selected \texttt{classifier\_layers} tuple in light text. The bottom \textsc{mean} row of each method's block reports the mean over the five seeds, of both the parameter count and the percentage of the base-LVLM's parameters.

\input{tables/params_table}

\subsection{Discussion}
\label{app:params_discussion}

Table~\ref{tab:app_exact_param_counts} makes the capacity of each benchmarked confidence estimator transparent and directly comparable. Three qualitative regimes are apparent. First, the three prompt-based methods (P(True), Self-Probing, Prompt Ensembles) introduce no trainable parameters at all and rely entirely on the frozen LVLM's own output behaviour. Second, CCPS occupies an extreme position at the other end of the spectrum of trained methods: its fixed two-stage convolutional head totals only $21{,}778$ parameters regardless of the base LVLM (well under $0.001\%$ of every backbone), because its input is already a compact $75$-channel trajectory-feature sequence rather than the high-dimensional LVLM hidden state. Third, the remaining trainable methods (SAPLMA, InternalInspector, P(I~Know), and \texttt{BICR}) occupy a moderate regime of a few hundred thousand to roughly eleven million trainable parameters, with the exact count driven by the LVLM hidden size $d_h$ (SAPLMA, P(I~Know), \texttt{BICR}) or by the fixed convolutional encoder alone (InternalInspector).

Within the moderate regime, InternalInspector is consistently the heaviest at $11{,}316{,}417$ parameters on every LVLM, ranging from $0.04\%$ of Gemma-3-27B and DeepSeek-VL2 up to $0.13\%$ of Qwen3-VL-8B, because its ResNet18-style CNN encoder dominates the count. \texttt{BICR} is substantially lighter: averaging the parameter counts across the five seeds, \texttt{BICR}'s MLP head ranges from $642{,}561$ parameters on Qwen3-VL-8B to $2{,}753{,}537$ on InternVL3.5-14B, which translates to a $4.1\times$ to $17.6\times$ reduction relative to InternalInspector across the five LVLMs (the smallest gap appearing on InternVL3.5-14B and the largest on Qwen3-VL-8B). Compared to its architectural counterpart P(I~Know), which uses an identical MLP formula but tends toward larger Optuna-selected widths, \texttt{BICR} is lighter on four of the five LVLMs (notably $4.7\times$ lighter on Qwen3-VL-8B, where Optuna selects $(128,64)$ for \texttt{BICR} on four of five seeds while selecting $(1024,512)$ or larger for P(I~Know) on the highest-parameter seeds), and is slightly heavier than P(I~Know) only on InternVL3.5-14B, where Optuna selects $(512,256)$ for \texttt{BICR} on every seed while P(I~Know) tends toward narrower $(512)$ heads. SAPLMA is the only method that is consistently lighter than \texttt{BICR}, at $0.7$M--$1.4$M parameters depending on $d_h$, owing to its fixed shallow architecture. The pattern across \texttt{BICR}'s selections is consistent with its design: the training-time ranking loss $\mathcal{L}_{\mathrm{rank}}$ provides a stronger learning signal than BCE alone, allowing smaller architectures to reach competitive performance, so the capacity advantage comes from the training objective rather than from the model size.

A separate point worth flagging is that \texttt{BICR}'s blank view contributes zero parameters to the deployed model, since only $\mathbf{h}_{\mathrm{base}}$ is processed at inference; the blank-view hidden state is consumed exclusively at training time by $\mathcal{L}_{\mathrm{rank}}$. \texttt{BICR} is therefore a strictly parameter-equivalent alternative to single-view probes like P(I~Know) at deployment, with the performance improvement coming entirely from the training signal.

Taken as a whole, every trainable method in our benchmark adds at most $\approx 0.13\%$ of the smallest base LVLM's parameter count, supporting our framing of these confidence estimators as genuinely lightweight additions to a frozen LVLM rather than as meaningful contributions to the deployed model's size or inference cost.

%% file: tables/params_table.tex
\begin{table*}[t]
\centering
\scriptsize
\setlength{\tabcolsep}{3pt}
\caption{Per-seed trainable-parameter counts and percentage of total base-LVLM parameters for every confidence-estimation method in our benchmark. Each cell reports the trainable parameter count of the saved checkpoint and, in parentheses, the count as a percentage of the total parameter count of the corresponding base LVLM. Percentages are rounded to two decimals; cells displayed as $0.00\%$ are non-zero but below $0.005\%$. SAPLMA, InternalInspector, and CCPS use fixed architectures, so their counts are identical across the five benchmark seeds $\{23, 42, 137, 2024, 3407\}$; for the Optuna-tuned methods (P(I~Know) and \texttt{BICR}), counts vary across seeds and the cell additionally reports the Optuna-selected \texttt{classifier\_layers} tuple. The bottom \textsc{mean} row of each method's block is the across-seed mean of both the parameter count and the percentage.}
\label{tab:app_exact_param_counts}
\resizebox{\textwidth}{!}{%
\begin{tabular}{l l r r r r r}
\toprule
\textbf{Method} & \textbf{Seed} & \textbf{Qwen3-VL-8B} & \textbf{LLaVA-Next-13B} & \textbf{InternVL3.5-14B} & \textbf{Gemma-3-27B} & \textbf{DeepSeek-VL2} \\
 & \textbf{(VLM total)} & \textit{8.77\,B} & \textit{13.35\,B} & \textit{15.12\,B} & \textit{27.43\,B} & \textit{27.48\,B} \\
\midrule
\multicolumn{7}{l}{\textit{Prompt-only methods (no trainable parameters)}} \\
P(True) & --- & $0$ & $0$ & $0$ & $0$ & $0$ \\
Self-Probing & --- & $0$ & $0$ & $0$ & $0$ & $0$ \\
Prompt Ensembles & --- & $0$ & $0$ & $0$ & $0$ & $0$ \\
\midrule
\multirow{6}{*}{\textbf{SAPLMA}} & 23 & $1{,}090{,}049$\,\tiny{(0.01\%)} & $1{,}352{,}193$\,\tiny{(0.01\%)} & $1{,}352{,}193$\,\tiny{(0.01\%)} & $1{,}417{,}729$\,\tiny{(0.01\%)} & $696{,}833$\,\tiny{(0.00\%)} \\
 & 42 & $1{,}090{,}049$\,\tiny{(0.01\%)} & $1{,}352{,}193$\,\tiny{(0.01\%)} & $1{,}352{,}193$\,\tiny{(0.01\%)} & $1{,}417{,}729$\,\tiny{(0.01\%)} & $696{,}833$\,\tiny{(0.00\%)} \\
 & 137 & $1{,}090{,}049$\,\tiny{(0.01\%)} & $1{,}352{,}193$\,\tiny{(0.01\%)} & $1{,}352{,}193$\,\tiny{(0.01\%)} & $1{,}417{,}729$\,\tiny{(0.01\%)} & $696{,}833$\,\tiny{(0.00\%)} \\
 & 2024 & $1{,}090{,}049$\,\tiny{(0.01\%)} & $1{,}352{,}193$\,\tiny{(0.01\%)} & $1{,}352{,}193$\,\tiny{(0.01\%)} & $1{,}417{,}729$\,\tiny{(0.01\%)} & $696{,}833$\,\tiny{(0.00\%)} \\
 & 3407 & $1{,}090{,}049$\,\tiny{(0.01\%)} & $1{,}352{,}193$\,\tiny{(0.01\%)} & $1{,}352{,}193$\,\tiny{(0.01\%)} & $1{,}417{,}729$\,\tiny{(0.01\%)} & $696{,}833$\,\tiny{(0.00\%)} \\
 & \textsc{mean} & $\mathbf{1{,}090{,}049}$\,\tiny{(\textbf{0.01\%})} & $\mathbf{1{,}352{,}193}$\,\tiny{(\textbf{0.01\%})} & $\mathbf{1{,}352{,}193}$\,\tiny{(\textbf{0.01\%})} & $\mathbf{1{,}417{,}729}$\,\tiny{(\textbf{0.01\%})} & $\mathbf{696{,}833}$\,\tiny{(\textbf{0.00\%})} \\
\midrule
\multirow{6}{*}{\textbf{InternalInspector}} & 23 & $11{,}316{,}417$\,\tiny{(0.13\%)} & $11{,}316{,}417$\,\tiny{(0.08\%)} & $11{,}316{,}417$\,\tiny{(0.07\%)} & $11{,}316{,}417$\,\tiny{(0.04\%)} & $11{,}316{,}417$\,\tiny{(0.04\%)} \\
 & 42 & $11{,}316{,}417$\,\tiny{(0.13\%)} & $11{,}316{,}417$\,\tiny{(0.08\%)} & $11{,}316{,}417$\,\tiny{(0.07\%)} & $11{,}316{,}417$\,\tiny{(0.04\%)} & $11{,}316{,}417$\,\tiny{(0.04\%)} \\
 & 137 & $11{,}316{,}417$\,\tiny{(0.13\%)} & $11{,}316{,}417$\,\tiny{(0.08\%)} & $11{,}316{,}417$\,\tiny{(0.07\%)} & $11{,}316{,}417$\,\tiny{(0.04\%)} & $11{,}316{,}417$\,\tiny{(0.04\%)} \\
 & 2024 & $11{,}316{,}417$\,\tiny{(0.13\%)} & $11{,}316{,}417$\,\tiny{(0.08\%)} & $11{,}316{,}417$\,\tiny{(0.07\%)} & $11{,}316{,}417$\,\tiny{(0.04\%)} & $11{,}316{,}417$\,\tiny{(0.04\%)} \\
 & 3407 & $11{,}316{,}417$\,\tiny{(0.13\%)} & $11{,}316{,}417$\,\tiny{(0.08\%)} & $11{,}316{,}417$\,\tiny{(0.07\%)} & $11{,}316{,}417$\,\tiny{(0.04\%)} & $11{,}316{,}417$\,\tiny{(0.04\%)} \\
 & \textsc{mean} & $\mathbf{11{,}316{,}417}$\,\tiny{(\textbf{0.13\%})} & $\mathbf{11{,}316{,}417}$\,\tiny{(\textbf{0.08\%})} & $\mathbf{11{,}316{,}417}$\,\tiny{(\textbf{0.07\%})} & $\mathbf{11{,}316{,}417}$\,\tiny{(\textbf{0.04\%})} & $\mathbf{11{,}316{,}417}$\,\tiny{(\textbf{0.04\%})} \\
\midrule
\multirow{6}{*}{\textbf{CCPS}} & 23 & $21{,}778$\,\tiny{(0.00\%)} & $21{,}778$\,\tiny{(0.00\%)} & $21{,}778$\,\tiny{(0.00\%)} & $21{,}778$\,\tiny{(0.00\%)} & $21{,}778$\,\tiny{(0.00\%)} \\
 & 42 & $21{,}778$\,\tiny{(0.00\%)} & $21{,}778$\,\tiny{(0.00\%)} & $21{,}778$\,\tiny{(0.00\%)} & $21{,}778$\,\tiny{(0.00\%)} & $21{,}778$\,\tiny{(0.00\%)} \\
 & 137 & $21{,}778$\,\tiny{(0.00\%)} & $21{,}778$\,\tiny{(0.00\%)} & $21{,}778$\,\tiny{(0.00\%)} & $21{,}778$\,\tiny{(0.00\%)} & $21{,}778$\,\tiny{(0.00\%)} \\
 & 2024 & $21{,}778$\,\tiny{(0.00\%)} & $21{,}778$\,\tiny{(0.00\%)} & $21{,}778$\,\tiny{(0.00\%)} & $21{,}778$\,\tiny{(0.00\%)} & $21{,}778$\,\tiny{(0.00\%)} \\
 & 3407 & $21{,}778$\,\tiny{(0.00\%)} & $21{,}778$\,\tiny{(0.00\%)} & $21{,}778$\,\tiny{(0.00\%)} & $21{,}778$\,\tiny{(0.00\%)} & $21{,}778$\,\tiny{(0.00\%)} \\
 & \textsc{mean} & $\mathbf{21{,}778}$\,\tiny{(\textbf{0.00\%})} & $\mathbf{21{,}778}$\,\tiny{(\textbf{0.00\%})} & $\mathbf{21{,}778}$\,\tiny{(\textbf{0.00\%})} & $\mathbf{21{,}778}$\,\tiny{(\textbf{0.00\%})} & $\mathbf{21{,}778}$\,\tiny{(\textbf{0.00\%})} \\
\midrule
\multirow{6}{*}{\textbf{P(I~Know)}} & 23 & $4{,}720{,}641$\,\tiny{(0.05\%; 1024,512)} & $2{,}622{,}465$\,\tiny{(0.02\%; 512)} & $2{,}622{,}465$\,\tiny{(0.02\%; 512)} & $2{,}753{,}537$\,\tiny{(0.01\%; 512)} & $1{,}311{,}745$\,\tiny{(0.00\%; 512)} \\
 & 42 & $1{,}081{,}857$\,\tiny{(0.01\%; 256,128)} & $2{,}622{,}465$\,\tiny{(0.02\%; 512)} & $2{,}622{,}465$\,\tiny{(0.02\%; 512)} & $1{,}376{,}769$\,\tiny{(0.01\%; 256)} & $1{,}311{,}745$\,\tiny{(0.00\%; 512)} \\
 & 137 & $2{,}229{,}249$\,\tiny{(0.03\%; 512,256)} & $2{,}753{,}537$\,\tiny{(0.02\%; 512,256)} & $1{,}311{,}233$\,\tiny{(0.01\%; 256)} & $1{,}409{,}537$\,\tiny{(0.01\%; 256,128)} & $1{,}442{,}817$\,\tiny{(0.01\%; 512,256)} \\
 & 2024 & $2{,}098{,}177$\,\tiny{(0.02\%; 512)} & $1{,}311{,}233$\,\tiny{(0.01\%; 256)} & $2{,}753{,}537$\,\tiny{(0.02\%; 512,256)} & $2{,}753{,}537$\,\tiny{(0.01\%; 512)} & $1{,}442{,}817$\,\tiny{(0.01\%; 512,256)} \\
 & 3407 & $4{,}851{,}713$\,\tiny{(0.06\%; 1024,512,256)} & $2{,}622{,}465$\,\tiny{(0.02\%; 512)} & $2{,}622{,}465$\,\tiny{(0.02\%; 512)} & $2{,}753{,}537$\,\tiny{(0.01\%; 512)} & $3{,}278{,}849$\,\tiny{(0.01\%; 1024,512,256)} \\
 & \textsc{mean} & $\mathbf{2{,}996{,}327}$\,\tiny{(\textbf{0.03\%})} & $\mathbf{2{,}386{,}433}$\,\tiny{(\textbf{0.02\%})} & $\mathbf{2{,}386{,}433}$\,\tiny{(\textbf{0.02\%})} & $\mathbf{2{,}209{,}383}$\,\tiny{(\textbf{0.01\%})} & $\mathbf{1{,}757{,}595}$\,\tiny{(\textbf{0.01\%})} \\
\midrule
\multirow{6}{*}{\textbf{\texttt{BICR (ours)}}} & 23 & $532{,}737$\,\tiny{(0.01\%; 128,64)} & $2{,}753{,}537$\,\tiny{(0.02\%; 512,256)} & $2{,}753{,}537$\,\tiny{(0.02\%; 512,256)} & $1{,}409{,}537$\,\tiny{(0.01\%; 256,128)} & $1{,}442{,}817$\,\tiny{(0.01\%; 512,256)} \\
 & 42 & $1{,}081{,}857$\,\tiny{(0.01\%; 256,128)} & $1{,}344{,}001$\,\tiny{(0.01\%; 256,128)} & $2{,}753{,}537$\,\tiny{(0.02\%; 512,256)} & $1{,}409{,}537$\,\tiny{(0.01\%; 256,128)} & $1{,}442{,}817$\,\tiny{(0.01\%; 512,256)} \\
 & 137 & $532{,}737$\,\tiny{(0.01\%; 128,64)} & $1{,}344{,}001$\,\tiny{(0.01\%; 256,128)} & $2{,}753{,}537$\,\tiny{(0.02\%; 512,256)} & $696{,}577$\,\tiny{(0.00\%; 128,64)} & $1{,}442{,}817$\,\tiny{(0.01\%; 512,256)} \\
 & 2024 & $532{,}737$\,\tiny{(0.01\%; 128,64)} & $663{,}809$\,\tiny{(0.00\%; 128,64)} & $2{,}753{,}537$\,\tiny{(0.02\%; 512,256)} & $1{,}409{,}537$\,\tiny{(0.01\%; 256,128)} & $1{,}442{,}817$\,\tiny{(0.01\%; 512,256)} \\
 & 3407 & $532{,}737$\,\tiny{(0.01\%; 128,64)} & $663{,}809$\,\tiny{(0.00\%; 128,64)} & $2{,}753{,}537$\,\tiny{(0.02\%; 512,256)} & $2{,}884{,}609$\,\tiny{(0.01\%; 512,256)} & $1{,}442{,}817$\,\tiny{(0.01\%; 512,256)} \\
 & \textsc{mean} & $\mathbf{642{,}561}$\,\tiny{(\textbf{0.01\%})} & $\mathbf{1{,}353{,}831}$\,\tiny{(\textbf{0.01\%})} & $\mathbf{2{,}753{,}537}$\,\tiny{(\textbf{0.02\%})} & $\mathbf{1{,}561{,}959}$\,\tiny{(\textbf{0.01\%})} & $\mathbf{1{,}442{,}817}$\,\tiny{(\textbf{0.01\%})} \\
\bottomrule
\end{tabular}}
\end{table*}

%% file: sections/appendix_design.tex
\section{Design Validation and Ablation Analysis}
\label{app:design}

This appendix presents the empirical evidence underlying each design choice in \texttt{BICR}. All experiments follow the benchmark evaluation protocol documented in Appendix~\ref{app:optuna}: 5~LVLMs $\times$ 5~seeds $\times$ 50~Optuna trials per (LVLM, seed) tuple, with metrics averaged across seeds (per-LVLM) or across LVLMs (cross-LVLM). Statistical significance is assessed via paired Wilcoxon signed-rank tests over the 25 (LVLM, seed) observations.

\subsection{Loss Component Ablation}
\label{app:design_loss}

\texttt{BICR}'s training objective (Eq.~\ref{eq:total_loss}) consists of three terms: $\mathcal{L}_{\mathrm{bce}}$, $\mathcal{L}_{\mathrm{brier}}$, and $\mathcal{L}_{\mathrm{rank}}$. We ablate each by removing it while keeping all other components unchanged, yielding four configurations: the full \texttt{BICR} model, $-\mathcal{L}_{\mathrm{brier}}$ (removing the calibration term), $-\mathcal{L}_{\mathrm{rank}}$ (removing the visual grounding ranking; note that this also removes the need for $\mathbf{h}_{\mathrm{blank}}$ entirely, as the blank view is only used by $\mathcal{L}_{\mathrm{rank}}$), and $\mathcal{L}_{\mathrm{bce}}$-only (removing both auxiliary terms).

\paragraph{Cross-LVLM results.} Table~\ref{tab:loss_ablation} reports the cross-LVLM average metrics for each ablation variant.

\begin{table*}[ht]
\centering
\small
\caption{Loss ablation for \texttt{BICR}, averaged across 5~LVLMs $\times$ 5~seeds. Each row removes one or more loss components. $\Delta$AUROC is the change relative to the full model. Best values in \textbf{bold}.}
\label{tab:loss_ablation}
\begin{tabular}{l c c c c r}
\toprule
\textbf{Variant} & \textbf{ECE}$\downarrow$ & \textbf{BS}$\downarrow$ & \textbf{AUCPR}$\uparrow$ & \textbf{AUROC}$\uparrow$ & $\Delta$ \\
\midrule
Full (\texttt{BICR})          & \textbf{7.1} & \textbf{18.4} & \textbf{87.4} & \textbf{78.6} & --- \\
$-\mathcal{L}_{\mathrm{brier}}$  & 8.5 & 19.0 & 87.1 & 78.0 & $-$0.6 \\
$-\mathcal{L}_{\mathrm{rank}}$   & 8.1 & 19.6 & 85.5 & 75.3 & $-$3.3 \\
$\mathcal{L}_{\mathrm{bce}}$ only & 9.2 & 19.9 & 85.5 & 75.2 & $-$3.4 \\
\bottomrule
\end{tabular}
\end{table*}

\paragraph{Key findings.} $\mathcal{L}_{\mathrm{rank}}$ is the critical component: removing it degrades AUROC by $3.3$ points ($p < 0.001$, paired Wilcoxon) and AUCPR by $1.9$ points ($p < 0.001$). This confirms that the blank-image contrastive signal is the primary driver of \texttt{BICR}'s discriminative advantage. $\mathcal{L}_{\mathrm{brier}}$ contributes a more modest calibration benefit on the cross-LVLM average ($\Delta\text{ECE} = -1.4$, $p = 0.059$; $\Delta\text{BS} = -0.6$, $p = 0.019$). Removing both auxiliary losses ($\mathcal{L}_{\mathrm{bce}}$-only) produces the worst configuration on every discrimination metric, with the degradation highly significant ($p < 0.005$).

\paragraph{Statistical significance, pooled aggregation.} Table~\ref{tab:loss_ablation_sig} reports the significance test results for each ablation comparison under cross-LVLM pooled aggregation.

\begin{table*}[ht]
\centering
\small
\caption{Statistical significance of loss ablation (paired Wilcoxon, $n{=}25$). Each cell shows the $p$-value; $^{***}$, $^{**}$, $^{*}$ denote $p < 0.001$, $p < 0.01$, $p < 0.05$; ``n.s.'' denotes $p \geq 0.05$.}
\label{tab:loss_ablation_sig}
\begin{tabular}{l c c c c}
\toprule
\textbf{Comparison} & \textbf{ECE} & \textbf{BS} & \textbf{AUCPR} & \textbf{AUROC} \\
\midrule
Full vs.\ $-\mathcal{L}_{\mathrm{brier}}$  & n.s. & 0.019$^{*}$ & n.s. & n.s. \\
Full vs.\ $-\mathcal{L}_{\mathrm{rank}}$   & n.s. & $<$0.001$^{***}$ & $<$0.001$^{***}$ & $<$0.001$^{***}$ \\
Full vs.\ $\mathcal{L}_{\mathrm{bce}}$ only & 0.004$^{**}$ & $<$0.001$^{***}$ & $<$0.001$^{***}$ & $<$0.001$^{***}$ \\
\bottomrule
\end{tabular}
\end{table*}

\paragraph{Statistical significance, equal-weight aggregation.} Table~\ref{tab:loss_ablation_sig_uw} reports the same Wilcoxon tests under per-dataset equal-weight aggregation. Two findings track the cross-LVLM pooled ablation: removing $\mathcal{L}_{\mathrm{rank}}$ produces highly significant degradation on every metric, and removing both auxiliary losses ($\mathcal{L}_{\mathrm{bce}}$-only) produces significant degradation on calibration metrics. Under this aggregation, removing $\mathcal{L}_{\mathrm{brier}}$ alone is not statistically significant on any metric (all $p > 0.05$), suggesting that the Brier term's contribution is concentrated on the larger datasets that dominate the pooled aggregation rather than on the smaller datasets that get equal weight here. Removing both auxiliary losses also weakens significance on AUCPR (n.s.) and AUROC ($p = 0.045$) compared to the much stronger pooled-aggregation effect, again indicating that auxiliary-loss benefits concentrate on the dominant datasets.

\begin{table*}[ht]
\centering
\small
\caption{Statistical significance of loss ablation under per-dataset equal-weight aggregation (paired Wilcoxon, $n{=}25$). Same conventions as Table~\ref{tab:loss_ablation_sig}.}
\label{tab:loss_ablation_sig_uw}
\begin{tabular}{l c c c c}
\toprule
\textbf{Comparison} & \textbf{ECE} & \textbf{BS} & \textbf{AUCPR} & \textbf{AUROC} \\
\midrule
Full vs.\ $-\mathcal{L}_{\mathrm{brier}}$  & n.s.             & n.s.             & n.s.             & n.s.             \\
Full vs.\ $-\mathcal{L}_{\mathrm{rank}}$   & $<$0.001$^{***}$ & $<$0.001$^{***}$ & 0.001$^{**}$     & $<$0.001$^{***}$ \\
Full vs.\ $\mathcal{L}_{\mathrm{bce}}$ only & $<$0.001$^{***}$ & $<$0.001$^{***}$ & n.s.             & 0.045$^{*}$      \\
\bottomrule
\end{tabular}
\end{table*}

\paragraph{Cluster-aware significance.} The paired Wilcoxon tests above treat each (LVLM, seed) tuple as an independent observation, but the 5 seeds within an LVLM share the same frozen weights and the same test set, so the truly independent unit is the LVLM. To verify the loss-ablation findings under a cluster-aware protocol, Table~\ref{tab:loss_ablation_cluster_aware} reports a cluster bootstrap (10{,}000 resamples) over LVLM-level seed-means with Holm-Bonferroni correction across the 4 metrics. The results sharpen rather than overturn the n=25 conclusions: $\mathcal{L}_{\mathrm{rank}}$'s discriminative contribution remains highly significant on AUCPR, AUROC, and BS ($p < 0.001$ on each), and the $\mathcal{L}_{\mathrm{bce}}$-only comparison shows the same pattern with comparable effect sizes. The contribution of $\mathcal{L}_{\mathrm{brier}}$ alone is the only place the picture softens: its ablation does not reach Holm-corrected significance under cluster-aware testing, consistent with the n=25 unweighted-aggregation result that $\mathcal{L}_{\mathrm{brier}}$'s benefit concentrates on specific datasets rather than as a uniform across-LVLM effect. ECE comparisons are also non-significant under this conservative protocol because the across-LVLM ECE distribution is heavy-tailed (Table~\ref{tab:loss_ablation_pervlm}), but the corresponding mean deltas are all in the expected direction.

\begin{table*}[ht]
\centering
\small
\caption{Cluster-aware significance of \texttt{BICR}'s loss ablation under pooled aggregation. Each row reports the mean per-LVLM delta (Full \texttt{BICR} minus ablated variant) across 5 LVLMs, with significance assessed by a cluster bootstrap (10{,}000 resamples) over LVLM-level seed-means and Holm-Bonferroni correction across the 4 metrics within each row. Significance level conventions match Table~\ref{tab:cluster_aware_significance}.}
\label{tab:loss_ablation_cluster_aware}
\resizebox{\textwidth}{!}{
\begin{tabular}{lcccccccc}
\toprule
\textbf{Comparison} & \multicolumn{2}{c}{\textbf{ECE $\downarrow$}} & \multicolumn{2}{c}{\textbf{BS $\downarrow$}} & \multicolumn{2}{c}{\textbf{AUCPR $\uparrow$}} & \multicolumn{2}{c}{\textbf{AUROC $\uparrow$}} \\
\cmidrule(lr){2-3} \cmidrule(lr){4-5} \cmidrule(lr){6-7} \cmidrule(lr){8-9}
 & \textbf{Mean $\Delta$} & \textbf{$p$} & \textbf{Mean $\Delta$} & \textbf{$p$} & \textbf{Mean $\Delta$} & \textbf{$p$} & \textbf{Mean $\Delta$} & \textbf{$p$} \\
\midrule
Full vs $-\mathcal{L}_{\mathrm{brier}}$ & -0.0140 & n.s. & -0.0061 & n.s. & +0.0032 & n.s. & +0.0059 & n.s. \\
Full vs $-\mathcal{L}_{\mathrm{rank}}$ & -0.0104 & n.s. & -0.0120 & $<$0.001$^{***}$ & +0.0189 & $<$0.001$^{***}$ & +0.0328 & $<$0.001$^{***}$ \\
Full vs $\mathcal{L}_{\mathrm{bce}}$ only & -0.0205 & n.s. & -0.0147 & $<$0.001$^{***}$ & +0.0193 & $<$0.001$^{***}$ & +0.0334 & $<$0.001$^{***}$ \\
\bottomrule
\end{tabular}}
\end{table*}

\paragraph{Per-LVLM breakdown.} Table~\ref{tab:loss_ablation_pervlm} presents the loss ablation results for each LVLM individually. The discrimination story is uniform: $\mathcal{L}_{\mathrm{rank}}$ improves AUROC and AUCPR on every LVLM, with the largest gains on LLaVA-NeXT ($\Delta$AUROC $= +6.0$) and DeepSeek-VL2 ($\Delta$AUROC $= +4.1$). Calibration trade-offs are more nuanced: on Qwen3-VL-8B, removing $\mathcal{L}_{\mathrm{rank}}$ yields a lower ECE than the full model (4.9 vs.\ 8.9), and on Gemma-3-27B removing $\mathcal{L}_{\mathrm{brier}}$ yields a slightly lower ECE (6.4 vs.\ 7.0). These per-LVLM ECE patterns reflect the fact that the auxiliary terms target the cross-LVLM average, where the calibration gain is consistent (Table~\ref{tab:loss_ablation}); on individual LVLMs, the rank loss occasionally trades a small amount of calibration error for its substantial discrimination gain.

\begin{table*}[ht]
\centering
\small
\caption{Per-LVLM loss ablation results (mean across 5 seeds). Best value per LVLM per metric in \textbf{bold}.}
\label{tab:loss_ablation_pervlm}
\begin{tabular}{l l c c c c}
\toprule
\textbf{LVLM} & \textbf{Variant} & \textbf{ECE}$\downarrow$ & \textbf{BS}$\downarrow$ & \textbf{AUCPR}$\uparrow$ & \textbf{AUROC}$\uparrow$ \\
\midrule
\multirow{4}{*}{Qwen3-VL-8B}
 & Full          & 8.9 & \textbf{17.5} & \textbf{90.1} & \textbf{80.1} \\
 & $-$Brier      & 8.5 & 19.0 & 87.1 & 78.0 \\
 & $-$Rank       & \textbf{4.9} & 17.7 & 88.7 & 77.2 \\
 & BCE only      & 7.3 & 18.2 & 88.7 & 77.2 \\
\midrule
\multirow{4}{*}{LLaVA-NeXT-13B}
 & Full          & \textbf{5.7} & \textbf{18.2} & \textbf{87.7} & \textbf{78.9} \\
 & $-$Brier      & 6.4 & 18.6 & 87.2 & 78.2 \\
 & $-$Rank       & 10.4 & 20.9 & 84.6 & 72.9 \\
 & BCE only      & 10.5 & 21.1 & 84.1 & 72.2 \\
\midrule
\multirow{4}{*}{InternVL3.5-14B}
 & Full          & \textbf{7.9} & \textbf{19.0} & \textbf{88.0} & \textbf{76.4} \\
 & $-$Brier      & 10.9 & 20.4 & 87.2 & 75.2 \\
 & $-$Rank       & 9.7 & 19.8 & 86.6 & 74.1 \\
 & BCE only      & 10.9 & 20.4 & 86.3 & 73.7 \\
\midrule
\multirow{4}{*}{DeepSeek-VL2}
 & Full          & \textbf{6.0} & \textbf{17.9} & \textbf{86.1} & \textbf{81.1} \\
 & $-$Brier      & 7.0 & 18.1 & 86.0 & 80.8 \\
 & $-$Rank       & 8.4 & 19.8 & 83.5 & 77.0 \\
 & BCE only      & 9.7 & 20.2 & 83.6 & 77.3 \\
\midrule
\multirow{4}{*}{Gemma-3-27B}
 & Full          & 7.0 & \textbf{19.6} & \textbf{85.1} & \textbf{76.6} \\
 & $-$Brier      & \textbf{6.4} & 19.3 & 85.4 & 77.0 \\
 & $-$Rank       & 7.3 & 19.9 & 84.2 & 75.3 \\
 & BCE only      & 7.3 & 19.7 & 84.8 & 75.9 \\
\bottomrule
\end{tabular}
\end{table*}

\subsection{\texorpdfstring{What $\mathcal{L}_{\mathrm{rank}}$ Teaches the Probe}{What Lrank Teaches the Probe}}
\label{app:design_behavior}

Beyond improving aggregate metrics, $\mathcal{L}_{\mathrm{rank}}$ fundamentally changes the probe's learned behavior. We analyze this by comparing the per-sample confidence outputs of the full \texttt{BICR} model against the $\mathcal{L}_{\mathrm{bce}}$-only baseline, using all 25 runs (5~LVLMs $\times$ 5~seeds).

\paragraph{Confidence separation.} Table~\ref{tab:conf_dist} compares the mean confidence assigned to correct versus incorrect samples.

\begin{table*}[ht]
\centering
\small
\caption{Confidence distributions across all 25 runs. ``Separation'' is the difference between correct and incorrect means; higher indicates better discrimination.}
\label{tab:conf_dist}
\begin{tabular}{l c c c}
\toprule
\textbf{Variant} & \textbf{Correct} & \textbf{Incorrect} & \textbf{Separation} \\
\midrule
Full (BICR)          & 70.2 & 41.6 & \textbf{28.7} \\
$-\mathcal{L}_{\mathrm{brier}}$  & 68.2 & 39.3 & 28.9 \\
$-\mathcal{L}_{\mathrm{rank}}$   & 72.2 & 49.2 & 23.0 \\
$\mathcal{L}_{\mathrm{bce}}$ only & 71.0 & 47.3 & 23.7 \\
\bottomrule
\end{tabular}
\end{table*}

Without $\mathcal{L}_{\mathrm{rank}}$, incorrect samples receive substantially higher confidence (49.2 vs.\ 41.6), reducing the separation by 20\%. $\mathcal{L}_{\mathrm{rank}}$ selectively suppresses confidence for incorrect predictions while maintaining high confidence for correct ones.

\paragraph{Calibration.} The effect is most pronounced in the high-confidence range. Table~\ref{tab:calib_bins} compares the reliability diagram bins where overconfidence is most harmful.

\begin{table*}[ht]
\centering
\small
\caption{Full reliability diagram across all 10 bins for each ablation variant, pooled across all 25 runs (5~LVLMs $\times$ 5~seeds). Comparing Full to $-\mathcal{L}_{\mathrm{rank}}$ isolates the contribution of the ranking loss in each bin (with $\mathcal{L}_{\mathrm{brier}}$ held constant); the $\mathcal{L}_{\mathrm{bce}}$-only column shows the joint effect of removing both auxiliary losses. ``Gap'' $= |\text{Pred.} - \text{Act.}|$; lower is better. All values are percentages, with the lowest gap per row bolded.}
\label{tab:calib_bins}
\begin{tabular}{lccccccccc}
\toprule
 & \multicolumn{3}{c}{\textbf{Full (\texttt{BICR})}} & \multicolumn{3}{c}{\textbf{$-\mathcal{L}_{\mathrm{rank}}$}} & \multicolumn{3}{c}{\textbf{$\mathcal{L}_{\mathrm{bce}}$ only}} \\
\cmidrule(lr){2-4} \cmidrule(lr){5-7} \cmidrule(lr){8-10}
\textbf{Bin} & Pred. & Act. & Gap & Pred. & Act. & Gap & Pred. & Act. & Gap \\
\midrule
$[0.0, 0.1)$ & 6.3 & 23.3 & 16.9 & 6.6 & 20.6 & \textbf{14.0} & 6.7 & 21.8 & 15.0 \\
$[0.1, 0.2)$ & 15.2 & 31.5 & \textbf{16.3} & 15.4 & 32.5 & 17.1 & 15.4 & 33.9 & 18.5 \\
$[0.2, 0.3)$ & 25.1 & 38.4 & \textbf{13.3} & 25.2 & 40.8 & 15.6 & 25.3 & 42.8 & 17.5 \\
$[0.3, 0.4)$ & 35.1 & 45.2 & \textbf{10.2} & 35.2 & 46.3 & 11.2 & 35.1 & 47.7 & 12.6 \\
$[0.4, 0.5)$ & 45.0 & 50.4 & 5.4 & 45.1 & 49.0 & \textbf{4.0} & 45.0 & 50.1 & 5.1 \\
$[0.5, 0.6)$ & 55.0 & 56.0 & \textbf{1.1} & 55.0 & 52.2 & 2.7 & 55.0 & 52.9 & 2.1 \\
$[0.6, 0.7)$ & 65.0 & 63.4 & \textbf{1.6} & 65.0 & 56.7 & 8.3 & 65.0 & 57.8 & 7.1 \\
$[0.7, 0.8)$ & 75.0 & 72.0 & \textbf{3.1} & 75.0 & 64.3 & 10.7 & 75.0 & 65.2 & 9.8 \\
$[0.8, 0.9)$ & 85.3 & 83.6 & \textbf{1.8} & 85.3 & 77.4 & 7.9 & 85.3 & 77.9 & 7.4 \\
$[0.9, 1.0]$ & 95.9 & 94.9 & \textbf{1.0} & 95.9 & 93.3 & 2.7 & 96.0 & 93.0 & 3.0 \\
\bottomrule
\end{tabular}
\end{table*}

The Full vs $-\mathcal{L}_{\mathrm{rank}}$ comparison isolates the ranking loss's contribution at each confidence level (with $\mathcal{L}_{\mathrm{brier}}$ present in both). The high-confidence range $[0.6, 0.9)$ is where the rank loss matters most: removing $\mathcal{L}_{\mathrm{rank}}$ produces severe overconfidence (a 75\% prediction corresponds to 64\% empirical accuracy, gap $= 10.7$\%), while keeping it in place tracks the diagonal much more closely (75\% prediction corresponds to 72\% accuracy, gap $= 3.1$\%). Across the four high-confidence bins $[0.6, 1.0]$ the gap reduction ranges from $2.7\times$ (in $[0.9, 1.0]$) to $5.2\times$ (in $[0.6, 0.7)$), with the largest improvements concentrated in precisely the confidence range where overconfidence is most consequential for downstream decision-making. The $\mathcal{L}_{\mathrm{bce}}$-only column confirms that removing both auxiliary losses does not produce calibration meaningfully worse than removing $\mathcal{L}_{\mathrm{rank}}$ alone in this range, indicating that $\mathcal{L}_{\mathrm{rank}}$ is the dominant driver of high-confidence calibration in \texttt{BICR}. On two of the low-confidence bins ($[0.0, 0.1)$ and $[0.4, 0.5)$), $-\mathcal{L}_{\mathrm{rank}}$ lands within 1--3 gap points of Full and occasionally edges it; this reflects the asymmetric design intent of $\mathcal{L}_{\mathrm{rank}}$, which targets overconfidence at high scores and is therefore expected to dominate in the high-confidence range where overconfidence is the practical risk, rather than in the low-confidence range where the failure mode is the opposite (underconfidence: predicted scores below empirical accuracy in all variants).

\begin{figure*}[t]
\centering
\includegraphics[width=\textwidth]{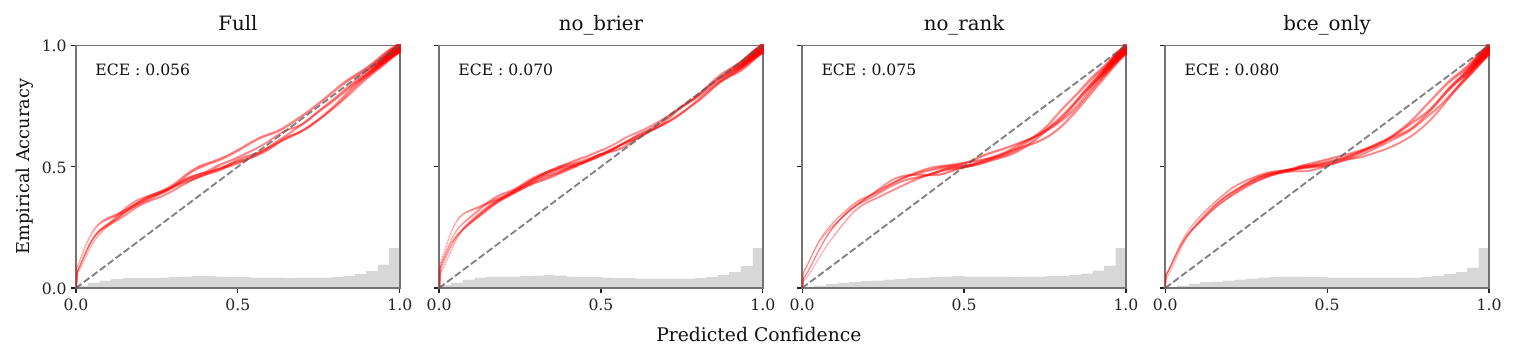}
\caption{Per-seed reliability diagrams for each ablation variant. Each panel shows one reliability curve per seed (5 translucent red curves) and a grey histogram in the lower portion of the panel showing the distribution of predicted confidence values pooled across all 5 LVLMs and 5 seeds (25 runs combined). The dashed diagonal marks perfect calibration. Cross-LVLM ECE values are shown in each panel: Full \texttt{BICR} achieves the lowest ECE ($0.056$); removing $\mathcal{L}_{\mathrm{rank}}$ raises ECE to $0.075$, and removing both auxiliary losses raises it further to $0.080$.}
\label{fig:calibration_ablation}
\end{figure*}

Figure~\ref{fig:calibration_ablation} visualizes this effect across the full reliability curve. Each panel shows one reliability curve per seed (5 curves total, drawn translucently), and a grey histogram in the lower portion of each panel showing the distribution of predicted confidence values pooled across all 5 LVLMs and 5 seeds. Moving from left to right (Full $\to$ $-$Brier $\to$ $-$Rank $\to$ BCE-only), the curves progressively bow further away from the diagonal in the low-to-mid confidence range, where empirical accuracy sits above predicted confidence (i.e., the probe is underconfident in that range, predicting low scores for samples that turn out correct more often than the score implies). The full \texttt{BICR} model tracks the diagonal closely across the full range; without $\mathcal{L}_{\mathrm{rank}}$ the curves develop a clear bow upward in the $[0, 0.5]$ region. The pooled histograms tell a complementary story: the full model spreads probability mass across the full range while the ablated variants (especially BCE-only) concentrate mass toward the high-confidence end.

\paragraph{Score utilization.} Without $\mathcal{L}_{\mathrm{rank}}$, the probe concentrates scores in the high-confidence range: 66.8\% of all samples receive scores above 0.5, and only 1.9\% fall below 0.1. With $\mathcal{L}_{\mathrm{rank}}$, the distribution broadens (60.1\% above 0.5, 4.1\% below 0.1), indicating that the probe has learned to use the full probability range and express genuine uncertainty when the model's prediction is unreliable. This broader utilization is a hallmark of well-calibrated confidence estimators.

\subsection{Optimized Hyperparameter Analysis}
\label{app:design_hparams}

The Optuna search jointly optimizes the loss weights $\beta$ ($\mathcal{L}_{\mathrm{brier}}$), $\lambda$ ($\mathcal{L}_{\mathrm{rank}}$), and the margin $\gamma$ alongside the architectural and optimizer hyperparameters. Table~\ref{tab:optuna_configs} reports the mean and range of these values across the 5 seeds for each LVLM, providing insight into the stability and LVLM-dependence of the optimized configurations.

\begin{table*}[ht]
\centering
\small
\caption{Optuna-optimized loss hyperparameters for \texttt{BICR} across LVLMs. Each cell shows the mean across 5 seeds, with the range [min, max] in parentheses. ``Arch.'' reports the most frequently selected \texttt{classifier\_layers} configuration across the 5 seeds.}
\label{tab:optuna_configs}
\begin{tabular}{l c c c c}
\toprule
\textbf{LVLM} & $\beta$ (\textbf{Brier weight}) & $\lambda$ (\textbf{Rank weight}) & $\gamma$ (\textbf{Margin}) & \textbf{Arch.} \\
\midrule
Qwen3-VL-8B       & 0.26 \footnotesize{[0.19, 0.39]} & 0.20 \footnotesize{[0.12, 0.24]} & 0.09 \footnotesize{[0.05, 0.17]} & (128, 64) \\
LLaVA-NeXT-13B    & 0.35 \footnotesize{[0.15, 0.49]} & 0.07 \footnotesize{[0.01, 0.25]} & 0.11 \footnotesize{[0.05, 0.14]} & (256, 128) \\
InternVL3.5-14B   & 0.16 \footnotesize{[0.00, 0.40]} & 0.11 \footnotesize{[0.02, 0.22]} & 0.12 \footnotesize{[0.07, 0.19]} & (512, 256) \\
DeepSeek-VL2      & 0.25 \footnotesize{[0.07, 0.48]} & 0.17 \footnotesize{[0.07, 0.25]} & 0.17 \footnotesize{[0.10, 0.23]} & (512, 256) \\
Gemma-3-27B       & 0.28 \footnotesize{[0.22, 0.32]} & 0.09 \footnotesize{[0.02, 0.14]} & 0.17 \footnotesize{[0.10, 0.21]} & (256, 128) \\
\bottomrule
\end{tabular}
\end{table*}

Three patterns are noteworthy. First, the Brier weight $\beta$ is consistently selected in a moderate range (mean 0.16--0.35), confirming that Optuna finds the calibration term beneficial but secondary to the classification objective. Second, the rank weight $\lambda$ is always positive and non-trivial (mean 0.07--0.20), demonstrating that the optimizer consistently allocates weight to the ranking signal across all LVLMs. Although one seed on LLaVA-NeXT-13B does select $\lambda = 0.01$, no LVLM's seed-level mean approaches the lower bound, supporting that $\mathcal{L}_{\mathrm{rank}}$ provides genuine training signal rather than being an artifact of the search space. Third, the margin $\gamma$ trends toward larger values for the LVLMs with the lowest baseline correctness rates (DeepSeek-VL2 at $\gamma = 0.17$, baseline 55.0\% correct; Gemma-3-27B at $\gamma = 0.17$, baseline 62.7\% correct), and toward smaller values for the LVLM with the highest correctness rate (Qwen3-VL-8B at $\gamma = 0.09$, baseline 68.7\% correct), although the relationship is not perfectly monotonic across the middle of the range. The Optuna-selected architectures vary across LVLMs from compact $(128, 64)$ heads on Qwen3-VL-8B to wider $(512, 256)$ on InternVL3.5-14B and DeepSeek-VL2, without a clean correspondence to LVLM hidden size; the architecture choice appears to depend more on interactions with the loss-coefficient settings than on $d_h$ alone.

\subsection{Choice of Null Image}
\label{app:design_blank_color}

\texttt{BICR} uses a solid black image as the null visual input $v_{\varnothing}$ throughout the main experiments. This choice is load-bearing: the rank loss $\mathcal{L}_{\mathrm{rank}}$ pushes $\sigma(W^{\top} \mathbf{h}_{\mathrm{base}})$ above $\sigma(W^{\top} \mathbf{h}_{\mathrm{blank}})$ for correctly answered samples, so the visual content of the null directly shapes the supervisory signal. To test whether the result is specific to black or whether \texttt{BICR} is robust to any visually-impoverished null, we systematically compare five null-image strategies, each chosen to isolate a specific axis along which a null can differ from the base image: \textbf{black} (current default; minimum visual signal, dark uniform field), \textbf{white} (minimum visual signal, bright uniform field; tests whether the effect comes from luminance or uniformity), \textbf{Gaussian noise} (uniform-random pixels; tests whether \texttt{BICR} responds to the absence of information or the absence of image-like structure), \textbf{blurred original} (low-frequency content preserved via Gaussian blur with radius~50, high-frequency content stripped; tests whether \texttt{BICR} responds to loss of detail or to total image absence), and \textbf{pixel-shuffled original} (per-pixel permutation; preserves the color histogram but destroys all spatial layout, testing whether the signal comes from spatial structure or color statistics).

\paragraph{Experimental setup.} For each of the four new null types we re-extract the \texttt{BICR} feature representation from scratch and re-train the \texttt{BICR} probe over the same five seeds $\{23, 42, 137, 2024, 3407\}$ used in the main experiments, then compare to the existing black-baseline checkpoints on the held-out test split. The two stochastic generators (Gaussian noise and pixel-shuffled) use a deterministic per-sample seed $\sigma = \mathrm{hash}_{32}(h_{\mathrm{id}}) \oplus s$, where $h_{\mathrm{id}}$ is the sample's hash identifier and $s = 42$ is a global null seed, so the same sample produces the same null pixels across reruns. To bound compute cost, this experiment is restricted to a single backbone (\texttt{Qwen3-VL-8B-Instruct}). All other forward-pass settings, training hyperparameters, Optuna budget (50 trials per seed), and evaluation protocol match the main \texttt{BICR} pipeline.

\paragraph{Test-set performance per null type.} Table~\ref{tab:null_ablation_wilcoxon} reports paired Wilcoxon signed-rank tests comparing each new null type against the black baseline, paired by seed ($n = 5$). With only five paired observations, the smallest $p$-value the Wilcoxon test can produce is $0.0625$, attained when all five seeds agree on the direction of the effect. Cells at this floor (marked $\dagger$) therefore represent the strongest possible evidence at this sample size: black wins (or loses) on every single seed. Cells above the floor (Gaussian noise on ECE at $p = 0.125$, blurred on ECE at $p = 0.188$) indicate that fewer than all five seeds agreed on the direction.

\begin{table*}[ht]
\centering
\small
\setlength{\tabcolsep}{4pt}
\caption{\texttt{BICR} test-set performance on \texttt{Qwen3-VL-8B-Instruct} under five null-image strategies for the blank view. Each cell is mean$\,\pm\,$std across 5 seeds; per-seed metrics are computed on all 30{,}514 pooled test samples. Best value per metric in \textbf{bold}.}
\label{tab:null_ablation_main}
\begin{tabular}{l r r r r r r}
\toprule
\textbf{Null type} & \textbf{ECE}\,$\downarrow$ & \textbf{BS}\,$\downarrow$ & \textbf{Acc}\,$\uparrow$ & \textbf{F1}\,$\uparrow$ & \textbf{AUCPR}\,$\uparrow$ & \textbf{AUROC}\,$\uparrow$ \\
\midrule
\textit{black (current)} & \textbf{0.0886}\,\tiny{$\pm$\,0.0167} & \textbf{0.1747}\,\tiny{$\pm$\,0.0034} & \textbf{0.7281}\,\tiny{$\pm$\,0.0044} & \textbf{0.7866}\,\tiny{$\pm$\,0.0068} & \textbf{0.9014}\,\tiny{$\pm$\,0.0009} & \textbf{0.8008}\,\tiny{$\pm$\,0.0019} \\
white                    & 0.1332\,\tiny{$\pm$\,0.0276}         & 0.2223\,\tiny{$\pm$\,0.0144}         & 0.6908\,\tiny{$\pm$\,0.0026}         & 0.3665\,\tiny{$\pm$\,0.0339}         & 0.5125\,\tiny{$\pm$\,0.0224}         & 0.6780\,\tiny{$\pm$\,0.0383} \\
Gaussian noise           & 0.1231\,\tiny{$\pm$\,0.0346}         & 0.2125\,\tiny{$\pm$\,0.0222}         & 0.6961\,\tiny{$\pm$\,0.0069}         & 0.3872\,\tiny{$\pm$\,0.0636}         & 0.5338\,\tiny{$\pm$\,0.0366}         & 0.7035\,\tiny{$\pm$\,0.0670} \\
blurred                  & 0.1255\,\tiny{$\pm$\,0.0311}         & 0.2153\,\tiny{$\pm$\,0.0146}         & 0.6919\,\tiny{$\pm$\,0.0052}         & 0.3273\,\tiny{$\pm$\,0.0997}         & 0.5231\,\tiny{$\pm$\,0.0160}         & 0.7007\,\tiny{$\pm$\,0.0277} \\
pixel-shuffled           & 0.1206\,\tiny{$\pm$\,0.0216}         & 0.2101\,\tiny{$\pm$\,0.0136}         & 0.6957\,\tiny{$\pm$\,0.0069}         & 0.3657\,\tiny{$\pm$\,0.0493}         & 0.5354\,\tiny{$\pm$\,0.0270}         & 0.7215\,\tiny{$\pm$\,0.0292} \\
\bottomrule
\end{tabular}
\end{table*}

\begin{table*}[ht]
\centering
\small
\setlength{\tabcolsep}{4pt}
\caption{Paired Wilcoxon signed-rank tests against the black baseline ($n=5$ seeds). $\dagger$ denotes unanimous direction across all five seeds (minimum attainable $p = 0.0625$ for $n = 5$).}
\label{tab:null_ablation_wilcoxon}
\begin{tabular}{l r r r r r r}
\toprule
\textbf{Comparison} & \textbf{ECE}\,$\downarrow$ & \textbf{Brier}\,$\downarrow$ & \textbf{Acc}\,$\uparrow$ & \textbf{F1}\,$\uparrow$ & \textbf{AUCPR}\,$\uparrow$ & \textbf{AUROC}\,$\uparrow$ \\
\midrule
vs.\ white          & $0.062\,\dagger$ & $0.062\,\dagger$ & $0.062\,\dagger$ & $0.062\,\dagger$ & $0.062\,\dagger$ & $0.062\,\dagger$ \\
vs.\ Gaussian noise & 0.125            & $0.062\,\dagger$ & $0.062\,\dagger$ & $0.062\,\dagger$ & $0.062\,\dagger$ & $0.062\,\dagger$ \\
vs.\ blurred        & 0.188            & $0.062\,\dagger$ & $0.062\,\dagger$ & $0.062\,\dagger$ & $0.062\,\dagger$ & $0.062\,\dagger$ \\
vs.\ pixel-shuffled & $0.062\,\dagger$ & $0.062\,\dagger$ & $0.062\,\dagger$ & $0.062\,\dagger$ & $0.062\,\dagger$ & $0.062\,\dagger$ \\
\bottomrule
\end{tabular}
\end{table*}

\paragraph{Black is the best of the five strategies on every metric.} The black baseline achieves the best mean performance on all six metrics across all four alternative null types. The margin is large in absolute terms: AUROC drops from 0.801 (black) to between 0.678 and 0.722 for the four alternatives, a $7.9$ to $12.3$ point gap; AUCPR collapses from 0.901 to roughly 0.51--0.54; F1 drops from 0.787 to roughly 0.33--0.39. Calibration also worsens: ECE roughly doubles ($0.089 \to 0.12$--$0.13$) and Brier rises by $\approx 0.04$. The direction of the effect is unanimous across the five paired seeds for every metric in $\{\mathrm{Brier}, \mathrm{Acc}, \mathrm{F1}, \mathrm{AUCPR}, \mathrm{AUROC}\}$ and for ECE under the white and pixel-shuffled comparisons (Wilcoxon $p = 0.0625$, the floor for $n = 5$). In the two cells where the ECE comparison falls slightly above the floor (Gaussian noise at $p = 0.125$ and blurred at $p = 0.188$), the mean still favors black.

\paragraph{Why the alternatives fail, by axis of variation.} The four alternative null types form an interpretable failure pattern that maps cleanly onto the four axes the comparison was designed to probe. \textbf{Luminance vs.\ uniformity:} white and black are both uniform fills, yet white loses 12.3 AUROC points relative to black. Uniformity alone is therefore not what makes black useful as a null; the LVLM produces meaningfully different hidden states for white than for black, and the latter are more useful as a contrast point. \textbf{Information content vs.\ image-likeness:} replacing the image with high-entropy random pixels (Gaussian noise) does not help, dropping AUROC by 9.7 points relative to black. The blank view should be \emph{information-poor} rather than \emph{information-different}; adding visual entropy without spatial structure is closer to ``a different image'' than to ``no image.'' \textbf{High-frequency detail vs.\ total absence:} heavy blurring strips edges, text, and object structure but preserves low-frequency content (broad color fields, average illumination), and still costs 10.0 AUROC points relative to black, suggesting that the residual low-frequency content is enough for the LVLM to maintain a non-null hidden state. \textbf{Spatial structure vs.\ color statistics:} pixel-shuffling preserves the image's full color histogram while scrambling all spatial layout. This is the least bad of the four alternatives (only 7.9 AUROC points worse than black), consistent with spatial structure carrying the bulk of the image-likeness signal; once layout is destroyed, the LVLM behaves closer to the all-black case, although the gap to black remains significant on every metric.

\paragraph{Design implication.} The simple solid-black null both produces the most distinct $\mathbf{h}_{\mathrm{blank}}$ from $\mathbf{h}_{\mathrm{base}}$ and provides the strongest training signal for the rank loss. The ablation supports the design choice in \texttt{BICR}: the null view should be information-poor in an absolute sense, not merely information-different. A null that still carries any image-like content, whether high-entropy noise, low-pass-filtered original, or shuffled color statistics, gives the LVLM enough to produce a non-null representation that the probe cannot exploit as effectively as it can the all-black contrast.

%% file: sections/appendix_extended_results.tex
\section{Extended Results}
\label{app:extended_results}

This appendix provides the complete set of results underlying the main-text summary in \S\ref{sec:results}. All trained methods report the mean across five seeds $\{23, 42, 137, 2024, 3407\}$, each with 50 Optuna hyperparameter trials. Metrics are computed on the shared subset of test samples present in all methods for a given LVLM, obtained by intersecting test \texttt{hash\_id}s across every confidence estimation method evaluated on that LVLM; per-LVLM and per-dataset shared counts are reported in Table~\ref{tab:shared_samples}. The intersection accounts for the small number of samples that drop from individual methods for benign reasons: prompt-based methods occasionally produce a malformed parse on a sample whose generated text does not match the expected response format, and Self-Probing requires a second LVLM forward pass that can fail on individual samples due to context-length overflow. Restricting evaluation to the intersection enforces apples-to-apples comparison: every metric, significance test, and figure in this appendix and in the main paper is computed on the same per-LVLM sample set across all methods. Best values per metric are \textbf{bolded}. Reported metrics are: Expected Calibration Error (ECE), Brier Score (BS), Accuracy (ACC), F1 Score (F1), Area Under Precision--Recall Curve (AUCPR), and Area Under ROC Curve (AUROC).

\input{tables/shared_samples}

\subsection{Per-LVLM Pooled Performance}
\label{app:results_pooled}

Table~\ref{tab:pooled_pervlm} reports the pooled aggregate performance of each method on each LVLM. In pooled evaluation, all test samples are combined into a single set and metrics are computed over the full shared subset per LVLM. \texttt{BICR} achieves the best AUCPR and AUROC on every LVLM (five of five), and the best calibration (ECE, BS) on three of five (LLaVA-NeXT-13B, InternVL3.5-14B, and DeepSeek-VL2); on the remaining two LVLMs the best ECE belongs to InternalInspector (Qwen3-VL-8B and Gemma-3-27B). 

\input{tables/pooled_per_vlm}

\subsection{Cross-LVLM Pooled Average}
\label{app:results_pooled_avg}

Table~\ref{tab:pooled_crossvlm} averages the per-LVLM pooled metrics from Table~\ref{tab:pooled_pervlm} across all five LVLMs, giving equal weight to each LVLM architecture. \texttt{BICR} achieves the best cross-LVLM average on five of six metrics (ECE 7.1, BS 18.4, ACC 71.5, AUCPR 87.5, AUROC 78.6); SAPLMA edges F1 (79.0 vs.\ 76.9). The next-best method on the discrimination metrics is P(I~Know), which \texttt{BICR} beats by $+1.2$ AUCPR and $+2.0$ AUROC points; on the calibration metrics, the next-best is InternalInspector on ECE (8.3, behind \texttt{BICR} by 1.2 points) and P(I~Know) on BS (19.4, behind by 1.0 point).

\input{tables/pooled_cross_vlm}

\subsection{Per-LVLM Unweighted Average Across Datasets}
\label{app:results_unweighted}

The pooled evaluation in \S\ref{app:results_pooled} is dominated by the largest datasets (GQA: 12{,}568 samples; POPE: 9{,}000 samples), which together constitute over 70\% of the test set. To check whether \texttt{BICR}'s advantage holds when each dataset contributes equally, Table~\ref{tab:uw_pervlm} reports per-dataset metrics averaged with equal weight, excluding datasets with fewer than 100 shared samples per LVLM (LLaVA-Wild is therefore dropped, leaving six of the seven datasets in the average for every LVLM). On this stricter aggregation, \texttt{BICR} achieves the best BS on every LVLM (five of five), the best ECE on three of five (LLaVA-NeXT-13B, InternVL3.5-14B, DeepSeek-VL2), and the best accuracy on every LVLM (five of five). Discrimination is more contested in this view: \texttt{BICR} leads on AUCPR and AUROC for LLaVA-NeXT-13B, P(I~Know) takes both metrics on Qwen3-VL-8B and Gemma-3-27B, P(True) takes both metrics on DeepSeek-VL2, and on InternVL3.5-14B P(True) takes AUCPR while Self-Probing takes AUROC (a known artifact of inference-only methods' near-saturated confidence collapsing into a high-AUC score on the smaller datasets that dominate the unweighted average; see \S\ref{app:results_calibration_perdataset}). The shift relative to the pooled view (\S\ref{app:results_pooled}, where \texttt{BICR} led AUCPR and AUROC on every LVLM) decomposes into two distinct effects when we look at the per-dataset breakdown (Table~\ref{tab:perdataset}). \texttt{BICR}'s discrimination lead is largest on GQA and POPE, the two largest datasets, which the pooled view amplifies and the equal-weight view de-emphasizes. On the harder grounding-bound datasets (GMAI-MMBench, MMMU-Pro, MME-Finance), the per-LVLM picture is more nuanced. On GMAI-MMBench, \texttt{BICR}'s discrimination gaps to the per-LVLM winner are small (typically 2--4 AUROC points) while its calibration is materially better; on MMMU-Pro, P(True) takes discrimination on several LVLMs, but this reflects the saturation artifact already discussed in \S\ref{app:results_calibration_perdataset} (inference-only methods collapse to a near-uniform high-confidence score that produces accidentally high AUROC on small datasets) rather than genuine probe-level superiority, and \texttt{BICR}'s calibration on MMMU-Pro is the best across LVLMs by a wide margin (cross-LVLM mean ECE 0.149--0.194 vs.\ next-best InternalInspector 0.270--0.343). This is what the unweighted view exposes: \texttt{BICR}'s strongest contribution on the visually-demanding datasets is calibration, not discrimination, and the equal-weight average amplifies these calibration gains, which is why \texttt{BICR}'s ECE and BS leads grow under equal weighting (Table~\ref{tab:uw_crossvlm}: ECE gap to InternalInspector grows from 1.2 to 4.2 points, BS gap from 1.5 to 3.7) even as discrimination tightens.

\input{tables/unweighted_per_vlm}

\subsection{Cross-LVLM Unweighted Average}
\label{app:results_unweighted_avg}

Table~\ref{tab:uw_crossvlm} averages the per-LVLM unweighted metrics from Table~\ref{tab:uw_pervlm} across all five LVLMs, giving equal weight to each LVLM architecture and each source dataset. \texttt{BICR} achieves the best calibration (ECE 13.8 vs.\ next-best II 18.0; BS 21.4 vs.\ next-best II 25.1) and the best accuracy (66.8 vs.\ next-best SAPLMA 59.6) by substantial margins. On discrimination, P(I~Know) narrowly edges \texttt{BICR} on AUROC (68.4 vs.\ 68.1) and AUCPR (64.8 vs.\ 64.2), with both gaps under one point and well within the across-LVLM standard deviations of either method. The trade-off this view exposes is informative: P(I~Know) reaches comparable discrimination to \texttt{BICR} but at a substantially worse calibration cost (ECE 21.3 vs.\ 13.8, a $7.5$-point gap; BS 25.7 vs.\ 21.4, a $4.3$-point gap), so the methods that compete with \texttt{BICR} on discrimination do so by being meaningfully more miscalibrated. SAPLMA leads F1 (63.7 vs.\ \texttt{BICR} 57.9). The headline takeaway from the unweighted view: even when the equal-weight aggregation strips out the dominance of the largest datasets, \texttt{BICR} retains a clear calibration and accuracy advantage and stays within sub-point distance of the strongest baseline on discrimination, while no baseline matches \texttt{BICR} on both axes at once.

\input{tables/unweighted_cross_vlm}

\subsection{Per-LVLM Per-Dataset Breakdown}
\label{app:results_perdataset}

Table~\ref{tab:perdataset} provides the full per-dataset breakdown for each LVLM. Within each LVLM block, results are grouped by dataset, with methods as rows and metrics as columns. The test set comprises seven source datasets: GQA (visual question answering), GMAI-MMBench (medical imaging), POPE (object hallucination detection), MME-Finance (financial document understanding), MMMU\_Pro 4-option and 10-option (multi-choice reasoning), and LLaVA-Wild (open-ended visual dialogue).

The headline pattern is on the multi-choice reasoning datasets where visual grounding is the bottleneck. On MMMU\_Pro 10-option \texttt{BICR} wins ECE on four of five LVLMs and accuracy on four of five (with the remaining LVLM on each metric a comfortable second), and the accuracy gap on DeepSeek-VL2 is striking (\texttt{BICR} 84.1 vs.\ next-best InternalInspector 64.2). On DeepSeek-VL2 specifically, MMMU\_Pro 4-opt and 10-opt are extreme low-base-rate regimes (correctness rates of $\sim$11--19\%) where \texttt{BICR}'s ACC advantage comes from the probe correctly assigning low confidence to the dominant incorrect class; the corresponding AUROC values fall below 0.5 in these cells, indicating that the score's ranking direction is unreliable on the (very few) correct samples even though the calibration gain to incorrect samples is large. We retain the bolding for ACC and ECE on these cells but caution against reading them as joint discrimination wins. MMMU\_Pro 4-option shows the same pattern with slightly narrower margins. GMAI-MMBench is more mixed: \texttt{BICR} leads accuracy on four of five LVLMs and ECE on two of five (InternVL and Gemma), but the per-LVLM picture is informative. On Qwen and Gemma, P(I~Know) takes both discrimination metrics (AUCPR/AUROC), and on DeepSeek Prompt Ensembles takes both, but in each of these cases the discrimination winner pays a sizable ECE cost relative to \texttt{BICR}: on Gemma, P(I~Know) leads on discrimination at ECE 26.2 while \texttt{BICR} sits at comparable discrimination with ECE 11.5; on InternVL, P(True) leads on discrimination (AUCPR 77.8, AUROC 68.7) at ECE 36.8 while \texttt{BICR} reaches comparable AUCPR 75.1 and AUROC 64.6 at ECE 11.3 and BS 24.2 (best on both); on LLaVA, CCPS leads discrimination (AUCPR 48.5, AUROC 63.6) at ECE 30.5, while P(I~Know) and \texttt{BICR} achieve nearly identical discrimination (AUCPR 47.7--47.8, AUROC 61.1--61.2) but \texttt{BICR}'s ECE (18.2) is materially better than P(I~Know)'s (28.5). On the larger near-saturated datasets (GQA, POPE), \texttt{BICR} is competitive but not dominant: P(I~Know), SAPLMA, and InternalInspector share calibration leadership on GQA; on POPE \texttt{BICR} matches P(I~Know) on AUCPR/AUROC across most LVLMs while P(I~Know) and InternalInspector edge calibration. F1 is unfavorable to \texttt{BICR} on several cells because the threshold-based metric rewards the high-recall regime that inference-only baselines and SAPLMA tend to occupy on imbalanced datasets; the threshold-free discrimination metrics (AUCPR, AUROC) tell a more consistent story across all settings.

\input{tables/per_dataset_longtable}

\subsection{Per-LVLM and Per-Dataset Reliability Diagrams}
\label{app:results_calibration_perdataset}

\begin{figure*}[t]
\centering
\includegraphics[width=\textwidth]{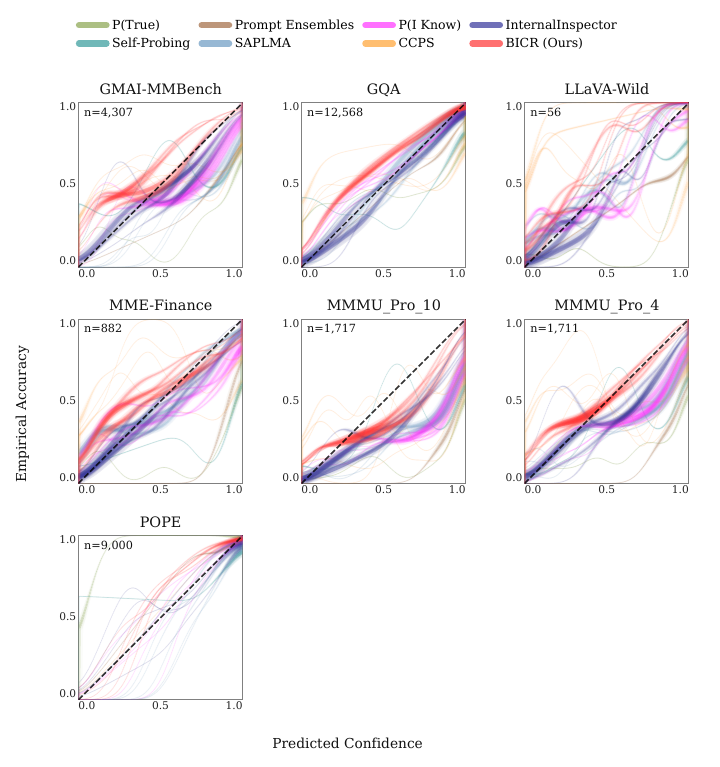}
\caption{Per-dataset reliability diagrams for \texttt{Qwen/Qwen3-VL-8B-Instruct}. Each panel corresponds to one of the seven source datasets ($n$ in panel titles denotes the number of shared test samples for this LVLM). Each method contributes five translucent curves in a method-specific color, one per seed, so the spread visualizes seed-to-seed variability. Visualization style influenced by reliability diagrams in \citet{nakkiran2025trainedtokenscalibratedconcepts}.}
\label{fig:cal_qwen}
\end{figure*}

\begin{figure*}[t]
\centering
\includegraphics[width=\textwidth]{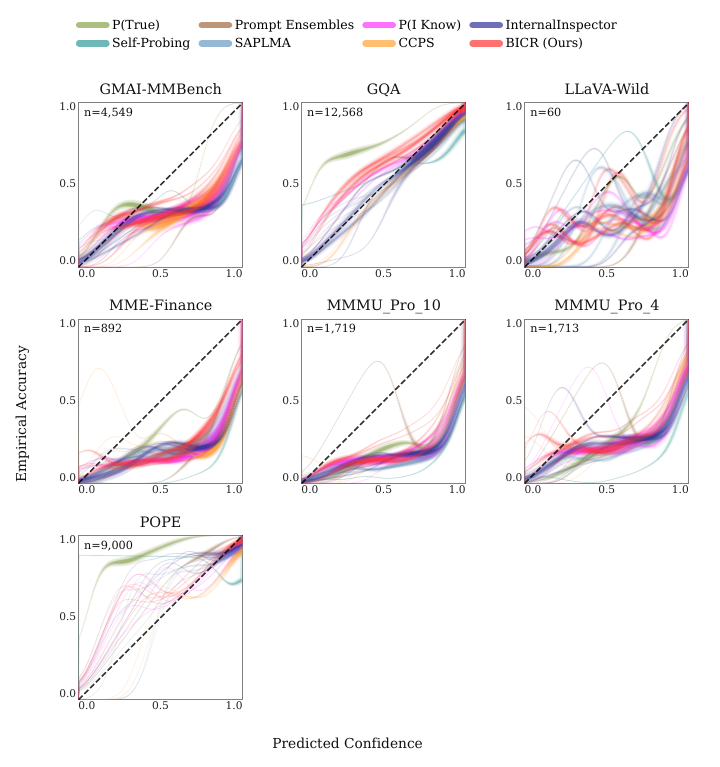}
\caption{Per-dataset reliability diagrams for \texttt{llava-hf/llava-v1.6-vicuna-13b-hf}. Plotting conventions match Figure~\ref{fig:cal_qwen}.}
\label{fig:cal_llava}
\end{figure*}

\begin{figure*}[t]
\centering
\includegraphics[width=\textwidth]{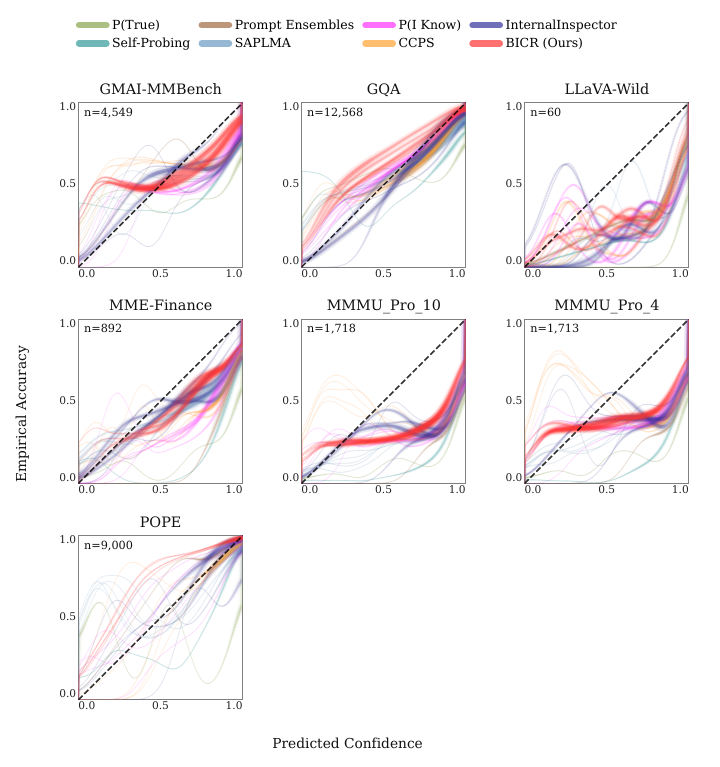}
\caption{Per-dataset reliability diagrams for \texttt{OpenGVLab/InternVL3\_5-14B-HF}. Plotting conventions match Figure~\ref{fig:cal_qwen}.}
\label{fig:cal_internvl}
\end{figure*}

\begin{figure*}[t]
\centering
\includegraphics[width=\textwidth]{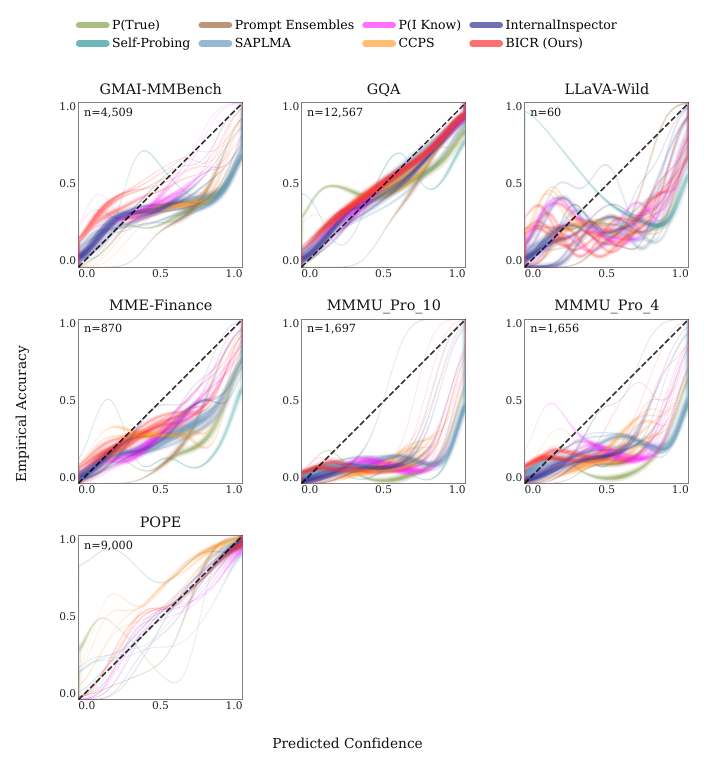}
\caption{Per-dataset reliability diagrams for \texttt{deepseek-ai/deepseek-vl2}. Plotting conventions match Figure~\ref{fig:cal_qwen}.}
\label{fig:cal_deepseek}
\end{figure*}

\begin{figure*}[t]
\centering
\includegraphics[width=\textwidth]{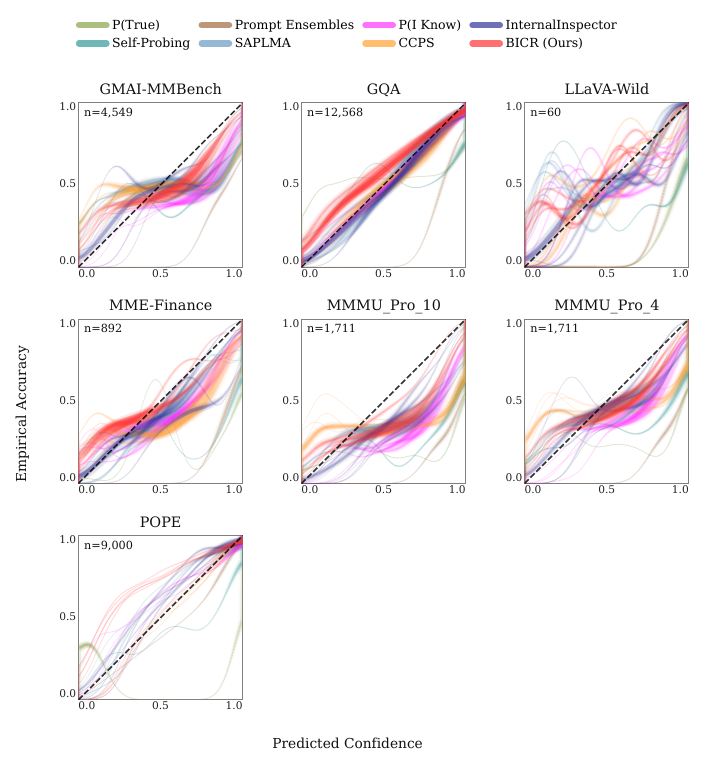}
\caption{Per-dataset reliability diagrams for \texttt{google/gemma-3-27b-it}. Plotting conventions match Figure~\ref{fig:cal_qwen}.}
\label{fig:cal_gemma}
\end{figure*}

\begin{figure*}[t]
\centering
\includegraphics[width=\textwidth]{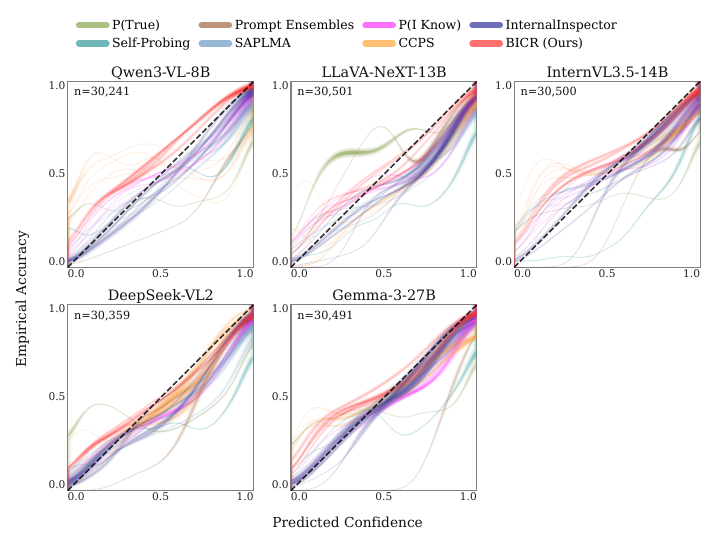}
\caption{Pooled per-LVLM reliability diagrams. Each panel fixes one LVLM and pools every (ground truth, confidence) pair across all seven source datasets, yielding one panel per LVLM ($n$ in panel titles is the size of the shared test subset for that LVLM). Each method contributes five translucent curves in a method-specific color, one per seed; line thickness within each curve is proportional to local bin density.}
\label{fig:cal_pooled_vlm}
\end{figure*}

\begin{figure*}[t]
\centering
\includegraphics[width=\textwidth]{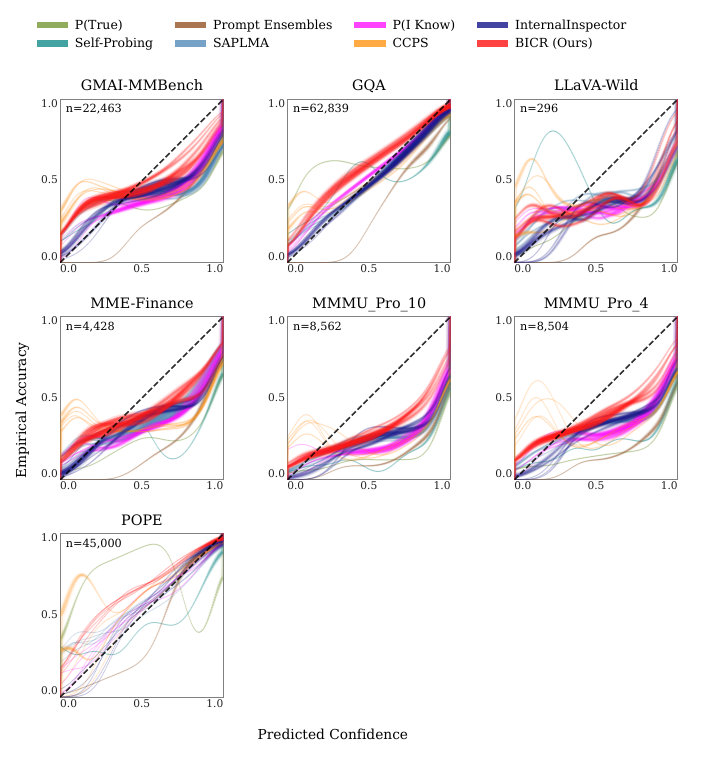}
\caption{Pooled per-dataset reliability diagrams. Each panel fixes one source dataset and pools every (ground truth, confidence) pair across all five LVLMs ($n$ in panel titles is the total number of pooled test samples). Each method contributes five translucent curves in a method-specific color, one per seed; line thickness within each curve is proportional to local bin density.}
\label{fig:cal_pooled_dataset}
\end{figure*}

The pooled calibration figure in the main text (\S\ref{sec:results}, Figure~\ref{fig:calibration}) collapses all five LVLMs and all seven source datasets into a single curve per method. To check whether \texttt{BICR}'s calibration advantage is uniform or concentrates in particular settings, this section splits the same per-bin reliability data along three complementary axes. Figures~\ref{fig:cal_qwen}--\ref{fig:cal_gemma} fix one LVLM per figure and split that LVLM's data by source dataset, yielding the finest-grained view (one panel per dataset, all methods overlaid). Figure~\ref{fig:cal_pooled_vlm} pools across all datasets to give one panel per LVLM, and Figure~\ref{fig:cal_pooled_dataset} pools across all five LVLMs to give one panel per dataset. 

\paragraph{How to read each panel.} Every panel is a reliability diagram. The dashed diagonal marks perfect calibration. Each method contributes \emph{five translucent curves} (one per seed) drawn in a method-specific color, so the spread of the five curves around their joint mean visualizes seed-to-seed variability for that method. Methods that produce a tight cluster of overlapping curves are seed-stable; methods that produce visibly spread or noisy curves are not. \emph{Line thickness within each curve is proportional to the local bin density}: thicker line segments correspond to confidence ranges where the method placed many test samples (and the curve location is therefore reliable), and thinner segments correspond to sparsely populated bins (where the curve location is dominated by sampling noise). A curve sitting \emph{above} the diagonal indicates that the method is under-confident at that confidence level (it is correct more often than its score implies), while a curve \emph{below} the diagonal indicates over-confidence. 

\paragraph{Where \texttt{BICR}'s calibration advantage concentrates.} Tabulating the best-ECE method on each (LVLM, dataset) cell of the per-LVLM figures (35 cells total: 5 LVLMs $\times$ 7 datasets), \texttt{BICR} achieves the lowest ECE on 13 cells, ranks in the top two on 21 cells, and in the top three on 26 cells of 35. Averaging ECE across the five LVLMs per dataset, \texttt{BICR} is the best-calibrated method on four of the seven datasets: GMAI-MMBench (cross-LVLM mean ECE 0.142, vs.\ next-best InternalInspector 0.168), MME-Finance (0.166 vs.\ InternalInspector 0.184), MMMU\_Pro 10-option (0.194 vs.\ InternalInspector 0.343), and MMMU\_Pro 4-option (0.149 vs.\ InternalInspector 0.270). On LLaVA-Wild ($n{\approx}56$--$60$ per LVLM) \texttt{BICR} (0.244) is essentially tied with InternalInspector (0.231), and on the two near-saturated datasets, POPE (binary object-presence detection) and GQA (large-scale VQA), internal-state baselines overtake \texttt{BICR} (POPE: P(I~Know) 0.029, InternalInspector 0.037, \texttt{BICR} 0.071; GQA: InternalInspector 0.044, P(I~Know) 0.046, \texttt{BICR} 0.096). This pattern is consistent with \texttt{BICR}'s design intent: enforcing visual contrast through blank-image ranking helps most where visual grounding is the bottleneck (medical imaging, document understanding, multi-choice reasoning), and helps less where the underlying task is so easy that nearly every method already places most probability mass in the rightmost bin.

\paragraph{Visible failure modes of baselines.} Three regular patterns are evident across the per-LVLM figures. (i) Inference-only methods (P(True), Self-Probing, Prompt Ensembles) collapse to the high-confidence corner on the harder datasets: their curves on GMAI-MMBench, MMMU\_Pro, and MME-Finance crowd into a narrow region close to $x{=}1$, reflecting confidence saturation rather than calibration. P(True) is particularly stark: its cross-LVLM mean ECE on POPE is 0.493, an artifact of placing virtually every sample into a single high-confidence bin while accuracy on POPE is around 0.5. (ii) CCPS swings systematically under-then-over the diagonal, producing characteristic S-shaped curves on Qwen3-VL-8B and LLaVA-NeXT-13B and inflating its cross-LVLM mean ECE on the multi-choice datasets to 0.513 (MMMU\_Pro 10-option) and 0.445 (MMMU\_Pro 4-option), a sign of distortion from its contrastive perturbation step; CCPS also displays the largest visible seed-to-seed spread of any trained method on the multi-choice and LLaVA-Wild panels. (iii) The LLaVA-Wild panel ($n{\approx}56$--$60$ per LVLM) is the noisiest in every figure: with so few samples per bin, even well-calibrated methods produce visibly jittered curves and substantial seed-to-seed disagreement; the panel should therefore be read as bounded by small-sample variance rather than as a clean miscalibration signal.

\clearpage

\clearpage

\clearpage

\clearpage

\clearpage

\paragraph{Pooled views, fixed-LVLM slice.} Figure~\ref{fig:cal_pooled_vlm} fixes the LVLM and pools every (ground truth, confidence) pair across all seven source datasets, yielding one panel per LVLM. The five LVLM subsets are within $\sim$1\% of each other in size ($n{=}30{,}241$ to $n{=}30{,}501$), so visual differences across panels reflect LVLM behaviour rather than sample-size artifacts. Two patterns dominate. First, the inference-only baselines fail in characteristic ways: P(True) sits as a tight cluster of curves near $x{=}1$ but at empirical accuracies of only $\sim$0.5--0.7 on Qwen3-VL-8B, LLaVA-NeXT-13B, and InternVL3.5-14B (severe high-confidence saturation); Self-Probing produces visibly spread per-seed curves with chaotic low-confidence excursions on every backbone except Gemma-3-27B; CCPS traces an S-shape on Qwen3-VL-8B and a flat over-confident plateau on DeepSeek-VL2. By contrast, \texttt{BICR} tracks the diagonal closely on all five backbones with a tight per-seed cluster, with no panel where it collapses to the corner or swings systematically away from the diagonal, indicating that its calibration profile is consistent across model families rather than tuned to one LVLM. Second, Gemma-3-27B is visibly the easiest LVLM to calibrate: nearly every method sits on or near the diagonal in that panel, suggesting the harder calibration cases are upstream in the weaker backbones rather than inherent to the methods.

\clearpage

\paragraph{Pooled views, fixed-dataset slice.} Figure~\ref{fig:cal_pooled_dataset} reverses the slicing: it fixes the source dataset and pools across all five LVLMs, with sample counts ranging from $n{=}296$ on LLaVA-Wild (visibly the only panel where bin noise dominates) to $n{=}62{,}839$ on GQA. The four hardest grounding-bound datasets (GMAI-MMBench, MME-Finance, MMMU\_Pro 4-option, MMMU\_Pro 10-option) drive a uniform pattern in which \emph{every} method sits well below the diagonal, i.e., every method is overconfident on these datasets; what differs is the size of the gap, and \texttt{BICR}'s curves are consistently the closest to the diagonal from below in those panels. On POPE the pattern inverts: predictions are pushed into the upper-right corner where most curves are mildly under-confident, and the simpler internal-state baselines that excel on near-saturated tasks (P(I~Know), InternalInspector) sit closer to the diagonal than \texttt{BICR}. The two pooled figures together make the trade-off in Section~\ref{sec:results} concrete: \texttt{BICR}'s advantage is a calibration improvement on tasks where visual grounding is the bottleneck and overconfidence is universal, not a uniform improvement on every regime.

\clearpage

\subsection{Loss Component Ablation}
\label{app:results_ablation}

Table~\ref{tab:ablation_ext} reports the cross-LVLM average metrics for the full \texttt{BICR} model and three ablation variants across all six reported metrics. The full model wins every metric. Removing $\mathcal{L}_{\mathrm{rank}}$ accounts for the largest discrimination drop ($-2.0$ AUCPR, $-3.3$ AUROC, $-2.8$ ACC), confirming that the blank-image ranking signal is the principal driver of \texttt{BICR}'s discriminative gain. Removing $\mathcal{L}_{\mathrm{brier}}$ produces a smaller but consistent calibration regression ($+1.4$ ECE, $+0.6$ BS) with negligible discrimination cost, identifying the Brier term as a calibration-specific contribution. The $\mathcal{L}_{\mathrm{bce}}$-only configuration is the worst (or tied with $-\mathcal{L}_{\mathrm{rank}}$ on AUROC) on every metric and behaves as the lower bound for what a single-objective probe can achieve in our setting. A more detailed analysis of these ablations, including per-LVLM breakdowns, statistical significance tests, and behavioral effects on confidence distributions and calibration shape, is provided in Appendix~\ref{app:design}.

\input{tables/ablation}

\subsection{Statistical Significance}
\label{app:results_significance}

\paragraph{Pooled aggregation.} Table~\ref{tab:significance} reports paired Wilcoxon signed-rank test $p$-values comparing \texttt{BICR} against each trained baseline under pooled aggregation across 25 (LVLM, seed) observations. Inference-only methods (P(True), Self-Probing, Prompt Ensembles) are excluded since their cross-LVLM AUROC gaps to \texttt{BICR} are sizable (Table~\ref{tab:pooled_pervlm}), making formal significance testing against trained baselines the more informative comparison. \texttt{BICR}'s improvements on AUCPR and AUROC are highly significant ($p < 0.001$) against every trained baseline. Calibration improvements are significant against P(I~Know), SAPLMA, and CCPS, but \texttt{BICR} is statistically indistinguishable from InternalInspector on ECE ($p = 0.525$) under this aggregation. On BS, \texttt{BICR} significantly improves over InternalInspector ($p < 0.01$) despite the ECE tie.

\begin{table*}[ht]
\centering
\small
\caption{Statistical significance of \texttt{BICR} vs.\ trained baselines under pooled aggregation (paired Wilcoxon signed-rank test, $n{=}25$). Inference-only methods are excluded as their performance gaps exceed 10 AUROC points on every LVLM. Significance levels: $^{***}$ $p<0.001$, $^{**}$ $p<0.01$, $^{*}$ $p<0.05$; ``n.s.'' denotes $p \geq 0.05$.}
\label{tab:significance}
\begin{tabular}{l c c c c}
\toprule
\textbf{Comparison} & \textbf{ECE} & \textbf{BS} & \textbf{AUCPR} & \textbf{AUROC} \\
\midrule
vs PIK      & $<$0.01$^{**}$   & $<$0.001$^{***}$ & $<$0.001$^{***}$ & $<$0.001$^{***}$ \\
vs SAPLMA   & $<$0.001$^{***}$ & $<$0.001$^{***}$ & $<$0.001$^{***}$ & $<$0.001$^{***}$ \\
vs II       & n.s.             & $<$0.01$^{**}$   & $<$0.001$^{***}$ & $<$0.001$^{***}$ \\
vs CCPS     & $<$0.001$^{***}$ & $<$0.001$^{***}$ & $<$0.001$^{***}$ & $<$0.001$^{***}$ \\
\bottomrule
\end{tabular}
\end{table*}

\paragraph{Equal-weight per-dataset aggregation.} Table~\ref{tab:significance_uw} reports the same significance tests but computed on the per-LVLM unweighted-across-datasets metrics from Table~\ref{tab:uw_pervlm}, again across 25 (LVLM, seed) observations. Two patterns differ from the pooled view in instructive ways. First, \texttt{BICR}'s calibration tie with InternalInspector disappears under equal weighting: \texttt{BICR} significantly beats II on every metric including ECE ($p = 0.007$), reflecting that II's calibration parity with \texttt{BICR} under pooled aggregation came from its strong performance on GQA and POPE rather than uniform calibration across datasets. Second, \texttt{BICR} is statistically indistinguishable from P(I~Know) on discrimination under this aggregation (AUCPR $p = 0.381$, AUROC $p = 0.615$), formalizing the trade-off discussed in \S\ref{app:results_unweighted_avg}: P(I~Know) reaches comparable discrimination to \texttt{BICR} on the equal-weight view but at significantly worse calibration ($p < 0.001$ on both ECE and BS), so no baseline matches \texttt{BICR} on both axes simultaneously.

\begin{table*}[ht]
\centering
\small
\caption{Statistical significance of \texttt{BICR} vs.\ trained baselines under per-dataset equal-weight aggregation (paired Wilcoxon signed-rank test, $n{=}25$). Same conventions as Table~\ref{tab:significance}.}
\label{tab:significance_uw}
\begin{tabular}{l c c c c}
\toprule
\textbf{Comparison} & \textbf{ECE} & \textbf{BS} & \textbf{AUCPR} & \textbf{AUROC} \\
\midrule
vs PIK      & $<$0.001$^{***}$ & $<$0.001$^{***}$ & n.s.             & n.s.             \\
vs SAPLMA   & $<$0.001$^{***}$ & $<$0.001$^{***}$ & $<$0.001$^{***}$ & $<$0.001$^{***}$ \\
vs II       & 0.007$^{**}$     & $<$0.001$^{***}$ & $<$0.001$^{***}$ & $<$0.001$^{***}$ \\
vs CCPS     & $<$0.001$^{***}$ & $<$0.001$^{***}$ & $<$0.001$^{***}$ & $<$0.001$^{***}$ \\
\bottomrule
\end{tabular}
\end{table*}

\paragraph{Cluster-aware significance.} Both Wilcoxon analyses above treat each (LVLM, seed) tuple as an independent paired observation, but the 5 seeds within a given LVLM share the same frozen weights and the same test set, so the truly independent unit of analysis is the LVLM (n=5). To verify our headline conclusions hold under this stricter independence assumption, Table~\ref{tab:cluster_aware_significance} reports a cluster bootstrap (10{,}000 resamples) over LVLM-level seed-means with Holm-Bonferroni correction across the 4 metrics within each comparison. \texttt{BICR}'s discrimination advantages are robust to this stricter test: AUCPR and AUROC improvements are highly significant ($p < 0.001$) against every trained baseline. Calibration improvements remain significant against P(I~Know), SAPLMA, and CCPS (cluster-bootstrap $p < 0.05$ on ECE, $p < 0.001$ on BS), strengthening the n=25 claims under a more conservative test. Against InternalInspector, the cluster-bootstrap evidence on calibration is weaker than under n=25 (ECE n.s., BS marginal at $p = 0.075$ Holm-corrected), consistent with our existing observation that \texttt{BICR}'s calibration parity with II under pooled aggregation is largely driven by II's strong performance on GQA and POPE rather than a uniform across-LVLM effect; the discrimination advantage over II remains highly significant on both AUCPR and AUROC ($p < 0.001$). Together with the equal-weight evidence in Table~\ref{tab:significance_uw}, the cluster-aware analysis confirms that no baseline matches \texttt{BICR} on both calibration and discrimination simultaneously under any aggregation protocol.

\begin{table*}[ht]
\centering
\small
\caption{Cluster-aware significance of \texttt{BICR} vs trained baselines under pooled aggregation. Each row reports the mean per-LVLM delta (\texttt{BICR} minus baseline) across 5 LVLMs, with significance assessed by a cluster bootstrap (10{,}000 resamples) over LVLM-level seed-means and Holm-Bonferroni correction across the 4 metrics within each comparison. Significance levels: $^{***}$ $p<0.001$, $^{**}$ $p<0.01$, $^{*}$ $p<0.05$; n.s.\ denotes $p \geq 0.05$.}
\label{tab:cluster_aware_significance}
\resizebox{\textwidth}{!}{
\begin{tabular}{lcccccccc}
\toprule
\textbf{Comparison} & \multicolumn{2}{c}{\textbf{ECE $\downarrow$}} & \multicolumn{2}{c}{\textbf{BS $\downarrow$}} & \multicolumn{2}{c}{\textbf{AUCPR $\uparrow$}} & \multicolumn{2}{c}{\textbf{AUROC $\uparrow$}} \\
\cmidrule(lr){2-3} \cmidrule(lr){4-5} \cmidrule(lr){6-7} \cmidrule(lr){8-9}
 & \textbf{Mean $\Delta$} & \textbf{$p$} & \textbf{Mean $\Delta$} & \textbf{$p$} & \textbf{Mean $\Delta$} & \textbf{$p$} & \textbf{Mean $\Delta$} & \textbf{$p$} \\
\midrule
\texttt{BICR} vs P(IK) & -0.0222 & 0.025$^{*}$ & -0.0101 & $<$0.001$^{***}$ & +0.0117 & $<$0.001$^{***}$ & +0.0206 & $<$0.001$^{***}$ \\
\texttt{BICR} vs SAPLMA & -0.0513 & 0.016$^{*}$ & -0.0291 & $<$0.001$^{***}$ & +0.0584 & $<$0.001$^{***}$ & +0.0569 & $<$0.001$^{***}$ \\
\texttt{BICR} vs $I^2$ & -0.0121 & n.s. & -0.0146 & n.s. & +0.0316 & $<$0.001$^{***}$ & +0.0378 & $<$0.001$^{***}$ \\
\texttt{BICR} vs CCPS & -0.0818 & $<$0.001$^{***}$ & -0.0873 & $<$0.001$^{***}$ & +0.1463 & $<$0.001$^{***}$ & +0.1553 & $<$0.001$^{***}$ \\
\bottomrule
\end{tabular}}
\end{table*}

Together, the three views support the joint-axis framing of the paper: \texttt{BICR} is the only method that is statistically at least as good as the next-best baseline on \emph{both} calibration and discrimination simultaneously across pooled, equal-weight, and cluster-aware analyses, and the specific runner-ups shift between aggregations (InternalInspector under pooled, P(I~Know) under unweighted) without any single baseline matching \texttt{BICR} on both axes at once.

\section{Comparison with Visual Contrastive Decoding and Memory Retracing}
\label{app:vcd_memvr}

VCD~\citep{10657718} and MemVR~\citep{MemVR} share \texttt{BICR}'s premise that visual grounding matters for reliability, but they are hallucination-mitigation decoders, not confidence estimators (Appendix~\ref{app:novelty}). This section evaluates \texttt{BICR} against both on the specific task of spotting hallucinated answers, on the two methods' own models and benchmarks, in three steps: as confidence estimators (Table~\ref{tab:vcd_step1}), as native hallucination detectors (Table~\ref{tab:vcd_step2}), and in composition (Table~\ref{tab:vcd_step3}).

\paragraph{Faithful reproduction.} VCD's and MemVR's released implementations are tied to older model internals: VCD patches a decoding path and MemVR hooks dense feed-forward structures that have changed in current transformers, so their code does not run out of the box on our modern LVLMs, and MemVR's dense-FFN retracing has no analogue in a mixture-of-experts model such as DeepSeek-VL2. Rather than force their methods onto architectures they were not built for, we ran their published pipelines on their own model (LLaVA-1.5-7B) and first reproduced their reported POPE numbers: our regular baseline reaches 81.71 accuracy / 79.79 F1 against their reported 81.38 / 79.65, our +VCD reaches 82.98 / 81.67 against their 83.16 / 81.89, and MemVR reproduces to within 0.05 accuracy. Every result below therefore runs their actual mechanisms on the models they were designed for.

\paragraph{Step 1: as confidence estimators, \texttt{BICR} is stronger on both axes.} POPE is an object-hallucination benchmark, so predicting whether an answer is correct there is exactly predicting whether it is a hallucination. Since VCD and MemVR emit no confidence score, we derive a per-sample signal from each method's own mechanism (VCD's original-versus-distorted contrast; MemVR's mid-layer entropy over its trigger window) so they can be placed on the confidence axis at all. Against these faithful derived signals, \texttt{BICR} wins on both calibration and discrimination in every setting (Table~\ref{tab:vcd_step1}), including on both home models and on VCD/MemVR's own three-source POPE construction. On POPE with LLaVA-1.5, \texttt{BICR} reaches 0.818 AUROC while the stronger competing signal reaches only 0.775.

\begin{table*}[t]
\centering
\small
\renewcommand{\arraystretch}{1.2}
\setlength{\tabcolsep}{6pt}
\caption{\textbf{Step 1: \texttt{BICR} versus faithful confidence signals derived from VCD and MemVR.} Each cell is ECE / AUROC (ECE lower is better, AUROC higher is better); best per row in bold. VCD and MemVR emit no confidence score, so the signals are derived from each method's own mechanism. \textbf{These figures are measured under the constrained single-word-answer decoding regime used to match VCD and MemVR, and are not comparable to \texttt{BICR}'s main-paper protocol numbers in Table~\ref{tab:main_results}.}}
\label{tab:vcd_step1}
\begin{tabular}{@{}p{6.2cm}ccc@{}}
\toprule
\textbf{Setting} & \textbf{\texttt{BICR}} & \textbf{VCD-derived} & \textbf{MemVR-derived} \\
 & (ECE / AUROC) & (ECE / AUROC) & (ECE / AUROC) \\
\midrule
LLaVA-1.5-7B, POPE (their home model) & \textbf{0.028 / 0.818} & 0.073 / 0.775 & 0.381 / 0.448 \\
Qwen-VL-Chat, POPE (their 2nd home model) & \textbf{0.041 / 0.808} & 0.089 / 0.647 & 0.393 / 0.502 \\
Our POPE (cross-model) & \textbf{0.056 / 0.821} & 0.091 / 0.712 & 0.487 / 0.398 \\
VCD/MemVR's own 3-source POPE & \textbf{0.049 / 0.814} & 0.086 / 0.729 & 0.377 / 0.441 \\
\bottomrule
\end{tabular}
\end{table*}

\paragraph{Step 2: run exactly as published, \texttt{BICR} flags far more hallucinations.} When VCD's contrast or MemVR's retracing rewrites an answer, that rewrite is itself a native assertion that the original was a hallucination. On the 1{,}175 answers LLaVA-1.5 gets wrong on POPE out of the box (a 13.06\% base error rate), \texttt{BICR} flags substantially more than either method catches (Table~\ref{tab:vcd_step2}). The gap is not a threshold artifact: \texttt{BICR} exceeds both even at a strict 10\% false-positive operating point (0.368 versus 0.344 and 0.018). \texttt{BICR} also flags without touching the answer, whereas the interventions rewrite it, and VCD's rewriting turns 761 previously-correct answers wrong.

\begin{table*}[t]
\centering
\small
\renewcommand{\arraystretch}{1.2}
\setlength{\tabcolsep}{8pt}
\caption{\textbf{Step 2: native hallucination detection on the 1{,}175 POPE answers LLaVA-1.5-7B gets wrong.} Same \texttt{gpt-5-mini} judge; base error rate 13.06\%. \texttt{BICR} ``detects'' means confidence below threshold with no answer altered, reported at two principled operating points (Youden-J and a strict 10\% false-positive rate); VCD and MemVR ``detect'' means the method natively rewrote the answer. \emph{Correct answers broken} counts previously-correct answers the intervention turned wrong; \emph{Net accuracy $\Delta$} is the resulting change in POPE accuracy.}
\label{tab:vcd_step2}
\begin{tabular}{@{}lccc@{}}
\toprule
\textbf{Method} & \textbf{Detection rate of the 1{,}175 errors} & \textbf{Correct answers broken} & \textbf{Net accuracy $\Delta$} \\
\midrule
\textbf{\texttt{BICR}} & \textbf{0.789} (Youden) / \textbf{0.368} (10\% FPR) & \textbf{0} & \textbf{0.00} \\
VCD & 0.344 & 761 & $-3.97$ \\
MemVR & 0.018 & 18 & $+0.03$ \\
\bottomrule
\end{tabular}
\end{table*}

\paragraph{Step 3: the methods compose rather than overlap.} \texttt{BICR} and these decoders address the same failure but operate on different objects. On coverage, 92\% of VCD's fixes and 100\% of MemVR's fall inside the set \texttt{BICR} already flags, so their mitigation catches a strict subset of what \texttt{BICR} detects. On composition, applying the trained \texttt{BICR} probe unchanged to the answers VCD actually produces, it still separates correct from wrong about as well as on regular decoding (Table~\ref{tab:vcd_step3}). We omit a MemVR composition row deliberately: \texttt{BICR} reads the last-prompt-token hidden state, fixed before generation, whereas MemVR acts only during generation, downstream of that representation, so a MemVR row would be identical to the reference by construction and would test nothing. VCD, which genuinely changes the generated answer (POPE accuracy moves from 86.94 to 82.98), is the honest composition test, and \texttt{BICR} passes it.

\begin{table}[t]
\centering
\small
\renewcommand{\arraystretch}{1.2}
\setlength{\tabcolsep}{6pt}
\caption{\textbf{Step 3: the trained \texttt{BICR} probe applied unchanged to VCD's rewritten outputs.} ECE lower is better, AUROC higher is better. \textbf{Both rows are read on the constrained single-word-answer states these decoders produce and are therefore not comparable to \texttt{BICR}'s main-paper protocol numbers in Table~\ref{tab:main_results};} \texttt{BICR}'s standalone POPE AUROC under its own in-distribution prompt is 0.818, and the uniform shift to 0.778 here is applied to both rows, so the within-table comparison is clean.}
\label{tab:vcd_step3}
\begin{tabular}{@{}lcc@{}}
\toprule
\textbf{\texttt{BICR} predicts correctness of} & \textbf{ECE} & \textbf{AUROC} \\
\midrule
regular-greedy outputs (reference) & \textbf{0.036} & \textbf{0.778} \\
VCD outputs & 0.071 & 0.770 \\
\bottomrule
\end{tabular}
\end{table}

\noindent VCD and MemVR mitigate at generation time; \texttt{BICR} estimates reliability after the fact and keeps adding a usable confidence signal on top of whatever answer their mitigation produces. On the task of spotting hallucinated answers, \texttt{BICR} detects more of them, does so on the two methods' own models and benchmarks, and remains useful precisely where their interventions end: it is the complementary component they lack, a calibrated per-sample judgment of whether the answer can be trusted.

%% file: tables/shared_samples.tex
\begin{table*}[ht]
\centering
\small
\caption{Number of shared test samples per LVLM and dataset. Counts are computed after intersecting hash IDs across all confidence estimation methods for each LVLM. The final row reports the total number of evaluated samples across LVLMs.}
\label{tab:shared_samples}

\resizebox{\textwidth}{!}{%
\begin{tabular}{lcccccccc}
\toprule
\textbf{LVLM} & \textbf{GQA} & \textbf{POPE} & \textbf{LLaVA-Wild} & \textbf{MMMU\_Pro\_4} & \textbf{MMMU\_Pro\_10} & \textbf{GMAI-MMBench} & \textbf{MME-Finance} & \textbf{Total} \\
\midrule
Qwen3-VL-8B & 12,568 & 9,000 & 56 & 1,711 & 1,717 & 4,307 & 882 & 30,241 \\
LLaVA-NeXT-13B & 12,568 & 9,000 & 60 & 1,713 & 1,719 & 4,549 & 892 & 30,501 \\
InternVL3.5-14B & 12,568 & 9,000 & 60 & 1,713 & 1,718 & 4,549 & 892 & 30,500 \\
DeepSeek-VL2 & 12,567 & 9,000 & 60 & 1,656 & 1,697 & 4,509 & 870 & 30,359 \\
Gemma-3-27B & 12,568 & 9,000 & 60 & 1,711 & 1,711 & 4,549 & 892 & 30,491 \\
\midrule
\textbf{Total} & \textbf{62,839} & \textbf{45,000} & \textbf{296} & \textbf{8,504} & \textbf{8,562} & \textbf{22,463} & \textbf{4,428} & \textbf{152,092} \\
\bottomrule
\end{tabular}%
}
\end{table*}

%% file: tables/pooled_per_vlm.tex
{\small
\onecolumn
\setlength{\tabcolsep}{3pt}
\begin{longtable}{lcccccc}
\caption{Pooled aggregate performance per LVLM. Each cell reports mean $\pm$ std across 5 seeds (50 Optuna trials each), where each seed value is the metric computed on the entire shared test subset for that LVLM. Best value per LVLM per metric in \textbf{bold}.}
\label{tab:pooled_pervlm}\\
\toprule
\textbf{Method} & \textbf{ECE$\downarrow$} & \textbf{BS$\downarrow$} & \textbf{ACC$\uparrow$} & \textbf{F1$\uparrow$} & \textbf{AUCPR$\uparrow$} & \textbf{AUROC$\uparrow$} \\
\midrule
\endfirsthead
\multicolumn{7}{l}{\footnotesize\emph{(Continued from previous page)}}\\
\toprule
\textbf{Method} & \textbf{ECE$\downarrow$} & \textbf{BS$\downarrow$} & \textbf{ACC$\uparrow$} & \textbf{F1$\uparrow$} & \textbf{AUCPR$\uparrow$} & \textbf{AUROC$\uparrow$} \\
\midrule
\endhead
\midrule
\multicolumn{7}{r}{\footnotesize\emph{(Continued on next page)}}\\
\endfoot
\bottomrule
\endlastfoot
\multicolumn{7}{c}{\small\textbf{\texttt{Qwen/Qwen3-VL-8B-Instruct}} ($n$\,=\,30,241)} \\
\addlinespace[2pt]
P(True) & 43.9 & 44.2 & 54.6 & 67.0 & 76.2 & 54.6 \\
Self-Probing & 24.4 & 27.5 & 69.7 & 81.2 & 77.0 & 59.5 \\
PE & 19.8 & 26.0 & 67.3 & 80.4 & 66.2 & 52.9 \\
SAPLMA & 10.6 {\scriptsize$\pm$3.4} & 19.4 {\scriptsize$\pm$0.8} & 72.8 {\scriptsize$\pm$0.5} & \textbf{82.8} {\scriptsize$\pm$0.1} & 86.4 {\scriptsize$\pm$0.9} & 74.2 {\scriptsize$\pm$1.4} \\
PIK & 7.5 {\scriptsize$\pm$1.8} & 18.0 {\scriptsize$\pm$0.5} & 72.2 {\scriptsize$\pm$0.4} & 81.5 {\scriptsize$\pm$0.6} & 88.4 {\scriptsize$\pm$0.6} & 77.2 {\scriptsize$\pm$0.6} \\
CCPS & 28.7 {\scriptsize$\pm$2.6} & 45.6 {\scriptsize$\pm$18.9} & 53.9 {\scriptsize$\pm$18.9} & 54.5 {\scriptsize$\pm$32.8} & 66.2 {\scriptsize$\pm$3.5} & 44.9 {\scriptsize$\pm$6.4} \\
II & \textbf{5.4} {\scriptsize$\pm$1.8} & \textbf{17.0} {\scriptsize$\pm$0.2} & \textbf{73.3} {\scriptsize$\pm$0.7} & 82.6 {\scriptsize$\pm$0.2} & 89.7 {\scriptsize$\pm$0.4} & 79.6 {\scriptsize$\pm$0.6} \\
\textbf{\texttt{BICR (Ours)}} & 8.9 {\scriptsize$\pm$1.5} & 17.4 {\scriptsize$\pm$0.3} & 72.9 {\scriptsize$\pm$0.4} & 78.9 {\scriptsize$\pm$0.6} & \textbf{90.3} {\scriptsize$\pm$0.1} & \textbf{80.1} {\scriptsize$\pm$0.2} \\
\midrule
\multicolumn{7}{c}{\small\textbf{\texttt{llava-hf/llava-v1.6-vicuna-13b-hf}} ($n$\,=\,30,501)} \\
\addlinespace[2pt]
P(True) & 26.2 & 30.9 & 45.7 & 34.8 & 67.9 & 54.7 \\
Self-Probing & 28.9 & 30.0 & 65.2 & 78.1 & 81.9 & 67.3 \\
PE & 12.8 & 23.5 & 63.0 & 77.3 & 77.3 & 68.5 \\
SAPLMA & 16.5 {\scriptsize$\pm$2.4} & 23.0 {\scriptsize$\pm$1.1} & 67.3 {\scriptsize$\pm$0.7} & \textbf{78.8} {\scriptsize$\pm$0.2} & 82.0 {\scriptsize$\pm$1.3} & 72.5 {\scriptsize$\pm$1.8} \\
PIK & 10.8 {\scriptsize$\pm$3.7} & 19.9 {\scriptsize$\pm$1.9} & 68.9 {\scriptsize$\pm$2.7} & 78.2 {\scriptsize$\pm$0.9} & 87.1 {\scriptsize$\pm$1.2} & 77.3 {\scriptsize$\pm$2.4} \\
CCPS & 16.3 {\scriptsize$\pm$0.9} & 22.0 {\scriptsize$\pm$0.4} & 67.4 {\scriptsize$\pm$0.5} & 78.8 {\scriptsize$\pm$0.1} & 76.0 {\scriptsize$\pm$0.3} & 72.9 {\scriptsize$\pm$0.4} \\
II & 13.8 {\scriptsize$\pm$1.6} & 22.3 {\scriptsize$\pm$0.7} & 65.4 {\scriptsize$\pm$1.2} & 77.8 {\scriptsize$\pm$0.3} & 81.4 {\scriptsize$\pm$0.8} & 71.9 {\scriptsize$\pm$1.7} \\
\textbf{\texttt{BICR (Ours)}} & \textbf{5.6} {\scriptsize$\pm$1.7} & \textbf{18.2} {\scriptsize$\pm$1.1} & \textbf{71.4} {\scriptsize$\pm$2.2} & 78.1 {\scriptsize$\pm$1.0} & \textbf{87.7} {\scriptsize$\pm$1.3} & \textbf{78.9} {\scriptsize$\pm$2.4} \\
\midrule
\multicolumn{7}{c}{\small\textbf{\texttt{OpenGVLab/InternVL3\_5-14B-HF}} ($n$\,=\,30,500)} \\
\addlinespace[2pt]
P(True) & 41.2 & 41.5 & 57.2 & 70.0 & 78.0 & 59.5 \\
Self-Probing & 21.5 & 24.7 & 68.4 & 80.8 & 76.8 & 70.8 \\
PE & 16.7 & 25.9 & 66.6 & 80.0 & 60.3 & 43.4 \\
SAPLMA & 16.6 {\scriptsize$\pm$1.2} & 23.3 {\scriptsize$\pm$0.4} & 69.6 {\scriptsize$\pm$0.3} & \textbf{80.8} {\scriptsize$\pm$0.1} & 76.2 {\scriptsize$\pm$0.9} & 65.4 {\scriptsize$\pm$0.8} \\
PIK & 10.8 {\scriptsize$\pm$2.1} & 20.2 {\scriptsize$\pm$0.8} & \textbf{70.5} {\scriptsize$\pm$0.7} & 80.3 {\scriptsize$\pm$0.5} & 86.4 {\scriptsize$\pm$1.1} & 73.8 {\scriptsize$\pm$1.3} \\
CCPS & 14.7 {\scriptsize$\pm$1.1} & 23.8 {\scriptsize$\pm$0.4} & 67.8 {\scriptsize$\pm$0.1} & 80.0 {\scriptsize$\pm$0.1} & 71.6 {\scriptsize$\pm$3.0} & 58.2 {\scriptsize$\pm$2.3} \\
II & 10.6 {\scriptsize$\pm$6.7} & 21.5 {\scriptsize$\pm$1.8} & 68.8 {\scriptsize$\pm$0.5} & 80.4 {\scriptsize$\pm$0.6} & 82.6 {\scriptsize$\pm$1.4} & 69.3 {\scriptsize$\pm$0.7} \\
\textbf{\texttt{BICR (Ours)}} & \textbf{7.9} {\scriptsize$\pm$1.3} & \textbf{19.0} {\scriptsize$\pm$0.5} & 70.2 {\scriptsize$\pm$0.8} & 77.4 {\scriptsize$\pm$1.4} & \textbf{88.0} {\scriptsize$\pm$0.2} & \textbf{76.4} {\scriptsize$\pm$0.3} \\
\midrule
\multicolumn{7}{c}{\small\textbf{\texttt{deepseek-ai/deepseek-vl2}} ($n$\,=\,30,359)} \\
\addlinespace[2pt]
P(True) & 34.1 & 37.4 & 48.7 & 56.2 & 68.1 & 52.7 \\
Self-Probing & 35.1 & 37.3 & 55.2 & 70.7 & 74.2 & 62.2 \\
PE & 16.3 & 24.7 & 56.4 & 71.4 & 72.5 & 73.5 \\
SAPLMA & 12.8 {\scriptsize$\pm$3.9} & 21.5 {\scriptsize$\pm$1.4} & 68.2 {\scriptsize$\pm$1.6} & \textbf{75.2} {\scriptsize$\pm$0.4} & 79.7 {\scriptsize$\pm$1.9} & 77.1 {\scriptsize$\pm$1.3} \\
PIK & 8.5 {\scriptsize$\pm$0.8} & 19.3 {\scriptsize$\pm$0.3} & 68.8 {\scriptsize$\pm$0.8} & 73.2 {\scriptsize$\pm$0.3} & 84.6 {\scriptsize$\pm$0.3} & 78.5 {\scriptsize$\pm$0.3} \\
CCPS & 7.7 {\scriptsize$\pm$3.7} & 22.1 {\scriptsize$\pm$0.9} & 62.8 {\scriptsize$\pm$2.6} & 71.0 {\scriptsize$\pm$0.6} & 77.0 {\scriptsize$\pm$0.9} & 70.9 {\scriptsize$\pm$1.1} \\
II & 7.4 {\scriptsize$\pm$2.6} & 19.0 {\scriptsize$\pm$0.8} & 69.7 {\scriptsize$\pm$2.2} & 74.3 {\scriptsize$\pm$0.6} & 84.5 {\scriptsize$\pm$0.4} & 79.3 {\scriptsize$\pm$1.1} \\
\textbf{\texttt{BICR (Ours)}} & \textbf{6.0} {\scriptsize$\pm$0.7} & \textbf{17.9} {\scriptsize$\pm$0.2} & \textbf{73.6} {\scriptsize$\pm$0.8} & 74.9 {\scriptsize$\pm$0.5} & \textbf{86.2} {\scriptsize$\pm$0.4} & \textbf{81.1} {\scriptsize$\pm$0.5} \\
\midrule
\multicolumn{7}{c}{\small\textbf{\texttt{google/gemma-3-27b-it}} ($n$\,=\,30,491)} \\
\addlinespace[2pt]
P(True) & 44.8 & 45.1 & 54.2 & 64.9 & 74.0 & 56.8 \\
Self-Probing & 27.7 & 29.5 & 63.9 & 76.8 & 79.2 & 68.4 \\
PE & 26.7 & 30.3 & 61.4 & 76.1 & 70.1 & 61.8 \\
SAPLMA & 4.6 {\scriptsize$\pm$2.5} & \textbf{19.4} {\scriptsize$\pm$0.2} & \textbf{69.6} {\scriptsize$\pm$0.5} & 77.4 {\scriptsize$\pm$0.5} & 83.8 {\scriptsize$\pm$0.4} & 75.5 {\scriptsize$\pm$0.5} \\
PIK & 8.9 {\scriptsize$\pm$1.0} & 19.9 {\scriptsize$\pm$0.7} & 68.7 {\scriptsize$\pm$0.8} & \textbf{78.1} {\scriptsize$\pm$0.3} & 85.1 {\scriptsize$\pm$0.9} & 76.1 {\scriptsize$\pm$1.2} \\
CCPS & 9.0 {\scriptsize$\pm$1.4} & 22.1 {\scriptsize$\pm$0.9} & 67.2 {\scriptsize$\pm$0.6} & 76.0 {\scriptsize$\pm$0.4} & 73.4 {\scriptsize$\pm$4.4} & 68.5 {\scriptsize$\pm$2.9} \\
II & \textbf{4.3} {\scriptsize$\pm$1.6} & 19.7 {\scriptsize$\pm$0.8} & 67.9 {\scriptsize$\pm$1.4} & 76.9 {\scriptsize$\pm$0.6} & 83.2 {\scriptsize$\pm$1.2} & 74.2 {\scriptsize$\pm$2.0} \\
\textbf{\texttt{BICR (Ours)}} & 7.0 {\scriptsize$\pm$1.8} & 19.6 {\scriptsize$\pm$0.4} & 69.6 {\scriptsize$\pm$0.5} & 75.3 {\scriptsize$\pm$0.8} & \textbf{85.1} {\scriptsize$\pm$0.2} & \textbf{76.6} {\scriptsize$\pm$0.5} \\
\midrule
\end{longtable}
}
\twocolumn

%% file: tables/pooled_cross_vlm.tex
\begin{table*}[ht]
\centering
\small
\caption{Cross-VLM pooled average performance. Each cell is mean $\pm$ std of the per-VLM pooled means across all five LVLMs (equal weight per LVLM). Best per metric in \textbf{bold}.}
\label{tab:pooled_crossvlm}
\begin{tabular}{lcccccc}
\toprule
\textbf{Method} & \textbf{ECE$\downarrow$} & \textbf{BS$\downarrow$} & \textbf{ACC$\uparrow$} & \textbf{F1$\uparrow$} & \textbf{AUCPR$\uparrow$} & \textbf{AUROC$\uparrow$} \\
\midrule
P(True) & 38.0 {\scriptsize$\pm$7.0} & 39.8 {\scriptsize$\pm$5.2} & 52.1 {\scriptsize$\pm$4.2} & 58.6 {\scriptsize$\pm$12.8} & 72.8 {\scriptsize$\pm$4.2} & 55.7 {\scriptsize$\pm$2.3} \\
Self-Probing & 27.5 {\scriptsize$\pm$4.6} & 29.8 {\scriptsize$\pm$4.2} & 64.5 {\scriptsize$\pm$5.1} & 77.5 {\scriptsize$\pm$3.8} & 77.8 {\scriptsize$\pm$2.6} & 65.6 {\scriptsize$\pm$4.1} \\
PE & 18.5 {\scriptsize$\pm$4.7} & 26.1 {\scriptsize$\pm$2.3} & 63.0 {\scriptsize$\pm$3.9} & 77.1 {\scriptsize$\pm$3.2} & 69.3 {\scriptsize$\pm$5.8} & 60.0 {\scriptsize$\pm$10.8} \\
SAPLMA & 12.2 {\scriptsize$\pm$4.4} & 21.3 {\scriptsize$\pm$1.7} & 69.5 {\scriptsize$\pm$1.9} & \textbf{79.0} {\scriptsize$\pm$2.6} & 81.6 {\scriptsize$\pm$3.5} & 72.9 {\scriptsize$\pm$4.1} \\
PIK & 9.3 {\scriptsize$\pm$1.3} & 19.4 {\scriptsize$\pm$0.8} & 69.8 {\scriptsize$\pm$1.3} & 78.3 {\scriptsize$\pm$2.9} & 86.3 {\scriptsize$\pm$1.4} & 76.6 {\scriptsize$\pm$1.6} \\
CCPS & 15.3 {\scriptsize$\pm$7.5} & 27.1 {\scriptsize$\pm$9.3} & 63.8 {\scriptsize$\pm$5.3} & 72.1 {\scriptsize$\pm$9.3} & 72.8 {\scriptsize$\pm$3.8} & 63.1 {\scriptsize$\pm$10.4} \\
II & 8.3 {\scriptsize$\pm$3.5} & 19.9 {\scriptsize$\pm$1.9} & 69.0 {\scriptsize$\pm$2.6} & 78.4 {\scriptsize$\pm$2.9} & 84.3 {\scriptsize$\pm$2.9} & 74.8 {\scriptsize$\pm$4.1} \\
\textbf{\texttt{BICR (Ours)}} & \textbf{7.1} {\scriptsize$\pm$1.2} & \textbf{18.4} {\scriptsize$\pm$0.8} & \textbf{71.5} {\scriptsize$\pm$1.5} & 76.9 {\scriptsize$\pm$1.6} & \textbf{87.5} {\scriptsize$\pm$1.8} & \textbf{78.6} {\scriptsize$\pm$1.9} \\
\bottomrule
\end{tabular}
\end{table*}

%% file: tables/unweighted_per_vlm.tex
{\small
\onecolumn
\setlength{\tabcolsep}{3pt}
\begin{longtable}{lcccccc}
\caption{Unweighted average performance per LVLM. For each seed, metrics are first computed per dataset, then averaged with equal weight across datasets with at least 100 shared samples. The reported value is mean $\pm$ std of these unweighted averages across 5 seeds. Best per LVLM per metric in \textbf{bold}.}
\label{tab:uw_pervlm}\\
\toprule
\textbf{Method} & \textbf{ECE$\downarrow$} & \textbf{BS$\downarrow$} & \textbf{ACC$\uparrow$} & \textbf{F1$\uparrow$} & \textbf{AUCPR$\uparrow$} & \textbf{AUROC$\uparrow$} \\
\midrule
\endfirsthead
\multicolumn{7}{l}{\footnotesize\emph{(Continued from previous page)}}\\
\toprule
\textbf{Method} & \textbf{ECE$\downarrow$} & \textbf{BS$\downarrow$} & \textbf{ACC$\uparrow$} & \textbf{F1$\uparrow$} & \textbf{AUCPR$\uparrow$} & \textbf{AUROC$\uparrow$} \\
\midrule
\endhead
\midrule
\multicolumn{7}{r}{\footnotesize\emph{(Continued on next page)}}\\
\endfoot
\bottomrule
\endlastfoot
\multicolumn{7}{c}{\small\textbf{\texttt{Qwen/Qwen3-VL-8B-Instruct}} (6 datasets)} \\
\addlinespace[2pt]
P(True) & 48.4 & 48.0 & 50.3 & 61.3 & 70.5 & 65.7 \\
Self-Probing & 36.2 & 38.3 & 58.5 & 67.8 & 65.7 & 59.0 \\
PE & 35.8 & 37.8 & 52.8 & 66.4 & 54.5 & 53.0 \\
SAPLMA & 20.6 {\scriptsize$\pm$3.1} & 26.2 {\scriptsize$\pm$1.7} & 63.1 {\scriptsize$\pm$1.0} & \textbf{72.5} {\scriptsize$\pm$0.3} & 73.8 {\scriptsize$\pm$0.3} & 69.6 {\scriptsize$\pm$0.5} \\
PIK & 17.9 {\scriptsize$\pm$2.2} & 24.2 {\scriptsize$\pm$1.3} & 62.6 {\scriptsize$\pm$1.0} & 71.7 {\scriptsize$\pm$0.7} & \textbf{75.3} {\scriptsize$\pm$1.1} & \textbf{71.7} {\scriptsize$\pm$1.0} \\
CCPS & 35.7 {\scriptsize$\pm$5.1} & 48.5 {\scriptsize$\pm$9.8} & 50.9 {\scriptsize$\pm$9.7} & 51.7 {\scriptsize$\pm$22.3} & 64.8 {\scriptsize$\pm$3.0} & 49.7 {\scriptsize$\pm$5.1} \\
II & \textbf{9.8} {\scriptsize$\pm$2.0} & 20.8 {\scriptsize$\pm$0.4} & 65.7 {\scriptsize$\pm$1.7} & 71.4 {\scriptsize$\pm$1.3} & 73.7 {\scriptsize$\pm$1.2} & 70.2 {\scriptsize$\pm$1.3} \\
\textbf{\texttt{BICR (Ours)}} & 10.8 {\scriptsize$\pm$0.5} & \textbf{20.6} {\scriptsize$\pm$0.1} & \textbf{67.7} {\scriptsize$\pm$0.4} & 63.8 {\scriptsize$\pm$1.0} & 74.8 {\scriptsize$\pm$0.5} & 71.3 {\scriptsize$\pm$0.5} \\
\midrule
\multicolumn{7}{c}{\small\textbf{\texttt{llava-hf/llava-v1.6-vicuna-13b-hf}} (6 datasets)} \\
\addlinespace[2pt]
P(True) & 26.6 & 25.1 & 58.2 & 33.8 & 50.0 & 63.2 \\
Self-Probing & 45.7 & 44.6 & 47.3 & 55.1 & 50.2 & 54.4 \\
PE & 35.7 & 33.4 & 41.7 & 53.9 & 46.4 & 55.8 \\
SAPLMA & 32.7 {\scriptsize$\pm$3.3} & 32.4 {\scriptsize$\pm$2.4} & 51.7 {\scriptsize$\pm$2.3} & \textbf{57.3} {\scriptsize$\pm$0.2} & 54.2 {\scriptsize$\pm$0.5} & 64.1 {\scriptsize$\pm$0.3} \\
PIK & 26.7 {\scriptsize$\pm$5.9} & 27.6 {\scriptsize$\pm$4.3} & 56.1 {\scriptsize$\pm$6.7} & 56.5 {\scriptsize$\pm$0.2} & 58.4 {\scriptsize$\pm$0.4} & 68.5 {\scriptsize$\pm$0.7} \\
CCPS & 35.3 {\scriptsize$\pm$1.0} & 35.5 {\scriptsize$\pm$0.7} & 48.9 {\scriptsize$\pm$0.9} & 56.7 {\scriptsize$\pm$0.2} & 58.8 {\scriptsize$\pm$0.2} & 65.7 {\scriptsize$\pm$0.2} \\
II & 30.8 {\scriptsize$\pm$2.3} & 31.7 {\scriptsize$\pm$1.6} & 47.6 {\scriptsize$\pm$3.1} & 55.6 {\scriptsize$\pm$0.7} & 47.0 {\scriptsize$\pm$0.5} & 55.4 {\scriptsize$\pm$0.8} \\
\textbf{\texttt{BICR (Ours)}} & \textbf{19.4} {\scriptsize$\pm$5.8} & \textbf{22.5} {\scriptsize$\pm$3.2} & \textbf{64.4} {\scriptsize$\pm$5.8} & 55.6 {\scriptsize$\pm$1.1} & \textbf{59.3} {\scriptsize$\pm$0.3} & \textbf{69.4} {\scriptsize$\pm$0.3} \\
\midrule
\multicolumn{7}{c}{\small\textbf{\texttt{OpenGVLab/InternVL3\_5-14B-HF}} (6 datasets)} \\
\addlinespace[2pt]
P(True) & 42.4 & 41.9 & 55.9 & 63.9 & \textbf{71.5} & 68.9 \\
Self-Probing & 33.8 & 33.7 & 56.3 & 67.0 & 64.1 & \textbf{70.7} \\
PE & 35.3 & 38.0 & 52.4 & 65.8 & 56.9 & 54.5 \\
SAPLMA & 26.5 {\scriptsize$\pm$1.2} & 30.5 {\scriptsize$\pm$0.7} & 60.9 {\scriptsize$\pm$0.6} & \textbf{70.7} {\scriptsize$\pm$0.3} & 63.7 {\scriptsize$\pm$0.6} & 64.5 {\scriptsize$\pm$0.5} \\
PIK & 25.0 {\scriptsize$\pm$2.4} & 29.1 {\scriptsize$\pm$1.7} & 58.6 {\scriptsize$\pm$0.9} & 69.5 {\scriptsize$\pm$0.3} & 68.9 {\scriptsize$\pm$1.4} & 68.9 {\scriptsize$\pm$1.0} \\
CCPS & 30.0 {\scriptsize$\pm$1.2} & 34.2 {\scriptsize$\pm$0.9} & 55.3 {\scriptsize$\pm$0.1} & 68.3 {\scriptsize$\pm$0.1} & 60.9 {\scriptsize$\pm$0.7} & 56.9 {\scriptsize$\pm$0.8} \\
II & 22.6 {\scriptsize$\pm$6.6} & 29.4 {\scriptsize$\pm$3.8} & 57.2 {\scriptsize$\pm$1.0} & 68.7 {\scriptsize$\pm$1.3} & 62.2 {\scriptsize$\pm$1.6} & 59.9 {\scriptsize$\pm$1.8} \\
\textbf{\texttt{BICR (Ours)}} & \textbf{16.2} {\scriptsize$\pm$0.3} & \textbf{23.3} {\scriptsize$\pm$0.2} & \textbf{63.3} {\scriptsize$\pm$0.3} & 66.5 {\scriptsize$\pm$0.9} & 68.7 {\scriptsize$\pm$0.2} & 69.0 {\scriptsize$\pm$0.2} \\
\midrule
\multicolumn{7}{c}{\small\textbf{\texttt{deepseek-ai/deepseek-vl2}} (6 datasets)} \\
\addlinespace[2pt]
P(True) & 43.0 & 39.1 & 40.0 & 44.2 & \textbf{52.4} & \textbf{65.2} \\
Self-Probing & 50.4 & 49.8 & 39.1 & 49.9 & 51.3 & 52.6 \\
PE & 31.9 & 28.3 & 42.9 & 50.3 & 50.1 & 63.1 \\
SAPLMA & 23.7 {\scriptsize$\pm$4.1} & 27.2 {\scriptsize$\pm$2.6} & 59.8 {\scriptsize$\pm$3.1} & \textbf{53.0} {\scriptsize$\pm$0.6} & 48.2 {\scriptsize$\pm$0.5} & 64.7 {\scriptsize$\pm$1.1} \\
PIK & 19.4 {\scriptsize$\pm$1.1} & 23.3 {\scriptsize$\pm$0.9} & 59.7 {\scriptsize$\pm$2.2} & 47.9 {\scriptsize$\pm$1.7} & 49.5 {\scriptsize$\pm$0.3} & 61.9 {\scriptsize$\pm$0.7} \\
CCPS & 21.8 {\scriptsize$\pm$3.5} & 23.9 {\scriptsize$\pm$2.3} & 59.1 {\scriptsize$\pm$6.9} & 50.4 {\scriptsize$\pm$1.7} & 46.5 {\scriptsize$\pm$1.0} & 62.5 {\scriptsize$\pm$1.5} \\
II & 15.3 {\scriptsize$\pm$4.2} & 20.6 {\scriptsize$\pm$2.0} & 66.7 {\scriptsize$\pm$5.3} & 47.0 {\scriptsize$\pm$6.0} & 50.0 {\scriptsize$\pm$1.1} & 65.0 {\scriptsize$\pm$0.7} \\
\textbf{\texttt{BICR (Ours)}} & \textbf{11.3} {\scriptsize$\pm$1.1} & \textbf{18.4} {\scriptsize$\pm$0.7} & \textbf{74.1} {\scriptsize$\pm$2.2} & 40.0 {\scriptsize$\pm$1.9} & 49.5 {\scriptsize$\pm$0.3} & 62.4 {\scriptsize$\pm$0.3} \\
\midrule
\multicolumn{7}{c}{\small\textbf{\texttt{google/gemma-3-27b-it}} (6 datasets)} \\
\addlinespace[2pt]
P(True) & 44.9 & 45.0 & 53.8 & 60.6 & 63.8 & 64.9 \\
Self-Probing & 34.1 & 35.2 & 55.4 & 64.8 & 60.8 & 62.2 \\
PE & 39.1 & 40.2 & 49.4 & 64.2 & 52.5 & 55.1 \\
SAPLMA & \textbf{11.1} {\scriptsize$\pm$1.2} & 22.6 {\scriptsize$\pm$0.4} & 62.7 {\scriptsize$\pm$1.1} & 64.9 {\scriptsize$\pm$1.8} & 65.7 {\scriptsize$\pm$0.3} & 65.4 {\scriptsize$\pm$0.3} \\
PIK & 17.3 {\scriptsize$\pm$2.2} & 24.2 {\scriptsize$\pm$1.3} & 60.2 {\scriptsize$\pm$2.0} & \textbf{69.0} {\scriptsize$\pm$0.5} & \textbf{72.2} {\scriptsize$\pm$0.9} & \textbf{71.0} {\scriptsize$\pm$0.9} \\
CCPS & 20.9 {\scriptsize$\pm$3.3} & 29.4 {\scriptsize$\pm$2.6} & 58.5 {\scriptsize$\pm$1.7} & 63.2 {\scriptsize$\pm$1.6} & 61.8 {\scriptsize$\pm$1.1} & 61.0 {\scriptsize$\pm$1.1} \\
II & 11.3 {\scriptsize$\pm$5.0} & 23.0 {\scriptsize$\pm$1.8} & 60.5 {\scriptsize$\pm$3.3} & 65.4 {\scriptsize$\pm$2.8} & 64.5 {\scriptsize$\pm$1.7} & 64.7 {\scriptsize$\pm$1.3} \\
\textbf{\texttt{BICR (Ours)}} & 11.5 {\scriptsize$\pm$1.2} & \textbf{22.1} {\scriptsize$\pm$0.3} & \textbf{64.4} {\scriptsize$\pm$1.4} & 63.7 {\scriptsize$\pm$1.6} & 69.0 {\scriptsize$\pm$1.1} & 68.3 {\scriptsize$\pm$0.7} \\
\midrule
\end{longtable}
}
\twocolumn

%% file: tables/unweighted_cross_vlm.tex
\begin{table*}[ht]
\centering
\small
\caption{Cross-VLM unweighted average performance. Each cell is mean $\pm$ std of the per-VLM unweighted means across all five LVLMs (equal weight per LVLM and per dataset). Best per metric in \textbf{bold}.}
\label{tab:uw_crossvlm}
\begin{tabular}{lcccccc}
\toprule
\textbf{Method} & \textbf{ECE$\downarrow$} & \textbf{BS$\downarrow$} & \textbf{ACC$\uparrow$} & \textbf{F1$\uparrow$} & \textbf{AUCPR$\uparrow$} & \textbf{AUROC$\uparrow$} \\
\midrule
P(True) & 41.1 {\scriptsize$\pm$7.5} & 39.8 {\scriptsize$\pm$7.9} & 51.7 {\scriptsize$\pm$6.4} & 52.8 {\scriptsize$\pm$11.8} & 61.6 {\scriptsize$\pm$9.0} & 65.6 {\scriptsize$\pm$1.9} \\
Self-Probing & 40.0 {\scriptsize$\pm$6.7} & 40.3 {\scriptsize$\pm$6.0} & 51.3 {\scriptsize$\pm$7.2} & 60.9 {\scriptsize$\pm$7.2} & 58.4 {\scriptsize$\pm$6.5} & 59.8 {\scriptsize$\pm$6.4} \\
PE & 35.6 {\scriptsize$\pm$2.3} & 35.5 {\scriptsize$\pm$4.2} & 47.9 {\scriptsize$\pm$4.7} & 60.1 {\scriptsize$\pm$6.7} & 52.1 {\scriptsize$\pm$3.6} & 56.3 {\scriptsize$\pm$3.5} \\
SAPLMA & 22.9 {\scriptsize$\pm$7.1} & 27.8 {\scriptsize$\pm$3.4} & 59.6 {\scriptsize$\pm$4.1} & \textbf{63.7} {\scriptsize$\pm$7.5} & 61.1 {\scriptsize$\pm$9.0} & 65.7 {\scriptsize$\pm$2.0} \\
PIK & 21.3 {\scriptsize$\pm$3.8} & 25.7 {\scriptsize$\pm$2.2} & 59.4 {\scriptsize$\pm$2.1} & 62.9 {\scriptsize$\pm$9.2} & \textbf{64.8} {\scriptsize$\pm$9.6} & \textbf{68.4} {\scriptsize$\pm$3.5} \\
CCPS & 28.7 {\scriptsize$\pm$6.4} & 34.3 {\scriptsize$\pm$8.2} & 54.5 {\scriptsize$\pm$4.0} & 58.1 {\scriptsize$\pm$6.8} & 58.5 {\scriptsize$\pm$6.3} & 59.2 {\scriptsize$\pm$5.5} \\
II & 18.0 {\scriptsize$\pm$7.8} & 25.1 {\scriptsize$\pm$4.6} & 59.5 {\scriptsize$\pm$6.9} & 61.6 {\scriptsize$\pm$9.0} & 59.5 {\scriptsize$\pm$9.8} & 63.0 {\scriptsize$\pm$5.0} \\
\textbf{\texttt{BICR (Ours)}} & \textbf{13.8} {\scriptsize$\pm$3.4} & \textbf{21.4} {\scriptsize$\pm$1.7} & \textbf{66.8} {\scriptsize$\pm$3.9} & 57.9 {\scriptsize$\pm$9.7} & 64.2 {\scriptsize$\pm$8.9} & 68.1 {\scriptsize$\pm$3.0} \\
\bottomrule
\end{tabular}
\end{table*}

%% file: tables/per_dataset_longtable.tex
{\small
\onecolumn
\setlength{\tabcolsep}{3pt}
\begin{longtable}{lcccccc}
\caption{Per-VLM per-dataset performance. Each block shows one LVLM with datasets as subgroups. For trained methods, each cell is mean $\pm$ std across 5 seeds (50 Optuna trials each). Best value per dataset per metric in \textbf{bold}.}
\label{tab:perdataset}\\
\toprule
\textbf{Method} & \textbf{ECE$\downarrow$} & \textbf{BS$\downarrow$} & \textbf{ACC$\uparrow$} & \textbf{F1$\uparrow$} & \textbf{AUCPR$\uparrow$} & \textbf{AUROC$\uparrow$} \\
\midrule
\endfirsthead
\multicolumn{7}{l}{\footnotesize\emph{(Continued from previous page)}}\\
\toprule
\textbf{Method} & \textbf{ECE$\downarrow$} & \textbf{BS$\downarrow$} & \textbf{ACC$\uparrow$} & \textbf{F1$\uparrow$} & \textbf{AUCPR$\uparrow$} & \textbf{AUROC$\uparrow$} \\
\midrule
\endhead
\midrule
\multicolumn{7}{r}{\footnotesize\emph{(Continued on next page)}}\\
\endfoot
\bottomrule
\endlastfoot
\multicolumn{7}{c}{\small\textbf{\texttt{Qwen/Qwen3-VL-8B-Instruct}}} \\
\addlinespace[2pt]
\multicolumn{7}{c}{\small\textit{GMAI-MMBench}} \\
\addlinespace[1pt]
P(True) & 45.8 & 45.7 & 54.0 & \textbf{69.6} & 68.7 & 60.6 \\
Self-Probing & 33.9 & 37.6 & 54.4 & 68.6 & 56.7 & 61.3 \\
PE & 31.4 & 34.7 & 53.0 & 69.2 & 56.2 & 53.6 \\
SAPLMA & 34.2 {\scriptsize$\pm$4.2} & 36.2 {\scriptsize$\pm$2.7} & 53.3 {\scriptsize$\pm$0.3} & 69.3 {\scriptsize$\pm$0.1} & 63.8 {\scriptsize$\pm$0.6} & 59.6 {\scriptsize$\pm$0.5} \\
PIK & 15.4 {\scriptsize$\pm$4.1} & 24.9 {\scriptsize$\pm$1.2} & 56.9 {\scriptsize$\pm$2.7} & 68.0 {\scriptsize$\pm$0.8} & \textbf{73.2} {\scriptsize$\pm$1.2} & \textbf{68.4} {\scriptsize$\pm$1.0} \\
CCPS & 34.6 {\scriptsize$\pm$16.8} & 48.8 {\scriptsize$\pm$2.3} & 50.1 {\scriptsize$\pm$2.1} & 47.0 {\scriptsize$\pm$25.1} & 57.0 {\scriptsize$\pm$7.0} & 50.5 {\scriptsize$\pm$2.6} \\
II & \textbf{10.7} {\scriptsize$\pm$4.4} & \textbf{24.6} {\scriptsize$\pm$1.0} & 57.4 {\scriptsize$\pm$1.9} & 67.9 {\scriptsize$\pm$1.6} & 68.0 {\scriptsize$\pm$1.9} & 65.1 {\scriptsize$\pm$1.9} \\
\textbf{\texttt{BICR (Ours)}} & 13.3 {\scriptsize$\pm$5.3} & 25.3 {\scriptsize$\pm$1.3} & \textbf{59.5} {\scriptsize$\pm$1.0} & 50.1 {\scriptsize$\pm$7.4} & 70.1 {\scriptsize$\pm$1.4} & 64.7 {\scriptsize$\pm$1.6} \\
\addlinespace[3pt]
\multicolumn{7}{c}{\small\textit{GQA}} \\
\addlinespace[1pt]
P(True) & 36.1 & 36.6 & 61.7 & 73.2 & 79.4 & 60.9 \\
Self-Probing & 23.6 & 27.6 & 70.1 & 81.7 & 76.1 & 54.5 \\
PE & 18.1 & 24.5 & 69.0 & 81.6 & 73.0 & 54.8 \\
SAPLMA & 3.7 {\scriptsize$\pm$1.6} & \textbf{16.3} {\scriptsize$\pm$0.2} & \textbf{75.6} {\scriptsize$\pm$0.5} & \textbf{83.7} {\scriptsize$\pm$0.4} & \textbf{89.3} {\scriptsize$\pm$0.2} & \textbf{79.9} {\scriptsize$\pm$0.3} \\
PIK & 5.7 {\scriptsize$\pm$1.4} & 17.5 {\scriptsize$\pm$0.3} & 73.0 {\scriptsize$\pm$0.8} & 80.5 {\scriptsize$\pm$1.2} & 88.7 {\scriptsize$\pm$0.5} & 78.1 {\scriptsize$\pm$0.5} \\
CCPS & 20.8 {\scriptsize$\pm$10.4} & 44.5 {\scriptsize$\pm$18.0} & 54.9 {\scriptsize$\pm$17.9} & 52.1 {\scriptsize$\pm$36.2} & 71.8 {\scriptsize$\pm$4.9} & 51.6 {\scriptsize$\pm$3.2} \\
II & \textbf{3.6} {\scriptsize$\pm$1.0} & 17.5 {\scriptsize$\pm$0.2} & 73.3 {\scriptsize$\pm$0.4} & 82.2 {\scriptsize$\pm$0.2} & 88.4 {\scriptsize$\pm$0.3} & 77.1 {\scriptsize$\pm$0.5} \\
\textbf{\texttt{BICR (Ours)}} & 11.7 {\scriptsize$\pm$1.3} & 18.6 {\scriptsize$\pm$0.4} & 70.6 {\scriptsize$\pm$0.8} & 76.5 {\scriptsize$\pm$1.1} & 89.2 {\scriptsize$\pm$0.2} & 78.6 {\scriptsize$\pm$0.3} \\
\addlinespace[3pt]
\multicolumn{7}{c}{\small\textit{LLaVA-Wild}} \\
\addlinespace[1pt]
P(True) & 32.8 & 31.5 & 67.9 & \textbf{71.9} & \textbf{85.9} & \textbf{82.8} \\
Self-Probing & 45.0 & 44.5 & 51.8 & 65.8 & 57.8 & 55.3 \\
PE & 38.6 & 39.8 & 46.4 & 63.4 & 52.3 & 57.8 \\
SAPLMA & 17.2 {\scriptsize$\pm$5.5} & 23.6 {\scriptsize$\pm$1.9} & 62.9 {\scriptsize$\pm$3.1} & 48.3 {\scriptsize$\pm$5.7} & 69.6 {\scriptsize$\pm$2.1} & 66.1 {\scriptsize$\pm$2.8} \\
PIK & 17.3 {\scriptsize$\pm$1.9} & 21.0 {\scriptsize$\pm$1.3} & 65.7 {\scriptsize$\pm$7.9} & 61.4 {\scriptsize$\pm$4.5} & 76.2 {\scriptsize$\pm$2.9} & 70.9 {\scriptsize$\pm$3.7} \\
CCPS & 38.1 {\scriptsize$\pm$20.0} & 50.3 {\scriptsize$\pm$10.4} & 48.9 {\scriptsize$\pm$10.0} & 42.1 {\scriptsize$\pm$22.9} & 52.7 {\scriptsize$\pm$13.1} & 45.8 {\scriptsize$\pm$15.5} \\
II & \textbf{12.8} {\scriptsize$\pm$4.3} & \textbf{20.6} {\scriptsize$\pm$1.1} & \textbf{70.4} {\scriptsize$\pm$1.8} & 67.3 {\scriptsize$\pm$3.1} & 73.9 {\scriptsize$\pm$6.1} & 73.6 {\scriptsize$\pm$3.3} \\
\textbf{\texttt{BICR (Ours)}} & 18.0 {\scriptsize$\pm$2.6} & 21.6 {\scriptsize$\pm$1.4} & 70.0 {\scriptsize$\pm$2.6} & 54.3 {\scriptsize$\pm$4.3} & 79.8 {\scriptsize$\pm$2.8} & 77.9 {\scriptsize$\pm$3.6} \\
\addlinespace[3pt]
\multicolumn{7}{c}{\small\textit{MME-Finance}} \\
\addlinespace[1pt]
P(True) & 44.3 & 43.7 & 54.4 & 69.2 & \textbf{82.7} & \textbf{79.6} \\
Self-Probing & 43.7 & 43.8 & 54.6 & \textbf{69.3} & 74.8 & 59.1 \\
PE & 40.4 & 40.9 & 51.6 & 68.1 & 59.0 & 57.9 \\
SAPLMA & 10.0 {\scriptsize$\pm$4.0} & \textbf{21.4} {\scriptsize$\pm$1.0} & \textbf{67.7} {\scriptsize$\pm$1.5} & 68.7 {\scriptsize$\pm$1.3} & 78.0 {\scriptsize$\pm$0.4} & 74.2 {\scriptsize$\pm$0.6} \\
PIK & 16.4 {\scriptsize$\pm$3.9} & 24.8 {\scriptsize$\pm$1.3} & 63.7 {\scriptsize$\pm$0.9} & 68.5 {\scriptsize$\pm$3.0} & 72.0 {\scriptsize$\pm$1.4} & 71.0 {\scriptsize$\pm$1.4} \\
CCPS & 32.8 {\scriptsize$\pm$14.3} & 46.4 {\scriptsize$\pm$7.0} & 52.8 {\scriptsize$\pm$6.8} & 47.3 {\scriptsize$\pm$26.2} & 58.2 {\scriptsize$\pm$8.9} & 51.9 {\scriptsize$\pm$9.9} \\
II & \textbf{6.7} {\scriptsize$\pm$2.2} & 22.9 {\scriptsize$\pm$0.9} & 62.1 {\scriptsize$\pm$2.3} & 64.9 {\scriptsize$\pm$4.4} & 69.3 {\scriptsize$\pm$2.7} & 68.0 {\scriptsize$\pm$3.0} \\
\textbf{\texttt{BICR (Ours)}} & 13.7 {\scriptsize$\pm$5.1} & 24.3 {\scriptsize$\pm$1.3} & 62.5 {\scriptsize$\pm$1.8} & 55.7 {\scriptsize$\pm$7.5} & 69.9 {\scriptsize$\pm$0.9} & 70.1 {\scriptsize$\pm$1.0} \\
\addlinespace[3pt]
\multicolumn{7}{c}{\small\textit{MMMU\_Pro\_10}} \\
\addlinespace[1pt]
P(True) & 60.4 & 59.1 & 37.9 & 42.2 & 46.3 & \textbf{68.6} \\
Self-Probing & 57.1 & 57.9 & 38.8 & 42.2 & 42.4 & 52.9 \\
PE & 65.0 & 61.1 & 23.9 & 38.6 & 19.9 & 42.5 \\
SAPLMA & 40.1 {\scriptsize$\pm$4.8} & 39.8 {\scriptsize$\pm$3.5} & 43.2 {\scriptsize$\pm$2.6} & 55.2 {\scriptsize$\pm$0.2} & 52.1 {\scriptsize$\pm$0.4} & 61.4 {\scriptsize$\pm$0.4} \\
PIK & 41.5 {\scriptsize$\pm$4.8} & 40.0 {\scriptsize$\pm$4.3} & 41.3 {\scriptsize$\pm$1.7} & 54.7 {\scriptsize$\pm$0.2} & 56.5 {\scriptsize$\pm$2.6} & 65.8 {\scriptsize$\pm$2.1} \\
CCPS & 53.1 {\scriptsize$\pm$5.8} & 57.9 {\scriptsize$\pm$3.4} & 41.5 {\scriptsize$\pm$3.1} & 50.0 {\scriptsize$\pm$2.2} & 54.3 {\scriptsize$\pm$5.1} & 49.5 {\scriptsize$\pm$1.7} \\
II & 19.7 {\scriptsize$\pm$3.9} & 25.5 {\scriptsize$\pm$1.5} & 56.8 {\scriptsize$\pm$5.5} & \textbf{56.2} {\scriptsize$\pm$0.9} & 56.2 {\scriptsize$\pm$3.0} & 67.2 {\scriptsize$\pm$1.5} \\
\textbf{\texttt{BICR (Ours)}} & \textbf{12.5} {\scriptsize$\pm$2.9} & \textbf{23.1} {\scriptsize$\pm$1.0} & \textbf{63.2} {\scriptsize$\pm$2.8} & 53.5 {\scriptsize$\pm$1.5} & \textbf{57.5} {\scriptsize$\pm$0.9} & 66.5 {\scriptsize$\pm$0.8} \\
\addlinespace[3pt]
\multicolumn{7}{c}{\small\textit{MMMU\_Pro\_4}} \\
\addlinespace[1pt]
P(True) & 54.2 & 53.1 & 44.1 & 50.9 & 53.0 & \textbf{69.0} \\
Self-Probing & 51.0 & 52.6 & 44.4 & 50.8 & 48.9 & 54.3 \\
PE & 58.3 & 55.9 & 30.7 & 47.0 & 25.7 & 42.0 \\
SAPLMA & 32.0 {\scriptsize$\pm$5.1} & 34.7 {\scriptsize$\pm$3.0} & 50.2 {\scriptsize$\pm$1.7} & \textbf{64.3} {\scriptsize$\pm$0.3} & 63.6 {\scriptsize$\pm$0.2} & 62.9 {\scriptsize$\pm$0.1} \\
PIK & 25.5 {\scriptsize$\pm$4.5} & 30.1 {\scriptsize$\pm$2.4} & 51.8 {\scriptsize$\pm$2.0} & 64.2 {\scriptsize$\pm$0.6} & 65.2 {\scriptsize$\pm$2.4} & 65.8 {\scriptsize$\pm$1.9} \\
CCPS & 44.3 {\scriptsize$\pm$4.8} & 51.2 {\scriptsize$\pm$1.3} & 48.1 {\scriptsize$\pm$1.1} & 57.6 {\scriptsize$\pm$3.7} & 60.6 {\scriptsize$\pm$4.8} & 49.9 {\scriptsize$\pm$0.9} \\
II & 15.0 {\scriptsize$\pm$3.7} & 25.4 {\scriptsize$\pm$1.0} & 55.7 {\scriptsize$\pm$2.9} & 63.1 {\scriptsize$\pm$1.8} & 64.7 {\scriptsize$\pm$2.1} & 65.9 {\scriptsize$\pm$1.6} \\
\textbf{\texttt{BICR (Ours)}} & \textbf{6.3} {\scriptsize$\pm$1.2} & \textbf{23.3} {\scriptsize$\pm$0.4} & \textbf{63.1} {\scriptsize$\pm$1.0} & 54.0 {\scriptsize$\pm$2.1} & \textbf{65.5} {\scriptsize$\pm$1.2} & 66.3 {\scriptsize$\pm$1.0} \\
\addlinespace[3pt]
\multicolumn{7}{c}{\small\textit{POPE}} \\
\addlinespace[1pt]
P(True) & 49.6 & 49.5 & 50.1 & 62.4 & 93.0 & 55.4 \\
Self-Probing & 7.7 & 10.2 & 88.6 & 94.0 & 95.6 & 72.0 \\
PE & \textbf{1.4} & 9.8 & 88.7 & 94.0 & 93.0 & 67.2 \\
SAPLMA & 3.7 {\scriptsize$\pm$1.5} & 8.7 {\scriptsize$\pm$0.3} & 88.7 {\scriptsize$\pm$0.1} & \textbf{94.0} {\scriptsize$\pm$0.0} & 96.0 {\scriptsize$\pm$0.4} & 79.7 {\scriptsize$\pm$2.2} \\
PIK & 2.8 {\scriptsize$\pm$0.7} & \textbf{8.2} {\scriptsize$\pm$0.2} & \textbf{88.9} {\scriptsize$\pm$0.5} & 94.0 {\scriptsize$\pm$0.3} & 96.3 {\scriptsize$\pm$0.2} & 81.2 {\scriptsize$\pm$1.1} \\
CCPS & 28.8 {\scriptsize$\pm$22.3} & 42.2 {\scriptsize$\pm$37.9} & 57.7 {\scriptsize$\pm$37.9} & 56.4 {\scriptsize$\pm$46.0} & 86.9 {\scriptsize$\pm$7.4} & 44.7 {\scriptsize$\pm$24.6} \\
II & 3.1 {\scriptsize$\pm$0.8} & 8.7 {\scriptsize$\pm$0.3} & 88.6 {\scriptsize$\pm$0.1} & 93.9 {\scriptsize$\pm$0.1} & 95.5 {\scriptsize$\pm$0.8} & 78.0 {\scriptsize$\pm$3.5} \\
\textbf{\texttt{BICR (Ours)}} & 7.5 {\scriptsize$\pm$1.0} & 9.1 {\scriptsize$\pm$0.2} & 87.4 {\scriptsize$\pm$0.4} & 92.9 {\scriptsize$\pm$0.2} & \textbf{96.4} {\scriptsize$\pm$0.2} & \textbf{81.6} {\scriptsize$\pm$0.5} \\
\addlinespace[3pt]
\midrule
\midrule
\multicolumn{7}{c}{\small\textbf{\texttt{llava-hf/llava-v1.6-vicuna-13b-hf}}} \\
\addlinespace[2pt]
\multicolumn{7}{c}{\small\textit{GMAI-MMBench}} \\
\addlinespace[1pt]
P(True) & \textbf{11.1} & \textbf{24.2} & \textbf{59.1} & 18.5 & 33.7 & 49.2 \\
Self-Probing & 47.2 & 47.2 & 38.1 & 52.1 & 29.0 & 49.5 \\
PE & 34.6 & 35.0 & 35.3 & 52.2 & 34.5 & 48.3 \\
SAPLMA & 43.4 {\scriptsize$\pm$4.8} & 43.5 {\scriptsize$\pm$3.7} & 38.0 {\scriptsize$\pm$1.3} & 51.8 {\scriptsize$\pm$0.6} & 36.9 {\scriptsize$\pm$0.6} & 52.1 {\scriptsize$\pm$0.6} \\
PIK & 28.5 {\scriptsize$\pm$9.8} & 32.0 {\scriptsize$\pm$5.3} & 47.7 {\scriptsize$\pm$8.3} & 50.9 {\scriptsize$\pm$1.4} & 47.8 {\scriptsize$\pm$0.4} & 61.1 {\scriptsize$\pm$0.5} \\
CCPS & 30.5 {\scriptsize$\pm$2.1} & 31.2 {\scriptsize$\pm$1.3} & 45.9 {\scriptsize$\pm$2.8} & \textbf{53.8} {\scriptsize$\pm$0.4} & \textbf{48.5} {\scriptsize$\pm$0.3} & \textbf{63.6} {\scriptsize$\pm$0.1} \\
II & 33.2 {\scriptsize$\pm$4.0} & 35.5 {\scriptsize$\pm$1.9} & 39.0 {\scriptsize$\pm$3.2} & 50.3 {\scriptsize$\pm$2.0} & 36.8 {\scriptsize$\pm$1.2} & 51.6 {\scriptsize$\pm$1.3} \\
\textbf{\texttt{BICR (Ours)}} & 18.2 {\scriptsize$\pm$7.5} & 26.7 {\scriptsize$\pm$3.1} & 56.9 {\scriptsize$\pm$5.9} & 46.0 {\scriptsize$\pm$6.8} & 47.7 {\scriptsize$\pm$0.7} & 61.2 {\scriptsize$\pm$0.6} \\
\addlinespace[3pt]
\multicolumn{7}{c}{\small\textit{GQA}} \\
\addlinespace[1pt]
P(True) & 37.2 & 35.2 & 36.4 & 22.8 & 75.0 & 55.8 \\
Self-Probing & 23.4 & 26.3 & 70.8 & 82.7 & 79.9 & 55.2 \\
PE & 6.9 & 20.8 & 70.4 & 82.6 & 78.1 & 59.9 \\
SAPLMA & 6.2 {\scriptsize$\pm$2.0} & 17.9 {\scriptsize$\pm$0.3} & \textbf{73.8} {\scriptsize$\pm$0.3} & 83.4 {\scriptsize$\pm$0.2} & 87.4 {\scriptsize$\pm$0.4} & 75.0 {\scriptsize$\pm$0.5} \\
PIK & 6.7 {\scriptsize$\pm$1.2} & 18.2 {\scriptsize$\pm$0.3} & 71.5 {\scriptsize$\pm$0.7} & 79.8 {\scriptsize$\pm$0.9} & 88.4 {\scriptsize$\pm$0.1} & 75.5 {\scriptsize$\pm$0.2} \\
CCPS & 5.9 {\scriptsize$\pm$0.8} & \textbf{17.6} {\scriptsize$\pm$0.1} & 73.7 {\scriptsize$\pm$0.1} & \textbf{83.4} {\scriptsize$\pm$0.1} & \textbf{88.5} {\scriptsize$\pm$0.0} & \textbf{76.1} {\scriptsize$\pm$0.0} \\
II & \textbf{5.1} {\scriptsize$\pm$1.1} & 19.3 {\scriptsize$\pm$0.2} & 71.1 {\scriptsize$\pm$0.5} & 82.0 {\scriptsize$\pm$0.7} & 83.3 {\scriptsize$\pm$0.3} & 68.5 {\scriptsize$\pm$0.7} \\
\textbf{\texttt{BICR (Ours)}} & 10.5 {\scriptsize$\pm$2.5} & 19.0 {\scriptsize$\pm$0.6} & 69.3 {\scriptsize$\pm$1.3} & 76.6 {\scriptsize$\pm$1.8} & 88.4 {\scriptsize$\pm$0.1} & 75.8 {\scriptsize$\pm$0.2} \\
\addlinespace[3pt]
\multicolumn{7}{c}{\small\textit{LLaVA-Wild}} \\
\addlinespace[1pt]
P(True) & \textbf{8.8} & \textbf{18.4} & \textbf{73.3} & 42.9 & \textbf{53.9} & 71.8 \\
Self-Probing & 57.8 & 54.9 & 33.3 & 47.4 & 42.0 & 51.1 \\
PE & 43.4 & 38.5 & 30.0 & 46.2 & 51.6 & \textbf{72.4} \\
SAPLMA & 32.1 {\scriptsize$\pm$2.7} & 31.6 {\scriptsize$\pm$1.9} & 55.0 {\scriptsize$\pm$3.7} & \textbf{53.5} {\scriptsize$\pm$3.1} & 35.5 {\scriptsize$\pm$1.3} & 63.8 {\scriptsize$\pm$2.2} \\
PIK & 37.5 {\scriptsize$\pm$1.7} & 37.0 {\scriptsize$\pm$1.3} & 45.7 {\scriptsize$\pm$2.3} & 45.4 {\scriptsize$\pm$1.1} & 47.7 {\scriptsize$\pm$3.3} & 58.2 {\scriptsize$\pm$1.2} \\
CCPS & 43.4 {\scriptsize$\pm$0.7} & 39.9 {\scriptsize$\pm$0.9} & 39.7 {\scriptsize$\pm$1.2} & 46.9 {\scriptsize$\pm$0.5} & 43.1 {\scriptsize$\pm$1.8} & 64.2 {\scriptsize$\pm$1.7} \\
II & 38.6 {\scriptsize$\pm$5.1} & 36.3 {\scriptsize$\pm$3.6} & 37.0 {\scriptsize$\pm$4.8} & 44.9 {\scriptsize$\pm$2.9} & 37.3 {\scriptsize$\pm$5.1} & 54.5 {\scriptsize$\pm$3.3} \\
\textbf{\texttt{BICR (Ours)}} & 31.1 {\scriptsize$\pm$6.0} & 30.3 {\scriptsize$\pm$4.0} & 54.0 {\scriptsize$\pm$9.0} & 47.8 {\scriptsize$\pm$5.2} & 48.8 {\scriptsize$\pm$3.8} & 60.6 {\scriptsize$\pm$2.8} \\
\addlinespace[3pt]
\multicolumn{7}{c}{\small\textit{MME-Finance}} \\
\addlinespace[1pt]
P(True) & \textbf{15.6} & \textbf{18.2} & \textbf{76.3} & 35.9 & 37.6 & 68.0 \\
Self-Probing & 70.0 & 66.1 & 23.0 & 36.0 & 40.4 & 58.9 \\
PE & 52.3 & 43.9 & 21.7 & 35.7 & 32.3 & 59.1 \\
SAPLMA & 40.4 {\scriptsize$\pm$5.3} & 34.8 {\scriptsize$\pm$4.1} & 39.9 {\scriptsize$\pm$5.6} & 37.0 {\scriptsize$\pm$1.4} & 31.8 {\scriptsize$\pm$1.1} & 60.9 {\scriptsize$\pm$1.7} \\
PIK & 45.9 {\scriptsize$\pm$2.8} & 39.2 {\scriptsize$\pm$2.4} & 37.7 {\scriptsize$\pm$4.5} & 37.8 {\scriptsize$\pm$0.7} & 38.5 {\scriptsize$\pm$1.0} & 66.1 {\scriptsize$\pm$1.0} \\
CCPS & 59.4 {\scriptsize$\pm$0.7} & 53.2 {\scriptsize$\pm$0.7} & 23.9 {\scriptsize$\pm$0.3} & 35.0 {\scriptsize$\pm$0.4} & 35.6 {\scriptsize$\pm$1.1} & 60.2 {\scriptsize$\pm$0.5} \\
II & 48.7 {\scriptsize$\pm$3.5} & 42.7 {\scriptsize$\pm$2.7} & 28.1 {\scriptsize$\pm$5.0} & 34.9 {\scriptsize$\pm$1.1} & 22.7 {\scriptsize$\pm$0.8} & 50.9 {\scriptsize$\pm$0.4} \\
\textbf{\texttt{BICR (Ours)}} & 36.5 {\scriptsize$\pm$7.2} & 31.0 {\scriptsize$\pm$5.9} & 49.4 {\scriptsize$\pm$8.0} & \textbf{40.8} {\scriptsize$\pm$2.1} & \textbf{41.8} {\scriptsize$\pm$1.4} & \textbf{69.2} {\scriptsize$\pm$0.8} \\
\addlinespace[3pt]
\multicolumn{7}{c}{\small\textit{MMMU\_Pro\_10}} \\
\addlinespace[1pt]
P(True) & \textbf{25.2} & \textbf{19.6} & \textbf{68.0} & \textbf{34.5} & 24.9 & \textbf{68.0} \\
Self-Probing & 62.9 & 59.2 & 30.5 & 30.4 & 28.0 & 56.0 \\
PE & 56.6 & 45.4 & 15.8 & 27.3 & 19.8 & 54.2 \\
SAPLMA & 54.9 {\scriptsize$\pm$4.2} & 46.4 {\scriptsize$\pm$4.0} & 31.3 {\scriptsize$\pm$4.6} & 33.1 {\scriptsize$\pm$0.8} & 33.8 {\scriptsize$\pm$1.2} & 67.0 {\scriptsize$\pm$0.6} \\
PIK & 39.6 {\scriptsize$\pm$12.8} & 33.2 {\scriptsize$\pm$10.7} & 46.8 {\scriptsize$\pm$16.8} & 33.4 {\scriptsize$\pm$1.7} & 34.6 {\scriptsize$\pm$0.8} & 63.0 {\scriptsize$\pm$1.4} \\
CCPS & 58.8 {\scriptsize$\pm$1.6} & 53.1 {\scriptsize$\pm$1.5} & 28.1 {\scriptsize$\pm$1.4} & 31.2 {\scriptsize$\pm$0.3} & \textbf{37.3} {\scriptsize$\pm$0.3} & 55.1 {\scriptsize$\pm$0.3} \\
II & 51.1 {\scriptsize$\pm$3.5} & 42.7 {\scriptsize$\pm$3.1} & 26.2 {\scriptsize$\pm$6.2} & 30.8 {\scriptsize$\pm$0.8} & 20.0 {\scriptsize$\pm$1.0} & 52.7 {\scriptsize$\pm$1.6} \\
\textbf{\texttt{BICR (Ours)}} & 25.5 {\scriptsize$\pm$11.9} & 23.3 {\scriptsize$\pm$6.2} & 64.1 {\scriptsize$\pm$12.8} & 34.3 {\scriptsize$\pm$1.1} & 35.8 {\scriptsize$\pm$0.3} & 64.4 {\scriptsize$\pm$0.7} \\
\addlinespace[3pt]
\multicolumn{7}{c}{\small\textit{MMMU\_Pro\_4}} \\
\addlinespace[1pt]
P(True) & \textbf{22.2} & \textbf{19.2} & \textbf{69.8} & 41.8 & 34.7 & \textbf{70.9} \\
Self-Probing & 59.7 & 56.6 & 33.7 & 35.6 & 30.2 & 56.5 \\
PE & 53.6 & 44.1 & 18.9 & 31.7 & 21.2 & 51.4 \\
SAPLMA & 45.4 {\scriptsize$\pm$4.3} & 41.1 {\scriptsize$\pm$3.2} & 39.4 {\scriptsize$\pm$4.0} & \textbf{45.1} {\scriptsize$\pm$0.6} & 42.3 {\scriptsize$\pm$0.8} & 65.1 {\scriptsize$\pm$0.5} \\
PIK & 36.6 {\scriptsize$\pm$10.9} & 34.6 {\scriptsize$\pm$8.4} & 44.7 {\scriptsize$\pm$11.7} & 43.7 {\scriptsize$\pm$1.0} & 44.5 {\scriptsize$\pm$0.8} & 62.7 {\scriptsize$\pm$1.3} \\
CCPS & 51.7 {\scriptsize$\pm$1.5} & 48.4 {\scriptsize$\pm$1.1} & 34.2 {\scriptsize$\pm$1.7} & 43.4 {\scriptsize$\pm$0.4} & \textbf{46.2} {\scriptsize$\pm$0.4} & 57.5 {\scriptsize$\pm$0.3} \\
II & 42.1 {\scriptsize$\pm$3.7} & 39.3 {\scriptsize$\pm$2.6} & 33.0 {\scriptsize$\pm$4.6} & 41.9 {\scriptsize$\pm$0.8} & 28.5 {\scriptsize$\pm$1.7} & 51.9 {\scriptsize$\pm$1.8} \\
\textbf{\texttt{BICR (Ours)}} & 23.0 {\scriptsize$\pm$10.2} & 26.0 {\scriptsize$\pm$4.7} & 59.0 {\scriptsize$\pm$9.6} & 42.5 {\scriptsize$\pm$2.0} & 45.4 {\scriptsize$\pm$0.2} & 63.7 {\scriptsize$\pm$0.6} \\
\addlinespace[3pt]
\multicolumn{7}{c}{\small\textit{POPE}} \\
\addlinespace[1pt]
P(True) & 48.3 & 34.1 & 39.8 & 49.2 & 94.2 & 67.1 \\
Self-Probing & 10.9 & 12.0 & 87.9 & 93.6 & 93.8 & 50.7 \\
PE & 10.4 & 11.2 & \textbf{88.4} & \textbf{93.8} & 92.3 & 61.9 \\
SAPLMA & 5.9 {\scriptsize$\pm$1.7} & 10.5 {\scriptsize$\pm$0.3} & 88.0 {\scriptsize$\pm$0.4} & 93.6 {\scriptsize$\pm$0.2} & 92.9 {\scriptsize$\pm$0.3} & 64.7 {\scriptsize$\pm$0.8} \\
PIK & 2.9 {\scriptsize$\pm$1.4} & \textbf{8.7} {\scriptsize$\pm$0.4} & 88.2 {\scriptsize$\pm$0.1} & 93.7 {\scriptsize$\pm$0.1} & \textbf{96.7} {\scriptsize$\pm$0.2} & \textbf{82.4} {\scriptsize$\pm$1.3} \\
CCPS & 5.5 {\scriptsize$\pm$0.3} & 9.2 {\scriptsize$\pm$0.1} & 87.9 {\scriptsize$\pm$0.3} & 93.5 {\scriptsize$\pm$0.2} & 96.6 {\scriptsize$\pm$0.0} & 81.6 {\scriptsize$\pm$0.3} \\
II & 4.5 {\scriptsize$\pm$1.3} & 10.6 {\scriptsize$\pm$0.3} & 88.2 {\scriptsize$\pm$0.2} & 93.7 {\scriptsize$\pm$0.1} & 90.7 {\scriptsize$\pm$0.8} & 56.9 {\scriptsize$\pm$3.1} \\
\textbf{\texttt{BICR (Ours)}} & \textbf{2.4} {\scriptsize$\pm$1.0} & 8.8 {\scriptsize$\pm$0.1} & 87.6 {\scriptsize$\pm$0.4} & 93.3 {\scriptsize$\pm$0.3} & 96.6 {\scriptsize$\pm$0.1} & 81.9 {\scriptsize$\pm$0.4} \\
\addlinespace[3pt]
\midrule
\midrule
\multicolumn{7}{c}{\small\textbf{\texttt{OpenGVLab/InternVL3\_5-14B-HF}}} \\
\addlinespace[2pt]
\multicolumn{7}{c}{\small\textit{GMAI-MMBench}} \\
\addlinespace[1pt]
P(True) & 36.8 & 37.0 & 60.5 & 74.8 & \textbf{77.8} & \textbf{68.7} \\
Self-Probing & 26.6 & 29.8 & 60.1 & \textbf{75.0} & 67.3 & 65.7 \\
PE & 23.8 & 29.4 & 60.0 & 75.0 & 64.9 & 56.7 \\
SAPLMA & 32.2 {\scriptsize$\pm$1.5} & 34.6 {\scriptsize$\pm$0.9} & 60.0 {\scriptsize$\pm$0.2} & 74.6 {\scriptsize$\pm$0.3} & 64.9 {\scriptsize$\pm$0.4} & 55.3 {\scriptsize$\pm$0.4} \\
PIK & 21.7 {\scriptsize$\pm$3.2} & 27.3 {\scriptsize$\pm$1.4} & 60.5 {\scriptsize$\pm$0.6} & 74.3 {\scriptsize$\pm$0.3} & 76.4 {\scriptsize$\pm$0.5} & 66.8 {\scriptsize$\pm$0.4} \\
CCPS & 28.7 {\scriptsize$\pm$1.4} & 33.4 {\scriptsize$\pm$0.7} & 57.4 {\scriptsize$\pm$0.4} & 71.6 {\scriptsize$\pm$0.5} & 56.9 {\scriptsize$\pm$2.3} & 46.4 {\scriptsize$\pm$2.3} \\
II & 17.5 {\scriptsize$\pm$10.2} & 27.8 {\scriptsize$\pm$3.4} & 59.4 {\scriptsize$\pm$1.9} & 72.2 {\scriptsize$\pm$3.8} & 65.9 {\scriptsize$\pm$4.1} & 57.4 {\scriptsize$\pm$5.0} \\
\textbf{\texttt{BICR (Ours)}} & \textbf{11.3} {\scriptsize$\pm$1.5} & \textbf{24.2} {\scriptsize$\pm$0.2} & \textbf{61.0} {\scriptsize$\pm$0.5} & 68.9 {\scriptsize$\pm$1.4} & 75.1 {\scriptsize$\pm$0.4} & 64.6 {\scriptsize$\pm$0.4} \\
\addlinespace[3pt]
\multicolumn{7}{c}{\small\textit{GQA}} \\
\addlinespace[1pt]
P(True) & 35.9 & 36.3 & 62.5 & 74.9 & 75.9 & 58.9 \\
Self-Probing & 21.6 & 25.8 & 68.3 & 81.1 & 71.3 & 60.3 \\
PE & 13.9 & 23.4 & 68.4 & \textbf{81.2} & 73.6 & 55.4 \\
SAPLMA & 11.9 {\scriptsize$\pm$1.4} & 21.8 {\scriptsize$\pm$0.4} & 69.4 {\scriptsize$\pm$0.3} & 80.0 {\scriptsize$\pm$0.4} & 79.3 {\scriptsize$\pm$0.6} & 67.4 {\scriptsize$\pm$0.6} \\
PIK & \textbf{4.8} {\scriptsize$\pm$1.1} & \textbf{18.0} {\scriptsize$\pm$0.2} & \textbf{72.1} {\scriptsize$\pm$0.4} & 80.0 {\scriptsize$\pm$0.5} & 87.8 {\scriptsize$\pm$0.3} & 76.5 {\scriptsize$\pm$0.5} \\
CCPS & 6.8 {\scriptsize$\pm$1.8} & 20.2 {\scriptsize$\pm$0.3} & 69.5 {\scriptsize$\pm$0.3} & 80.8 {\scriptsize$\pm$0.3} & 82.5 {\scriptsize$\pm$0.3} & 69.1 {\scriptsize$\pm$0.6} \\
II & 6.1 {\scriptsize$\pm$3.2} & 19.2 {\scriptsize$\pm$0.5} & 70.4 {\scriptsize$\pm$1.1} & 81.0 {\scriptsize$\pm$0.7} & 85.3 {\scriptsize$\pm$1.0} & 72.8 {\scriptsize$\pm$1.7} \\
\textbf{\texttt{BICR (Ours)}} & 11.1 {\scriptsize$\pm$3.9} & 19.4 {\scriptsize$\pm$1.1} & 69.4 {\scriptsize$\pm$1.9} & 75.5 {\scriptsize$\pm$2.8} & \textbf{88.1} {\scriptsize$\pm$0.1} & \textbf{77.1} {\scriptsize$\pm$0.1} \\
\addlinespace[3pt]
\multicolumn{7}{c}{\small\textit{LLaVA-Wild}} \\
\addlinespace[1pt]
P(True) & 62.3 & 60.8 & 36.7 & 42.4 & 49.2 & \textbf{75.1} \\
Self-Probing & 65.2 & 60.7 & 25.0 & 40.0 & \textbf{59.6} & 65.9 \\
PE & 57.8 & 52.0 & 25.0 & 40.0 & 30.6 & 57.5 \\
SAPLMA & 49.5 {\scriptsize$\pm$2.8} & 46.6 {\scriptsize$\pm$2.4} & 43.7 {\scriptsize$\pm$2.2} & \textbf{45.7} {\scriptsize$\pm$1.3} & 32.3 {\scriptsize$\pm$2.9} & 61.2 {\scriptsize$\pm$1.9} \\
PIK & 46.0 {\scriptsize$\pm$3.5} & 40.4 {\scriptsize$\pm$3.4} & 35.0 {\scriptsize$\pm$3.5} & 39.7 {\scriptsize$\pm$1.2} & 32.4 {\scriptsize$\pm$4.2} & 57.6 {\scriptsize$\pm$5.1} \\
CCPS & 49.1 {\scriptsize$\pm$1.9} & 43.4 {\scriptsize$\pm$2.3} & 31.3 {\scriptsize$\pm$1.9} & 38.7 {\scriptsize$\pm$1.8} & 38.7 {\scriptsize$\pm$2.7} & 58.2 {\scriptsize$\pm$4.1} \\
II & 43.0 {\scriptsize$\pm$9.6} & 37.7 {\scriptsize$\pm$9.3} & 34.7 {\scriptsize$\pm$12.6} & 40.3 {\scriptsize$\pm$5.6} & 36.6 {\scriptsize$\pm$8.8} & 60.9 {\scriptsize$\pm$7.0} \\
\textbf{\texttt{BICR (Ours)}} & \textbf{36.1} {\scriptsize$\pm$2.8} & \textbf{32.9} {\scriptsize$\pm$2.2} & \textbf{44.7} {\scriptsize$\pm$4.1} & 39.9 {\scriptsize$\pm$1.4} & 34.5 {\scriptsize$\pm$3.4} & 54.0 {\scriptsize$\pm$3.6} \\
\addlinespace[3pt]
\multicolumn{7}{c}{\small\textit{MME-Finance}} \\
\addlinespace[1pt]
P(True) & 38.7 & 37.5 & 59.0 & \textbf{71.4} & \textbf{83.4} & \textbf{81.8} \\
Self-Probing & 40.6 & 39.4 & 51.8 & 68.1 & 78.3 & 75.8 \\
PE & 38.8 & 38.7 & 51.6 & 68.0 & 73.2 & 69.4 \\
SAPLMA & 21.9 {\scriptsize$\pm$1.3} & 28.2 {\scriptsize$\pm$0.7} & 59.1 {\scriptsize$\pm$0.9} & 69.6 {\scriptsize$\pm$0.7} & 63.1 {\scriptsize$\pm$0.8} & 65.9 {\scriptsize$\pm$0.4} \\
PIK & 29.1 {\scriptsize$\pm$3.4} & 30.7 {\scriptsize$\pm$2.4} & 55.1 {\scriptsize$\pm$2.4} & 69.2 {\scriptsize$\pm$0.9} & 67.4 {\scriptsize$\pm$1.9} & 70.3 {\scriptsize$\pm$1.2} \\
CCPS & 34.7 {\scriptsize$\pm$1.2} & 36.0 {\scriptsize$\pm$0.8} & 52.6 {\scriptsize$\pm$0.4} & 68.3 {\scriptsize$\pm$0.2} & 67.1 {\scriptsize$\pm$1.4} & 64.5 {\scriptsize$\pm$1.0} \\
II & 16.5 {\scriptsize$\pm$8.1} & 27.8 {\scriptsize$\pm$3.7} & 55.4 {\scriptsize$\pm$2.9} & 66.9 {\scriptsize$\pm$1.9} & 61.1 {\scriptsize$\pm$2.9} & 59.5 {\scriptsize$\pm$3.2} \\
\textbf{\texttt{BICR (Ours)}} & \textbf{14.5} {\scriptsize$\pm$1.9} & \textbf{24.8} {\scriptsize$\pm$0.7} & \textbf{61.2} {\scriptsize$\pm$1.4} & 69.7 {\scriptsize$\pm$0.4} & 64.5 {\scriptsize$\pm$0.7} & 67.8 {\scriptsize$\pm$0.8} \\
\addlinespace[3pt]
\multicolumn{7}{c}{\small\textit{MMMU\_Pro\_10}} \\
\addlinespace[1pt]
P(True) & 48.7 & 47.2 & 49.9 & 45.3 & \textbf{46.2} & \textbf{74.7} \\
Self-Probing & 55.9 & 50.6 & 33.0 & 39.6 & 32.0 & 67.4 \\
PE & 67.6 & 63.8 & 22.5 & 36.7 & 18.4 & 41.9 \\
SAPLMA & 47.2 {\scriptsize$\pm$2.0} & 45.9 {\scriptsize$\pm$1.8} & 43.1 {\scriptsize$\pm$1.8} & \textbf{50.5} {\scriptsize$\pm$0.5} & 36.5 {\scriptsize$\pm$1.2} & 60.4 {\scriptsize$\pm$1.1} \\
PIK & 49.3 {\scriptsize$\pm$4.4} & 47.7 {\scriptsize$\pm$4.3} & 35.3 {\scriptsize$\pm$2.1} & 47.1 {\scriptsize$\pm$0.4} & 40.3 {\scriptsize$\pm$4.1} & 56.7 {\scriptsize$\pm$3.1} \\
CCPS & 56.1 {\scriptsize$\pm$2.2} & 53.5 {\scriptsize$\pm$2.4} & 31.3 {\scriptsize$\pm$0.2} & 46.5 {\scriptsize$\pm$0.2} & 33.0 {\scriptsize$\pm$1.3} & 52.3 {\scriptsize$\pm$0.9} \\
II & 46.5 {\scriptsize$\pm$9.5} & 46.4 {\scriptsize$\pm$8.3} & 33.4 {\scriptsize$\pm$3.1} & 46.0 {\scriptsize$\pm$2.1} & 29.3 {\scriptsize$\pm$5.4} & 44.9 {\scriptsize$\pm$7.2} \\
\textbf{\texttt{BICR (Ours)}} & \textbf{28.5} {\scriptsize$\pm$1.3} & \textbf{31.3} {\scriptsize$\pm$0.8} & \textbf{50.1} {\scriptsize$\pm$1.4} & 46.5 {\scriptsize$\pm$0.4} & 42.3 {\scriptsize$\pm$0.7} & 59.3 {\scriptsize$\pm$0.5} \\
\addlinespace[3pt]
\multicolumn{7}{c}{\small\textit{MMMU\_Pro\_4}} \\
\addlinespace[1pt]
P(True) & 44.6 & 43.5 & 53.7 & 51.6 & \textbf{55.9} & \textbf{77.6} \\
Self-Probing & 51.9 & 47.2 & 37.2 & 45.2 & 38.6 & 70.3 \\
PE & 63.6 & 60.6 & 26.4 & 41.7 & 21.7 & 41.3 \\
SAPLMA & 40.4 {\scriptsize$\pm$2.0} & 41.3 {\scriptsize$\pm$1.6} & 48.7 {\scriptsize$\pm$1.5} & \textbf{57.4} {\scriptsize$\pm$0.6} & 43.8 {\scriptsize$\pm$1.6} & 60.2 {\scriptsize$\pm$1.4} \\
PIK & 39.9 {\scriptsize$\pm$4.4} & 41.7 {\scriptsize$\pm$3.5} & 41.5 {\scriptsize$\pm$1.5} & 53.9 {\scriptsize$\pm$0.9} & 43.6 {\scriptsize$\pm$3.4} & 53.8 {\scriptsize$\pm$3.4} \\
CCPS & 51.0 {\scriptsize$\pm$2.1} & 49.7 {\scriptsize$\pm$1.9} & 35.4 {\scriptsize$\pm$0.1} & 50.5 {\scriptsize$\pm$0.2} & 35.3 {\scriptsize$\pm$1.1} & 45.0 {\scriptsize$\pm$1.3} \\
II & 41.1 {\scriptsize$\pm$8.7} & 43.0 {\scriptsize$\pm$7.1} & 39.3 {\scriptsize$\pm$2.0} & 54.2 {\scriptsize$\pm$1.1} & 36.3 {\scriptsize$\pm$4.9} & 46.1 {\scriptsize$\pm$6.6} \\
\textbf{\texttt{BICR (Ours)}} & \textbf{19.9} {\scriptsize$\pm$1.5} & \textbf{29.0} {\scriptsize$\pm$0.6} & \textbf{54.2} {\scriptsize$\pm$1.5} & 48.6 {\scriptsize$\pm$0.9} & 44.2 {\scriptsize$\pm$0.9} & 55.7 {\scriptsize$\pm$0.8} \\
\addlinespace[3pt]
\multicolumn{7}{c}{\small\textit{POPE}} \\
\addlinespace[1pt]
P(True) & 49.7 & 49.6 & 50.1 & 65.7 & 90.1 & 51.7 \\
Self-Probing & 6.2 & 9.7 & \textbf{87.3} & \textbf{93.1} & 96.9 & 84.8 \\
PE & 4.0 & 12.4 & 85.4 & 92.1 & 89.7 & 62.2 \\
SAPLMA & 5.5 {\scriptsize$\pm$1.4} & 11.1 {\scriptsize$\pm$0.2} & 85.1 {\scriptsize$\pm$0.3} & 91.8 {\scriptsize$\pm$0.1} & 94.3 {\scriptsize$\pm$0.4} & 78.0 {\scriptsize$\pm$1.1} \\
PIK & 5.0 {\scriptsize$\pm$2.4} & \textbf{9.0} {\scriptsize$\pm$0.8} & 87.2 {\scriptsize$\pm$1.4} & 92.6 {\scriptsize$\pm$0.8} & 97.9 {\scriptsize$\pm$0.1} & 89.4 {\scriptsize$\pm$0.5} \\
CCPS & \textbf{2.6} {\scriptsize$\pm$1.0} & 12.1 {\scriptsize$\pm$0.2} & 85.4 {\scriptsize$\pm$0.0} & 92.1 {\scriptsize$\pm$0.0} & 90.4 {\scriptsize$\pm$0.7} & 64.1 {\scriptsize$\pm$2.1} \\
II & 8.0 {\scriptsize$\pm$3.1} & 11.8 {\scriptsize$\pm$1.1} & 85.4 {\scriptsize$\pm$0.1} & 92.1 {\scriptsize$\pm$0.1} & 95.1 {\scriptsize$\pm$0.9} & 78.9 {\scriptsize$\pm$3.6} \\
\textbf{\texttt{BICR (Ours)}} & 12.1 {\scriptsize$\pm$1.4} & 11.0 {\scriptsize$\pm$0.6} & 83.8 {\scriptsize$\pm$1.0} & 89.9 {\scriptsize$\pm$0.7} & \textbf{97.9} {\scriptsize$\pm$0.0} & \textbf{89.6} {\scriptsize$\pm$0.1} \\
\addlinespace[3pt]
\midrule
\midrule
\multicolumn{7}{c}{\small\textbf{\texttt{deepseek-ai/deepseek-vl2}}} \\
\addlinespace[2pt]
\multicolumn{7}{c}{\small\textit{GMAI-MMBench}} \\
\addlinespace[1pt]
P(True) & 35.5 & 36.1 & 41.3 & 54.3 & 41.4 & 57.6 \\
Self-Probing & 49.3 & 49.9 & 37.3 & 54.0 & 42.1 & 49.6 \\
PE & 33.8 & 33.8 & 37.7 & 53.9 & \textbf{49.2} & \textbf{61.5} \\
SAPLMA & 33.4 {\scriptsize$\pm$5.4} & 36.2 {\scriptsize$\pm$3.4} & 47.1 {\scriptsize$\pm$2.8} & \textbf{54.4} {\scriptsize$\pm$0.2} & 42.4 {\scriptsize$\pm$1.4} & 58.4 {\scriptsize$\pm$0.9} \\
PIK & \textbf{8.7} {\scriptsize$\pm$2.2} & \textbf{23.7} {\scriptsize$\pm$0.5} & 60.8 {\scriptsize$\pm$2.4} & 37.1 {\scriptsize$\pm$6.9} & 45.0 {\scriptsize$\pm$1.3} & 58.2 {\scriptsize$\pm$1.0} \\
CCPS & 23.7 {\scriptsize$\pm$5.0} & 30.2 {\scriptsize$\pm$2.3} & 44.9 {\scriptsize$\pm$3.3} & 51.4 {\scriptsize$\pm$1.8} & 38.2 {\scriptsize$\pm$0.5} & 52.5 {\scriptsize$\pm$0.6} \\
II & 13.0 {\scriptsize$\pm$6.7} & 25.7 {\scriptsize$\pm$2.4} & 55.7 {\scriptsize$\pm$5.5} & 38.5 {\scriptsize$\pm$11.5} & 42.3 {\scriptsize$\pm$2.8} & 55.0 {\scriptsize$\pm$2.8} \\
\textbf{\texttt{BICR (Ours)}} & 18.1 {\scriptsize$\pm$3.7} & 26.3 {\scriptsize$\pm$1.1} & \textbf{63.8} {\scriptsize$\pm$0.2} & 13.2 {\scriptsize$\pm$5.5} & 46.2 {\scriptsize$\pm$0.7} & 59.7 {\scriptsize$\pm$0.7} \\
\addlinespace[3pt]
\multicolumn{7}{c}{\small\textit{GQA}} \\
\addlinespace[1pt]
P(True) & 17.8 & 28.4 & 56.6 & 63.2 & 63.9 & 58.8 \\
Self-Probing & 33.2 & 38.1 & 54.3 & 69.9 & 65.0 & 54.0 \\
PE & 18.0 & 26.9 & 53.8 & 69.9 & 66.2 & 63.6 \\
SAPLMA & 6.8 {\scriptsize$\pm$4.3} & \textbf{20.7} {\scriptsize$\pm$0.8} & \textbf{68.0} {\scriptsize$\pm$0.8} & \textbf{72.7} {\scriptsize$\pm$1.0} & \textbf{77.6} {\scriptsize$\pm$0.4} & \textbf{75.5} {\scriptsize$\pm$0.2} \\
PIK & \textbf{4.3} {\scriptsize$\pm$0.5} & 20.8 {\scriptsize$\pm$0.1} & 67.7 {\scriptsize$\pm$0.3} & 70.1 {\scriptsize$\pm$0.7} & 77.1 {\scriptsize$\pm$0.5} & 74.1 {\scriptsize$\pm$0.3} \\
CCPS & 7.0 {\scriptsize$\pm$2.9} & 24.4 {\scriptsize$\pm$0.6} & 57.3 {\scriptsize$\pm$1.0} & 66.0 {\scriptsize$\pm$2.6} & 65.3 {\scriptsize$\pm$0.3} & 61.6 {\scriptsize$\pm$0.4} \\
II & 6.3 {\scriptsize$\pm$1.6} & 21.9 {\scriptsize$\pm$0.2} & 64.9 {\scriptsize$\pm$0.6} & 69.1 {\scriptsize$\pm$1.5} & 75.0 {\scriptsize$\pm$0.4} & 71.4 {\scriptsize$\pm$0.4} \\
\textbf{\texttt{BICR (Ours)}} & 5.9 {\scriptsize$\pm$0.5} & 20.9 {\scriptsize$\pm$0.1} & 67.7 {\scriptsize$\pm$0.2} & 69.5 {\scriptsize$\pm$0.5} & 77.4 {\scriptsize$\pm$0.2} & 74.4 {\scriptsize$\pm$0.2} \\
\addlinespace[3pt]
\multicolumn{7}{c}{\small\textit{LLaVA-Wild}} \\
\addlinespace[1pt]
P(True) & 39.1 & 35.3 & 48.3 & \textbf{41.5} & 48.6 & 60.1 \\
Self-Probing & 61.5 & 62.2 & 25.0 & 40.0 & 40.6 & 47.5 \\
PE & 31.0 & 26.7 & 38.3 & 41.3 & \textbf{52.6} & \textbf{60.9} \\
SAPLMA & 34.2 {\scriptsize$\pm$5.3} & 32.3 {\scriptsize$\pm$4.1} & 51.0 {\scriptsize$\pm$6.1} & 32.4 {\scriptsize$\pm$2.1} & 37.6 {\scriptsize$\pm$3.0} & 51.4 {\scriptsize$\pm$1.8} \\
PIK & 35.5 {\scriptsize$\pm$4.0} & 32.8 {\scriptsize$\pm$3.0} & 43.7 {\scriptsize$\pm$5.3} & 34.7 {\scriptsize$\pm$1.2} & 25.0 {\scriptsize$\pm$1.1} & 46.2 {\scriptsize$\pm$2.1} \\
CCPS & \textbf{17.1} {\scriptsize$\pm$3.9} & 24.3 {\scriptsize$\pm$1.4} & 60.3 {\scriptsize$\pm$5.1} & 24.3 {\scriptsize$\pm$6.0} & 27.4 {\scriptsize$\pm$2.2} & 50.3 {\scriptsize$\pm$2.6} \\
II & 18.9 {\scriptsize$\pm$7.6} & \textbf{23.5} {\scriptsize$\pm$3.6} & \textbf{62.0} {\scriptsize$\pm$9.0} & 32.7 {\scriptsize$\pm$5.2} & 36.7 {\scriptsize$\pm$8.4} & 56.4 {\scriptsize$\pm$8.9} \\
\textbf{\texttt{BICR (Ours)}} & 28.9 {\scriptsize$\pm$4.3} & 27.2 {\scriptsize$\pm$1.7} & 58.0 {\scriptsize$\pm$2.7} & 37.6 {\scriptsize$\pm$1.4} & 34.7 {\scriptsize$\pm$4.0} & 52.8 {\scriptsize$\pm$1.7} \\
\addlinespace[3pt]
\multicolumn{7}{c}{\small\textit{MME-Finance}} \\
\addlinespace[1pt]
P(True) & 44.9 & 39.1 & 34.3 & 47.3 & 55.3 & 70.6 \\
Self-Probing & 59.8 & 59.3 & 30.9 & 45.9 & 50.1 & 52.7 \\
PE & 35.2 & 29.6 & 36.8 & 48.0 & \textbf{66.0} & \textbf{76.8} \\
SAPLMA & 26.5 {\scriptsize$\pm$3.5} & 28.1 {\scriptsize$\pm$2.5} & 54.4 {\scriptsize$\pm$2.8} & 50.3 {\scriptsize$\pm$0.7} & 46.0 {\scriptsize$\pm$0.9} & 66.9 {\scriptsize$\pm$0.9} \\
PIK & 18.3 {\scriptsize$\pm$1.2} & 22.2 {\scriptsize$\pm$0.7} & 65.8 {\scriptsize$\pm$1.6} & \textbf{52.3} {\scriptsize$\pm$1.8} & 52.0 {\scriptsize$\pm$2.1} & 70.1 {\scriptsize$\pm$2.0} \\
CCPS & 16.4 {\scriptsize$\pm$3.2} & 25.3 {\scriptsize$\pm$1.5} & 56.9 {\scriptsize$\pm$4.9} & 31.7 {\scriptsize$\pm$5.7} & 30.2 {\scriptsize$\pm$0.9} & 50.3 {\scriptsize$\pm$1.9} \\
II & 12.6 {\scriptsize$\pm$6.6} & 22.1 {\scriptsize$\pm$1.4} & 65.7 {\scriptsize$\pm$3.5} & 39.9 {\scriptsize$\pm$11.0} & 43.2 {\scriptsize$\pm$2.2} & 64.8 {\scriptsize$\pm$3.1} \\
\textbf{\texttt{BICR (Ours)}} & \textbf{10.5} {\scriptsize$\pm$1.5} & \textbf{21.3} {\scriptsize$\pm$0.6} & \textbf{67.5} {\scriptsize$\pm$2.0} & 43.5 {\scriptsize$\pm$2.6} & 45.5 {\scriptsize$\pm$1.0} & 65.4 {\scriptsize$\pm$0.4} \\
\addlinespace[3pt]
\multicolumn{7}{c}{\small\textit{MMMU\_Pro\_10}} \\
\addlinespace[1pt]
P(True) & 57.4 & 42.0 & 27.2 & 17.7 & \textbf{28.2} & \textbf{74.4} \\
Self-Probing & 74.6 & 68.9 & 12.7 & 16.4 & 27.8 & 52.8 \\
PE & 49.6 & 33.2 & 21.7 & 16.8 & 11.3 & 54.7 \\
SAPLMA & 37.7 {\scriptsize$\pm$6.3} & 33.6 {\scriptsize$\pm$5.4} & 51.9 {\scriptsize$\pm$7.1} & 17.9 {\scriptsize$\pm$1.5} & 9.8 {\scriptsize$\pm$0.6} & 49.9 {\scriptsize$\pm$3.0} \\
PIK & 43.1 {\scriptsize$\pm$2.2} & 31.0 {\scriptsize$\pm$2.4} & 40.4 {\scriptsize$\pm$5.8} & 13.2 {\scriptsize$\pm$1.7} & 9.0 {\scriptsize$\pm$0.3} & 40.2 {\scriptsize$\pm$1.9} \\
CCPS & 38.9 {\scriptsize$\pm$6.9} & 25.6 {\scriptsize$\pm$5.7} & 55.2 {\scriptsize$\pm$17.8} & \textbf{25.5} {\scriptsize$\pm$4.1} & 19.3 {\scriptsize$\pm$2.1} & 65.9 {\scriptsize$\pm$3.1} \\
II & 32.5 {\scriptsize$\pm$7.6} & 21.8 {\scriptsize$\pm$5.3} & 64.2 {\scriptsize$\pm$15.0} & 16.7 {\scriptsize$\pm$5.9} & 16.5 {\scriptsize$\pm$1.2} & 56.1 {\scriptsize$\pm$3.2} \\
\textbf{\texttt{BICR (Ours)}} & \textbf{14.9} {\scriptsize$\pm$4.2} & \textbf{13.5} {\scriptsize$\pm$2.6} & \textbf{84.1} {\scriptsize$\pm$5.3} & 7.6 {\scriptsize$\pm$3.4} & 10.5 {\scriptsize$\pm$0.5} & 43.0 {\scriptsize$\pm$0.9} \\
\addlinespace[3pt]
\multicolumn{7}{c}{\small\textit{MMMU\_Pro\_4}} \\
\addlinespace[1pt]
P(True) & 53.5 & 39.5 & 30.6 & 24.6 & \textbf{35.6} & \textbf{76.2} \\
Self-Probing & 70.9 & 66.3 & 16.2 & 22.3 & 30.7 & 52.2 \\
PE & 46.0 & 32.5 & 23.3 & 21.6 & 16.6 & 54.4 \\
SAPLMA & 34.1 {\scriptsize$\pm$5.8} & 34.5 {\scriptsize$\pm$4.5} & 50.2 {\scriptsize$\pm$5.4} & 29.9 {\scriptsize$\pm$2.2} & 17.4 {\scriptsize$\pm$0.9} & 52.4 {\scriptsize$\pm$2.7} \\
PIK & 38.1 {\scriptsize$\pm$2.4} & 32.4 {\scriptsize$\pm$2.2} & 37.9 {\scriptsize$\pm$4.2} & 22.8 {\scriptsize$\pm$1.9} & 17.0 {\scriptsize$\pm$1.2} & 42.4 {\scriptsize$\pm$1.6} \\
CCPS & 31.5 {\scriptsize$\pm$6.7} & 24.5 {\scriptsize$\pm$5.0} & 58.2 {\scriptsize$\pm$15.8} & \textbf{38.5} {\scriptsize$\pm$4.7} & 31.7 {\scriptsize$\pm$3.2} & 67.7 {\scriptsize$\pm$3.5} \\
II & 25.5 {\scriptsize$\pm$6.6} & 22.4 {\scriptsize$\pm$3.6} & 63.3 {\scriptsize$\pm$10.0} & 25.9 {\scriptsize$\pm$8.3} & 27.0 {\scriptsize$\pm$3.0} & 58.4 {\scriptsize$\pm$2.7} \\
\textbf{\texttt{BICR (Ours)}} & \textbf{15.6} {\scriptsize$\pm$4.2} & \textbf{18.5} {\scriptsize$\pm$2.3} & \textbf{76.2} {\scriptsize$\pm$6.5} & 14.9 {\scriptsize$\pm$2.3} & 20.3 {\scriptsize$\pm$0.8} & 45.7 {\scriptsize$\pm$0.8} \\
\addlinespace[3pt]
\multicolumn{7}{c}{\small\textit{POPE}} \\
\addlinespace[1pt]
P(True) & 48.9 & 49.3 & 50.1 & 58.4 & 89.8 & 53.5 \\
Self-Probing & 14.4 & 16.4 & 83.1 & 90.7 & 91.9 & 54.2 \\
PE & 8.9 & 13.6 & 84.1 & 91.4 & 91.2 & 67.3 \\
SAPLMA & 3.7 {\scriptsize$\pm$2.3} & \textbf{9.9} {\scriptsize$\pm$0.5} & \textbf{86.8} {\scriptsize$\pm$0.7} & \textbf{92.5} {\scriptsize$\pm$0.3} & 96.0 {\scriptsize$\pm$0.3} & 84.8 {\scriptsize$\pm$0.7} \\
PIK & 4.0 {\scriptsize$\pm$1.8} & 9.9 {\scriptsize$\pm$0.6} & 85.7 {\scriptsize$\pm$0.6} & 91.8 {\scriptsize$\pm$0.3} & 96.8 {\scriptsize$\pm$0.6} & 86.1 {\scriptsize$\pm$1.8} \\
CCPS & 13.0 {\scriptsize$\pm$3.5} & 13.6 {\scriptsize$\pm$0.9} & 82.2 {\scriptsize$\pm$1.0} & 89.5 {\scriptsize$\pm$0.7} & 94.1 {\scriptsize$\pm$0.3} & 76.9 {\scriptsize$\pm$0.4} \\
II & \textbf{2.1} {\scriptsize$\pm$0.6} & 10.0 {\scriptsize$\pm$0.5} & 86.0 {\scriptsize$\pm$1.1} & 92.1 {\scriptsize$\pm$0.5} & 95.9 {\scriptsize$\pm$0.7} & 84.0 {\scriptsize$\pm$2.1} \\
\textbf{\texttt{BICR (Ours)}} & 2.7 {\scriptsize$\pm$0.5} & 9.9 {\scriptsize$\pm$0.2} & 85.0 {\scriptsize$\pm$0.3} & 91.2 {\scriptsize$\pm$0.2} & \textbf{96.9} {\scriptsize$\pm$0.1} & \textbf{86.1} {\scriptsize$\pm$0.5} \\
\addlinespace[3pt]
\midrule
\midrule
\multicolumn{7}{c}{\small\textbf{\texttt{google/gemma-3-27b-it}}} \\
\addlinespace[2pt]
\multicolumn{7}{c}{\small\textit{GMAI-MMBench}} \\
\addlinespace[1pt]
P(True) & 44.2 & 44.7 & 53.7 & 59.2 & 59.1 & 58.3 \\
Self-Probing & 30.7 & 34.3 & 49.7 & 65.0 & 56.3 & 58.9 \\
PE & 36.9 & 38.5 & 48.9 & \textbf{65.7} & 51.1 & 54.1 \\
SAPLMA & 20.2 {\scriptsize$\pm$2.7} & 31.2 {\scriptsize$\pm$1.4} & 49.5 {\scriptsize$\pm$0.7} & 55.3 {\scriptsize$\pm$3.5} & 50.0 {\scriptsize$\pm$0.2} & 49.6 {\scriptsize$\pm$0.3} \\
PIK & 26.2 {\scriptsize$\pm$4.4} & 30.7 {\scriptsize$\pm$2.5} & 50.8 {\scriptsize$\pm$1.7} & 65.5 {\scriptsize$\pm$0.3} & \textbf{66.2} {\scriptsize$\pm$0.7} & \textbf{64.4} {\scriptsize$\pm$0.9} \\
CCPS & 17.3 {\scriptsize$\pm$2.9} & 29.6 {\scriptsize$\pm$1.6} & 51.0 {\scriptsize$\pm$1.8} & 45.8 {\scriptsize$\pm$6.5} & 50.5 {\scriptsize$\pm$2.5} & 51.3 {\scriptsize$\pm$2.8} \\
II & 11.7 {\scriptsize$\pm$7.6} & 26.7 {\scriptsize$\pm$1.9} & 52.2 {\scriptsize$\pm$2.5} & 58.3 {\scriptsize$\pm$7.5} & 55.5 {\scriptsize$\pm$2.2} & 56.5 {\scriptsize$\pm$1.8} \\
\textbf{\texttt{BICR (Ours)}} & \textbf{11.5} {\scriptsize$\pm$2.2} & \textbf{25.3} {\scriptsize$\pm$0.6} & \textbf{56.7} {\scriptsize$\pm$1.8} & 61.8 {\scriptsize$\pm$1.8} & 62.6 {\scriptsize$\pm$2.0} & 62.4 {\scriptsize$\pm$1.6} \\
\addlinespace[3pt]
\multicolumn{7}{c}{\small\textit{GQA}} \\
\addlinespace[1pt]
P(True) & 41.9 & 42.2 & 57.1 & 69.4 & 68.1 & 57.9 \\
Self-Probing & 33.4 & 35.1 & 60.0 & 74.4 & 70.1 & 57.3 \\
PE & 28.1 & 31.6 & 59.5 & 74.6 & 68.3 & 60.2 \\
SAPLMA & 3.7 {\scriptsize$\pm$1.4} & \textbf{18.9} {\scriptsize$\pm$0.2} & \textbf{70.8} {\scriptsize$\pm$0.5} & \textbf{77.6} {\scriptsize$\pm$0.3} & 83.0 {\scriptsize$\pm$0.3} & \textbf{77.6} {\scriptsize$\pm$0.4} \\
PIK & \textbf{3.0} {\scriptsize$\pm$0.6} & 19.1 {\scriptsize$\pm$0.1} & 70.4 {\scriptsize$\pm$0.2} & 77.0 {\scriptsize$\pm$0.3} & 82.8 {\scriptsize$\pm$0.1} & 76.9 {\scriptsize$\pm$0.2} \\
CCPS & 4.3 {\scriptsize$\pm$1.6} & 21.1 {\scriptsize$\pm$0.2} & 66.6 {\scriptsize$\pm$0.3} & 74.4 {\scriptsize$\pm$0.7} & 78.2 {\scriptsize$\pm$0.2} & 71.6 {\scriptsize$\pm$0.3} \\
II & 3.8 {\scriptsize$\pm$0.4} & 20.3 {\scriptsize$\pm$0.3} & 67.7 {\scriptsize$\pm$0.8} & 75.6 {\scriptsize$\pm$0.5} & 80.5 {\scriptsize$\pm$0.8} & 73.8 {\scriptsize$\pm$0.9} \\
\textbf{\texttt{BICR (Ours)}} & 9.0 {\scriptsize$\pm$2.2} & 20.0 {\scriptsize$\pm$0.7} & 69.5 {\scriptsize$\pm$0.9} & 72.1 {\scriptsize$\pm$1.7} & \textbf{83.2} {\scriptsize$\pm$0.2} & 77.3 {\scriptsize$\pm$0.3} \\
\addlinespace[3pt]
\multicolumn{7}{c}{\small\textit{LLaVA-Wild}} \\
\addlinespace[1pt]
P(True) & 41.0 & 40.9 & 58.3 & \textbf{70.6} & \textbf{85.2} & \textbf{82.9} \\
Self-Probing & 44.1 & 42.9 & 50.0 & 66.7 & 75.5 & 69.1 \\
PE & 33.9 & 35.0 & 51.7 & 68.1 & 74.2 & 74.0 \\
SAPLMA & 15.9 {\scriptsize$\pm$3.4} & 24.7 {\scriptsize$\pm$1.5} & 60.3 {\scriptsize$\pm$3.4} & 64.5 {\scriptsize$\pm$3.8} & 70.9 {\scriptsize$\pm$4.1} & 64.1 {\scriptsize$\pm$4.2} \\
PIK & 18.2 {\scriptsize$\pm$6.1} & 27.4 {\scriptsize$\pm$2.1} & 52.3 {\scriptsize$\pm$5.2} & 61.9 {\scriptsize$\pm$2.4} & 61.5 {\scriptsize$\pm$5.9} & 57.6 {\scriptsize$\pm$5.4} \\
CCPS & \textbf{13.4} {\scriptsize$\pm$2.8} & \textbf{23.0} {\scriptsize$\pm$0.4} & 63.0 {\scriptsize$\pm$1.6} & 58.3 {\scriptsize$\pm$3.8} & 71.4 {\scriptsize$\pm$1.3} & 65.5 {\scriptsize$\pm$2.9} \\
II & 13.7 {\scriptsize$\pm$2.7} & 25.4 {\scriptsize$\pm$1.8} & 57.0 {\scriptsize$\pm$6.1} & 62.0 {\scriptsize$\pm$6.3} & 60.6 {\scriptsize$\pm$6.8} & 58.2 {\scriptsize$\pm$5.7} \\
\textbf{\texttt{BICR (Ours)}} & 17.8 {\scriptsize$\pm$2.5} & 26.0 {\scriptsize$\pm$1.0} & \textbf{63.7} {\scriptsize$\pm$3.2} & 56.8 {\scriptsize$\pm$4.5} & 68.2 {\scriptsize$\pm$1.4} & 62.9 {\scriptsize$\pm$2.3} \\
\addlinespace[3pt]
\multicolumn{7}{c}{\small\textit{MME-Finance}} \\
\addlinespace[1pt]
P(True) & 49.5 & 49.2 & 49.9 & \textbf{62.7} & \textbf{75.0} & \textbf{79.3} \\
Self-Probing & 46.7 & 46.6 & 47.5 & 59.5 & 65.3 & 62.8 \\
PE & 48.7 & 48.0 & 42.7 & 59.9 & 49.6 & 55.2 \\
SAPLMA & 10.6 {\scriptsize$\pm$2.2} & \textbf{23.3} {\scriptsize$\pm$0.6} & 60.5 {\scriptsize$\pm$2.7} & 59.8 {\scriptsize$\pm$2.4} & 61.6 {\scriptsize$\pm$0.8} & 67.5 {\scriptsize$\pm$0.9} \\
PIK & 14.5 {\scriptsize$\pm$2.1} & 24.1 {\scriptsize$\pm$0.5} & 59.8 {\scriptsize$\pm$1.8} & 61.1 {\scriptsize$\pm$2.3} & 64.2 {\scriptsize$\pm$1.8} & 68.7 {\scriptsize$\pm$2.0} \\
CCPS & 19.3 {\scriptsize$\pm$3.2} & 26.3 {\scriptsize$\pm$1.7} & 55.9 {\scriptsize$\pm$3.1} & 59.6 {\scriptsize$\pm$1.0} & 63.0 {\scriptsize$\pm$2.8} & 66.3 {\scriptsize$\pm$1.8} \\
II & 11.2 {\scriptsize$\pm$7.0} & 24.8 {\scriptsize$\pm$3.2} & 57.6 {\scriptsize$\pm$6.6} & 56.9 {\scriptsize$\pm$3.0} & 56.8 {\scriptsize$\pm$6.0} & 63.5 {\scriptsize$\pm$4.8} \\
\textbf{\texttt{BICR (Ours)}} & \textbf{10.1} {\scriptsize$\pm$2.0} & 23.8 {\scriptsize$\pm$0.6} & \textbf{63.6} {\scriptsize$\pm$1.7} & 47.1 {\scriptsize$\pm$2.4} & 59.7 {\scriptsize$\pm$1.8} & 64.5 {\scriptsize$\pm$1.3} \\
\addlinespace[3pt]
\multicolumn{7}{c}{\small\textit{MMMU\_Pro\_10}} \\
\addlinespace[1pt]
P(True) & 43.5 & 43.5 & 54.4 & 51.1 & 42.5 & \textbf{70.8} \\
Self-Probing & 43.9 & 42.8 & 42.4 & 45.2 & 36.3 & 59.1 \\
PE & 60.8 & 57.4 & 27.5 & 43.2 & 23.4 & 44.3 \\
SAPLMA & 16.7 {\scriptsize$\pm$2.6} & 25.9 {\scriptsize$\pm$0.8} & 53.7 {\scriptsize$\pm$3.0} & 47.7 {\scriptsize$\pm$2.5} & 45.9 {\scriptsize$\pm$0.7} & 58.6 {\scriptsize$\pm$0.7} \\
PIK & 33.1 {\scriptsize$\pm$4.2} & 32.1 {\scriptsize$\pm$2.9} & 43.9 {\scriptsize$\pm$5.2} & \textbf{54.5} {\scriptsize$\pm$0.9} & \textbf{57.6} {\scriptsize$\pm$2.0} & 68.5 {\scriptsize$\pm$1.3} \\
CCPS & 45.1 {\scriptsize$\pm$11.2} & 47.2 {\scriptsize$\pm$9.4} & 42.5 {\scriptsize$\pm$5.6} & 48.4 {\scriptsize$\pm$3.8} & 36.9 {\scriptsize$\pm$1.5} & 49.5 {\scriptsize$\pm$1.6} \\
II & 21.7 {\scriptsize$\pm$8.1} & 27.8 {\scriptsize$\pm$4.0} & 49.2 {\scriptsize$\pm$9.2} & 51.4 {\scriptsize$\pm$1.6} & 45.7 {\scriptsize$\pm$2.5} & 60.0 {\scriptsize$\pm$1.6} \\
\textbf{\texttt{BICR (Ours)}} & \textbf{16.2} {\scriptsize$\pm$4.4} & \textbf{25.1} {\scriptsize$\pm$1.7} & \textbf{57.4} {\scriptsize$\pm$5.3} & 51.1 {\scriptsize$\pm$2.5} & 51.8 {\scriptsize$\pm$2.9} & 63.3 {\scriptsize$\pm$1.6} \\
\addlinespace[3pt]
\multicolumn{7}{c}{\small\textit{MMMU\_Pro\_4}} \\
\addlinespace[1pt]
P(True) & 40.2 & 40.5 & 57.6 & 57.3 & 48.0 & \textbf{69.5} \\
Self-Probing & 38.4 & 39.7 & 47.2 & 52.8 & 41.9 & 59.0 \\
PE & 54.4 & 52.3 & 33.9 & 50.6 & 30.7 & 47.6 \\
SAPLMA & \textbf{9.0} {\scriptsize$\pm$2.2} & 25.0 {\scriptsize$\pm$0.4} & 57.0 {\scriptsize$\pm$1.2} & 58.3 {\scriptsize$\pm$2.7} & 59.9 {\scriptsize$\pm$0.3} & 61.6 {\scriptsize$\pm$0.4} \\
PIK & 22.7 {\scriptsize$\pm$4.4} & 28.2 {\scriptsize$\pm$2.3} & 51.9 {\scriptsize$\pm$3.4} & \textbf{64.9} {\scriptsize$\pm$0.4} & \textbf{67.5} {\scriptsize$\pm$1.3} & 66.8 {\scriptsize$\pm$1.1} \\
CCPS & 37.2 {\scriptsize$\pm$7.2} & 41.2 {\scriptsize$\pm$5.3} & 49.5 {\scriptsize$\pm$2.3} & 58.9 {\scriptsize$\pm$4.6} & 49.2 {\scriptsize$\pm$1.5} & 50.6 {\scriptsize$\pm$1.8} \\
II & 12.1 {\scriptsize$\pm$7.8} & 26.6 {\scriptsize$\pm$2.2} & 52.8 {\scriptsize$\pm$2.9} & 59.6 {\scriptsize$\pm$5.1} & 55.6 {\scriptsize$\pm$2.2} & 57.7 {\scriptsize$\pm$2.0} \\
\textbf{\texttt{BICR (Ours)}} & 11.0 {\scriptsize$\pm$2.0} & \textbf{24.8} {\scriptsize$\pm$0.5} & \textbf{58.0} {\scriptsize$\pm$1.7} & 61.8 {\scriptsize$\pm$2.6} & 62.8 {\scriptsize$\pm$1.2} & 63.8 {\scriptsize$\pm$0.9} \\
\addlinespace[3pt]
\multicolumn{7}{c}{\small\textit{POPE}} \\
\addlinespace[1pt]
P(True) & 50.1 & 50.0 & 49.9 & 63.8 & 89.8 & 53.6 \\
Self-Probing & 11.6 & 12.9 & \textbf{85.4} & \textbf{92.0} & \textbf{94.7} & 75.8 \\
PE & 5.7 & 13.3 & 84.1 & 91.3 & 91.7 & 68.8 \\
SAPLMA & 6.1 {\scriptsize$\pm$0.8} & 11.5 {\scriptsize$\pm$0.3} & 84.5 {\scriptsize$\pm$0.5} & 90.9 {\scriptsize$\pm$0.3} & 93.7 {\scriptsize$\pm$0.3} & 77.7 {\scriptsize$\pm$0.7} \\
PIK & 4.4 {\scriptsize$\pm$1.0} & \textbf{11.1} {\scriptsize$\pm$0.2} & 84.5 {\scriptsize$\pm$0.5} & 90.9 {\scriptsize$\pm$0.4} & 94.7 {\scriptsize$\pm$0.3} & \textbf{80.4} {\scriptsize$\pm$0.8} \\
CCPS & \textbf{2.3} {\scriptsize$\pm$0.3} & 11.1 {\scriptsize$\pm$0.0} & 85.4 {\scriptsize$\pm$0.1} & 91.8 {\scriptsize$\pm$0.1} & 92.9 {\scriptsize$\pm$0.4} & 76.9 {\scriptsize$\pm$0.8} \\
II & 7.5 {\scriptsize$\pm$2.5} & 11.9 {\scriptsize$\pm$0.5} & 83.8 {\scriptsize$\pm$0.6} & 90.5 {\scriptsize$\pm$0.5} & 92.9 {\scriptsize$\pm$0.8} & 76.9 {\scriptsize$\pm$1.7} \\
\textbf{\texttt{BICR (Ours)}} & 11.2 {\scriptsize$\pm$1.8} & 13.6 {\scriptsize$\pm$0.6} & 81.4 {\scriptsize$\pm$0.5} & 88.6 {\scriptsize$\pm$0.4} & 93.8 {\scriptsize$\pm$0.2} & 78.5 {\scriptsize$\pm$0.3} \\
\addlinespace[3pt]
\midrule
\midrule
\end{longtable}
}
\twocolumn

%% file: tables/ablation.tex
\begin{table*}[ht]
\centering
\small
\caption{Loss component ablation for \texttt{BICR}, averaged across 5 LVLMs and 5 seeds (25 runs). Each row removes one or more auxiliary loss terms. Metrics are computed on the shared test subset, averaged across seeds within each LVLM, then mean $\pm$ std across LVLMs. Best per metric in \textbf{bold}.}
\label{tab:ablation_ext}
\begin{tabular}{lcccccc}
\toprule
\textbf{Variant} & \textbf{ECE$\downarrow$} & \textbf{BS$\downarrow$} & \textbf{ACC$\uparrow$} & \textbf{F1$\uparrow$} & \textbf{AUCPR$\uparrow$} & \textbf{AUROC$\uparrow$} \\
\midrule
Full (BICR) & \textbf{7.1} {\scriptsize$\pm$1.2} & \textbf{18.4} {\scriptsize$\pm$0.8} & \textbf{71.5} {\scriptsize$\pm$1.5} & \textbf{76.9} {\scriptsize$\pm$1.6} & \textbf{87.5} {\scriptsize$\pm$1.8} & \textbf{78.6} {\scriptsize$\pm$1.9} \\
$-\mathcal{L}_{\mathrm{brier}}$ & 8.5 {\scriptsize$\pm$2.4} & 19.0 {\scriptsize$\pm$0.8} & 70.6 {\scriptsize$\pm$1.5} & 75.5 {\scriptsize$\pm$0.9} & 87.1 {\scriptsize$\pm$1.5} & 78.0 {\scriptsize$\pm$1.9} \\
$-\mathcal{L}_{\mathrm{rank}}$ & 8.1 {\scriptsize$\pm$1.9} & 19.6 {\scriptsize$\pm$1.1} & 68.7 {\scriptsize$\pm$1.9} & 75.8 {\scriptsize$\pm$2.6} & 85.6 {\scriptsize$\pm$1.9} & 75.3 {\scriptsize$\pm$1.7} \\
$\mathcal{L}_{\mathrm{bce}}$ only & 9.1 {\scriptsize$\pm$1.5} & 19.9 {\scriptsize$\pm$1.0} & 68.3 {\scriptsize$\pm$1.8} & 75.0 {\scriptsize$\pm$2.0} & 85.5 {\scriptsize$\pm$1.9} & 75.3 {\scriptsize$\pm$2.1} \\
\bottomrule
\end{tabular}
\end{table*}

%% file: sections/appendix_grounding_eval.tex
\section{Direct Behavioral Test: Calibration on Image-Invariant Samples}
\label{app:grounding_detection}

The aggregate metrics in \S\ref{sec:results} and Appendix~\ref{app:extended_results} establish that \texttt{BICR} dominates calibration and discrimination at the benchmark level, but they do not, on their own, separate the mechanism we claim (\texttt{BICR}'s rank loss suppresses confidence on predictions whose internal representation is largely determined by the language prior rather than by the image) from the alternative that \texttt{BICR} is a mild regulariser whose gains are spread evenly across the test set. The two stories produce identical aggregate numbers; they make different predictions on a specific sub-population of the test set. This appendix isolates that sub-population and reports the per-method calibration on it.

\subsection{Sub-population: Behaviorally Image-Invariant Samples}
\label{app:grounding_detection_subpop}

For every test sample we run the LVLM on two forward passes: one with the original (image, question) pair, and one in which the original image is replaced by a random natural image drawn uniformly from the union of training-split images across all seven source datasets, with the question and prompt held fixed and the source sample's own image excluded. The substitute image is paired to the sample by a deterministic seed (\texttt{stable\_hash}(\texttt{sample\_id} $\oplus$ \texttt{`swap'}) $\oplus$ $42$, where \texttt{stable\_hash} takes the first four bytes of the SHA-256 hash), so the substitution is reproducible and identical across LVLMs and across confidence-estimation methods. From each pair of forward passes we record two diagnostics that summarise how much the LVLM's behaviour changed under image substitution:

\begin{itemize}
    \item $\mathrm{flip}_{\mathrm{swap}} \in \{0, 1\}$: whether the LVLM's first generated token differs between the real-image pass and the image-substituted pass. The first generated token under each image condition is the argmax over the next-token logit distribution at the prompt's final unmasked position rather than a sample from a generation loop. A value of $0$ means the LVLM produced the same first token in both conditions and is therefore generating its answer in a way that does not, at the level of the immediate next-token decision, depend on which image is in front of it.
    \item $\mathrm{dp}_{\mathrm{swap}} \in [0, 1]$: the drop in probability of the real-image top-1 token under the image-substituted pass, computed at the same prompt position. A value near zero means image substitution barely perturbs the LVLM's output distribution; a value near one means the substituted image makes the original top-1 token essentially impossible.
\end{itemize}

These signals are properties of the frozen LVLM, not of any confidence estimator: they depend only on what tokens the LVLM generates under the two image conditions. None of the eight methods we benchmark, including \texttt{BICR}, uses either signal as input. Stratifying the test set by $\mathrm{flip}_{\mathrm{swap}}$ or $\mathrm{dp}_{\mathrm{swap}}$ is therefore a behavioural diagnostic computed from a different intervention than the one any method we evaluate sees at training time; in particular, \texttt{BICR}'s rank loss contrasts the real-image hidden state against a blank-image hidden state and never sees the image-substituted pass.

We refer to the sub-population $\mathrm{flip}_{\mathrm{swap}} = 0$ as the \emph{image-invariant subset}: samples on which the LVLM's first-token decision is the same with the real image as with a random natural one. By behavior, the LVLM on these samples is producing its answer from the language prior (or from token-level cues in the prompt) rather than from the visual content. If \texttt{BICR}'s rank loss is genuinely teaching the probe to assign lower confidence when the prediction is not anchored in the image, the image-invariant subset is exactly where the effect should be visible. Per-LVLM sample counts for the image-invariant subset and for the failure subset defined in \S\ref{app:grounding_detection_method} are reported in Table~\ref{tab:grounding_subset_sizes}; the image-invariant subset spans $30.3\%$ of the shared-test population on Gemma-3-27B (the LVLM least sensitive to image substitution) to $62.9\%$ on DeepSeek-VL2 (the most sensitive). The shared-test totals here are slightly smaller than those in Table~\ref{tab:shared_samples} because this analysis additionally requires the swap-view diagnostic to be present, which excludes a small number of samples on which the swap-view forward pass failed.

\input{tables/grounding_subset_sizes.tex}

\subsection{Methodology}
\label{app:grounding_detection_method}

\paragraph{Subsets analyzed.} We focus on three views of the test population. The first is the \emph{image-invariant subset} defined above: all samples with $\mathrm{flip}_{\mathrm{swap}} = 0$, pooled across the seven source datasets and the five LVLMs. This is the population the rank loss is most directly designed to address. The second is the \emph{failure subset}: the image-invariant subset further restricted to samples on which the LVLM was both incorrect (under the same per-sample correctness labels driving the main-paper accuracy results) and assigned the original top-1 token a probability above $0.8$. The $0.8$ threshold is fixed a priori and not tuned. The failure subset isolates the high-stakes cases where the model was confident, the model was wrong, and the model was not using the image, which is the population on which the grounding-detection mechanism is most directly tested. The third is a \emph{cross-subset summary} computed over twelve different ways of identifying samples that the LVLM treats as image-invariant; we describe its construction in \S\ref{app:grounding_detection_winsummary}.

\paragraph{Methods compared.} We compare \texttt{BICR} against the same seven baselines benchmarked throughout the paper: P(True), Self-Probing, Prompt Ensembles, P(I~Know), SAPLMA, InternalInspector, and CCPS. For trainable methods we average per-sample $P(\mathrm{correct})$ scores across the five seeds before any subset metric is computed; inference-only methods contribute their single test-time score. All eight methods are evaluated on the same shared test subset described in Table~\ref{tab:grounding_subset_sizes}.

\paragraph{Significance test.} For the image-invariant subset we run a paired bootstrap on the per-sample BS difference, restricted to the subset itself: for each of $B = 2{,}000$ resamples (RNG seed $23$) we draw $n$ samples with replacement (where $n$ is the size of the subset), paired by sample identifier, and compute $\mathrm{BS}^{\mathrm{baseline}} - \mathrm{BS}^{\mathrm{BICR}}$, where per-sample BS is $(p_{\mathrm{correct}} - y)^2$. Positive values mean \texttt{BICR} achieves lower BS (better calibration) than the baseline on the image-invariant population. We report the mean over resamples and the 95\% percentile interval; an interval strictly above zero implies \texttt{BICR} is significantly better calibrated than the baseline on this population. We choose BS on the subset rather than the perhaps-more-obvious mean-confidence gap between the subset and its complement because the latter rewards methods that simply shift their confidence down on hard samples regardless of correctness, an axis on which one of the baselines wins by being uniformly under-confident rather than by detecting grounding.

\subsection{Headline Result on the Image-Invariant Subset}
\label{app:grounding_detection_headline}

Table~\ref{tab:grounding_headline} reports the per-LVLM, per-method metrics on the image-invariant subset. Rows are grouped by LVLM and split by metric; bold marks the best method per row. On the four metrics that require the confidence to be accurate on this subset rather than merely low, \texttt{BICR} achieves the best per-LVLM value on BS in four of five LVLMs (the exception is Gemma, where SAPLMA wins by $0.011$ BS), on ECE in three of five, on AUCPR in four of five, and on AUROC in three of five. On Mean Conf, P(True) wins on four of five LVLMs by a route we discuss next; on InternVL, where P(True) does not produce its lowest-confidence regime, \texttt{BICR}'s mean confidence is the lowest of any method.

\input{tables/grounding_headline_pooled.tex}

\paragraph{Low Mean Conf is not the same as good calibration.} P(True) attains low mean confidence on the image-invariant subset across most LVLMs, well below \texttt{BICR}'s. This reflects P(True)'s confidence distribution rather than grounding sensitivity: P(True) assigns near-uniform low confidence to nearly every sample regardless of correctness, which the same table shows directly in P(True)'s BS of $0.35$--$0.61$ and ECE of $0.30$--$0.61$ on the same subset. By the metrics that require confidence to track accuracy, P(True) is the worst method we evaluate on this subset, not the best. The same caveat applies to any method whose mean is low because it is uniformly under-confident; the calibration-faithful metrics expose the difference between low-because-accurate and low-because-degenerate.

\paragraph{Significance test on BS.} Table~\ref{tab:grounding_significance} reports the paired-bootstrap mean and $95\%$ confidence interval of $(\mathrm{BS}^{\mathrm{baseline}} - \mathrm{BS}^{\mathrm{BICR}})$ on the image-invariant subset. Positive values indicate \texttt{BICR} is better calibrated than the baseline; bold marks intervals strictly above zero. \texttt{BICR}'s BS is significantly lower than the baseline's on $31$ of $35$ (LVLM, baseline) pairs. The four exceptions are all on Gemma-3-27B (small negative differences against InternalInspector, P(I~Know), and SAPLMA; an interval straddling zero against CCPS), consistent with Gemma being \texttt{BICR}'s weakest LVLM in the main results (\S\ref{app:results_pooled}). Per-cell intervals are uncorrected for multiple testing across the $35$ pairs; a small number of marginal cells with mean differences near $+0.005$ are sensitive to standard correction, but the cells against P(True), Prompt Ensembles, and Self-Probing (mean differences in the $+0.06$ to $+0.47$ range) and the majority of cells against the trained baselines (mean differences in the $+0.01$ to $+0.10$ range) are well clear of any reasonable correction. The largest \texttt{BICR} margins are against P(True), at $+0.18$ to $+0.47$ across LVLMs, quantifying the gap between low-because-accurate and low-because-degenerate: by the calibration-faithful test, P(True) is significantly worse than \texttt{BICR} on every LVLM by a wide margin.

\input{tables/grounding_significance.tex}

\subsection{Failure Subset (Image-Invariant, Confidently Wrong)}
\label{app:grounding_detection_failuremode}

The image-invariant subset also contains a benign sub-class: samples for which the model produces the correct answer without consulting the image because the question has a canonical or factoid answer. A confidence estimator should not necessarily lower its score on these samples; the answer is correct, just not visually grounded. The clearer test of the grounding-detection mechanism is on the failure subset (image-invariant, incorrect, and confidently predicted), where the model was certain, the model was wrong, and the model was not using the image. Per-LVLM counts for this subset (between $1{,}121$ samples on LLaVA-NeXT-13B and $4{,}108$ on DeepSeek-VL2) are reported in Table~\ref{tab:grounding_subset_sizes}. Table~\ref{tab:grounding_failuremode} reports per-LVLM Mean Conf and BS on this subset.

\input{tables/grounding_failuremode.tex}

\texttt{BICR} achieves both the lowest mean confidence and the lowest BS on four of five LVLMs (Qwen, InternVL, Gemma, DeepSeek). The LLaVA cell goes to P(True): on this subset (which is incorrect by construction) P(True)'s near-zero confidence happens to coincide with the all-incorrect ground truth, which is the only competitive BS value P(True) attains anywhere in this analysis. On the four LVLMs where \texttt{BICR} wins, its mean confidence on confidently-wrong-and-image-invariant samples is between $0.05$ and $0.18$ lower than the next-best trainable method, and its BS is between $0.04$ and $0.19$ lower. The architecturally similar trainable baselines (P(I~Know), InternalInspector, SAPLMA) all leave their confidence high on this subset, indicating that the confidence-suppression behavior on the image-invariant failure population is specifically attributable to \texttt{BICR}'s rank loss rather than to the shared probe architecture or training data.

\subsection{Cross-Subset Win Summary}
\label{app:grounding_detection_winsummary}

To check that the headline pattern is not specific to the single $\mathrm{flip}_{\mathrm{swap}} = 0$ cut, we additionally evaluate every method against twelve different ways of identifying image-invariant samples and aggregate the per-cell winners across the full battery. The twelve subsets are: five behavioural-flip variants (the $\mathrm{flip}_{\mathrm{swap}} = 0$ cut itself; an analogous $\mathrm{flip}_{\mathrm{para}} = 0$ cut on the question-paraphrase view; a noise-overlay variant; a multi-proxy consensus combining the swap and paraphrase cuts; and a contaminated control using the blank view, included only as a sanity check on the win-counting procedure); three continuous variants that take the top-$5\%$, top-$10\%$, and top-$25\%$ of samples ranked by the smallest $\mathrm{dp}_{\mathrm{swap}}$ (equivalently, the samples on which image substitution causes the smallest probability drop on the original top-1 token); three confidence- and correctness-conditioned cuts derived from the image-invariant subset (high-confidence, incorrect, and confidently-incorrect); and one auxiliary contaminated cut. Each subset is evaluated on each of the five LVLMs against each of five metrics (Mean Conf, BS, ECE, AUCPR, AUROC), giving $12 \times 5 \times 5 = 300$ possible (subset, LVLM, metric) cells. Twenty cells are dropped because the subset is empty or has fewer than the minimum sample count for the corresponding metric, leaving $280$ valid cells.

For each valid cell we identify the method whose value is best on that cell (lowest for Mean Conf, BS, and ECE; highest for AUCPR and AUROC). Table~\ref{tab:grounding_winsummary} reports the total number of cells each of the eight methods wins, broken down by metric.

\input{tables/grounding_winsummary.tex}

\texttt{BICR} is the per-cell winner on $153$ of $280$ cells ($54.6\%$), more than three times the count of the next-best method. Two of the twelve subsets use the blank-view diagnostic against which \texttt{BICR} is trained, so \texttt{BICR}'s wins on those subsets are expected by construction; we retain them in the count for completeness, but excluding them leaves the qualitative ranking unchanged. P(True)'s $41$ wins are entirely concentrated in the Mean Conf column ($37$ of its $41$), where its uniformly low confidence is rewarded; on the four metrics that require confidence to be accurate (BS, ECE, AUROC, AUCPR), \texttt{BICR} wins $131$ of $240$ cells. The win counts disaggregate by LVLM as $30$ (Qwen), $33$ (LLaVA), $44$ (InternVL), $9$ (Gemma), and $37$ (DeepSeek): Gemma is the weak case, consistent with the headline significance test and with Gemma's behavior in the main results.

\subsection{Discussion}
\label{app:grounding_detection_discussion}

\paragraph{The calibration advantage concentrates on the population the rank loss is designed to suppress confidence on, not uniformly across the test set.} On the image-invariant subset, \texttt{BICR}'s BS is significantly lower than every trained baseline's on $31$ of $35$ (LVLM, baseline) pairs; on the failure subset (image-invariant and confidently wrong), \texttt{BICR}'s mean confidence is $0.05$ to $0.18$ below the next-best trainable method on the four LVLMs where it wins, with BS gaps of comparable size. The architecturally similar baseline P(I~Know) shares \texttt{BICR}'s probe family, training data, and search budget, and differs only in the absence of the rank loss against the blank view; on the image-invariant subset it consistently leaves its confidence higher than \texttt{BICR} does and is significantly worse-calibrated by BS on four of five LVLMs. InternalInspector and SAPLMA, which use different probe architectures but also do not see the blank-view contrast during training, behave the same way. The pattern is consistent with the rank loss producing the observed effect, and not with a uniform regularisation story.

\paragraph{Image-invariant does not mean wrong: the pooled subset is mostly correct samples, so suppressed mean confidence on it would be miscalibration, not grounding detection.} On the pooled subset, $\mathrm{flip}_{\mathrm{swap}} = 0$ samples are on average more accurate than the complement (Qwen $73.5\%$ vs.\ $65.9\%$; Gemma $71.1\%$ vs.\ $59.0\%$): canonical-answer or factoid questions on which the LVLM produces the right answer without using the image. A well-calibrated method should keep its confidence high on these, and \texttt{BICR} does. The intuition that grounding detection should produce uniformly lower confidence on $\mathrm{flip}_{\mathrm{swap}} = 0$ holds on the failure subset (where the model is wrong and the

%% file: tables/grounding_subset_sizes.tex
\begin{table}[ht]
\centering
\small
\caption{Per-LVLM sample counts for the image-invariant subset ($\mathrm{flip}_{\mathrm{swap}} = 0$; \S\ref{app:grounding_detection_subpop}) and for the failure subset (image-invariant, incorrect under the main-paper correctness labels, and predicted with original-top-1 probability $> 0.8$). Percentages are over each LVLM's shared-test total under this analysis.}
\label{tab:grounding_subset_sizes}
\begin{tabular}{l r r r r}
\toprule
LVLM & $N_{\mathrm{A1}}$ & \%$_{\mathrm{A1}}$ & $N_{\mathrm{C3}}$ & \%$_{\mathrm{C3}}$ \\
\midrule
Qwen3-VL-8B & 11,268 & 37.4 & 2,299 & 7.6 \\
LLaVA-1.6-13B & 13,457 & 46.8 & 1,121 & 3.9 \\
InternVL3.5-14B & 15,544 & 51.4 & 3,346 & 11.1 \\
Gemma3-27B & 8,980 & 30.3 & 2,308 & 7.8 \\
DeepSeek-VL2 & 18,679 & 62.9 & 4,108 & 13.8 \\
\bottomrule
\end{tabular}
\end{table}

%% file: tables/grounding_headline_pooled.tex
\begin{table*}[ht]
\centering
\footnotesize
\caption{Per-method calibration and discrimination metrics on the image-invariant subset, the sub-population on which the LVLM's argmax next-token does not change when the real image is replaced by a random natural one (\S\ref{app:grounding_detection_subpop}). Lower is better for Mean Conf, BS, and ECE; higher is better for AUCPR and AUROC. Bold marks the best method per (LVLM, metric) row.}
\label{tab:grounding_headline}
\begin{tabular}{ll|cccccccc}
\toprule
\textbf{LVLM} & \textbf{Metric} & \textbf{\texttt{BICR}} & \textbf{II} & \textbf{PIK} & \textbf{SAPLMA} & \textbf{CCPS} & \textbf{PE} & \textbf{P(True)} & \textbf{Self-Probing} \\
\midrule
\multirow{5}{*}{Qwen3-VL-8B} & Conf & 0.690 & 0.810 & 0.805 & 0.836 & 0.616 & 0.875 & \textbf{0.397} & 0.941 \\
 & Brier & \textbf{0.143} & 0.148 & 0.157 & 0.173 & 0.239 & 0.212 & 0.612 & 0.237 \\
 & ECE & \textbf{0.045} & 0.075 & 0.076 & 0.104 & 0.233 & 0.139 & 0.611 & 0.215 \\
 & AUCPR & \textbf{0.932} & 0.929 & 0.921 & 0.905 & 0.636 & 0.755 & 0.700 & 0.781 \\
 & AUROC & 0.835 & \textbf{0.835} & 0.807 & 0.776 & 0.351 & 0.585 & 0.376 & 0.605 \\
\midrule
\multirow{5}{*}{LLaVA-1.6-13B} & Conf & 0.689 & 0.801 & 0.766 & 0.826 & 0.845 & 0.761 & \textbf{0.340} & 0.919 \\
 & Brier & \textbf{0.169} & 0.218 & 0.191 & 0.225 & 0.233 & 0.221 & 0.349 & 0.285 \\
 & ECE & \textbf{0.051} & 0.159 & 0.126 & 0.185 & 0.204 & 0.100 & 0.302 & 0.275 \\
 & AUCPR & \textbf{0.897} & 0.857 & 0.891 & 0.846 & 0.742 & 0.793 & 0.611 & 0.753 \\
 & AUROC & \textbf{0.818} & 0.775 & 0.806 & 0.759 & 0.727 & 0.680 & 0.441 & 0.683 \\
\midrule
\multirow{5}{*}{InternVL3.5-14B} & Conf & \textbf{0.625} & 0.767 & 0.744 & 0.795 & 0.794 & 0.835 & 0.710 & 0.866 \\
 & Brier & \textbf{0.187} & 0.202 & 0.198 & 0.220 & 0.234 & 0.254 & 0.448 & 0.232 \\
 & ECE & \textbf{0.067} & 0.104 & 0.094 & 0.142 & 0.136 & 0.163 & 0.445 & 0.198 \\
 & AUCPR & \textbf{0.880} & 0.857 & 0.865 & 0.782 & 0.725 & 0.620 & 0.752 & 0.791 \\
 & AUROC & \textbf{0.769} & 0.735 & 0.745 & 0.687 & 0.604 & 0.452 & 0.549 & 0.717 \\
\midrule
\multirow{5}{*}{Gemma3-27B} & Conf & 0.663 & 0.725 & 0.756 & 0.732 & 0.745 & 0.888 & \textbf{0.501} & 0.935 \\
 & Brier & 0.164 & 0.158 & 0.158 & \textbf{0.153} & 0.167 & 0.228 & 0.565 & 0.242 \\
 & ECE & 0.070 & 0.036 & 0.048 & \textbf{0.029} & 0.036 & 0.175 & 0.563 & 0.232 \\
 & AUCPR & 0.902 & 0.901 & \textbf{0.907} & 0.903 & 0.865 & 0.831 & 0.741 & 0.819 \\
 & AUROC & 0.803 & 0.802 & 0.806 & \textbf{0.809} & 0.771 & 0.677 & 0.418 & 0.703 \\
\midrule
\multirow{5}{*}{DeepSeek-VL2} & Conf & 0.572 & 0.643 & 0.641 & 0.651 & 0.596 & 0.718 & \textbf{0.452} & 0.896 \\
 & Brier & \textbf{0.177} & 0.182 & 0.189 & 0.198 & 0.224 & 0.251 & 0.416 & 0.372 \\
 & ECE & 0.053 & 0.083 & 0.082 & 0.097 & \textbf{0.046} & 0.157 & 0.367 & 0.341 \\
 & AUCPR & \textbf{0.862} & 0.861 & 0.852 & 0.818 & 0.741 & 0.708 & 0.597 & 0.628 \\
 & AUROC & \textbf{0.814} & 0.813 & 0.797 & 0.786 & 0.679 & 0.700 & 0.430 & 0.599 \\
\bottomrule
\end{tabular}
\end{table*}

%% file: tables/grounding_significance.tex
\begin{table*}[ht]
\centering
\footnotesize
\caption{Paired-bootstrap mean and 95\% CI of $(\mathrm{BS}^{\text{baseline}} - \mathrm{BS}^{\text{BICR}})$ on the image-invariant subset, $B = 2{,}000$ resamples (RNG seed $23$). Positive values mean \texttt{BICR} has lower BS (better calibration) than the baseline on image-invariant samples. \textbf{Bold} marks 95\% CIs strictly above zero (\texttt{BICR} significantly better calibrated).}
\label{tab:grounding_significance}
\resizebox{\textwidth}{!}{
\begin{tabular}{lccccc}
\toprule
\textbf{Baseline} & \textbf{Qwen3-VL-8B} & \textbf{LLaVA-1.6-13B} & \textbf{InternVL3.5-14B} & \textbf{Gemma3-27B} & \textbf{DeepSeek-VL2} \\
\midrule
II & \textbf{+0.005 [+0.00, +0.01]} & \textbf{+0.049 [+0.05, +0.05]} & \textbf{+0.014 [+0.01, +0.02]} & -0.007 [-0.01, -0.00] & \textbf{+0.005 [+0.00, +0.01]} \\
PIK & \textbf{+0.014 [+0.01, +0.02]} & \textbf{+0.022 [+0.02, +0.02]} & \textbf{+0.011 [+0.01, +0.01]} & -0.007 [-0.01, -0.00] & \textbf{+0.012 [+0.01, +0.01]} \\
SAPLMA & \textbf{+0.030 [+0.03, +0.03]} & \textbf{+0.057 [+0.05, +0.06]} & \textbf{+0.033 [+0.03, +0.04]} & -0.011 [-0.01, -0.01] & \textbf{+0.021 [+0.02, +0.02]} \\
CCPS & \textbf{+0.095 [+0.09, +0.10]} & \textbf{+0.064 [+0.06, +0.07]} & \textbf{+0.047 [+0.04, +0.05]} & +0.003 [-0.00, +0.01] & \textbf{+0.048 [+0.04, +0.05]} \\
PE & \textbf{+0.069 [+0.06, +0.07]} & \textbf{+0.056 [+0.05, +0.06]} & \textbf{+0.068 [+0.06, +0.07]} & \textbf{+0.064 [+0.06, +0.07]} & \textbf{+0.074 [+0.07, +0.08]} \\
P(True) & \textbf{+0.469 [+0.46, +0.48]} & \textbf{+0.180 [+0.17, +0.19]} & \textbf{+0.260 [+0.25, +0.27]} & \textbf{+0.401 [+0.39, +0.41]} & \textbf{+0.239 [+0.23, +0.25]} \\
Self-Probing & \textbf{+0.094 [+0.09, +0.10]} & \textbf{+0.117 [+0.11, +0.12]} & \textbf{+0.045 [+0.04, +0.05]} & \textbf{+0.079 [+0.07, +0.09]} & \textbf{+0.195 [+0.19, +0.20]} \\
\bottomrule
\end{tabular}
}
\end{table*}

%% file: tables/grounding_failuremode.tex
\begin{table*}[ht]
\centering
\footnotesize
\caption{Failure subset (image-invariant, incorrect, and confidently predicted): the image-invariant subset further restricted to samples on which the LVLM was both incorrect and assigned the original top-1 token a probability above $0.8$. Lower is better for both Mean Conf and BS. Bold marks the best method per (LVLM, metric) cell.}
\label{tab:grounding_failuremode}
\begin{tabular}{lcc|cc|cc|cc|cc}
\toprule
 & \multicolumn{2}{c|}{\textbf{Qwen3-VL-8B}} & \multicolumn{2}{c|}{\textbf{LLaVA-1.6-13B}} & \multicolumn{2}{c|}{\textbf{InternVL3.5-14B}} & \multicolumn{2}{c|}{\textbf{Gemma3-27B}} & \multicolumn{2}{c}{\textbf{DeepSeek-VL2}} \\
\cmidrule(lr){2-3} \cmidrule(lr){4-5} \cmidrule(lr){6-7} \cmidrule(lr){8-9} \cmidrule(lr){10-11}
\textbf{Method} & Conf & Brier & Conf & Brier & Conf & Brier & Conf & Brier & Conf & Brier \\
\midrule
\textbf{\texttt{BICR}} & \textbf{0.490} & \textbf{0.289} & 0.709 & 0.541 & \textbf{0.484} & \textbf{0.292} & \textbf{0.442} & \textbf{0.275} & \textbf{0.471} & \textbf{0.291} \\
II & 0.682 & 0.491 & 0.770 & 0.605 & 0.709 & 0.519 & 0.583 & 0.369 & 0.547 & 0.336 \\
PIK & 0.673 & 0.496 & 0.793 & 0.661 & 0.653 & 0.477 & 0.579 & 0.394 & 0.528 & 0.336 \\
SAPLMA & 0.746 & 0.600 & 0.790 & 0.659 & 0.719 & 0.574 & 0.549 & 0.357 & 0.519 & 0.336 \\
CCPS & 0.640 & 0.445 & 0.909 & 0.830 & 0.770 & 0.621 & 0.614 & 0.422 & 0.560 & 0.329 \\
PE & 0.869 & 0.757 & 0.745 & 0.559 & 0.848 & 0.723 & 0.873 & 0.763 & 0.712 & 0.511 \\
P(True) & 0.517 & 0.509 & \textbf{0.364} & \textbf{0.151} & 0.817 & 0.790 & 0.622 & 0.613 & 0.483 & 0.321 \\
Self-Probing & 0.911 & 0.870 & 0.920 & 0.865 & 0.772 & 0.678 & 0.873 & 0.816 & 0.868 & 0.799 \\
\bottomrule
\end{tabular}
\end{table*}

%% file: tables/grounding_winsummary.tex
\begin{table*}[ht]
\centering
\footnotesize
\caption{Per-method per-cell win counts across the cross-subset battery defined in \S\ref{app:grounding_detection_winsummary}: $12$ subset definitions for image-invariance, $5$ metrics (Mean Conf, BS, ECE, AUROC, AUCPR), and $5$ LVLMs, for $280$ valid cells in total. Each cell contributes one win to the method whose value is best (lowest for Mean Conf, BS, ECE; highest for AUROC, AUCPR). \textbf{Bold} marks the most-winning method per metric column and overall.}
\label{tab:grounding_winsummary}
\begin{tabular}{lcccccc}
\toprule
\textbf{Method} & \textbf{Mean Conf} & \textbf{BS} & \textbf{ECE} & \textbf{AUROC} & \textbf{AUCPR} & \textbf{Total} \\
\midrule
\textbf{\texttt{BICR}} & 22 & \textbf{38} & \textbf{35} & \textbf{28} & \textbf{30} & \textbf{153} \\
II & 0 & 4 & 4 & 7 & 6 & 21 \\
PIK & 0 & 9 & 5 & 11 & 14 & 39 \\
SAPLMA & 0 & 6 & 6 & 3 & 0 & 15 \\
CCPS & 1 & 1 & 6 & 1 & 0 & 9 \\
PE & 0 & 0 & 2 & 0 & 0 & 2 \\
P(True) & \textbf{37} & 2 & 2 & 0 & 0 & 41 \\
Self-Probing & 0 & 0 & 0 & 0 & 0 & 0 \\
\bottomrule
\end{tabular}
\end{table*}

%% file: references.bib
@inproceedings{ccps,
    title = "Calibrating {LLM} Confidence by Probing Perturbed Representation Stability",
    author = "Khanmohammadi, Reza  and
      Miahi, Erfan  and
      Mardikoraem, Mehrsa  and
      Kaur, Simerjot  and
      Brugere, Ivan  and
      Smiley, Charese  and
      Thind, Kundan S  and
      Ghassemi, Mohammad M.",
    editor = "Christodoulopoulos, Christos  and
      Chakraborty, Tanmoy  and
      Rose, Carolyn  and
      Peng, Violet",
    booktitle = "Proceedings of the 2025 Conference on Empirical Methods in Natural Language Processing",
    month = nov,
    year = "2025",
    address = "Suzhou, China",
    publisher = "Association for Computational Linguistics",
    url = "https://aclanthology.org/2025.emnlp-main.530/",
    doi = "10.18653/v1/2025.emnlp-main.530",
    pages = "10448--10514",
    ISBN = "979-8-89176-332-6",
}

@inproceedings{saplma,
    title = "The Internal State of an {LLM} Knows When It{'}s Lying",
    author = "Azaria, Amos  and
      Mitchell, Tom",
    editor = "Bouamor, Houda  and
      Pino, Juan  and
      Bali, Kalika",
    booktitle = "Findings of the Association for Computational Linguistics: EMNLP 2023",
    month = dec,
    year = "2023",
    address = "Singapore",
    publisher = "Association for Computational Linguistics",
    url = "https://aclanthology.org/2023.findings-emnlp.68/",
    doi = "10.18653/v1/2023.findings-emnlp.68",
    pages = "967--976"
}

@misc{promptensemble,
      title={Confidence Calibration in Vision-Language-Action Models}, 
      author={Thomas P Zollo and Richard Zemel},
      year={2025},
      eprint={2507.17383},
      archivePrefix={arXiv},
      primaryClass={cs.RO},
      url={https://arxiv.org/abs/2507.17383}, 
}

@misc{kadavath,
      title={Language Models (Mostly) Know What They Know}, 
      author={Saurav Kadavath and Tom Conerly and Amanda Askell and Tom Henighan and Dawn Drain and Ethan Perez and Nicholas Schiefer and Zac Hatfield-Dodds and Nova DasSarma and Eli Tran-Johnson and Scott Johnston and Sheer El-Showk and Andy Jones and Nelson Elhage and Tristan Hume and Anna Chen and Yuntao Bai and Sam Bowman and Stanislav Fort and Deep Ganguli and Danny Hernandez and Josh Jacobson and Jackson Kernion and Shauna Kravec and Liane Lovitt and Kamal Ndousse and Catherine Olsson and Sam Ringer and Dario Amodei and Tom Brown and Jack Clark and Nicholas Joseph and Ben Mann and Sam McCandlish and Chris Olah and Jared Kaplan},
      year={2022},
      eprint={2207.05221},
      archivePrefix={arXiv},
      primaryClass={cs.CL},
      url={https://arxiv.org/abs/2207.05221}, 
}

@inproceedings{
self-probing,
title={Can {LLM}s Express Their Uncertainty? An Empirical Evaluation of Confidence Elicitation in {LLM}s},
author={Miao Xiong and Zhiyuan Hu and Xinyang Lu and YIFEI LI and Jie Fu and Junxian He and Bryan Hooi},
booktitle={The Twelfth International Conference on Learning Representations},
year={2024},
url={https://openreview.net/forum?id=gjeQKFxFpZ}
}

@inproceedings{GQA,
  title={{GQA}: A new dataset for real-world visual reasoning and compositional question answering},
  author={Hudson, Drew A and Manning, Christopher D},
  booktitle={Proceedings of the IEEE/CVF conference on computer vision and pattern recognition},
  pages={6700--6709},
  year={2019}
}

@inproceedings{POPE,
    title = "Evaluating Object Hallucination in Large Vision-Language Models",
    author = "Li, Yifan  and
      Du, Yifan  and
      Zhou, Kun  and
      Wang, Jinpeng  and
      Zhao, Xin  and
      Wen, Ji-Rong",
    editor = "Bouamor, Houda  and
      Pino, Juan  and
      Bali, Kalika",
    booktitle = "Proceedings of the 2023 Conference on Empirical Methods in Natural Language Processing",
    month = dec,
    year = "2023",
    address = "Singapore",
    publisher = "Association for Computational Linguistics",
    url = "https://aclanthology.org/2023.emnlp-main.20/",
    doi = "10.18653/v1/2023.emnlp-main.20",
    pages = "292--305",
}

@inproceedings{GMAI-MMBench,
author = {Chen, Pengcheng and Ye, Jin and Wang, Guoan and Li, Yanjun and Deng, Zhongying and Li, Wei and Li, Tianbin and Duan, Haodong and Huang, Ziyan and Su, Yanzhou and Wang, Benyou and Zhang, Shaoting and Fu, Bin and Cai, Jianfei and Zhuang, Bohan and Seibel, Eric J and Qiao, Yu and He, Junjun},
title = {{GMAI-MMBench}: A Comprehensive Multimodal Evaluation Benchmark Towards General Medical {AI}},
year = {2024},
isbn = {9798331314385},
publisher = {Curran Associates Inc.},
address = {Red Hook, NY, USA},
booktitle = {Proceedings of the 38th International Conference on Neural Information Processing Systems},
articleno = {2992},
numpages = {101},
location = {Vancouver, BC, Canada},
series = {NIPS '24}
}

@inproceedings{MME-Finance,
author = {Gan, Ziliang and Zhang, Dong and Li, Haohan and Wu, Yang and Lin, Xueyuan and Liu, Ji and Wu, Haipang and Fu, Chaoyou and Xu, Zenglin and Zhang, Rongjunchen and Dai, Yong},
title = {{MME-Finance}: A Multimodal Finance Benchmark for Expert-level Understanding and Reasoning},
year = {2025},
isbn = {9798400720352},
publisher = {Association for Computing Machinery},
address = {New York, NY, USA},
url = {https://doi.org/10.1145/3746027.3758230},
doi = {10.1145/3746027.3758230},
booktitle = {Proceedings of the 33rd ACM International Conference on Multimedia},
pages = {12867–12874},
numpages = {8},
location = {Dublin, Ireland},
series = {MM '25}
}

@inproceedings{MMMU,
    title = "{MMMU}-Pro: A More Robust Multi-discipline Multimodal Understanding Benchmark",
    author = "Yue, Xiang  and
      Zheng, Tianyu  and
      Ni, Yuansheng  and
      Wang, Yubo  and
      Zhang, Kai  and
      Tong, Shengbang  and
      Sun, Yuxuan  and
      Yu, Botao  and
      Zhang, Ge  and
      Sun, Huan  and
      Su, Yu  and
      Chen, Wenhu  and
      Neubig, Graham",
    editor = "Che, Wanxiang  and
      Nabende, Joyce  and
      Shutova, Ekaterina  and
      Pilehvar, Mohammad Taher",
    booktitle = "Proceedings of the 63rd Annual Meeting of the Association for Computational Linguistics (Volume 1: Long Papers)",
    month = jul,
    year = "2025",
    address = "Vienna, Austria",
    publisher = "Association for Computational Linguistics",
    url = "https://aclanthology.org/2025.acl-long.736/",
    doi = "10.18653/v1/2025.acl-long.736",
    pages = "15134--15186",
    ISBN = "979-8-89176-251-0",
}

@inproceedings{LlavaWild,
author = {Liu, Haotian and Li, Chunyuan and Wu, Qingyang and Lee, Yong Jae},
title = {Visual instruction tuning},
year = {2023},
publisher = {Curran Associates Inc.},
address = {Red Hook, NY, USA},
booktitle = {Proceedings of the 37th International Conference on Neural Information Processing Systems},
articleno = {1516},
numpages = {25},
location = {New Orleans, LA, USA},
series = {NIPS '23}
}

@misc{qwen3,
      title={Qwen3 Technical Report}, 
      author={Qwen Team},
      year={2025},
      eprint={2505.09388},
      archivePrefix={arXiv},
      primaryClass={cs.CL},
      url={https://arxiv.org/abs/2505.09388}, 
}

@INPROCEEDINGS{llava,
  author={Liu, Haotian and Li, Chunyuan and Li, Yuheng and Lee, Yong Jae},
  booktitle={2024 IEEE/CVF Conference on Computer Vision and Pattern Recognition (CVPR)}, 
  title={Improved Baselines with Visual Instruction Tuning}, 
  year={2024},
  volume={},
  number={},
  pages={26296--26306},
  doi={10.1109/CVPR52733.2024.02484}}

@article{internvl3_5,
  title={InternVL3.5: Advancing Open-Source Multimodal Models in Versatility, Reasoning, and Efficiency},
  author={Wang, Weiyun and Gao, Zhangwei and Gu, Lixin and Pu, Hengjun and Cui, Long and Wei, Xingguang and Liu, Zhaoyang and Jing, Linglin and Ye, Shenglong and Shao, Jie and others},
  journal={arXiv preprint arXiv:2508.18265},
  year={2025}
}

@misc{gemma3,
      title={{Gemma} 3 Technical Report},
      author={{Gemma Team} and Aishwarya Kamath and Johan Ferret and Shreya Pathak and Nino Vieillard and Ramona Merhej and Sarah Perrin and Tatiana Matejovicova and Alexandre Ramé and Morgane Rivière and Louis Rouillard and Thomas Mesnard and Geoffrey Cideron and Jean-Bastien Grill and Sabela Ramos and Edouard Yvinec and Michelle Casbon and Etienne Pot and Ivo Penchev and Gaël Liu and Francesco Visin and Kathleen Kenealy and Lucas Beyer and Xiaohai Zhai and Anton Tsitsulin and Robert Busa-Fekete and Alex Feng and Noveen Sachdeva and Benjamin Coleman and Yi Gao and Basil Mustafa and Iain Barr and Emilio Parisotto and David Tian and Matan Eyal and Colin Cherry and Jan-Thorsten Peter and Danila Sinopalnikov and Surya Bhupatiraju and Rishabh Agarwal and Mehran Kazemi and Dan Malkin and Ravin Kumar and David Vilar and Idan Brusilovsky and Jiaming Luo and Andreas Steiner and Abe Friesen and Abhanshu Sharma and Abheesht Sharma and Adi Mayrav Gilady and Adrian Goedeckemeyer and Alaa Saade and Alex Feng and Alexander Kolesnikov and Alexei Bendebury and Alvin Abdagic and Amit Vadi and András György and André Susano Pinto and Anil Das and Ankur Bapna and Antoine Miech and Antoine Yang and Antonia Paterson and Ashish Shenoy and Ayan Chakrabarti and Bilal Piot and Bo Wu and Bobak Shahriari and Bryce Petrini and Charlie Chen and Charline Le Lan and Christopher A. Choquette-Choo and CJ Carey and Cormac Brick and Daniel Deutsch and Danielle Eisenbud and Dee Cattle and Derek Cheng and Dimitris Paparas and Divyashree Shivakumar Sreepathihalli and Doug Reid and Dustin Tran and Dustin Zelle and Eric Noland and Erwin Huizenga and Eugene Kharitonov and Frederick Liu and Gagik Amirkhanyan and Glenn Cameron and Hadi Hashemi and Hanna Klimczak-Plucińska and Harman Singh and Harsh Mehta and Harshal Tushar Lehri and Hussein Hazimeh and Ian Ballantyne and Idan Szpektor and Ivan Nardini and Jean Pouget-Abadie and Jetha Chan and Joe Stanton and John Wieting and Jonathan Lai and Jordi Orbay and Joseph Fernandez and Josh Newlan and Ju-Yeong Ji and Jyotinder Singh and Kat Black and Kathy Yu and Kevin Hui and Kiran Vodrahalli and Klaus Greff and Linhai Qiu and Marcella Valentine and Marina Coelho and Marvin Ritter and Matt Hoffman and Matthew Watson and Mayank Chaturvedi and Michael Moynihan and Min Ma and Nabila Babar and Natasha Noy and Nathan Byrd and Nick Roy and Nikola Momchev and Nilay Chauhan and Noveen Sachdeva and Oskar Bunyan and Pankil Botarda and Paul Caron and Paul Kishan Rubenstein and Phil Culliton and Philipp Schmid and Pier Giuseppe Sessa and Pingmei Xu and Piotr Stanczyk and Pouya Tafti and Rakesh Shivanna and Renjie Wu and Renke Pan and Reza Rokni and Rob Willoughby and Rohith Vallu and Ryan Mullins and Sammy Jerome and Sara Smoot and Sertan Girgin and Shariq Iqbal and Shashir Reddy and Shruti Sheth and Siim Põder and Sijal Bhatnagar and Sindhu Raghuram Panyam and Sivan Eiger and Susan Zhang and Tianqi Liu and Trevor Yacovone and Tyler Liechty and Uday Kalra and Utku Evci and Vedant Misra and Vincent Roseberry and Vlad Feinberg and Vlad Kolesnikov and Woohyun Han and Woosuk Kwon and Xi Chen and Yinlam Chow and Yuvein Zhu and Zichuan Wei and Zoltan Egyed and Victor Cotruta and Minh Giang and Phoebe Kirk and Anand Rao and Kat Black and Nabila Babar and Jessica Lo and Erica Moreira and Luiz Gustavo Martins and Omar Sanseviero and Lucas Gonzalez and Zach Gleicher and Tris Warkentin and Vahab Mirrokni and Evan Senter and Eli Collins and Joelle Barral and Zoubin Ghahramani and Raia Hadsell and Yossi Matias and D. Sculley and Slav Petrov and Noah Fiedel and Noam Shazeer and Oriol Vinyals and Jeff Dean and Demis Hassabis and Koray Kavukcuoglu and Clement Farabet and Elena Buchatskaya and Jean-Baptiste Alayrac and Rohan Anil and Dmitry Lepikhin and Sebastian Borgeaud and Olivier Bachem and Armand Joulin and Alek Andreev and Cassidy Hardin and Robert Dadashi and Léonard Hussenot},
      year={2025},
      eprint={2503.19786},
      archivePrefix={arXiv},
      primaryClass={cs.CL},
      url={https://arxiv.org/abs/2503.19786}, 
}

@misc{deepseek-vl2,
      title={DeepSeek-VL2: Mixture-of-Experts Vision-Language Models for Advanced Multimodal Understanding}, 
      author={Zhiyu Wu and Xiaokang Chen and Zizheng Pan and Xingchao Liu and Wen Liu and Damai Dai and Huazuo Gao and Yiyang Ma and Chengyue Wu and Bingxuan Wang and Zhenda Xie and Yu Wu and Kai Hu and Jiawei Wang and Yaofeng Sun and Yukun Li and Yishi Piao and Kang Guan and Aixin Liu and Xin Xie and Yuxiang You and Kai Dong and Xingkai Yu and Haowei Zhang and Liang Zhao and Yisong Wang and Chong Ruan},
      year={2024},
      eprint={2412.10302},
      archivePrefix={arXiv},
      primaryClass={cs.CV},
      url={https://arxiv.org/abs/2412.10302}, 
}

@inproceedings{
PHSV,
title={Reasoning Models Know When They{\textquoteright}re Right: Probing Hidden States for Self-Verification},
author={Anqi Zhang and Yulin Chen and Jane Pan and Chen Zhao and Aurojit Panda and Jinyang Li and He He},
booktitle={Second Conference on Language Modeling},
year={2025},
url={https://openreview.net/forum?id=O6I0Av7683}
}

@inproceedings{calibration-tuning,
    title = "Calibration-Tuning: Teaching Large Language Models to Know What They Don{'}t Know",
    author = "Kapoor, Sanyam  and
      Gruver, Nate  and
      Roberts, Manley  and
      Pal, Arka  and
      Dooley, Samuel  and
      Goldblum, Micah  and
      Wilson, Andrew",
    editor = {V{\'a}zquez, Ra{\'u}l  and
      Celikkanat, Hande  and
      Ulmer, Dennis  and
      Tiedemann, J{\"o}rg  and
      Swayamdipta, Swabha  and
      Aziz, Wilker  and
      Plank, Barbara  and
      Baan, Joris  and
      de Marneffe, Marie-Catherine},
    booktitle = "Proceedings of the 1st Workshop on Uncertainty-Aware NLP (UncertaiNLP 2024)",
    month = mar,
    year = "2024",
    address = "St Julians, Malta",
    publisher = "Association for Computational Linguistics",
    url = "https://aclanthology.org/2024.uncertainlp-1.1/",
    pages = "1--14",
}

@misc{emit,
      title={Finetuning Language Models to Emit Linguistic Expressions of Uncertainty}, 
      author={Arslan Chaudhry and Sridhar Thiagarajan and Dilan Gorur},
      year={2024},
      eprint={2409.12180},
      archivePrefix={arXiv},
      primaryClass={cs.CL},
      url={https://arxiv.org/abs/2409.12180}, 
}

@inproceedings{rmcb,
    title = "How Reliable are Confidence Estimators for Large Reasoning Models? A Systematic Benchmark on High-Stakes Domains",
    author = "Khanmohammadi, Reza  and
      Miahi, Erfan  and
      Kaur, Simerjot  and
      Smiley, Charese  and
      Brugere, Ivan  and
      Thind, Kundan S  and
      Ghassemi, Mohammad M.",
    editor = "Demberg, Vera  and
      Inui, Kentaro  and
      Marquez, Llu{\'i}s",
    booktitle = "Proceedings of the 19th Conference of the {E}uropean Chapter of the {A}ssociation for {C}omputational {L}inguistics (Volume 1: Long Papers)",
    month = mar,
    year = "2026",
    address = "Rabat, Morocco",
    publisher = "Association for Computational Linguistics",
    url = "https://aclanthology.org/2026.eacl-long.78/",
    doi = "10.18653/v1/2026.eacl-long.78",
    pages = "1669--1754",
    ISBN = "979-8-89176-380-7",
}

@inproceedings{MSCOCO,
  title={Microsoft COCO: Common Objects in Context},
  author={Tsung-Yi Lin and Michael Maire and Serge J. Belongie and James Hays and Pietro Perona and Deva Ramanan and Piotr Doll{\'a}r and C. Lawrence Zitnick},
  booktitle={European Conference on Computer Vision},
  year={2014},
}

@inproceedings{internalinspector,
    title = "{I}nternal{I}nspector $I^2$: Robust Confidence Estimation in {LLM}s through Internal States",
    author = "Beigi, Mohammad  and
      Shen, Ying  and
      Yang, Runing  and
      Lin, Zihao  and
      Wang, Qifan  and
      Mohan, Ankith  and
      He, Jianfeng  and
      Jin, Ming  and
      Lu, Chang-Tien  and
      Huang, Lifu",
    editor = "Al-Onaizan, Yaser  and
      Bansal, Mohit  and
      Chen, Yun-Nung",
    booktitle = "Findings of the Association for Computational Linguistics: EMNLP 2024",
    month = nov,
    year = "2024",
    address = "Miami, Florida, USA",
    publisher = "Association for Computational Linguistics",
    url = "https://aclanthology.org/2024.findings-emnlp.751/",
    doi = "10.18653/v1/2024.findings-emnlp.751",
    pages = "12847--12865",
}

@inproceedings{optuna,
  title={{O}ptuna: A Next-Generation Hyperparameter Optimization Framework},
  author={Akiba, Takuya and Sano, Shotaro and Yanase, Toshihiko and Ohta, Takeru and Koyama, Masanori},
  booktitle={The 25th ACM SIGKDD International Conference on Knowledge Discovery \& Data Mining},
  pages={2623--2631},
  year={2019}
}

@inproceedings{vl-calibration,
    title = "{VL}-Calibration: Decoupled Confidence Calibration for Large Vision-Language Models Reasoning",
    author = "Xiao, Wenyi  and
      XU, Xinchi  and
      Gan, Leilei",
    editor = "Liakata, Maria  and
      Moreira, Viviane P.  and
      Zhang, Jiajun  and
      Jurgens, David",
    booktitle = "Proceedings of the 64th Annual Meeting of the {A}ssociation for {C}omputational {L}inguistics (Volume 1: Long Papers)",
    month = jul,
    year = "2026",
    address = "San Diego, California, United States",
    publisher = "Association for Computational Linguistics",
    url = "https://aclanthology.org/2026.acl-long.2074/",
    doi = "10.18653/v1/2026.acl-long.2074",
    pages = "44791--44815",
    ISBN = "979-8-89176-390-6"
}

@inproceedings{
self-consistency,
title={Self-Consistency Improves Chain of Thought Reasoning in Language Models},
author={Xuezhi Wang and Jason Wei and Dale Schuurmans and Quoc V Le and Ed H. Chi and Sharan Narang and Aakanksha Chowdhery and Denny Zhou},
booktitle={The Eleventh International Conference on Learning Representations },
year={2023},
url={https://openreview.net/forum?id=1PL1NIMMrw}
}

@inproceedings{
zhou2025hademif,
title={HaDeMiF: Hallucination Detection and Mitigation in Large Language Models},
author={Xiaoling Zhou and Mingjie Zhang and Zhemg Lee and Wei Ye and Shikun Zhang},
booktitle={The Thirteenth International Conference on Learning Representations},
year={2025},
url={https://openreview.net/forum?id=VwOYxPScxB}
}

@inproceedings{tempscaling,
author = {Guo, Chuan and Pleiss, Geoff and Sun, Yu and Weinberger, Kilian Q.},
title = {On calibration of modern neural networks},
year = {2017},
publisher = {JMLR.org},
booktitle = {Proceedings of the 34th International Conference on Machine Learning - Volume 70},
pages = {1321–1330},
numpages = {10},
location = {Sydney, NSW, Australia},
series = {ICML'17}
}

@misc{Wani2024HiddenInstability,
      title={Same Answer, Different Representations: Hidden instability in VLMs}, 
      author={Farooq Ahmad Wani and Alessandro Suglia and Rohit Saxena and Aryo Pradipta Gema and Wai-Chung Kwan and Fazl Barez and Maria Sofia Bucarelli and Fabrizio Silvestri and Pasquale Minervini},
      year={2026},
      eprint={2602.06652},
      archivePrefix={arXiv},
      primaryClass={cs.AI},
      url={https://arxiv.org/abs/2602.06652}, 
}

@misc{dang2026instinct,
      title={Instinct vs. Reflection: Unifying Token and Verbalized Confidence in Multimodal Large Models}, 
      author={Yunkai Dang and Yifan Jiang and Yizhu Jiang and Anqi Chen and Wenbin Li and Yang Gao},
      year={2026},
      eprint={2604.17274},
      archivePrefix={arXiv},
      primaryClass={cs.CV},
      url={https://arxiv.org/abs/2604.17274}, 
}

@misc{Zhao2025SemanticCalibration,
      title={Object-Level Verbalized Confidence Calibration in Vision-Language Models via Semantic Perturbation}, 
      author={Yunpu Zhao and Rui Zhang and Junbin Xiao and Ruibo Hou and Jiaming Guo and Zihao Zhang and Yifan Hao and Yunji Chen},
      year={2025},
      eprint={2504.14848},
      archivePrefix={arXiv},
      primaryClass={cs.CV},
      url={https://arxiv.org/abs/2504.14848}, 
}

@inproceedings{
yang2025mitigating,
title={Mitigating Hallucination in Large Vision-Language Models via Modular Attribution and Intervention},
author={Tianyun Yang and Ziniu Li and Juan Cao and Chang Xu},
booktitle={The Thirteenth International Conference on Learning Representations},
year={2025},
url={https://openreview.net/forum?id=Bjq4W7P2Us}
}

@inproceedings{Woo2025AVISC,
    title = "Don{'}t Miss the Forest for the Trees: Attentional Vision Calibration for Large Vision Language Models",
    author = "Woo, Sangmin  and
      Kim, Donguk  and
      Jang, Jaehyuk  and
      Choi, Yubin  and
      Kim, Changick",
    editor = "Che, Wanxiang  and
      Nabende, Joyce  and
      Shutova, Ekaterina  and
      Pilehvar, Mohammad Taher",
    booktitle = "Findings of the Association for Computational Linguistics: ACL 2025",
    month = jul,
    year = "2025",
    address = "Vienna, Austria",
    publisher = "Association for Computational Linguistics",
    url = "https://aclanthology.org/2025.findings-acl.99/",
    doi = "10.18653/v1/2025.findings-acl.99",
    pages = "1927--1951",
    ISBN = "979-8-89176-256-5",
}

@inproceedings{chen2025unveilinguncertainty,
    title = "Unveiling Uncertainty: A Deep Dive into Calibration and Performance of Multimodal Large Language Models",
    author = "Chen, Zijun  and
      Hu, Wenbo  and
      He, Guande  and
      Deng, Zhijie  and
      Zhang, Zheng  and
      Hong, Richang",
    editor = "Rambow, Owen  and
      Wanner, Leo  and
      Apidianaki, Marianna  and
      Al-Khalifa, Hend  and
      Eugenio, Barbara Di  and
      Schockaert, Steven",
    booktitle = "Proceedings of the 31st International Conference on Computational Linguistics",
    month = jan,
    year = "2025",
    address = "Abu Dhabi, UAE",
    publisher = "Association for Computational Linguistics",
    url = "https://aclanthology.org/2025.coling-main.208/",
    pages = "3095--3109",
}

@inproceedings{xuan2025seeingbelieving,
    title = "Seeing is Believing, but How Much? A Comprehensive Analysis of Verbalized Calibration in Vision-Language Models",
    author = "Xuan, Weihao  and
      Zeng, Qingcheng  and
      Qi, Heli  and
      Wang, Junjue  and
      Yokoya, Naoto",
    editor = "Christodoulopoulos, Christos  and
      Chakraborty, Tanmoy  and
      Rose, Carolyn  and
      Peng, Violet",
    booktitle = "Proceedings of the 2025 Conference on Empirical Methods in Natural Language Processing",
    month = nov,
    year = "2025",
    address = "Suzhou, China",
    publisher = "Association for Computational Linguistics",
    url = "https://aclanthology.org/2025.emnlp-main.74/",
    doi = "10.18653/v1/2025.emnlp-main.74",
    pages = "1408--1450",
    ISBN = "979-8-89176-332-6",
}

@inproceedings{li2024referencefree,
    title = "Reference-free Hallucination Detection for Large Vision-Language Models",
    author = "Li, Qing  and
      Geng, Jiahui  and
      Lyu, Chenyang  and
      Zhu, Derui  and
      Panov, Maxim  and
      Karray, Fakhri",
    editor = "Al-Onaizan, Yaser  and
      Bansal, Mohit  and
      Chen, Yun-Nung",
    booktitle = "Findings of the Association for Computational Linguistics: EMNLP 2024",
    month = nov,
    year = "2024",
    address = "Miami, Florida, USA",
    publisher = "Association for Computational Linguistics",
    url = "https://aclanthology.org/2024.findings-emnlp.262/",
    doi = "10.18653/v1/2024.findings-emnlp.262",
    pages = "4542--4551",
}

@inproceedings{Du2026MedicalVQA,
author = {Du, Yuetian and Wang, Yucheng and Kong, Ming and Liang, Tian and Long, Qiang and Chen, Bingdi and Zhu, Qiang},
title = {Confidence Calibration for Multimodal {LLM}s: An Empirical Study Through Medical {VQA}},
year = {2025},
isbn = {978-3-032-04977-3},
publisher = {Springer-Verlag},
address = {Berlin, Heidelberg},
url = {https://doi.org/10.1007/978-3-032-04978-0_9},
doi = {10.1007/978-3-032-04978-0_9},
booktitle = {Medical Image Computing and Computer Assisted Intervention – MICCAI 2025: 28th International Conference, Daejeon, South Korea, September 23–27, 2025, Proceedings, Part VI},
pages = {89–99},
numpages = {11},
location = {Daejeon, Korea (Republic of)}
}

@inproceedings{tian-etal-2023-just,
    title = "Just Ask for Calibration: Strategies for Eliciting Calibrated Confidence Scores from Language Models Fine-Tuned with Human Feedback",
    author = "Tian, Katherine  and
      Mitchell, Eric  and
      Zhou, Allan  and
      Sharma, Archit  and
      Rafailov, Rafael  and
      Yao, Huaxiu  and
      Finn, Chelsea  and
      Manning, Christopher",
    editor = "Bouamor, Houda  and
      Pino, Juan  and
      Bali, Kalika",
    booktitle = "Proceedings of the 2023 Conference on Empirical Methods in Natural Language Processing",
    month = dec,
    year = "2023",
    address = "Singapore",
    publisher = "Association for Computational Linguistics",
    url = "https://aclanthology.org/2023.emnlp-main.330/",
    doi = "10.18653/v1/2023.emnlp-main.330",
    pages = "5433--5442",
}

@inproceedings{zeng-etal-2025-thinking,
    title = "Thinking Out Loud: Do Reasoning Models Know When They{'}re Right?",
    author = "Zeng, Qingcheng  and
      Xuan, Weihao  and
      Cui, Leyang  and
      Voigt, Rob",
    editor = "Christodoulopoulos, Christos  and
      Chakraborty, Tanmoy  and
      Rose, Carolyn  and
      Peng, Violet",
    booktitle = "Proceedings of the 2025 Conference on Empirical Methods in Natural Language Processing",
    month = nov,
    year = "2025",
    address = "Suzhou, China",
    publisher = "Association for Computational Linguistics",
    url = "https://aclanthology.org/2025.emnlp-main.73/",
    doi = "10.18653/v1/2025.emnlp-main.73",
    pages = "1394--1407",
    ISBN = "979-8-89176-332-6",
}

@article{jiang-etal-2021-know,
    title = "How Can We Know When Language Models Know? On the Calibration of Language Models for Question Answering",
    author = "Jiang, Zhengbao  and
      Araki, Jun  and
      Ding, Haibo  and
      Neubig, Graham",
    editor = "Roark, Brian  and
      Nenkova, Ani",
    journal = "Transactions of the Association for Computational Linguistics",
    volume = "9",
    year = "2021",
    address = "Cambridge, MA",
    publisher = "MIT Press",
    url = "https://aclanthology.org/2021.tacl-1.57/",
    doi = "10.1162/tacl_a_00407",
    pages = "962--977",
}

@inproceedings{
nakkiran2025trainedtokenscalibratedconcepts,
title={Trained on Tokens, Calibrated on Concepts: The Emergence of Semantic Calibration in {LLM}s},
author={Preetum Nakkiran and Arwen Bradley and Adam Golinski and Eugene Ndiaye and Michael Kirchhof and Sinead Williamson},
booktitle={The Fourteenth International Conference on Learning Representations},
year={2026},
url={https://openreview.net/forum?id=0sCyk9Tr5J}
}

@inproceedings{vllm,
author = {Kwon, Woosuk and Li, Zhuohan and Zhuang, Siyuan and Sheng, Ying and Zheng, Lianmin and Yu, Cody Hao and Gonzalez, Joseph and Zhang, Hao and Stoica, Ion},
title = {Efficient Memory Management for Large Language Model Serving with PagedAttention},
year = {2023},
isbn = {9798400702297},
publisher = {Association for Computing Machinery},
address = {New York, NY, USA},
url = {https://doi.org/10.1145/3600006.3613165},
doi = {10.1145/3600006.3613165},
booktitle = {Proceedings of the 29th Symposium on Operating Systems Principles},
pages = {611–626},
numpages = {16},
location = {Koblenz, Germany},
series = {SOSP '23}
}

@INPROCEEDINGS{10657718,
  author={Leng, Sicong and Zhang, Hang and Chen, Guanzheng and Li, Xin and Lu, Shijian and Miao, Chunyan and Bing, Lidong},
  booktitle={2024 IEEE/CVF Conference on Computer Vision and Pattern Recognition (CVPR)}, 
  title={Mitigating Object Hallucinations in Large Vision-Language Models through Visual Contrastive Decoding}, 
  year={2024},
  volume={},
  number={},
  pages={13872-13882},
  doi={10.1109/CVPR52733.2024.01316}}

@misc{VL-Uncertainty,
      title={VL-Uncertainty: Detecting Hallucination in Large Vision-Language Model via Uncertainty Estimation}, 
      author={Ruiyang Zhang and Hu Zhang and Zhedong Zheng},
      year={2024},
      eprint={2411.11919},
      archivePrefix={arXiv},
      primaryClass={cs.CV},
      url={https://arxiv.org/abs/2411.11919}, 
}

@inproceedings{
Chain-of-Embedding,
title={Understanding Language Prior of {LVLM}s by Contrasting Chain-of-Embedding},
author={Lin Long and Changdae Oh and Seongheon Park and Sharon Li},
booktitle={The Fourteenth International Conference on Learning Representations},
year={2026},
url={https://openreview.net/forum?id=wnTnwWKu72}
}

@INPROCEEDINGS{Devils,
  author={Jiang, Zhangqi and Chen, Junkai and Zhu, Beier and Luo, Tingjin and Shen, Yankun and Yang, Xu},
  booktitle={2025 IEEE/CVF Conference on Computer Vision and Pattern Recognition (CVPR)}, 
  title={Devils in Middle Layers of Large Vision-Language Models: Interpreting, Detecting and Mitigating Object Hallucinations via Attention Lens}, 
  year={2025},
  volume={},
  number={},
  pages={25004-25014},
  doi={10.1109/CVPR52734.2025.02328}}

@inproceedings{
Hidden,
title={Hidden in plain sight: {VLM}s overlook their visual representations},
author={Stephanie Fu and Tyler Bonnen and Devin Guillory and Trevor Darrell},
booktitle={Second Conference on Language Modeling},
year={2025},
url={https://openreview.net/forum?id=qQb1JLrwol}
}

@inproceedings{
geometric-mean1,
title={Semantic Uncertainty: Linguistic Invariances for Uncertainty Estimation in Natural Language Generation},
author={Lorenz Kuhn and Yarin Gal and Sebastian Farquhar},
booktitle={The Eleventh International Conference on Learning Representations },
year={2023},
url={https://openreview.net/forum?id=VD-AYtP0dve}
}

@inproceedings{
geometric-mean2,
title={Uncertainty Estimation in Autoregressive Structured Prediction},
author={Andrey Malinin and Mark Gales},
booktitle={International Conference on Learning Representations},
year={2021},
url={https://openreview.net/forum?id=jN5y-zb5Q7m}
}

@inproceedings{MemVR,
author = {Zou, Xin and Wang, Yizhou and Yan, Yibo and lyu, Yuanhuiyi and Zheng, Kening and Huang, Sirui and Chen, Junkai and Jiang, Peijie and Liu, Jia and Tang, Chang and Hu, Xuming},
title = {Look twice before you answer: memory-space visual retracing for hallucination mitigation in multimodal large language models},
year = {2025},
publisher = {JMLR.org},
booktitle = {Proceedings of the 42nd International Conference on Machine Learning},
articleno = {3254},
numpages = {27},
location = {Vancouver, Canada},
series = {ICML'25}
}
